\documentclass[10pt,twocolumn,letterpaper]{article}

\usepackage{iccv}
\usepackage{times}
\usepackage{epsfig}
\usepackage{graphicx}
\usepackage{amsmath}
\usepackage{amssymb}
\usepackage{mathtools}
\usepackage{resizegather}
\usepackage{marvosym}
\usepackage[linesnumbered,ruled]{algorithm2e}
\usepackage{dsfont}
\usepackage{enumitem}
\usepackage{caption}
\usepackage{multirow}
\usepackage{amsthm}
\usepackage[section]{placeins}
\usepackage{color}
\usepackage[table]{xcolor}
\definecolor{visda_color_0}{rgb}{0.9922,0.0,0.0235}
\definecolor{visda_color_1}{rgb}{0.9922,0.4235,0.0314}
\definecolor{visda_color_2}{rgb}{1.0,1.0,0.3098}
\definecolor{visda_color_3}{rgb}{0.4471,1.0,0.0275}
\definecolor{visda_color_4}{rgb}{0.1686,1.0,0.0980}
\definecolor{visda_color_5}{rgb}{0.1333,1.0,0.4275}
\definecolor{visda_color_6}{rgb}{0.1333,1.0,1.0}
\definecolor{visda_color_7}{rgb}{0.0510,0.4,1.0}
\definecolor{visda_color_8}{rgb}{0.0,0.0,1.0000}
\definecolor{visda_color_9}{rgb}{0.4196,0.0,1.0}
\definecolor{visda_color_10}{rgb}{0.9882,0.0,1.0}
\definecolor{visda_color_11}{rgb}{0.9922,0.0,0.4275}
\definecolor{city_color_0}{rgb}{0.0,0.0,0.0}
\definecolor{city_color_1}{rgb}{0.5020,0.2510,0.5020}
\definecolor{city_color_2}{rgb}{0.9569,0.1373,0.9098}
\definecolor{city_color_3}{rgb}{0.2745,0.2745,0.2745}
\definecolor{city_color_4}{rgb}{0.4000,0.4000,0.6118}
\definecolor{city_color_5}{rgb}{0.7451,0.6000,0.6000}
\definecolor{city_color_6}{rgb}{0.6000,0.6000,0.6000}
\definecolor{city_color_7}{rgb}{0.9804,0.6667,0.1176}
\definecolor{city_color_8}{rgb}{0.8627,0.8627,0.0000}
\definecolor{city_color_9}{rgb}{0.4196,0.5569,0.1373}
\definecolor{city_color_10}{rgb}{0.5961,0.9843,0.5961}
\definecolor{city_color_11}{rgb}{0.2745,0.5098,0.7059}
\definecolor{city_color_12}{rgb}{0.8627,0.0784,0.2353}
\definecolor{city_color_13}{rgb}{1.0000,0.0000,0.0000}
\definecolor{city_color_14}{rgb}{0.0000,0.0000,0.5569}
\definecolor{city_color_15}{rgb}{0.0000,0.0000,0.2745}
\definecolor{city_color_16}{rgb}{0.0000,0.2353,0.3922}
\definecolor{city_color_17}{rgb}{0.0000,0.3137,0.3922}
\definecolor{city_color_18}{rgb}{0.0000,0.0000,0.9020}
\definecolor{city_color_19}{rgb}{0.4667,0.0431,0.1255}
\definecolor{citecolor}{RGB}{119,185,0}
\usepackage[hypertexnames=false,pagebackref=true,breaklinks=true,letterpaper=true,colorlinks,citecolor=citecolor,bookmarks=false]{hyperref}

\theoremstyle{plain}

\newtheorem{proposition}{Proposition}

\newcommand{\MYhref}[3][blue]{\href{#2}{\color{#1}{#3}}}%

\usepackage{amsmath,amsfonts,bm}

\def\eqref#1{equation~\ref{#1}}

\def\1{\bm{1}}

\DeclareMathAlphabet{\mathsfit}{\encodingdefault}{\sfdefault}{m}{sl}
\SetMathAlphabet{\mathsfit}{bold}{\encodingdefault}{\sfdefault}{bx}{n}

\DeclareMathOperator*{\argmax}{arg\,max}

\iccvfinalcopy 

\def\iccvPaperID{0774} 

\ificcvfinal\fi

\begin{document}

\title{Confidence Regularized Self-Training}

\author{
Yang Zou$^{1*}$~~~~
Zhiding Yu$^{2}$\thanks{The authors contributed equally.}~~~~
Xiaofeng Liu$^{1}$~~~~
B.V.K. Vijaya Kumar$^{1}$~~~~
Jinsong Wang$^{3}$\thanks{Work done during the affiliation with General Motors R\&D.}\\
$^{1}$~Carnegie Mellon University~~~~
$^{2}$~NVIDIA~~~~
$^{3}$~General Motors R\&D\\
{\small\Letter~~\tt
\MYhref[black]{mailto:yzou2@andrew.cmu.edu}{yzou2@andrew.cmu.edu}, 
\MYhref[black]{mailto:zhidingy@nvidia.com}{zhidingy@nvidia.com}, 
\MYhref[black]{mailto:liuxiaofeng@cmu.edu}{liuxiaofeng@cmu.edu}}
}

\maketitle

\renewcommand{\thefootnote}{\Letter}
\footnotetext{Contact emails of corresponding authors.}
\renewcommand*{\thefootnote}{\arabic{footnote}}

\ificcvfinal\thispagestyle{empty}\fi

\begin{abstract}
Recent advances in domain adaptation show that deep self-training presents a powerful means for unsupervised domain adaptation. These methods often involve an iterative process of predicting on target domain and then taking the confident predictions as pseudo-labels for retraining. However, since pseudo-labels can be noisy, self-training can put overconfident label belief on wrong classes, leading to deviated solutions with propagated errors. To address the problem, we propose a confidence regularized self-training (CRST) framework, formulated as regularized self-training. Our method treats pseudo-labels as continuous latent variables jointly optimized via alternating optimization. We propose two types of confidence regularization: label regularization (LR) and model regularization (MR). CRST-LR generates soft pseudo-labels while CRST-MR encourages the smoothness on network output. Extensive experiments on image classification and semantic segmentation show that CRSTs outperform their non-regularized counterpart with state-of-the-art performance. The code and models of this work are available at \href{https://github.com/yzou2/CRST}{https://github.com/yzou2/CRST}.
\end{abstract}

\section{Introduction}\label{sec:intro}
Transferring knowledge learned by deep neural networks from label-rich domains to a new target domain is an important but challenging problem. Such domain change naturally occurs in many applications, such as synthetic data training~\cite{peng2018visda,richter2016playing} and simulation for robotics/autonomous driving. The existence of cross-domain differences often leads to considerably decreased model performance, and unsupervised domain adaptation (UDA) aims to address this problem by adapting source model to target domain with the aid of unlabeled target data. To this end, a predominant stream of adversarial learning based UDA methods were proposed to reduce the discrepancy between source and target domain features
\cite{chen2018domain,chen2017no,hoffman2018cycada,kim2019unsupervised,long2018conditional,murez2018image,pinheiro2018unsupervised,saito2018adversarial,sankaranarayanan2018generate,Tsai_adaptseg_2018}.
\begin{figure}[!t]
\includegraphics[width=\linewidth]{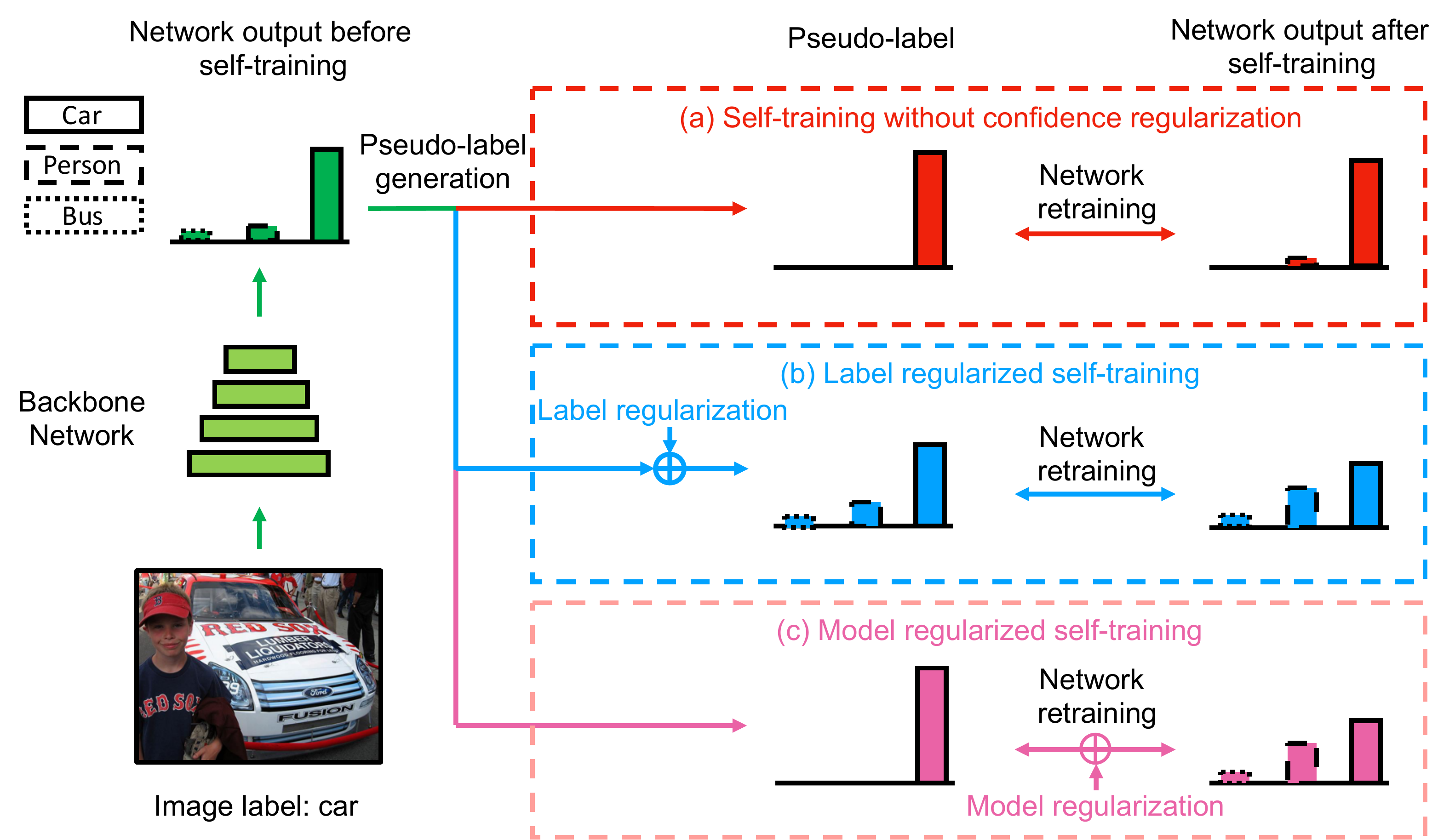}
\caption{Illustration of proposed confidence regularization. (a) Self-training without confidence regularization generates and retrains with hard pseudo-labels, resulting in sharp network output. (b) Label regularized self-training introduces soft pseudo-labels, therefore enabling outputs to be smooth. (c) Model regularized self-training also retrains with hard pseudo-labels, but incorporates a regularizer to directly promote output smoothness.}\label{fig:teaser}
\vspace{-2mm}
\end{figure}

More recently, self-training with networks emerged as a promising alternative towards domain adaptation \cite{busto2018open, Chen_2019_CVPR,inoue2018cross, lee2013pseudo, saito2017asymmetric, shu2018dirt, Zou_2018_ECCV}. Self-training iteratively generates a set of one-hot (or hard) pseudo-labels corresponding to large selection scores (i.e., prediction confidence) in target domain, and then retrains network based on these pseudo-labels with target data. Recently,~\cite{Zou_2018_ECCV} proposes class-balanced self-training (CBST) and formulates self-training as a unified loss minimization with pseudo-labels that can be solved in an end-to-end manner. Instead of reducing domain gap by minimizing both the task loss and domain adversarial loss, the self-training loss implicitly encourages cross-domain feature alignment for each class by learning from both labeled source data and pseudo-labeled target data.

Early work~\cite{lee2013pseudo} shows that the essence of deep self-training is entropy minimization - pushing network output to be as sharp as hard pseudo-label. However, $100\%$ accuracy cannot always be guaranteed for pseudo-labels. Trusting all selected pseudo-labels as ``ground truth'' by encoding them as hard labels can lead to overconfident mistakes and propagated errors. In addition, semantic labels of natural images can be highly ambiguous. Taking a sample image from VisDA17~\cite{peng2018visda} (see Fig. \ref{fig:teaser}) as an example: both person and car dominate significant portions of this image. Enforcing a model to be very confident on only one of the class during training can hurt the learning behavior~\cite{bagherinezhad2018label}, particularly within the under-determined context of UDA.

The above issues motivate us to prevent infinite entropy minimization in self-training via confidence regularization. A natural idea is to generate soft pseudo-label that redistributes a certain amount of confidence to other classes. Learning with soft pseudo-labels attenuates the misleading effect brought by incorrect or ambiguous supervision. Alternatively, to achieve the same goal, one can also encourage the smoothness of output probabilities and prevent overconfident prediction in network training. Both ideas are illustrated in Fig. \ref{fig:teaser}. At high-level, the major goal of CRST is still aligned with entropy minimization. However, the confidence regularization serves as a safety measure to prevent infinite entropy minimization and degraded performance.

In this work, we choose CBST \cite{Zou_2018_ECCV} as a state-of-the-art non-regularized self-training baseline, and propose a variety of specific confidence regularizers to comprehensively validate CRST. Our contributions are listed as follows:

\begin{itemize}[noitemsep,topsep=0pt]
\item In Section~\ref{sec:ccbst}, We generalize CBST to continuous CBST as a necessary preliminary for introducing our CRST, where we relax the feasible space of pseudo-labels from one-hot vectors to a probability simplex, 
\item In Section~\ref{subsec:lr}, we introduce label regularized self-training (CRST-LR). CRST-LR generates soft pseudo-labels for self-training. Specifically, we propose a label entropy regularizer (LRENT). In Section \ref{subsec:mr}, we introduce model regularized self-training (CRST-MR). CRST-MR introduces an output smoothing regularizer to network training. Specifically, we introduce three model regularizers, including $L_2$ (MRL2), entropy (MRENT), and KLD (MRKLD).
\item In Section~\ref{sec:theory}, we investigate theoretical properties of CRST, and prove that CRST is equivalent to regularized Classification Maximum Likelihood which can be solved via Classification Expectation Maximization (CEM). We also prove the convergence of CRST, and show that LRENT-regularized pseudo-label is equivalent to a generalized softmax with temperature~\cite{hinton2015distilling}.
\item In Section~\ref{sec:exp}, we comprehensively evaluate CRST on multiple domain adaptation tasks, including image classification (visDA17/Office-31) and semantic segmentation (GTA5/SYNTHIA $\rightarrow$ Cityscapes). We demonstrate state-of-the-art or competitive results from the proposed framework, and discuss the comparison between different regularizers in Section~\ref{sec:discuss}. We also show that LR+MR may benefit self-training.
\end{itemize}

\section{Related works}\label{sec:related}
\noindent\textbf{Self-training: } Self-training has been widely investigated in semi-supervised learning \cite{yarowsky1995unsupervised,amini2002semi,grandvalet2005semi}. An overview of different self-training techniques is presented in~\cite{triguero2015self}. Recent interests in self-training were revitalized with deep neural networks~\cite{lee2013pseudo}. A subtle difference between self-training on fixed features and deep self-training is that the latter involves the learning of embeddings which renders greater flexibility towards domain alignment than classifier-level adaptation. Within this context, \cite{Zou_2018_ECCV} proposed class-balanced self-training and achieved state-of-the-art performance in cross-domain semantic segmentation.

\noindent\textbf{Domain adaptation: } (Unsupervised) domain adaptation (UDA) has recently gained considerable interests. For UDA with deep networks, a major principle is to let the network learn domain invariant embeddings by minimizing the cross-domain difference of feature distributions with certain criteria. Examples of these methods include maximum mean discrepancy (MMD)~\cite{long2015learning,tzeng2014deep}, deep correlation alignment (CORAL)~\cite{sun2016deep}, sliced Wasserstein discrepancy \cite{lee2019sliced}, adversarial learning at input-level~\cite{hoffman2018cycada,gong2019dlow,dundar2020domain}, feature level~\cite{chen2019learning, ganin2015unsupervised,hoffman2018cycada,liu2019feature,saito2018adversarial,tzeng2017adversarial,wu2018dcan}, output space level~\cite{Tsai_adaptseg_2018}, and a variety of follow up works~\cite{chen2017no,long2018conditional,pinheiro2018unsupervised,sankaranarayanan2018generate} etc. Open set domain adaptation \cite{panareda2017open,saito2018open} focuses on the problem where classes are not totally shared between source and target domains. More recently, there have been multiple deep self-training/pseudo-label based methods that are proposed for domain adaptation~\cite{busto2018open,han2019unsupervised,inoue2018cross,saito2017asymmetric,shu2018dirt,Zou_2018_ECCV}.

\noindent\textbf{Semi-supervised learning (SSL): }There exist a natural strong connection between domain adaptation and semi-supervised learning with their problem definitions. A series of teacher-student based approaches have been recently proposed for both SSL \cite{laine2016temporal, tarvainen2017mean,luo2018smooth} and UDA problems\cite{french2018self}.

\noindent\textbf{Noisy label learning: }
Self-training can also be regarded as noisy label learning \cite{natarajan2013learning,reed2015training,sukhbaatar2014learning,yu2018seal} due to potential mistakes on pseudo-labels. \cite{reed2015training} introduced a bootstrapping method for noisy label learning. \cite{sukhbaatar2014learning} proposed an extra noise layer into the network adapting the network outputs to match the noisy label distribution.

\noindent\textbf{Network regularization: }
Regularization is a typical approach in supervised neural network training to avoid overfitting. Besides the standard weight decay, typical regularization techniques include label smoothing \cite{Goodfellow-et-al-2016,szegedy2016rethinking,liu2019wass}, network output regularization \cite{pereyra2017regularizing}, knowledge distillation \cite{hinton2015distilling}. Yet few principled research have considered regularized self-training within the context of SSL/UDA.

\section{Continuous class-balanced self-training}\label{sec:ccbst}
In this section, we review the class-balanced self-training algorithm in~\cite{Zou_2018_ECCV} and reformulate it under a continuous framework. Specifically, for an UDA problem, we have access to the labeled source samples $(\mathbf{x}_s,\mathbf{y}_s)$ from source domain $\{\mathbf{X}_S, \mathbf{Y}_S\}$, and target samples $\mathbf{x}_t$ from unlabeled target domain data $\mathbf{X}_T$. Any target label $\hat{\mathbf{y}}_t=(\hat{y}_t^{(1)},...,\hat{y}_t^{(K)})$ from $\hat{\mathbf{Y}}_T$ is unknown. $K$ is the total number of classes. We define the network weights as $\mathbf{w}$ and $p(k|\mathbf{x};\mathbf{w})$ as the classifier's softmax probability for class $k$.
 
CBST is a self-training framework that performs joint network learning and pseudo-label estimation under a unified loss minimization problem. The pseudo-labels are treated as discrete learnable latent variables being either one-hot or all-zero. Here, we first relax the pseudo-label variables to continuous domain, as shown in Eq. (\ref{cbst}):
\begin{equation}\label{cbst}
\begin{split}
\underset{\mathbf{w},\hat{\mathbf{Y}}_T}{\mathop{\min }}\,&\mathcal{L}_{CB}(\mathbf{w}, \hat{\mathbf{Y}}) = -\sum\limits_{s\in S}{\sum\limits_{k=1}^K{y_s^{(k)}}\log p(k|\mathbf{x}_s;\mathbf{w})}\\
&-\sum\limits_{t\in T}\sum\limits_{k=1}^{K}\hat{y}_{t}^{(k)}\log \frac{p(k|{\mathbf{x}_{t}};\mathbf{w})}{\lambda_k}\\
s.t.~&~\hat{\mathbf{y}}_t\in \Delta^{K-1}\cup \{\mathbf{0}\},~\forall t
\end{split}
\end{equation}
The feasible set is the union of $\{\mathbf{0}\}$ and a probability simplex $\Delta^{K-1}$. The continuous CBST is solved by alternating optimization based on the following \textbf{a)}, \textbf{b)} steps:\\
\noindent \textbf{a) Pseudo-label generation} \label{a)} ~ Fix $\mathbf{w}$ and solve:
\begin{equation}\label{cbst_a}
\begin{split}
\underset{\hat{\mathbf{Y}}_T}{\mathop{\min}}\,&-\sum\limits_{{{t}}\in {{T}}}{\sum\limits_{k=1}^{K}{\hat{y}_{t}^{(k)}}\log \frac{p(k|{\mathbf{x}_{t}};\mathbf{w})}{\lambda_k}}\\
s.t.~&~\hat{y}_{t}\in\Delta^{K-1}\cup \{\mathbf{0}\},~\forall t
\end{split}
\end{equation}
\textbf{b) Network retraining} \label{b)} ~ Fix $\hat{\mathbf{Y}}_T$ and solve:
\begin{equation}\label{cbst_b}
\begin{split}
\underset{\mathbf{w}}{\mathop{\min }}\,&-\sum\limits_{{{s}}\in {{S}}}{\sum\limits_{k=1}^{K}{y_{s}^{(k)}}\log p(k|{\mathbf{x}_{s}};\mathbf{w})}\\
&-\sum\limits_{{{t}}\in {{T}}}{\sum\limits_{k=1}^{K}{\hat{y}_{t}^{(k)}}\log p(k|{\mathbf{x}_{t}};\mathbf{w})} 
\end{split}
\end{equation}

We define going through step \textbf{a)} and \textbf{b)} once as one ``\textbf{self-training round}''. For solving step \textbf{a)}, there is a global optimizer for arbitrary $\hat{\mathbf{y}}_t=(\hat{y}_t^{(1)},...,\hat{y}_t^{(K)})$ as follows.
\begin{equation}\label{cbst_a_solver}
\hat{y}_{t}^{(k)*}=\left\{
\begin{aligned}
1, &~\text{if}~k=\argmax_{c}\{\frac {p(c|{\mathbf{x}_{t}};\mathbf{w})}{\lambda_c}\}\\
& ~~~~~ \text{and} ~~~ p(k|\mathbf{x}_t;\mathbf{w})>\lambda_k\\
0, &~\mathrm{otherwise}
\end{aligned}
\right.
\end{equation}

For solving step \textbf{b)}, one can use typical gradient-based methods such as mini-batch gradient descent. Intuitively, solving \textbf{a)} by (\ref{cbst_a_solver}) is actually conducting pseudo-label learning and selection simultaneously. Note that $\hat{\mathbf{y}}_{t}^{*}$ in (\ref{cbst_a_solver}) not only can be one-hot, but also can be a zero vector $\mathbf{0}$. For each target sample $(\mathbf{x}_t,\hat{\mathbf{y}}_{t}^{*})$, if $\hat{\mathbf{y}}_{t}^{*}$ is an one-hot, the sample is selected for model retraining. If $\hat{\mathbf{y}}_{t}^{*}=\mathbf{0}$, this sample is not chosen. Specifically, $\lambda_{k}$ is a parameter controlling sample selection. If a sample's predication is relatively confident with $p(k^*|\mathbf{x}_t;\mathbf{w}) > \lambda_{k^*}$, it is selected and labeled as class $k^* = \argmax_{k}\{\frac {p(k|{\mathbf{x}_{t}};\mathbf{w})}{\lambda_k}\}$. The less confident ones with $p(k^*|\mathbf{x}_t;\mathbf{w}) \leq \lambda_{k^*}$ are not selected.

$\lambda_{k}$ are critical parameters to control pseudo-label learning and selection. The same class-balanced $\lambda_{k}$ strategy introduced in \cite{Zou_2018_ECCV} is adopted for all self-training methods in this work. $\lambda_{k}$ for each class $k$ is determined by a single portion parameter $p$ which indicts how many samples we want to select in target domain. Specifically, we define the confidence for a sample as the max of its output softmax probabilities. For each class $k$, $\lambda_{k}$ is determined by the confidence value selecting the most confident $p$ portion of class $k$ predictions in the entire target set. We emphasize that only one parameter $p$ is used to determine all $\lambda_k$'s. Practically, we gradually increase $p$ to incorporate more pseudo-labels for each additional round. For detailed algorithm, we recommend to read Algorithm 2 in \cite{Zou_2018_ECCV}.

\noindent\textbf{Remark:} The only difference between CBST and continuous CBST lies in the feasible set where continuous CBST has a probability simplex while CBST has a set of one-hot vectors. Although the feasible set relaxization does not change the solutions of CBST and the pseudo-labels are still one-hot vectors, continuous CBST allows generating soft pseudo-labels if specific regularizers are introduced into pseudo-label generation. Thus it serves as the basis for our proposed label regularized self-training.
\section{Confidence regularized self-training}\label{sec:crst}
As mentioned in Section~\ref{sec:intro}, we leverage confidence regularization to prevent the over-minimization of entropy that could lead to degraded performance in self-training. Below, we introduce the general definition of CRST:
\begin{align}\label{crst}
\underset{\mathbf{w},{\hat{\mathbf{Y}}_T}}{\mathop{\min }}\,&\mathcal{L}_{CR}(\mathbf{w},\hat{\mathbf{Y}}_T) = \mathcal{L}_{CB}(\mathbf{w},\hat{\mathbf{Y}}_{T}) + \alpha\mathcal{R}_{C}(\mathbf{w},\hat{\mathbf{Y}}_{T})\notag\\
&=-\sum\limits_{s\in S}\sum\limits_{k=1}^K y_s^{(k)}\log p(k|{\mathbf{x}_{s}};\mathbf{w})\notag\\
&-\sum\limits_{t\in T}\Big[\sum\limits_{k=1}^{K}\hat{y}_{t}^{(k)}\log \frac{p(k|{\mathbf{x}_{t}};\mathbf{w})}{\lambda_k} - \alpha r_c(\mathbf{w},\hat{\mathbf{y}}_t)\Big]\notag\\
s.t.~&~\hat{y}_t\in~\Delta^{(K-1)}\cup\{\mathbf{0}\},~\forall t
\end{align}
${\mathcal{R}_{C}(\mathbf{w},{{{\hat{\mathbf{Y}}}}_{T}})}=\sum_{{{t}}\in {{T}}}r_c(\mathbf{w},\hat{\mathbf{y}}_t)$ is the confidence regularizer and $\alpha \ge 0$ is the weight coefficient. Similar to CBST, the optimization algorithm of CRST can be formulated as alternatively taking step \textbf{a)} pseudo-label generation and step \textbf{b)} network retraining. In this paper, we introduce two types of CRST frameworks: label regularized self-training and model regularized self-training.

\subsection{Label regularization}\label{subsec:lr}
The label regularizer has a general form of  $\mathcal{R}_{C}(\hat{\mathbf{Y}}_T)=\sum_{t\in T}r_c(\hat{\mathbf{y}}_t)$ and only depends on pseudo-labels $\{\hat{\mathbf{y}}_t\}$. With fixed $\mathbf{w}$, the pseudo-label generation in step \textbf{a)} of CRST-LR is defined as follows:
\begin{equation}\label{crst_lr_a}
\begin{split}
\underset{\hat{\mathbf{Y}}_T}{\mathop{\min}}\,&-\sum\limits_{t\in T}\Big[\sum\limits_{k=1}^{K}\hat{y}_{t}^{(k)}\log \frac{p(k|{\mathbf{x}_{t}};\mathbf{w})}{\lambda_k} - \alpha r_c(\hat{\mathbf{y}}_t)\Big]\\
s.t.~&~\hat{y}_t\in\Delta^{(K-1)}\cup \{\mathbf{0}\},~\forall t
\end{split}
\end{equation}

The global minimizer of (\ref{crst_lr_a}) can be found via a two-stage optimization given the special structure of the feasible space. The first stage involves minimizing (\ref{crst_lr_a}) within $\Delta^{(K-1)}$ only, which gives $\hat{\mathbf{y}}_t^\dag$. The second stage is to select between $\hat{\mathbf{y}}_t^\dag$ or $\mathbf{0}$ by checking which leads to a lower cost:
\begin{equation}
\hat{\mathbf{y}}_t^*=\left\{
\begin{aligned}
\hat{\mathbf{y}}_t^\dag, &~~\text{if}~~\mathcal{C}(\hat{\mathbf{y}}_t^\dag)<\mathcal{C}(\mathbf{0})\\
\mathbf{0}~~, &~~\text{otherwise}
\end{aligned}
\right.
\end{equation}
where $\mathcal{C}(\hat{\mathbf{y}}_t)$ is the cost of a single sample $t$ in (\ref{crst_lr_a}):
\begin{equation}\label{plcro_0}
\mathcal{C}(\hat{\mathbf{y}}_t)= -\hat{y}_{t}^{(k)} \sum\limits_{k=1}^{K}\log \frac{p(k|{\mathbf{x}_{t}};\mathbf{w})}{\lambda_k} + \alpha r_c(\hat{\mathbf{y}}_t)
\end{equation}
Note that the above regularized term prefers selecting pseudo-labels with certain smoothness rather than sparse ones. In addition, CRST-LR and CBST share the same network retraining strategy in step \textbf{b)}.

Specifically, we introduce a negative entropy label regularizer (LRENT) in Table \ref{table:regs} with its definition and the corresponding solution of $\hat{\mathbf{y}}_t^\dag$. For clarity, we write ${p(k|{\mathbf{x}_{t}};\mathbf{w})}$ as ${p(k|{\mathbf{x}_{t}})}$ for short. $\hat{\mathbf{y}}_t^\dag$ can be obtained via solving with a Lagrangian multiplier (KKT conditions)~\cite{boyd2004convex}. The detailed derivations are shown in Section \ref{sec:derivlrent} of the Appendix.

\subsection{Model regularization}\label{subsec:mr}
The model regularizer has a general form of  ${\mathcal{R}_{C}({\mathbf{w}})}=\sum_{{{t}}\in {{T}}}r_c(p(\mathbf{x}_t;\mathbf{w}))$ where $p(\mathbf{x}_t;\mathbf{w})$ is the network softmax output probabilites. Compared to CBST, CRST-MR has the same hard pseudo-label generation process. But in network retraining of step \textbf{b)}, CRST-MR uses a cross-entropy loss regularized by an output smoothness encouraging term. We define the optimization problem in step \textbf{b)} as follows:
\begin{equation}
\begin{split}\label{pcro}
\underset{\mathbf{w}}{\mathop{\min }}\,
& -\sum\limits_{{{s}}\in {{S}}}{\sum\limits_{k=1}^{K}{y_{s}^{(k)}}\log p(k|{\mathbf{x}_{s}};\mathbf{w})}\\
& -\sum\limits_{{{t}}\in {{T}}}[\sum\limits_{k=1}^{K}{\hat{y}_{t}^{(k)}}\log p(k|{\mathbf{x}_{t}};\mathbf{w})-\alpha r_c(p(\mathbf{x}_t;\mathbf{w}))]
\end{split}
\end{equation}

Specifically, we introduce three model regularizers in Table \ref{table:regs} based on $L_2$, negative entropy and KLD between uniform distribution $\mathbf{u}$ and softmax output. The gradients w.r.t. softmax logits $z_i$ are also provided. $H(\mathbf{p})$ is the entropy.
\begin{table}[!t]
\centering
\resizebox{0.48\textwidth}{!}{
\centering
\begin{tabular}{c|c|c}
\hline
& Regularizer & Pseudo-label solution (LR)/Gradient (MR)\\
\hline
LRENT & ${\sum\limits_{k=1}^{K}{\hat{y}_t^{(k)}\log~ (\hat{y}_t^{(k)}})}$  & $\begin{aligned}
\hat{y}_{t}^{(i)\dag}=\frac{{({\frac{p(i|{\mathbf{x}_{t}})}{\lambda_k}})^{\frac{1}{\alpha }}}}{\sum\limits_{k=1}^{K}{{({\frac{p(k|{\mathbf{x}_{t}})}{\lambda_k}})^{\frac{1}{\alpha }}}}}
\end{aligned} $                    \\ \hline \hline
MRL2     & 
$\sum\limits_{k=1}^{K}p(k|\mathbf{x}_t)^2 $ 
& ${2\sum\limits_{k=1}^{K}{p^{2}(k|{\mathbf{x}_{t}})[{{\delta }_{ki}}-{{p}}(i|{\mathbf{x}_{t}})]},} 
   {~~\delta_{ki}=\mathds{1}[k=i]} $                \\ \hline
 MRENT & ${\sum\limits_{k=1}^{K}{p(k|\mathbf{x}_t)\log p(k|\mathbf{x}_t)}}$ & $p(i|{\mathbf{x}_{t}}) [\log p(i|{\mathbf{x}_{t}})+H(p({\mathbf{x}_{t}})) ]$  \\ \hline
MRKLD    & ${-\sum\limits_{k=1}^{K}{\frac{1}{K}\log p(k|\mathbf{x}_t)}}$           & $p(i|\mathbf{x}_t)-\frac{1}{K} $  \\ \hline
\end{tabular}
}
\caption{List of proposed regularizers with corresponding pseudo-label solution or gradients w.r.t. softmax logit $z_i$.}
\label{table:regs}
\end{table}

\section{Theoretical properties}\label{sec:theory}
\subsection{A probabilistic view of CRST}\label{prob_explan}
There exists an inherent connection between the CRST and some probabilistic models. Specifically, the CRST self-training algorithm can be interpreted as an instance of classification expectation maximization~\cite{amini2002semi}:
\begin{proposition}\label{prop:cem}
	CRST can be modeled as a regularized classification maximum likelihood (RCML) problem optimized via classification expectation maximization.
\end{proposition}
\begin{proof}
	Please refer to Section~\ref{sec:proofprop1} of the Appendix.
\end{proof}
\begin{proposition}\label{prop:convergence}
	Given pre-determined $\lambda_k$, CRST is convergent under certain conditions.
\end{proposition}
\begin{proof}
	Please refer to Section~\ref{sec:proofprop2} of the Appendix.
\end{proof}

\subsection{Soft pseudo-label in LRENT}
There is an intrinsic connection between the soft pseudo-label of LRENT (given in Table \ref{table:regs}) and softmax with temperature. Softmax with temperature \cite{hinton2015distilling} is a common approach in neural network for scaling softmax probabilities with applications in knowledge distillation \cite{hinton2015distilling}, model calibration \cite{pmlr-v70-guo17a}, etc. Typically, networks produce categorical probabilities by a softmax activation layer to convert
the logit $z_i$ for each class into a probability $p(i)$. And the softmax with temperature introduces a positive temperature $\alpha$ to scale its smoothness as follows.
\begin{align}
p(i)=\frac{{{e}^{\frac{{{z}_{i}}}{\alpha }}}}{\sum_{k=1,...,K}{{{e}^{\frac{{{z}_{k}}}{\alpha }}}}}
\end{align}

For high temperature ($\alpha \rightarrow \infty$), the new distribution is softened as a uniform distribution that has the highest entropy and uncertainty. For temperature $\alpha = 1$, we recover the original softmax probabilities. For low temperature ($\alpha \rightarrow 0$), the distribution collapses to a sparse one-hot vector with all probability on the class with the most original softmax probability. Now we draw the connection of soft pseudo-label in LRENT to softmax with temperature:
\begin{proposition}\label{prop:prop3}
	If $\lambda_k$ are equal for all $k$, the soft pseudo-label of LRENT given in Table \ref{table:regs} is exactly the same as softmax with temperature.
\end{proposition}
\begin{proof}
	\begin{displaymath}
		\label{softmaxtemp}
		\begin{split}
		\hat{y}_{t}^{(i)*}&=\frac{{({\frac{p(i|{{x}_{t}})}{\lambda_k}})^{\frac{1}{\alpha }}}}{\sum_{k}{{({\frac{p(k|{{x}_{t}})}{\lambda_k}})^{\frac{1}{\alpha }}}}}=\frac{p{{(k|{{x}_{t}})}^{\frac{1}{\alpha }}}}{\sum_{k}{p{{(k|{{x}_{t}})}^{\frac{1}{\alpha }}}}}=\frac{{{[\frac{{{e}^{{{z}_{i}}}}}{\sum_{q}{{{e}^{{{z}_{q}}}}}}]}^{\frac{1}{\alpha }}}}{\sum_{k}{{{[\frac{{{e}^{{{z}_{k}}}}}{\sum_{q}{{{e}^{{{z}_{q}}}}}}]}^{\frac{1}{\alpha }}}}}\\
		&=\frac{{{({{e}^{{{z}_{i}}}})}^{\frac{1}{\alpha }}}}{\sum_{k}{{{({{e}^{{{z}_{k}}}})}^{\frac{1}{\alpha }}}}}=\frac{{{e}^{\frac{{{z}_{i}}}{\alpha }}}}{\sum_{k}{{{e}^{\frac{{{z}_{k}}}{\alpha }}}}}
		\end{split}
	\end{displaymath}
\end{proof}
The soft pseudo-label of LRENT can be regarded as a generalized softmax with temperature. In self-training, if selected properly, $\lambda_k$ can help to generate class-balanced soft pseudo-labels.

\begin{proposition}\label{prop:prop4}
	\textit{KLD model confidence regulared self-training is equivalent to self-training with pseudo-label uniformly smoothed by $\epsilon=(K\alpha-\alpha)/(K+K\alpha)$, where $\alpha$ is the regularizer weight.}
\end{proposition}
\begin{proof}
	Please refer to Section~\ref{sec:proofprop4} of the Appendix.
\end{proof}

\begin{proposition}\label{prop:prop5}
$D_{KL}(p(\mathbf{x}_t)||\mathbf{u})$ KLD model regularizer (the reverse of the proposed $D_{KL}(\mathbf{u}||p(\mathbf{x}_t))$ KLD regularizer) is equivalent to entropy model regularizer $-H(p(\mathbf{x}_t))$, where $\mathbf{u}$ is the uniform distribution.
\end{proposition}
\begin{proof}
	Please refer to Section~\ref{sec:proofprop5} of the Appendix.
\end{proof}

\section{Experiments}\label{sec:exp}
In this section, we conduct comprehensive evaluation on different domain adaptation tasks.\\
\noindent\textbf{Adaptation for image classification:} We consider two adaptation benchmarks: 1) VisDA17 \cite{peng2018visda} and 2) Office-31 \cite{saenko2010adapting}. VisDA17 contains $152,409$ 2D synthetic images of 12 classes in the source training set and $55,400$ real images from MS-COCO~\cite{lin2014microsoft} as the target domain validation set. Office-31 is a small-scale dataset containing images of $31$ classes from three domains - Amazon (A), Webcam (W) and DSLR (D). Each domain contains $2,817$, $795$ and $498$ images respectively. We follow the standard protocol in~\cite{saenko2010adapting,sankaranarayanan2018generate} and evaluate on six transfer tasks $A \rightarrow W$, $D \rightarrow W$, $W \rightarrow D$, $A \rightarrow D$, $D \rightarrow A$, and $W \rightarrow A$.\\
\noindent\textbf{Adaptation for semantic segmentation:} We consider two popular synthetic-to-real adaptation scenarios: 1) GTA5 \cite{richter2016playing} to Cityscapes \cite{cordts2016cityscapes}, and
2) SYNTHIA \cite{ros2016synthia} to Cityscapes. The GTA5 dataset includes $24,966$ images rendered by GTA5 game engine. For SYNTHIA, we choose
SYNTHIA-RAND-CITYSCAPES which includes $9,400$ labeled images. Following the standard protocols~\cite{hoffman2018cycada,Tsai_adaptseg_2018}, we adapt the model to the Cityscapes training set and evaluate the performance on the validation set.

To comprehensively demonstrate the improvement of CRST, we report the performance of CRST with all regularizers and compare with CBST in each task.

\subsection{Implementation details}
\noindent\textbf{Image classification: } For VisDA17/Office-31, we implement CBST/CRSTs using PyTorch~\cite{paszke2017automatic} and choose ResNet-101/ResNet-50~\cite{he2016deep} as backbones. For fair comparison, we compare to other works with the same backbone networks. Both backbones are pre-trained on ImageNet~\cite{deng2009imagenet}, and then fine-tuned on source domain using SGD, with learning rate $1\times10^{-3}$, weight decay $5\times10^{-4}$, momentum $0.9$ and batch size $32$. For self-training, we apply the same training strategy but a different learning rate $1\times10^{-4}$.\\
\noindent\textbf{Semantic segmentation: } For semantic segmentation, we further consider DeepLabv2~\cite{chen2018deeplab} as a backbone besides the ResNet-38 backbone in~\cite{Zou_2018_ECCV}. For experiments with DeepLabv2, we implement CBST/CRSTs using PyTorch, while following the MXNet~\cite{chen2015mxnet} implementation of~\cite{Zou_2018_ECCV} for experiments with ResNet-38. DeepLabv2 is pre-trained on ImageNet and fine-tuned on source domain using SGD, with learning rate $2.5\times10^{-4}$, weight decay $5\times10^{-4}$, momentum $0.9$, batch size $2$, patch size $512\times1024$, multi-scale training augmentation ($0.5 - 1.5$) and horizontal flipping. In self-training, we apply SGD with learning rate of $5\times10^{-5}$. For fair comparison, we unify the total number of self-training rounds to be 3, each with 2 re-training epochs.

\subsection{Domain adaptation for image classification}
\noindent \textbf{VisDA17:}
We present the results on VisDA17 in Table \ref{table:visda17} in terms of per-class accuracy and mean accuracy. For each proposed approach, we report the averages and standard deviations of the evaluation results over $5$ runs. Note that both MRKLD and LRENT outperform the non-regularized CBST, whereas MRL2 and MRENT show slightly worse results. Among CRSTs with single regularizer, MRKLD achieves the best performance with considerable improvement. The combination of MRKLD and LRENT further outperforms single regularizers and other recently proposed methods. The result even outperforms certain methods with stronger backbones (ResNet-152)~\cite{pinheiro2018unsupervised,sankaranarayanan2018generate}.\\
\noindent \textbf{Office-31:}
We compare the performance of different methods on Office-31 with the same backbone ResNet-50 in Table \ref{table:office}. All CRSTs achieve similar results that outperform the baseline CBST. In addition, MRKLD+LRENT again outperforms single regularizers, achieving comparable or better performance compared with other recent methods.

\begin{table*}[!t]
	\captionsetup{font=small}
	\centering
	\resizebox{\linewidth}{!}{
	\centering
	\begin{tabular}{c|cccccccccccc|c}
		\hline
		Method & Aero & Bike & Bus & Car & Horse & Knife & Motor & Person & Plant & Skateboard & Train & Truck & Mean\\
		\hline
		Source \cite{saito2018adversarial} & 55.1 & 53.3 & 61.9 & 59.1 & 80.6 & 17.9 & 79.7 & 31.2 & 81.0 & 26.5 & 73.5 & 8.5 & 52.4\\
		MMD \cite{long2015learning} & 87.1 & 63.0 & 76.5 & 42.0 & 90.3 & 42.9 & 85.9 & 53.1 & 49.7 & 36.3 & 85.8 & 20.7 & 61.1\\
		DANN \cite{ganin2016domain} & 81.9 & 77.7 & 82.8 & 44.3 & 81.2 & 29.5 & 65.1 & 28.6 & 51.9 & 54.6 & 82.8 & 7.8 & 57.4\\ 
		ENT \cite{grandvalet2005semi} & 80.3 & 75.5 & 75.8 & 48.3 & 77.9 & 27.3 & 69.7 & 40.2 & 46.5 & 46.6 & 79.3 & 16.0 & 57.0\\
		MCD \cite{saito2017maximum} & 87.0 & 60.9 & \textbf{83.7} & 64.0 & 88.9 & 79.6 & 84.7 & \textbf{76.9} & \textbf{88.6} & 40.3 & 83.0 & 25.8 & 71.9\\
		ADR \cite{saito2018adversarial} & 87.8 & 79.5 & \textbf{83.7} & 65.3 & \textbf{92.3} & 61.8 & \textbf{88.9} & 73.2 & 87.8 & 60.0 & \textbf{85.5} & {32.3} & 74.8\\  
		SimNet-Res152 \cite{pinheiro2018unsupervised} & \textbf{94.3} & 82.3 & 73.5 & 47.2 & 87.9 & 49.2 & 75.1 & 79.7 & 85.3 & 68.5 & 81.1 & 50.3 & 72.9\\
		GTA-Res152 \cite{sankaranarayanan2018generate} & - & - & - & - & - & - & - & - & - & - & - & - & 77.1\\
		\hline
		Source-Res101 & 68.7 & 36.7 & 61.3 & \textbf{70.4} & 67.9 & 5.9 & 82.6 & 25.5 & 75.6 & 29.4 & 83.8 &  10.9 & 51.6\\
		CBST & 87.2$\pm$2.4 & 78.8$\pm$1.0 & 56.5$\pm$2.2 & 55.4$\pm$3.6 & 85.1$\pm$1.4 & 79.2$\pm$10.3 & 83.8$\pm$0.4 &  77.7$\pm$4.0 & 82.8$\pm$2.8 & 88.8$\pm$3.2 & 69.0$\pm$2.9 & 72.0$\pm$3.8 & 76.4$\pm$0.9 \\
		MRL2 & 87.0$\pm$2.9 & 79.5$\pm$1.9 & 57.1$\pm$3.2 & 54.7$\pm$2.9 & 85.5$\pm$1.1 & 78.1$\pm$11.7 & 83.0$\pm$1.5 &  77.7$\pm$3.7 & 82.4$\pm$1.7 & 88.6$\pm$2.7 & 69.1$\pm$2.2 &  71.8$\pm$3.0 & 76.2$\pm$1.0        \\
		MRENT & 87.1$\pm$2.7 & 78.3$\pm$0.7 & 56.1$\pm$4.0 & 54.4$\pm$2.7 & 84.4$\pm$2.3 & 79.9$\pm$10.6 & 83.7$\pm$1.1 &  77.9$\pm$4.4 & 82.7$\pm$2.4 &  87.4$\pm$2.8 & 70.0$\pm$1.4 &  72.8$\pm$3.3 & 76.2$\pm$0.8 \\
		MRKLD & 87.3$\pm$2.5 & 79.4$\pm$1.9 & 60.5$\pm$2.4 & 59.7$\pm$2.5 & 87.6$\pm$1.4 & \textbf{82.4$\pm$4.4} & 86.5$\pm$1.1 & 78.4$\pm$2.6 & 84.6$\pm$1.7 & 86.4$\pm$2.8 & 72.5$\pm$2.4 &   69.8$\pm$2.5 & 77.9$\pm$0.5\\
		LRENT & 87.7$\pm$2.4 & 78.7$\pm$0.8 & 57.3$\pm$3.3 & 54.5$\pm$4.0 & 84.8$\pm$1.7 & 79.7$\pm$10.3 & 84.2$\pm$1.4 &  77.4$\pm$3.7 & 83.1$\pm$1.5 & \textbf{88.3$\pm$2.6} & 70.9$\pm$2.1 &   \textbf{72.6$\pm$2.4} & 76.6$\pm$0.9\\
		MRKLD+LRENT & 88.0$\pm$0.6 & 79.2$\pm$2.2 & 61.0$\pm$3.1 & 60.0$\pm$1.0 & 87.5$\pm$1.2 & 81.4$\pm$5.6 & 86.3$\pm$1.5 & 78.8$\pm$2.1 & 85.6$\pm$0.9 & 86.6$\pm$2.5 & 73.9$\pm$1.3 &   68.8$\pm$2.3 & \textbf{78.1$\pm$0.2}\\
		\hline
	\end{tabular}
	}
	\vspace{-2mm}
	\caption{Experimental results on VisDA17.}
	\label{table:visda17}
	\vspace{-2mm}
\end{table*}

\begin{table}[]
	\centering
	\resizebox{0.95\linewidth}{!}{
	\begin{tabular}{c|cccccc|c}
		\hline
		Method & A$\rightarrow$W & D$\rightarrow$W & W$\rightarrow$D & A$\rightarrow$D & D$\rightarrow$A & W$\rightarrow$A & Mean\\
		\hline
		ResNet-50 \cite{he2016deep} & 68.4$\pm$0.2 & 96.7$\pm$0.1 & 99.3$\pm$0.1 & 68.9$\pm$0.2 & 62.5$\pm$0.3 & 60.7$\pm$0.3 & 76.1\\
		DAN \cite{long2015learning} & 80.5$\pm$0.4 & 97.1$\pm$0.2 & 99.6$\pm$0.1 & 78.6$\pm$0.2 & 63.6$\pm$0.3 & 62.8$\pm$0.2 & 80.4\\
		RTN \cite{long2016unsupervised} & 84.5$\pm$0.2 & 96.8$\pm$0.1 & 99.4$\pm$0.1 & 77.5$\pm$0.3 & 66.2$\pm$0.2 & 64.8$\pm$0.3 & 81.6\\
		DANN \cite{ganin2016domain} & 82.0$\pm$0.4 & 96.9$\pm$0.2 & 99.1$\pm$0.1 & 79.7$\pm$0.4 & 68.2$\pm$0.4 & 67.4$\pm$0.5 & 82.2\\
		ADDA \cite{tzeng2017adversarial} & 86.2$\pm$0.5 & 96.2$\pm$0.3 & 98.4$\pm$0.3 & 77.8$\pm$0.3 & 69.5$\pm$0.4 & 68.9$\pm$0.5 & 82.9\\
		JAN \cite{long2017deep} & 85.4$\pm$0.3 & 97.4$\pm$0.2 & 99.8$\pm$0.2 & 84.7$\pm$0.3 & 68.6$\pm$0.3 & 70.0$\pm$0.4 & 84.3\\
		GTA \cite{sankaranarayanan2018generate} & \textbf{89.5$\pm$0.5} & 97.9$\pm$0.3 & 99.8$\pm$0.4 & 87.7$\pm$0.5 & 72.8$\pm$0.3 & 71.4$\pm$0.4 & 86.5\\
		\hline
		CBST & 87.8$\pm$0.8 & 98.5$\pm$0.1 & \textbf{100$\pm$0.0} & 86.5$\pm$1.0 & 71.2$\pm$0.4 & 70.9$\pm$0.7 & 85.8\\
		MRL2 & 88.4$\pm$0.2 & 98.6$\pm$0.1 & \textbf{100$\pm$0.0} & 87.7$\pm$0.9 & 71.8$\pm$0.2 & \textbf{72.1$\pm$0.2} & 86.4\\
		MRENT & 88.0$\pm$0.4 & 98.6$\pm$0.1 & \textbf{100$\pm$0.0} & 87.4$\pm$0.8 & \textbf{72.7$\pm$0.2} & 71.0$\pm$0.4 & 86.4\\
		MRKLD & 88.4$\pm$0.9 & 98.7$\pm$0.1 & \textbf{100$\pm$0.0} & 88.0$\pm$0.9 & 71.7$\pm$0.8 & 70.9$\pm$0.4 & 86.3\\
		LRENT & 88.6$\pm$0.4 & 98.7$\pm$0.1 & \textbf{100$\pm$0.0} & \textbf{89.0$\pm$0.8} & 72.0$\pm$0.6 & 71.0$\pm$0.3 & 86.6\\
		MRKLD+LRENT & 89.4$\pm$0.7 & \textbf{98.9$\pm$0.4} & \textbf{100$\pm$0.0} & 88.7$\pm$0.8 & 72.6$\pm$0.7 & 70.9$\pm$0.5 & \textbf{86.8}\\
		\hline
	\end{tabular}
	}
	\vspace{-2mm}
	\caption{Experimental results on Office-31.}
	\label{table:office}
	\vspace{-2mm}
\end{table}

\subsection{Domain adaptation for semantic segmentation}
\noindent\textbf{GTA5 $\rightarrow$ Cityscapes:} Table~\ref{table:gtacity} shows the adaptation performance of CRSTs and other comparing methods. On a DeepLabv2 backbone, one could see that MRKLD achieves the best result outperforming previous state-of-the-art. In addition, Fig. \ref{fig:gta2city} visualizes the adapted prediction results obtained by CBST and CRSTs on Cityscapes validation set. Fig. \ref{fig:plgta2city} further compares the pseudo-label maps in the second round of self-training. On a wide ResNet-38 backbone, all CRSTs outperform the baseline CBST and we achieve the state-of-the-art system-level performance with the spatial priors (SP) and multi-scale testing (MST) from~\cite{Zou_2018_ECCV}.

\noindent\textbf{SYNTHIA $\rightarrow$ Cityscapes:} Table \ref{table:syncity} shows the adaptation results where CRSTs again show the performance on par with or better than the baseline CBST. In particular, MRKLD maintains the best performance among all regularizers and outperforms the previous state-of-the-art~\cite{Zou_2018_ECCV}.

\subsection{Parameter analysis}
$p$ is an important parameter controling the pseudo-label generation and selection sensitivity. We adopt the same $p$ policy as~\cite{Zou_2018_ECCV} where we start $p$ from $20\%$, and empirically add $5\%$ to $p$ in each additional self-training round. We conduct a sensitivity analysis for portion $p$ similar to~\cite{Zou_2018_ECCV}, where we consider the starting portion $p_0$ and the incremental portion $\Delta p$ on a difficult task of Office-31: W $\rightarrow$ A. Table \ref{table:p} shows that CRSTs are not sensitive to $p_0$ and  $\Delta p$. 

In CRST, the coefficient $\alpha$ is an important parameter that balances the weight between self-training loss and confidence regularizer. In all the experiments, we unify $\alpha$ to be $0.025,0.1,0.1,0.25$ for MRL2, MRENT, MRKLD and LRENT, respectively. Note that various regularizers have different $\alpha$ due to their intrinsic differences. We also present the sensitivity analysis of $\alpha$ on W $\rightarrow$ A in Table \ref{table:alpha}. We can see all CRSTs are not sensitive to $\alpha$ in certain intervals. 
\section{Discussion}\label{sec:discuss}
\subsection{How does confidence regularization work?}
\begin{table*}[!t]
	\centering
	\resizebox{\linewidth}{!}{
	\begin{tabular}{c|c|ccccccccccccccccccc|c}
		\hline
		Method & Backbone & Road & SW & Build & Wall & Fence & Pole & TL & TS & Veg. & Terrain & Sky & PR & Rider & Car & Truck & Bus & Train & Motor & Bike & mIoU\\
		\hline
		Source & \multirow{2}{0.1\linewidth}{\centering{DRN-26}} & 42.7 & 26.3 & 51.7 & 5.5 & 6.8 & 13.8 & 23.6 & 6.9  & 75.5 & 11.5 & 36.8 & 49.3 & 0.9 & 46.7 & 3.4 & 5.0 & 0.0 & 5.0 & 1.4  & 21.7\\
		CyCADA~\cite{hoffman2018cycada} & & 79.1 & 33.1 & 77.9 & 23.4 & 17.3 & 32.1 & 33.3 & 31.8 & 81.5 & 26.7 & 69.0 & 62.8 & 14.7 & 74.5 & 20.9 & 25.6 & 6.9 & 18.8 & 20.4 & 39.5\\
		\hline
		Source & \multirow{2}{0.1\linewidth}{\centering{DRN-105}} & 36.4 & 14.2 & 67.4 & 16.4 & 12.0 & 20.1 & 8.7 & 0.7 & 69.8 & 13.3 & 56.9 & 37.0 & 0.4 & 53.6 & 10.6 & 3.2 & 0.2 & 0.9 & 0.0 & 22.2\\
		MCD~\cite{saito2017maximum} & & 90.3 & 31.0 & 78.5 & 19.7 & 17.3 & 28.6 & 30.9 & 16.1 & 83.7 & 30.0 & 69.1 & 58.5 & 19.6 & 81.5 & 23.8 & 30.0 & 5.7 & 25.7 & 14.3 & 39.7\\
		\hline
		Source & \multirow{2}{0.1\linewidth}{\centering{DeepLabv2}} & 75.8 & 16.8 & 77.2 & 12.5 & 21.0 & 25.5 & 30.1 & 20.1 & 81.3 & 24.6 & 70.3 & 53.8 & 26.4 & 49.9 & 17.2 & 25.9 & 6.5 & 25.3 & 36.0 & 36.6\\
		AdaptSegNet~\cite{Tsai_adaptseg_2018} & & 86.5 & 36.0 & 79.9& 23.4 & 23.3 & 23.9 & 35.2 & 14.8 & 83.4 & 33.3 & 75.6 & 58.5 & 27.6 & 73.7 & 32.5 & 35.4 & 3.9 & 30.1 & 28.1 & 42.4\\ \hline
		AdvEnt~\cite{vu2019advent} & DeepLabv2 & 89.4 & 33.1 & \textbf{81.0} & 26.6 & 26.8 & 27.2 & 33.5 & 24.7 & 83.9 & \textbf{36.7} & 78.8 & 58.7 & 30.5 & 84.8 & 38.5 & 44.5 & 1.7 & 31.6 & 32.4 & 45.5 \\
		\hline
		Source & \multirow{2}{0.1\linewidth}{\centering{DeepLabv2}} &  - & - & - & - & - & - & - & - & -& - & - & - & - & - & - & - & - & - & - & 29.2\\
		FCAN~\cite{zhang2018fully} &  & - & - & - & - & - & - & - & - & -& - & - & - & - & - & - & - & - & - & - & 46.6 \\
		\hline
		Source & \multirow{6}{0.1\linewidth}{\centering{DeepLabv2}} & 71.3 & 19.2 & 69.1 & 18.4 & 10.0 & 35.7 & 27.3 &  6.8 & 79.6 & 24.8 & 72.1 & 57.6 & 19.5 & 55.5 & 15.5 & 15.1 & 11.7 & 21.1 & 12.0 & 33.8\\
		CBST & & 91.8 & 53.5 & 80.5 & 32.7 & 21.0 & 34.0 & 28.9 & 20.4 & 83.9 & 34.2 & 80.9 & 53.1 & 24.0 & 82.7 & 30.3 & 35.9 & 16.0 & 25.9 & 42.8 & 45.9\\
		MRL2 & & \textbf{91.9} & 55.2 & 80.9 & 32.1 & 21.5 & 36.7 & 30.0 & 19.0 & 84.8 & 34.9 & 80.1 & 56.1 & 23.8 & 83.9 & 28.0 & 29.4 & 20.5 & 24.0 & 40.3 & 46.0\\
		MRENT & & 91.8 & 53.4 & 80.6 & 32.6 & 20.8 & 34.3 & 29.7 & 21.0 & 84.0 & 34.1 & 80.6 & 53.9 & 24.6 & 82.8 & 30.8 & 34.9 & 16.6 & 26.4 & 42.6 & 46.1\\
		MRKLD & & 91.0 & \textbf{55.4} & 80.0 & 33.7 & 21.4 & 37.3 & 32.9 & 24.5 & \textbf{85.0} & 34.1 & 80.8 & 57.7 & 24.6 & 84.1 & 27.8 & 30.1 & 26.9 & 26.0 & 42.3 & 47.1\\
		LRENT & & 91.8 & 53.5 & 80.5 & 32.7 & 21.0 & 34.0 & 29.0 & 20.3 & 83.9 & 34.2 & \textbf{80.9} & 53.1 & 23.9 & 82.7 & 30.2 & 35.6 & 16.3 & 25.9 & 42.8 & 45.9\\
		\hline
		Source & \multirow{6}{0.1\linewidth}{\centering{ResNet-38}} & 70.0 & 23.7 & 67.8 & 15.4 & 18.1 & 40.2 & 41.9 & 25.3 & 78.8 & 11.7 & 31.4 & 62.9 & 29.8 & 60.1 & 21.5 & 26.8 & 7.7 & 28.1 & 12.0 & 35.4\\
		CBST~\cite{Zou_2018_ECCV} & & 86.8  & 46.7 & 76.9 & 26.3 & 24.8  & 42.0 & 46.0 & 38.6 & 80.7 & 15.7 & 48.0 & 57.3 & 27.9 & 78.2 & 24.5 & 49.6 & 17.7 & 25.5 & 45.1 & 45.2\\
		MRL2 & & 84.4 & 52.7 & 74.7 & 38.0 & \textbf{32.2} & 43.7 & \textbf{53.7} & 38.6 & 73.9 & 24.4 & 64.4 & 45.6 & 24.6 & 63.2 & 3.22 & 31.9 & \textbf{45.9} & 44.2 & 34.8 & 46.0\\
		MRENT & & 84.6 & 49.5 & 73.9 & \textbf{35.8} & 25.1 & \textbf{46.2} & 53.3 & 43.3 & 75.2 & 24.2 & 63.8 & 48.2 & \textbf{33.8} & 65.7 & 2.89 & 32.6 & 39.2 & \textbf{50.0} & 34.7 & 46.4\\
		MRKLD & & 84.5 & 47.7 & 74.1 & 27.9 & 22.1 & 43.8 & 46.5 & 37.8 & 83.7 & 22.7 & 56.1 & 56.8 & 26.8  & 81.7 & 22.5 & 46.2 & 27.5 & 32.3 & \textbf{47.9} & 46.8\\
		LRENT & & 80.3 & 40.8 & 65.8 & 24.6 & 30.5 & 43.1 & 49.5 & 40.3 & 82.1 & 26.0 & 54.6 & 59.4 & 32.1 & 68.0 & 31.9 & 30.0 & 21.9 & 44.8 & 46.7 & 45.9\\
		\hline
		CBST-SP & \multirow{3}{0.1\linewidth}{\centering{ResNet-38}} & 85.6 & 55.1 & 76.9 & 26.8 & 23.4 & 38.9 & 47.1 & \textbf{46.9} & 83.4 & 25.5 & 68.7 & 45.6 & 15.7 & 79.7 & 27.7 & 50.3 & 38.2 & 33.4 & 44.6 & 48.1\\
		MRKLD-SP & & 90.8 & 46.0 & 79.9 & 27.4 & 23.3 & 42.3 & 46.2 & 40.9 & 83.5 & 19.2 & 59.1 & 63.5 & 30.8 & 83.5 & 36.8 & \textbf{52.0} & 28.0 & 36.8 & 46.4 & 49.2\\
		MRKLD-SP-MST & & 91.7 & 45.1 & 80.9 & 29.0 & 23.4 & 43.8 & 47.1 & 40.9 & 84.0 & 20.0 & 60.6 & \textbf{64.0} & 31.9 & \textbf{85.8} & \textbf{39.5} & 48.7 & 25.0 & 38.0 & 47.0 & \textbf{49.8}\\
		\hline
	\end{tabular}
	}
	\vspace{-2mm}
	\caption{Experimental results on GTA5 $\rightarrow$ Cityscapes.}
	\label{table:gtacity}
\end{table*}

\begin{table*}[!t]
	\centering
	\resizebox{\linewidth}{!}{
	\begin{tabular}{c|c|cccccccccccccccc|c|c}
		\hline
		Method & Backbone & Road & SW & Build & Wall* & Fence* & Pole* & TL & TS & Veg. & Sky & PR & Rider & Car & Bus & Motor & Bike & mIoU & mIoU*\\
		\hline
		Source & \multirow{2}{0.1\linewidth}{\centering{DRN-105}} & 14.9 & 11.4 & 58.7 & 1.9 & 0.0 & 24.1 & 1.2 & 6.0 & 68.8 & 76.0 & 54.3 & 7.1 & 34.2 & 15.0 & 0.8 & 0.0 & 23.4 & 26.8\\
		MCD \cite{saito2017maximum} & & 84.8 & \textbf{43.6} & 79.0 & 3.9 & 0.2 & 29.1 & 7.2 & 5.5 & 83.8 & 83.1 & 51.0 & 11.7 & 79.9 & 27.2 & 6.2 & 0.0 & 37.3  & 43.5\\
		\hline
		Source & \multirow{2}{0.1\linewidth}{\centering{DeepLabv2}} & 55.6 & 23.8 & 74.6 & $-$ & $-$ & $-$ & 6.1 & 12.1 & 74.8 & 79.0 & 55.3 & 19.1 & 39.6 & 23.3 & 13.7 & 25.0 & $-$ & 38.6\\
		AdaptSegNet \cite{Tsai_adaptseg_2018} & & 84.3 & 42.7 & 77.5 & $-$ & $-$ & $-$ & 4.7 & 7.0 & 77.9 & 82.5 & 54.3 & 21.0 & 72.3 & 32.2 & 18.9 & 32.3 & $-$ & 46.7\\
		\hline
		AdvEnt \cite{vu2019advent} & DeepLabv2 & \textbf{85.6} & 42.2 & \textbf{79.7} & 8.7 & 0.4 & 25.9 & 5.4 & 8.1 & 80.4 & \textbf{84.1} & 57.9 & 23.8 & 73.3 & \textbf{36.4} & 14.2 & 33.0 & 41.2 & 48.0\\
		\hline
		Source & \multirow{2}{0.1\linewidth}{\centering{ResNet-38}} & 32.6 & 21.5 & 46.5 & 4.8 & 0.1 & 26.5 & 14.8 & 13.1 & 70.8 & 60.3 & 56.6 & 3.5 & 74.1 & 20.4 & 8.9 & 13.1 & 29.2 & 33.6\\
		CBST~\cite{Zou_2018_ECCV} & & 53.6 & 23.7 & 75.0 & 12.5 & 0.3 & 36.4 & 23.5 & 26.3 & \textbf{84.8} & 74.7 & \textbf{67.2} & 17.5 & \textbf{84.5} & 28.4 & 15.2 & \textbf{55.8} & 42.5 & 48.4\\
		\hline
		Source & \multirow{6}{0.1\linewidth}{\centering{DeepLabv2}} & 64.3 & 21.3 & 73.1 & 2.4 & 1.1 & 31.4 & 7.0 & 27.7 & 63.1 & 67.6 & 42.2 & 19.9 & 73.1 & 15.3 & 10.5 & 38.9 & 34.9 & 40.3\\
		CBST & & 68.0 & 29.9 & 76.3 & 10.8 & 1.4 & 33.9 & 22.8 & \textbf{29.5} & 77.6 & 78.3 & 60.6 & 28.3 & 81.6 & 23.5 & 18.8 & 39.8 & 42.6 & 48.9\\
		MRL2 &  & 63.4 & 27.1 & 76.4 & 14.2 & 1.4 & 35.2 & \textbf{23.6} & 29.4 & 78.5 & 77.8 & 61.4 & \textbf{29.5} & 82.2 & 22.8 & 18.9 & 42.3 & 42.8 & 48.7\\
		MRENT & & 69.6 & 32.6 & 75.8 & 12.2 & \textbf{1.8} & 35.3 & 23.3 & \textbf{29.5} & 77.7 & 78.9 & 60.0 & 28.5 & 81.5 & 25.9 & 19.6 & 41.8 & 43.4 & 49.6\\
		MRKLD & & 67.7 & 32.2 & 73.9 & 10.7 & 1.6 & \textbf{37.4} & 22.2 & 31.2 & 80.8 & 80.5 & 60.8 & 29.1 & 82.8 & 25.0 & 19.4 & 45.3 & \textbf{43.8} & \textbf{50.1}\\
		LRENT & & 65.6 & 30.3 & 74.6 & \textbf{13.8} & 1.5 & 35.8 & 23.1 & 29.1 & 77.0 & 77.5 & 60.1 & 28.5 & 82.2 & 22.6 & \textbf{20.1} & 41.9 & 42.7 & 48.7\\
		\hline
	\end{tabular}
	}
	\vspace{-2mm}
	\caption{Experimental results on SYNTHIA $\rightarrow$ Cityscapes.}
	\label{table:syncity}
\end{table*}

Confidence regularization smooths the output by lowering the confidence (the max of output softmax) and raising the probability level of other classes. Such smoothing helps to reduce the confidence on false positives (FP), although the confidence of certain true positives (TP) may also decrease. To see the change w/wo CR, we compare CBST vs MRKLD/LRENT (DeepLabv2) on GTA5 $\rightarrow$ Cityscapes, by presenting their per-class mean confidence of TP ($C_{TP}$), mean confidence of FP ($C_{FP}$) and the $C_{TP}/C_{FP}$ ratios at the end of first round in Table \ref{ConfComp}. For both TP and FP, the confidence of MRKLD/LRENT are lower than CBST, but either MRKLD or LRENT outperforms CBST on almost all per-class ratios and mean ratios. This intuitively illustrates how confidence regularization benefits self-training.
\begin{figure*}[!t]
	\centering
	\resizebox{0.98\textwidth}{!}{
	\begin{tabular}{@{}cccccccccc@{}}
		\cellcolor{city_color_1}\textcolor{white}{~~road~~} &
		\cellcolor{city_color_2}~~sidewalk~~&
		\cellcolor{city_color_3}\textcolor{white}{~~building~~} &
		\cellcolor{city_color_4}\textcolor{white}{~~wall~~} &
		\cellcolor{city_color_5}~~fence~~ &
		\cellcolor{city_color_6}~~pole~~ &
		\cellcolor{city_color_7}~~traffic lgt~~ &
		\cellcolor{city_color_8}~~traffic sgn~~ &
		\cellcolor{city_color_9}~~vegetation~~ & 
		\cellcolor{city_color_0}\textcolor{white}{~~ignored~~}\\
		\cellcolor{city_color_10}~~terrain~~ &
		\cellcolor{city_color_11}~~sky~~ &
		\cellcolor{city_color_12}\textcolor{white}{~~person~~} &
		\cellcolor{city_color_13}\textcolor{white}{~~rider~~} &
		\cellcolor{city_color_14}\textcolor{white}{~~car~~} &
		\cellcolor{city_color_15}\textcolor{white}{~~truck~~} &
		\cellcolor{city_color_16}\textcolor{white}{~~bus~~} &
		\cellcolor{city_color_17}\textcolor{white}{~~train~~} &
		\cellcolor{city_color_18}\textcolor{white}{~~motorcycle~~} &
		\cellcolor{city_color_19}\textcolor{white}{~~bike~~}
	\vspace{0.5mm}
	\end{tabular}
	}
	\vspace{0.5mm}
	\includegraphics[width=0.138\textwidth]{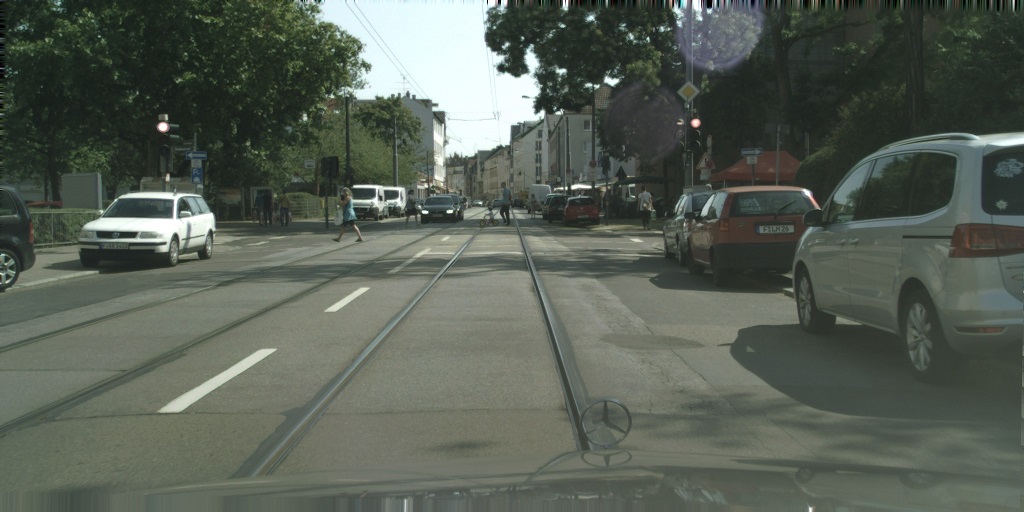}
	\includegraphics[width=0.138\textwidth]{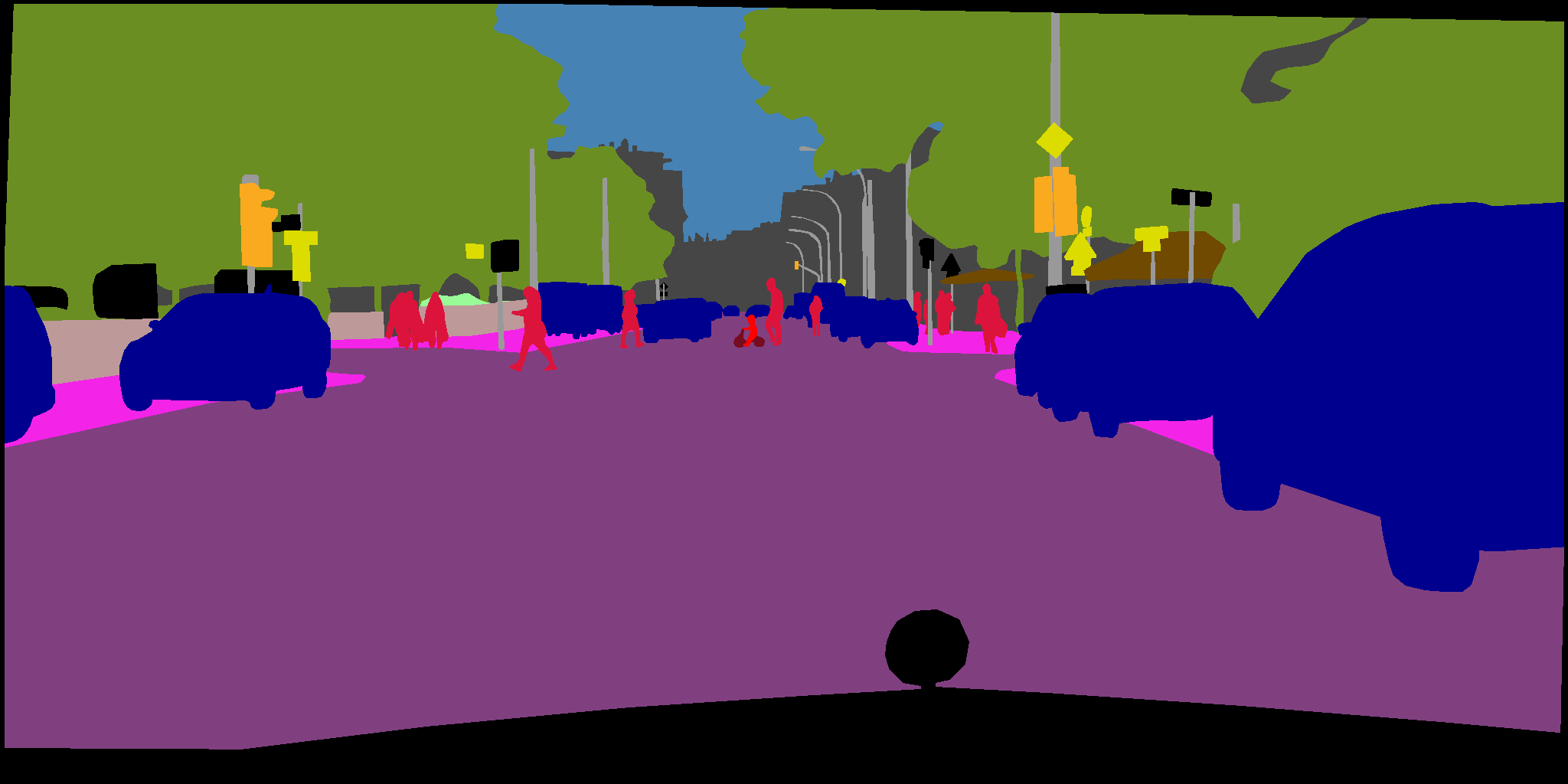}
	\includegraphics[width=0.138\textwidth]{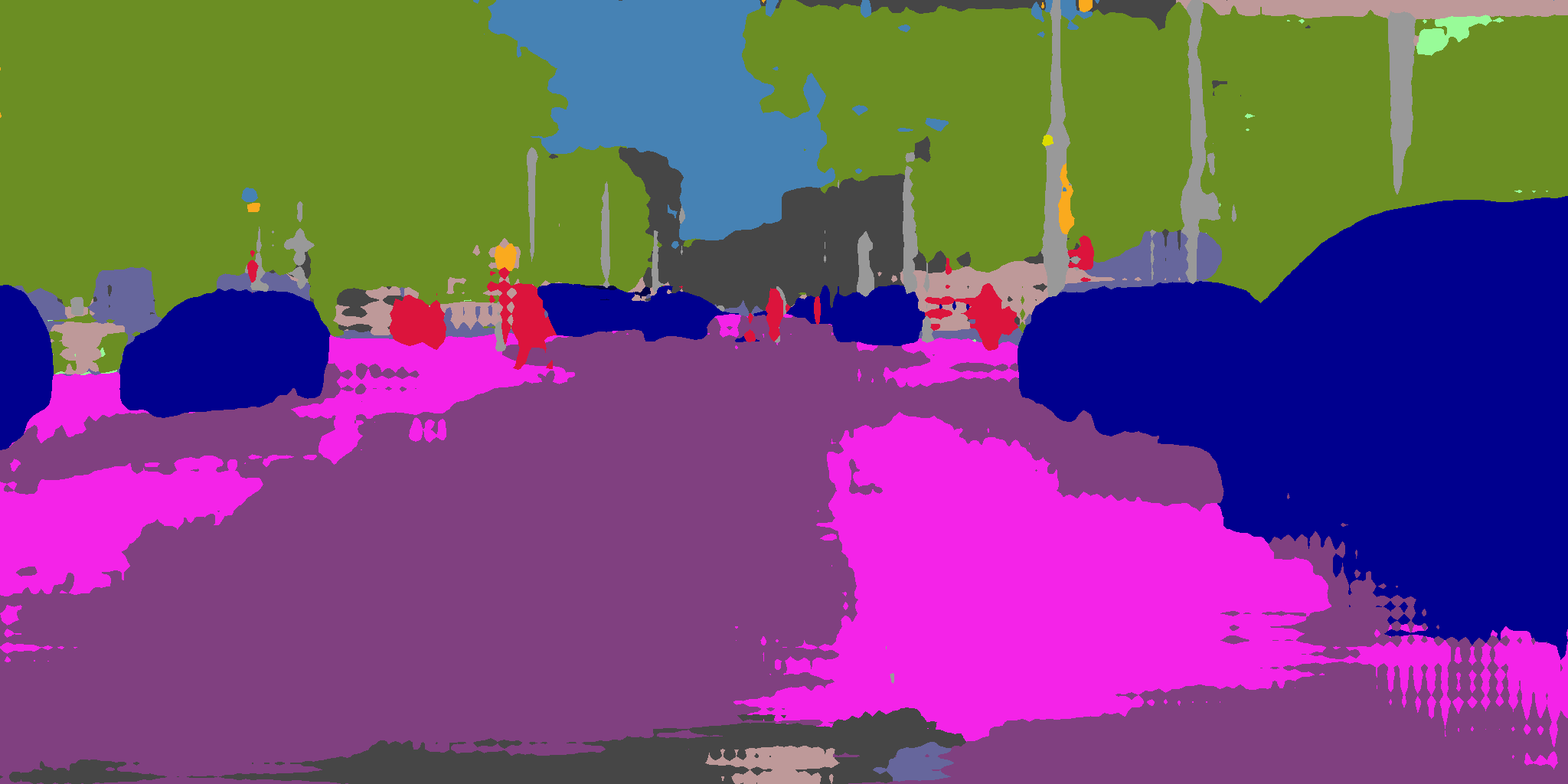}
	\includegraphics[width=0.138\textwidth]{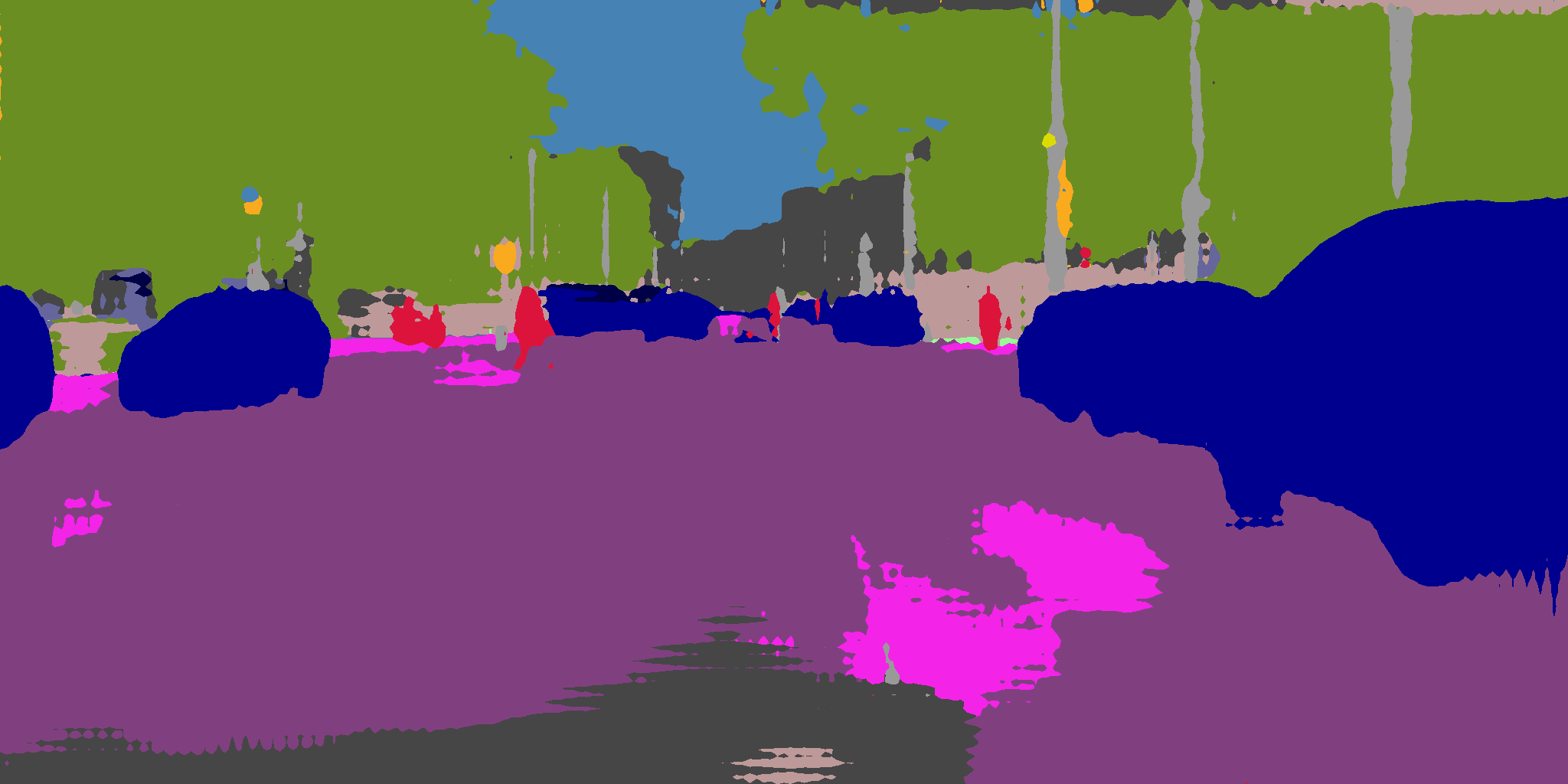}
	\includegraphics[width=0.138\textwidth]{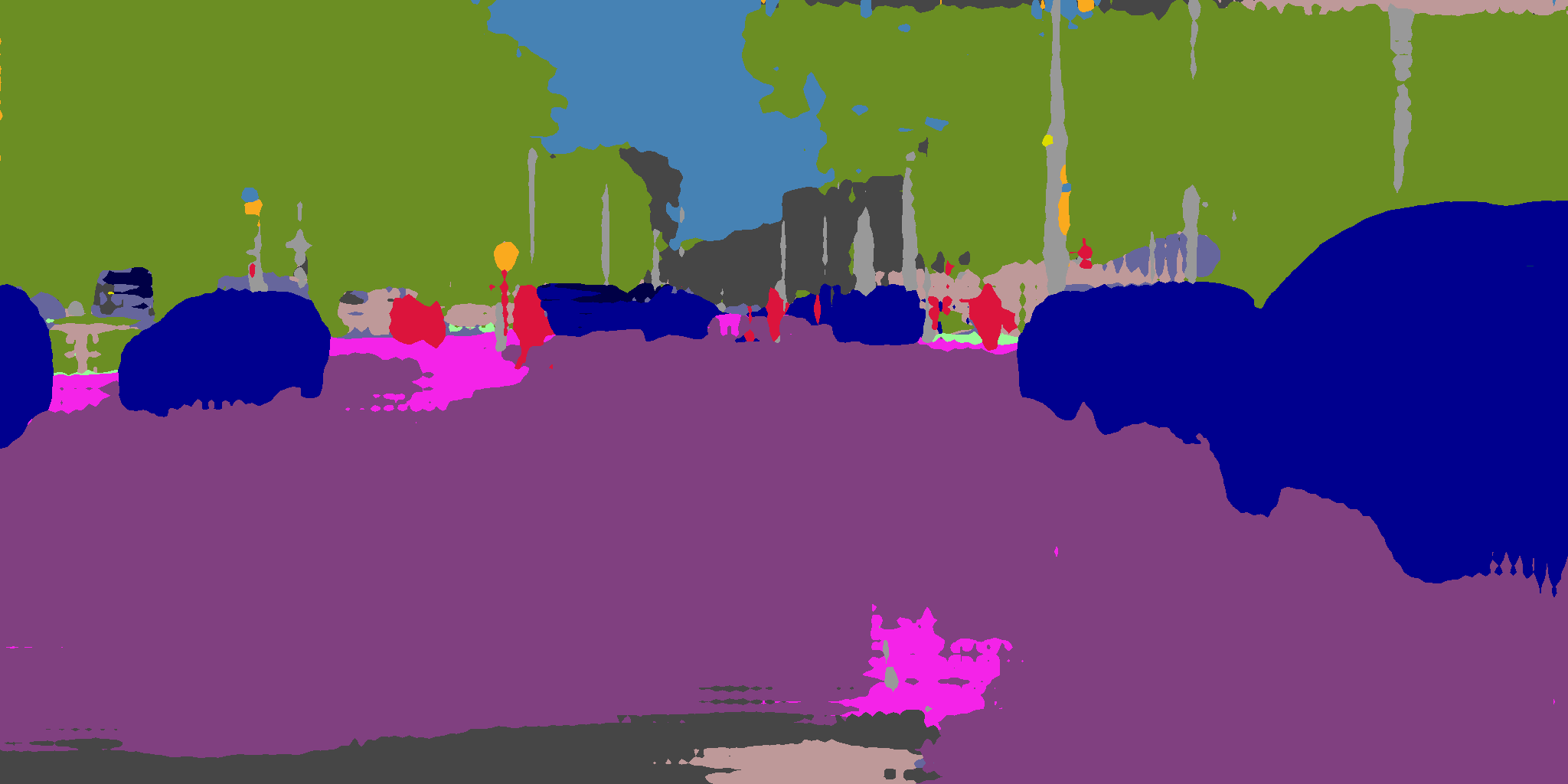}
	\includegraphics[width=0.138\textwidth]{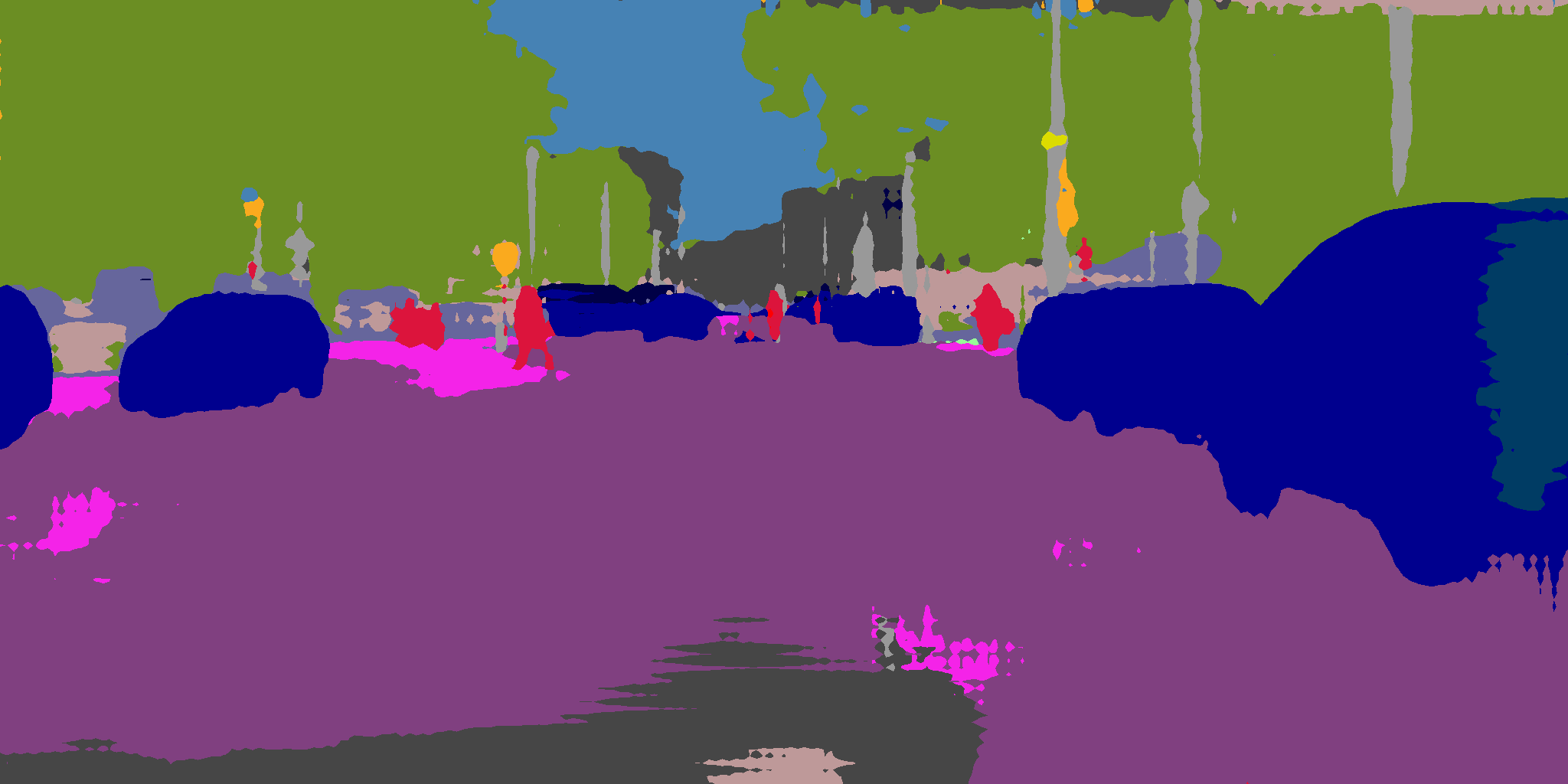}
	\includegraphics[width=0.138\textwidth]{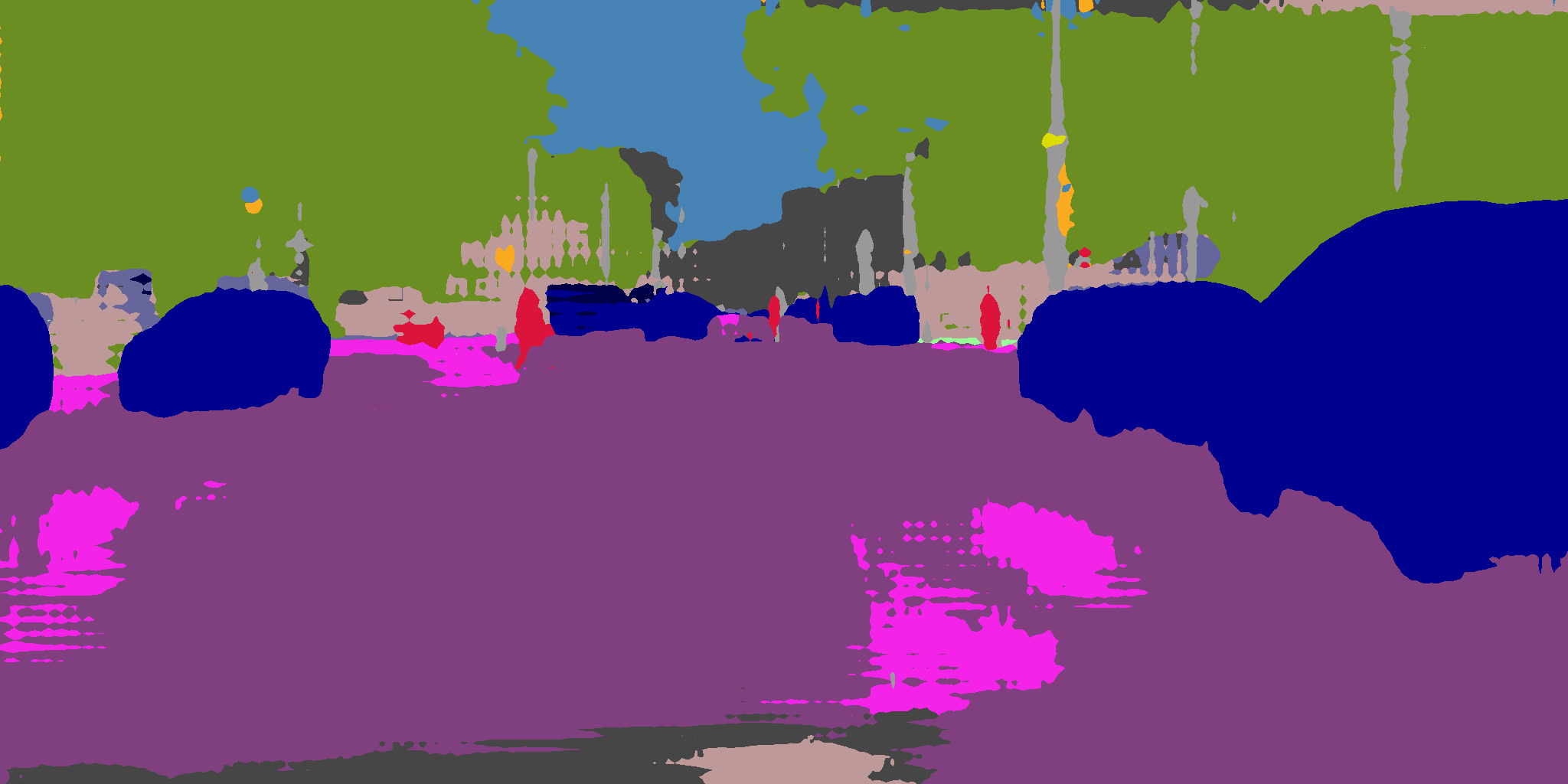}\\
	\includegraphics[width=0.138\textwidth]{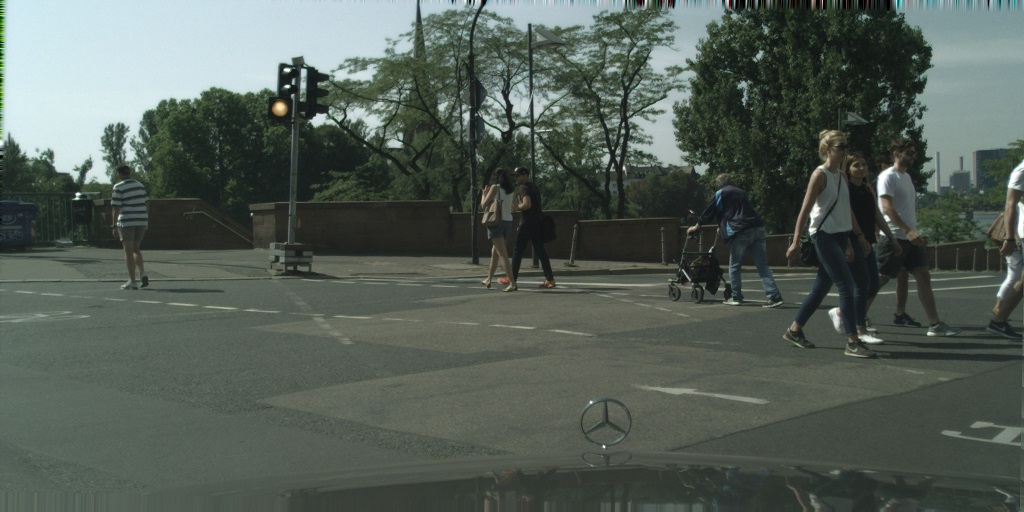}
	\includegraphics[width=0.138\textwidth]{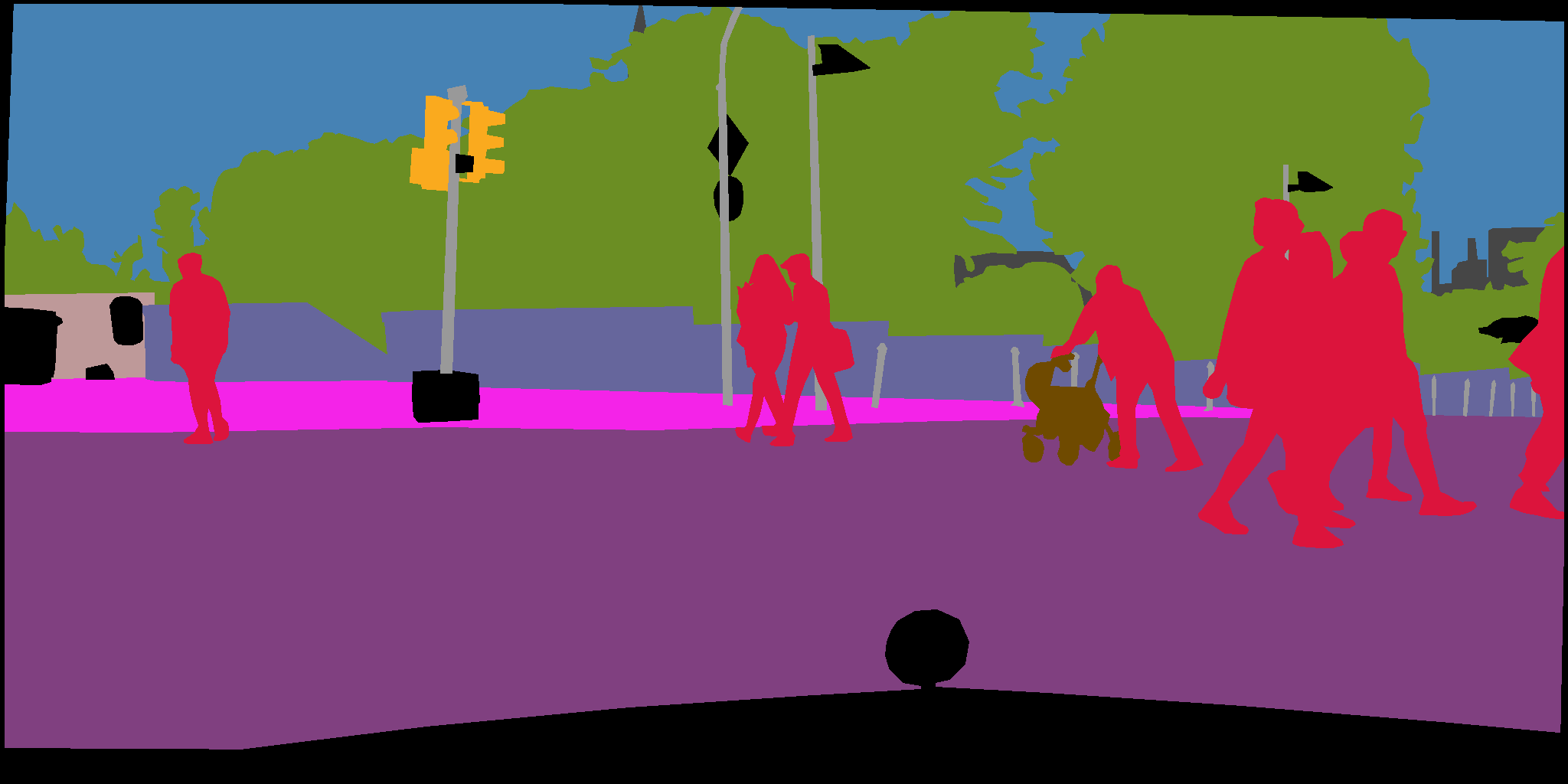}
	\includegraphics[width=0.138\textwidth]{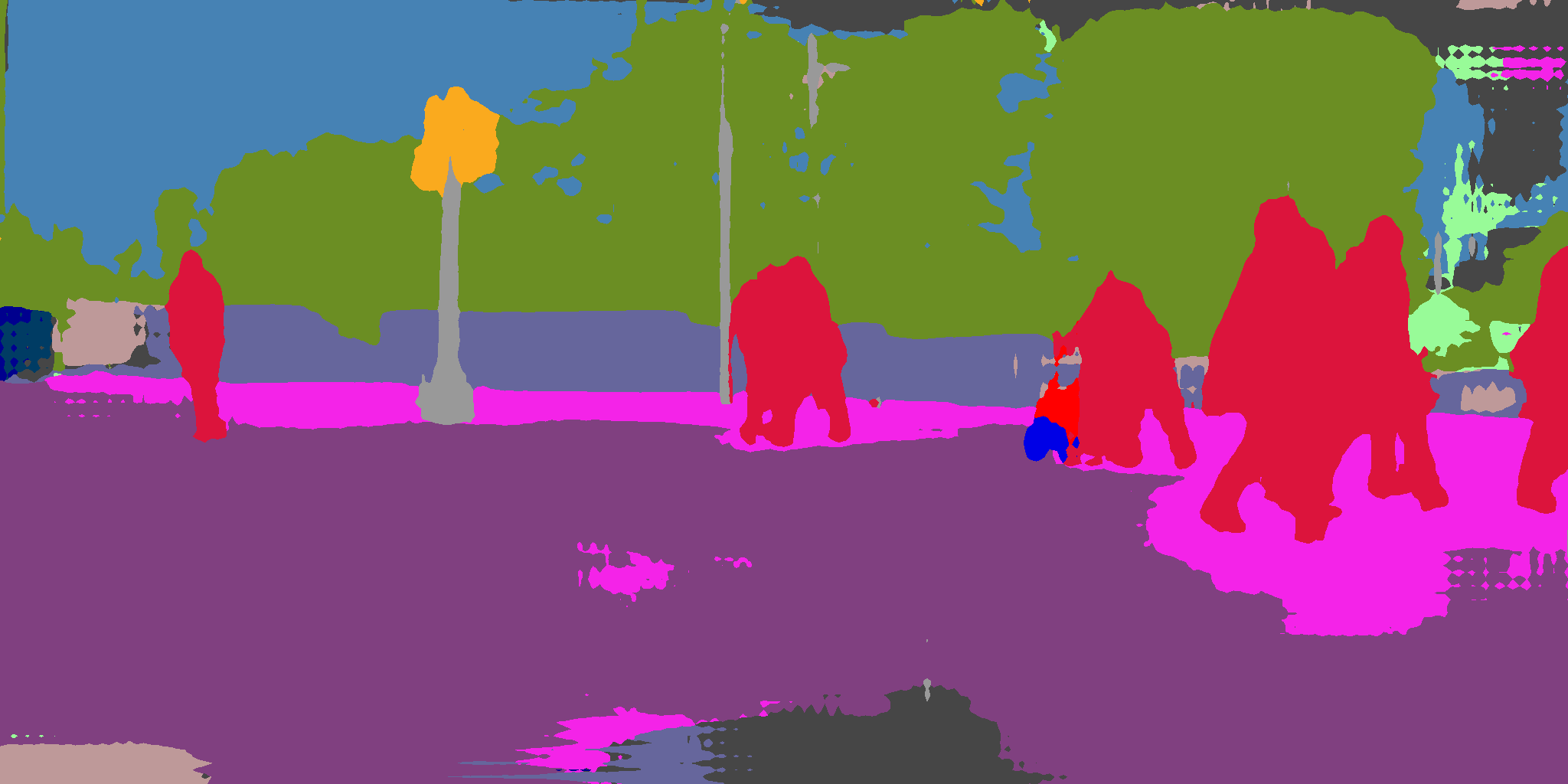}
	\includegraphics[width=0.138\textwidth]{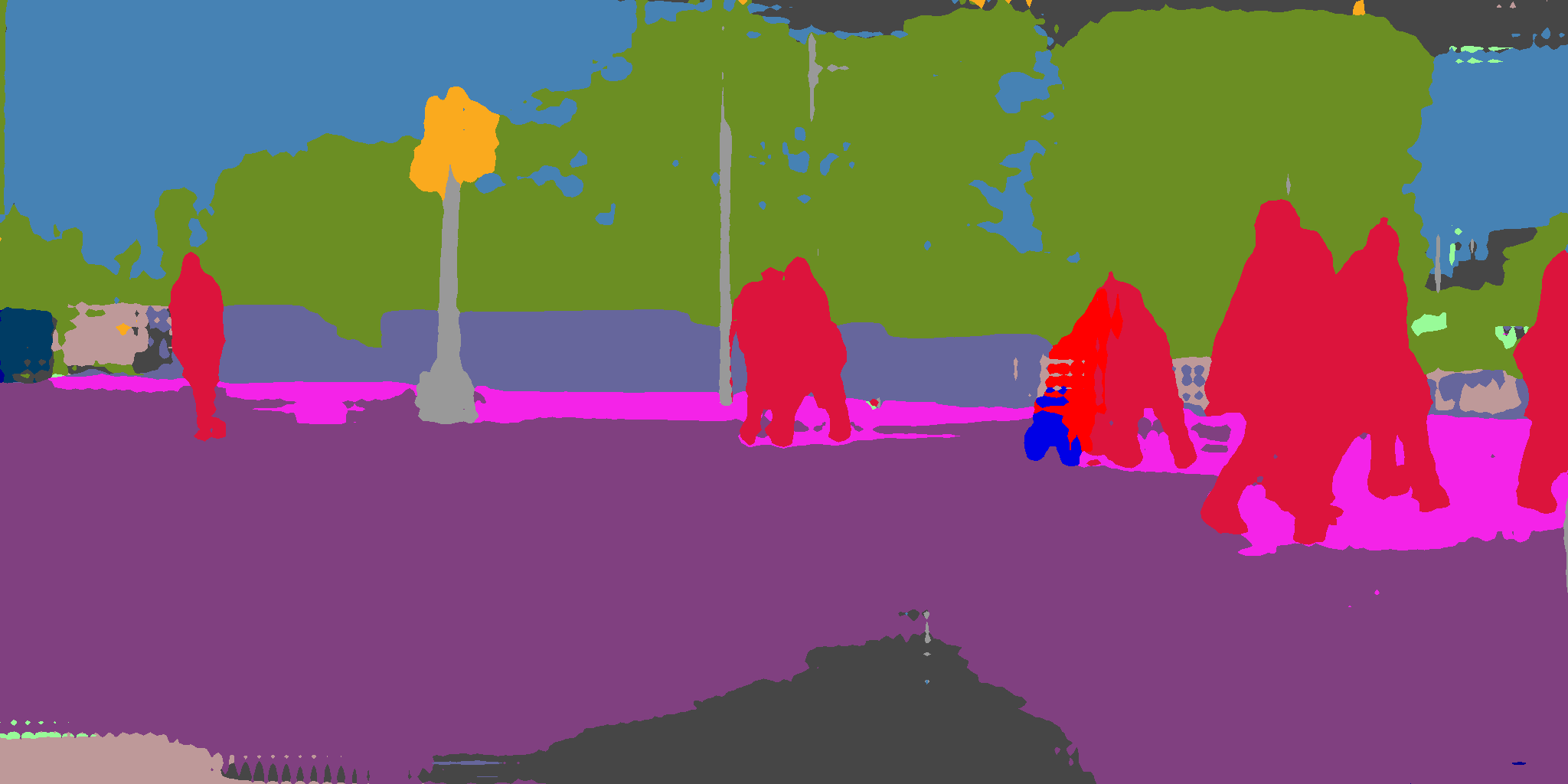}
	\includegraphics[width=0.138\textwidth]{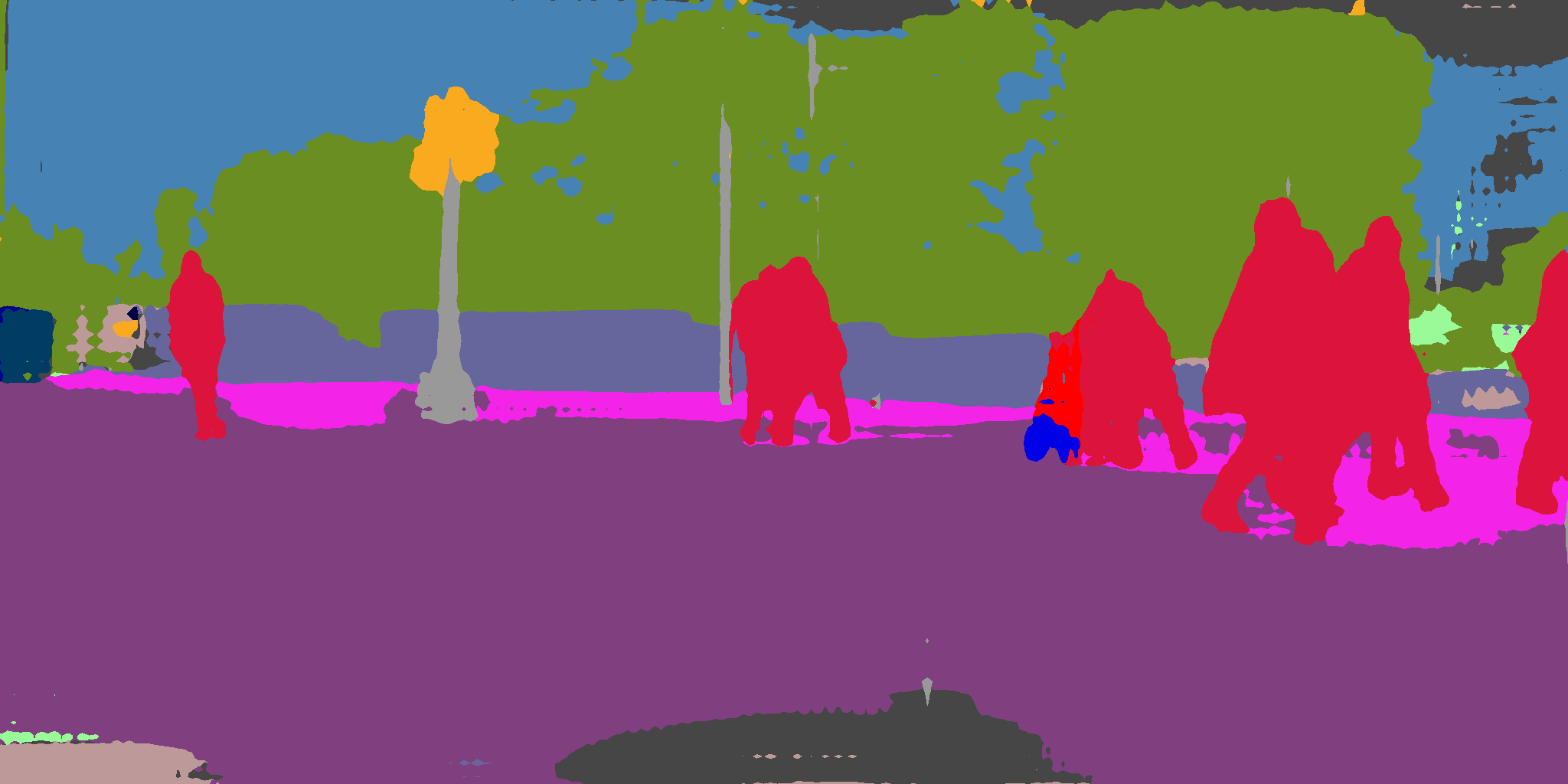}
	\includegraphics[width=0.138\textwidth]{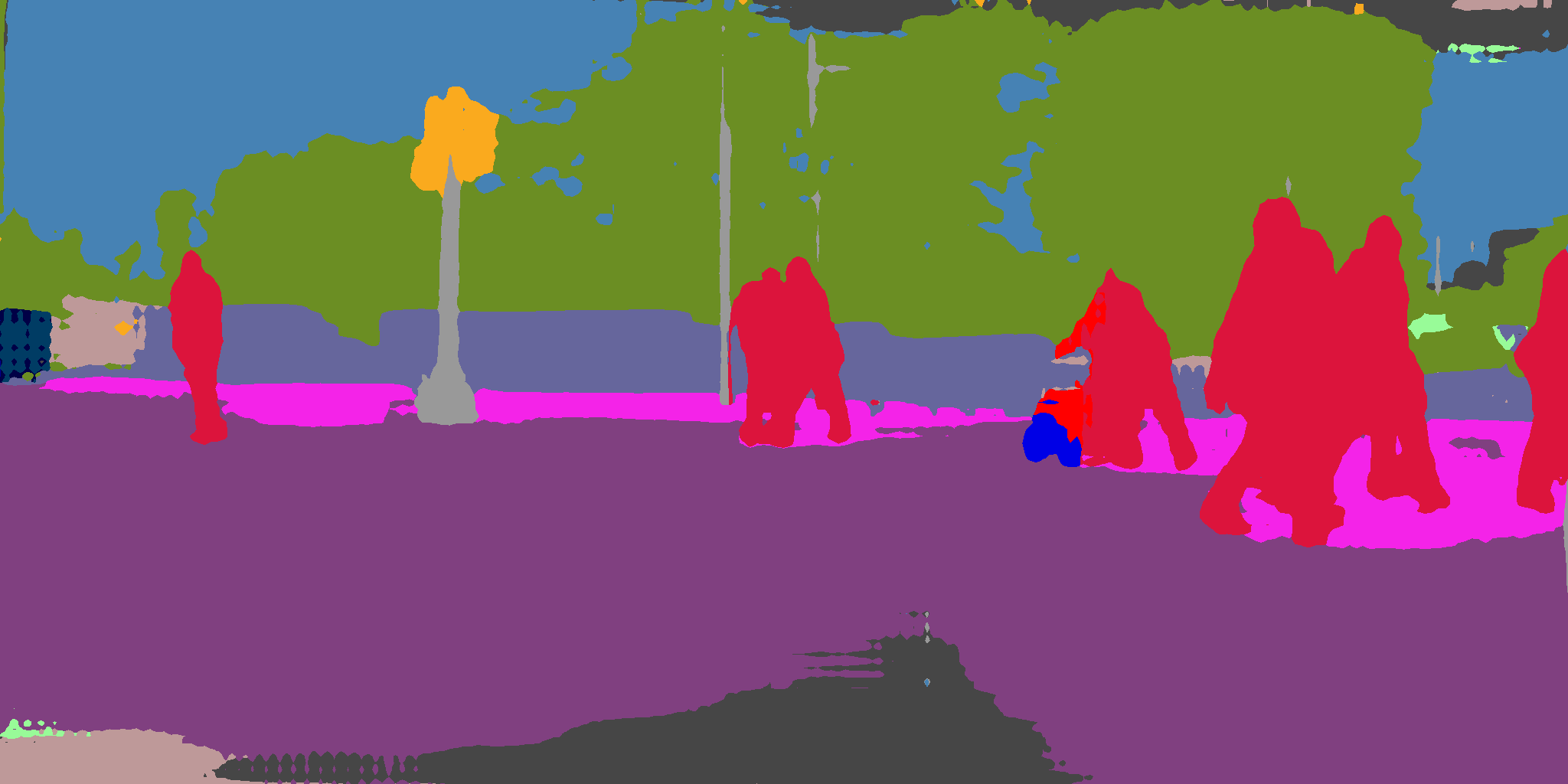}
	\includegraphics[width=0.138\textwidth]{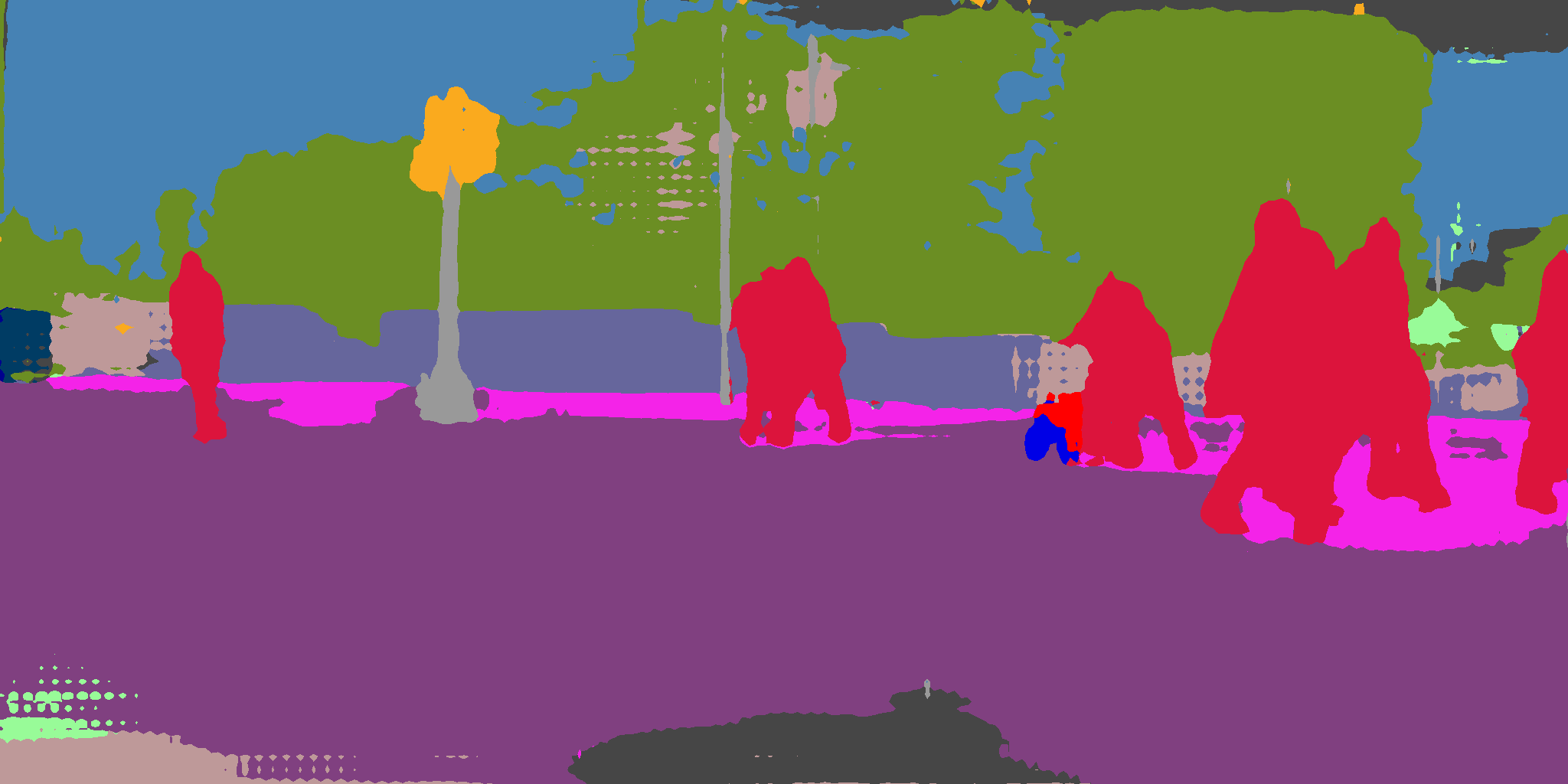}\\
	\vspace{-2mm}
	\caption{Adaptation results on GTA5 $\rightarrow$ Cityscapes. Rows correspond to sample images in Cityscapes. From left to right, columns correspond to original images, ground truth, and predication results of CBST, MRL2, MRENT, MRKLD, LRENT.}
	\label{fig:gta2city}
\end{figure*}

\begin{figure*}[!t]
	\centering
	\includegraphics[width=0.138\textwidth]{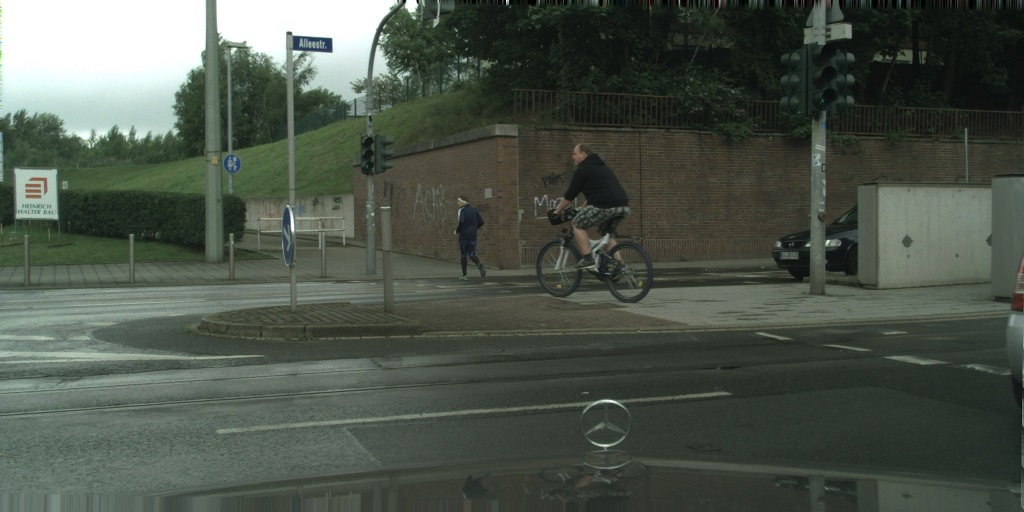}
	\includegraphics[width=0.138\textwidth]{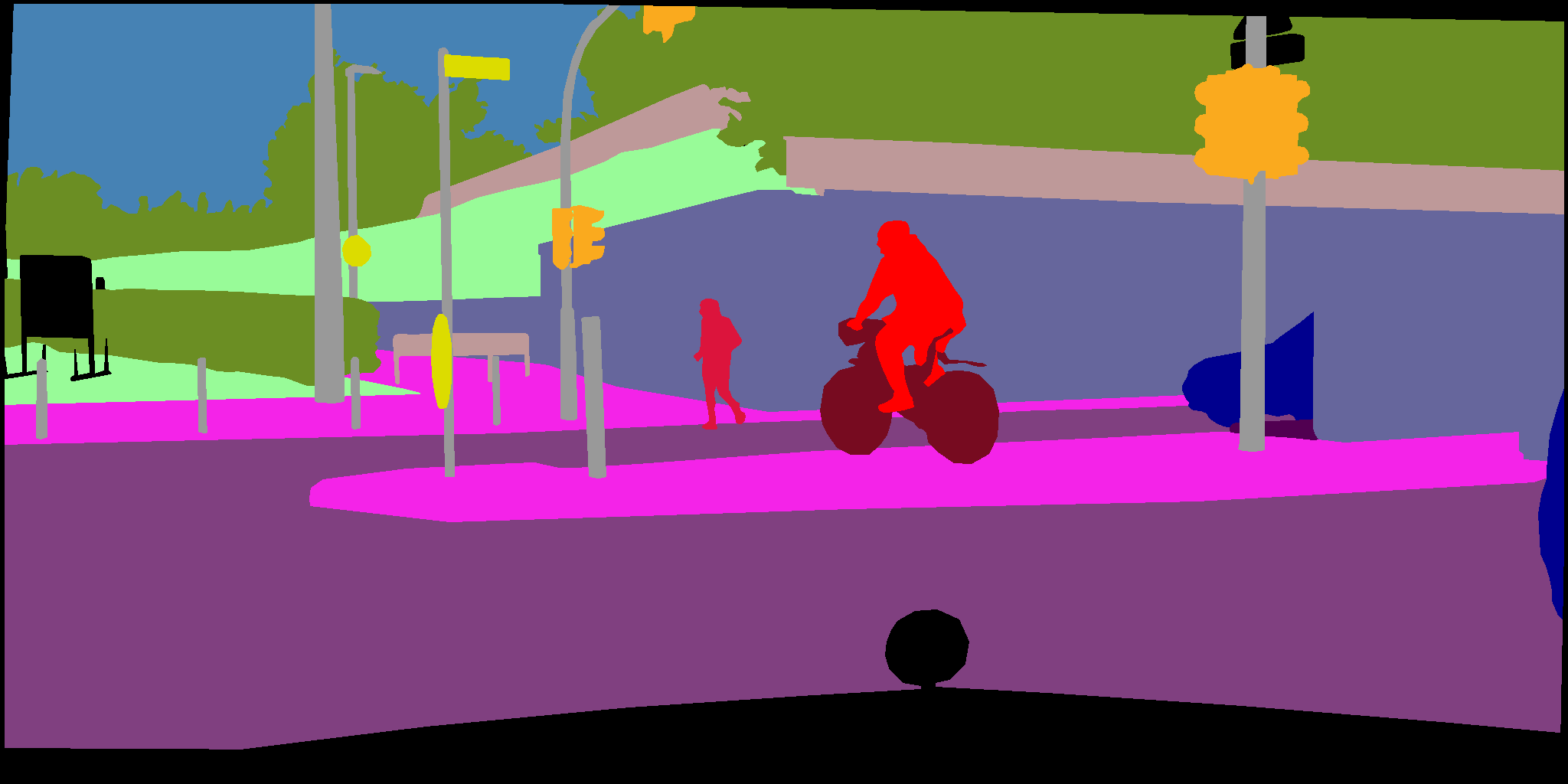}
	\includegraphics[width=0.138\textwidth]{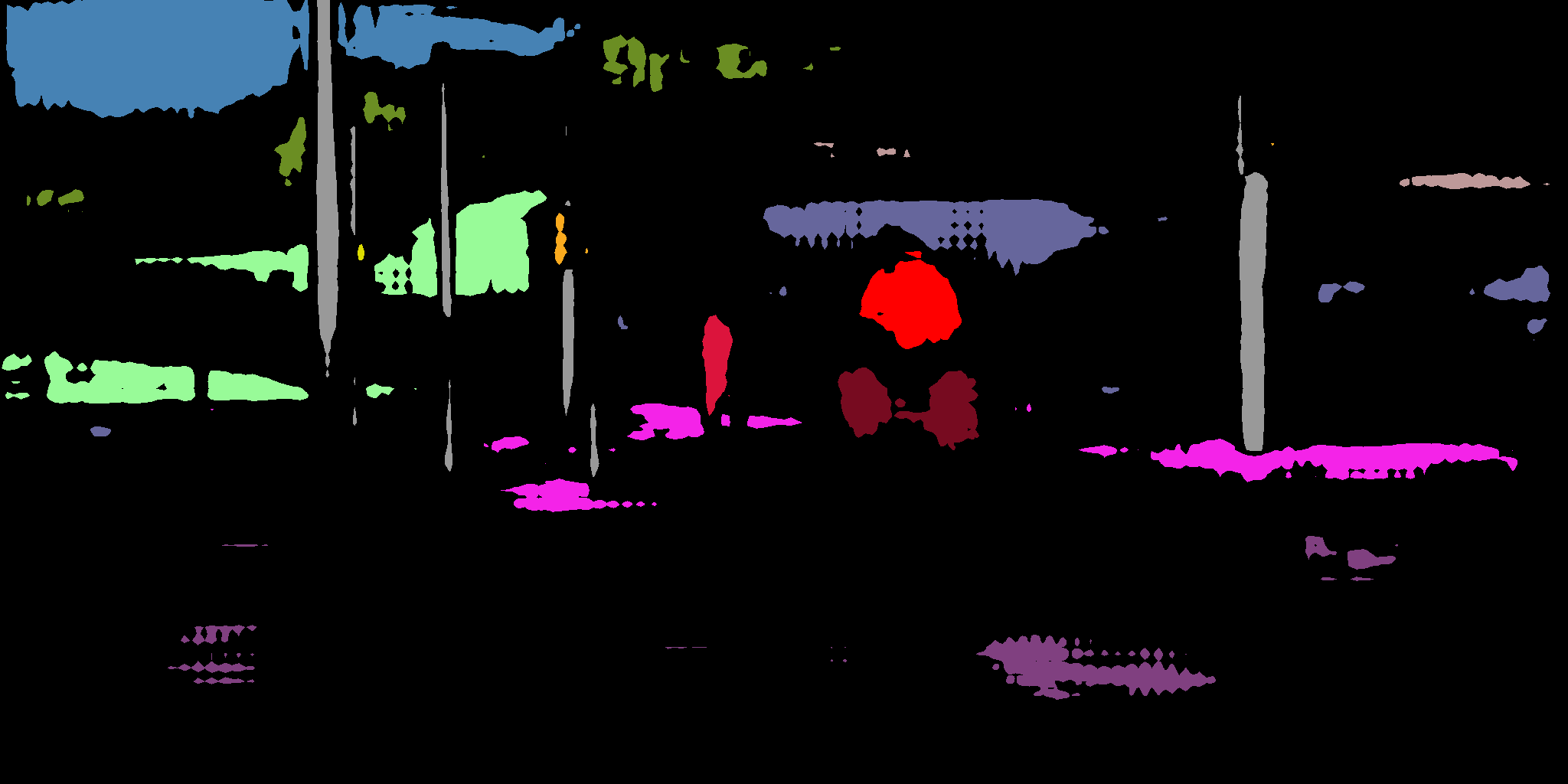}
	\includegraphics[width=0.138\textwidth]{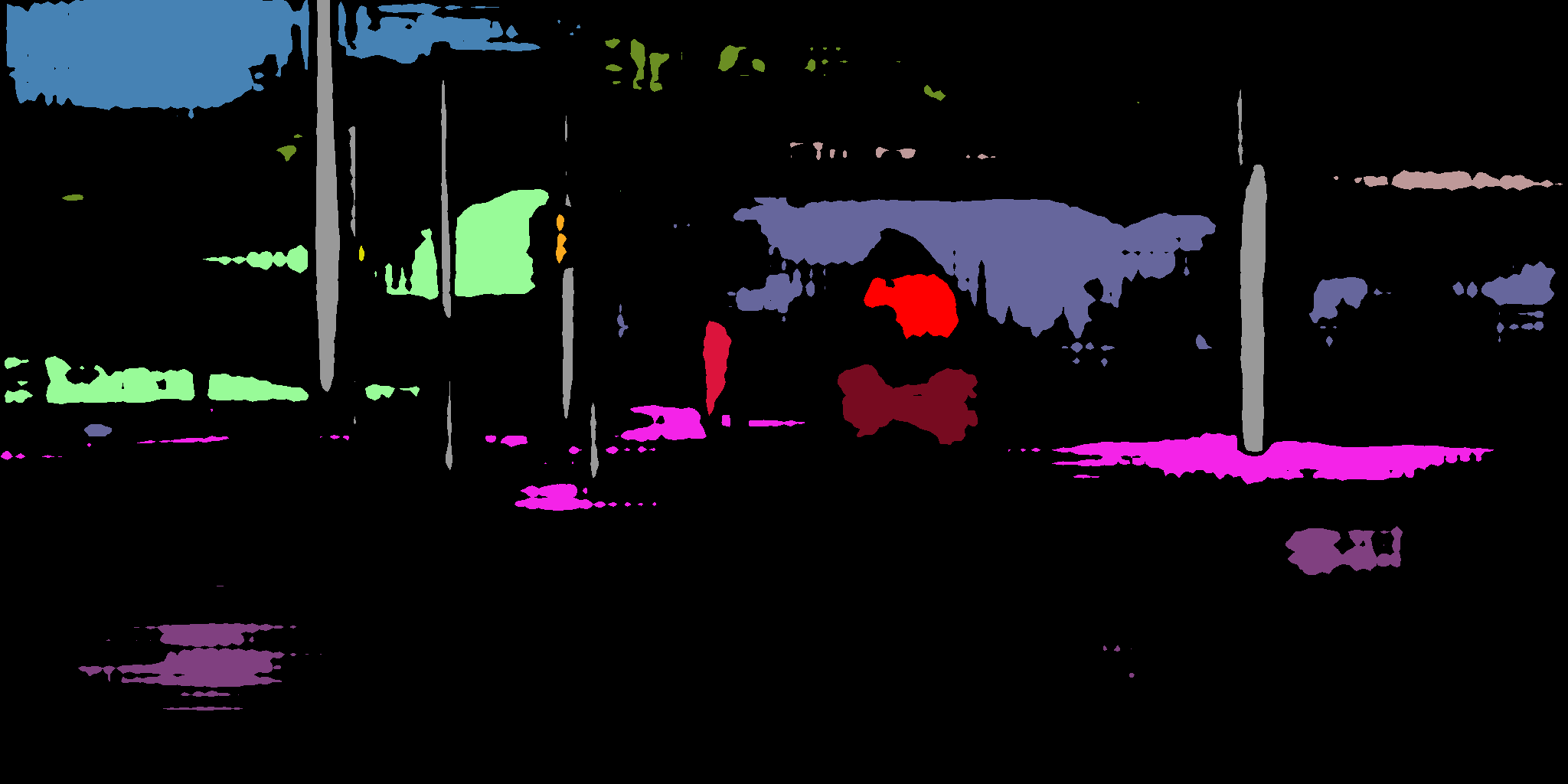}
	\includegraphics[width=0.138\textwidth]{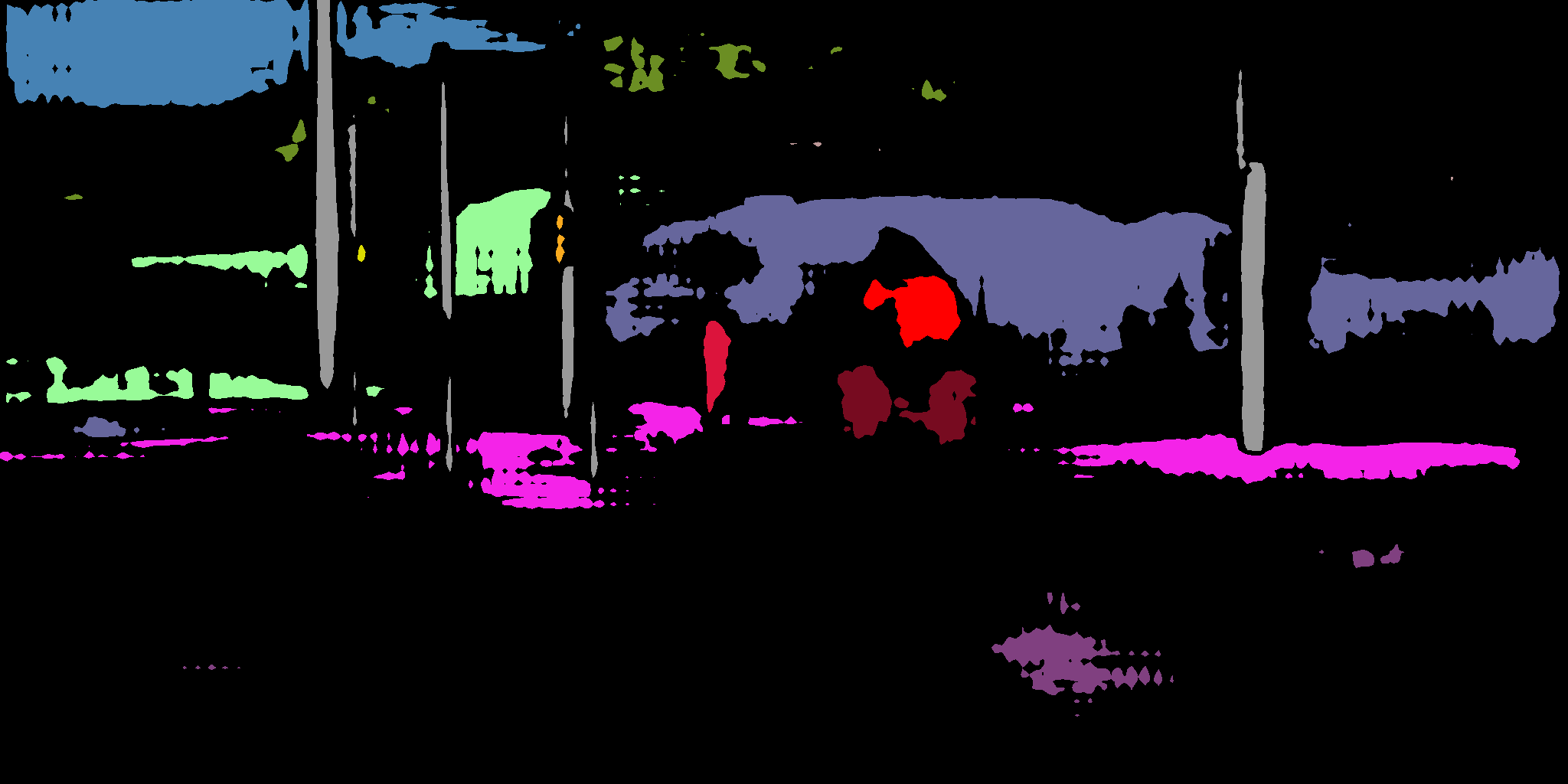}
	\includegraphics[width=0.138\textwidth]{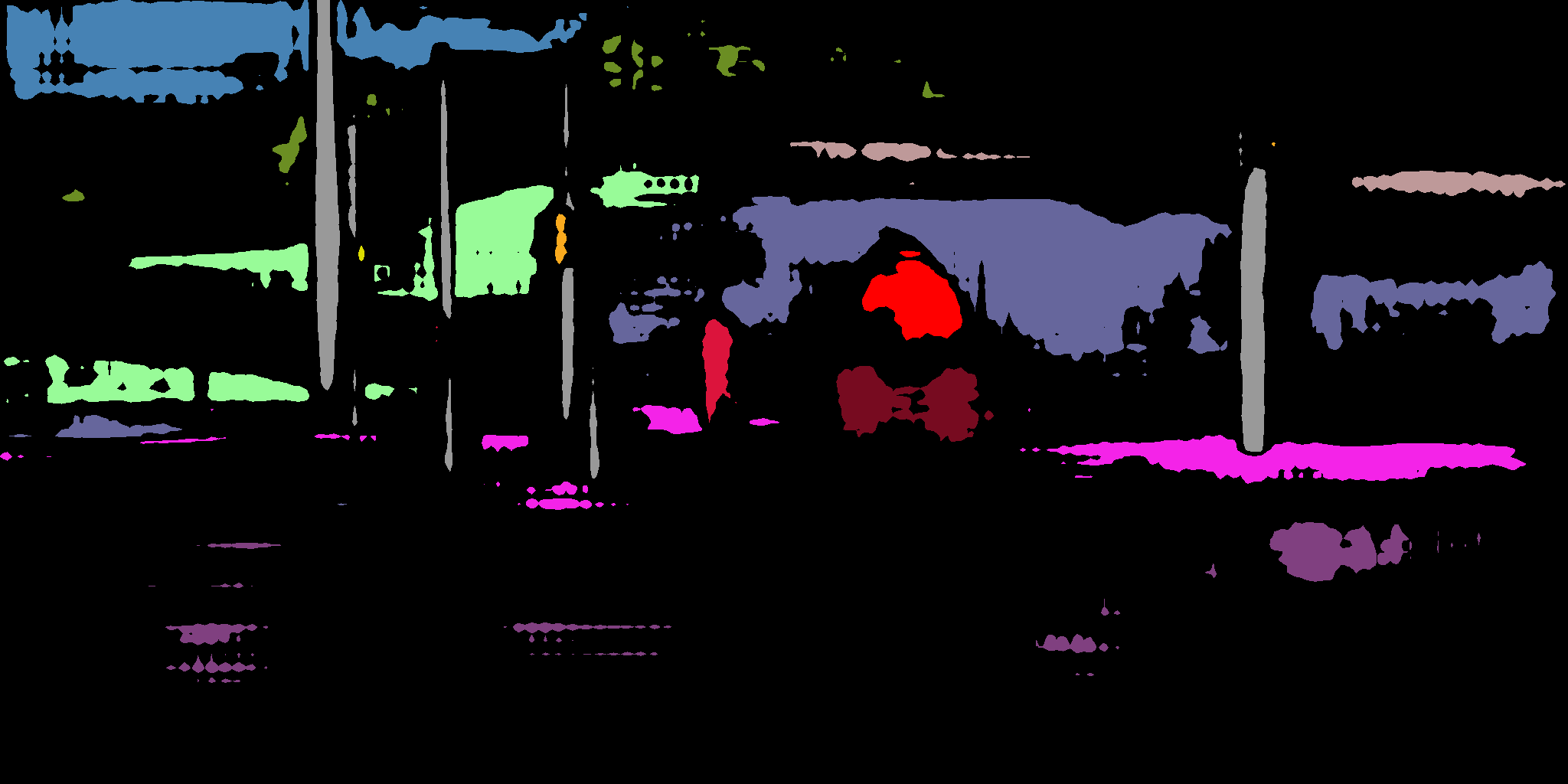}
	\includegraphics[width=0.138\textwidth]{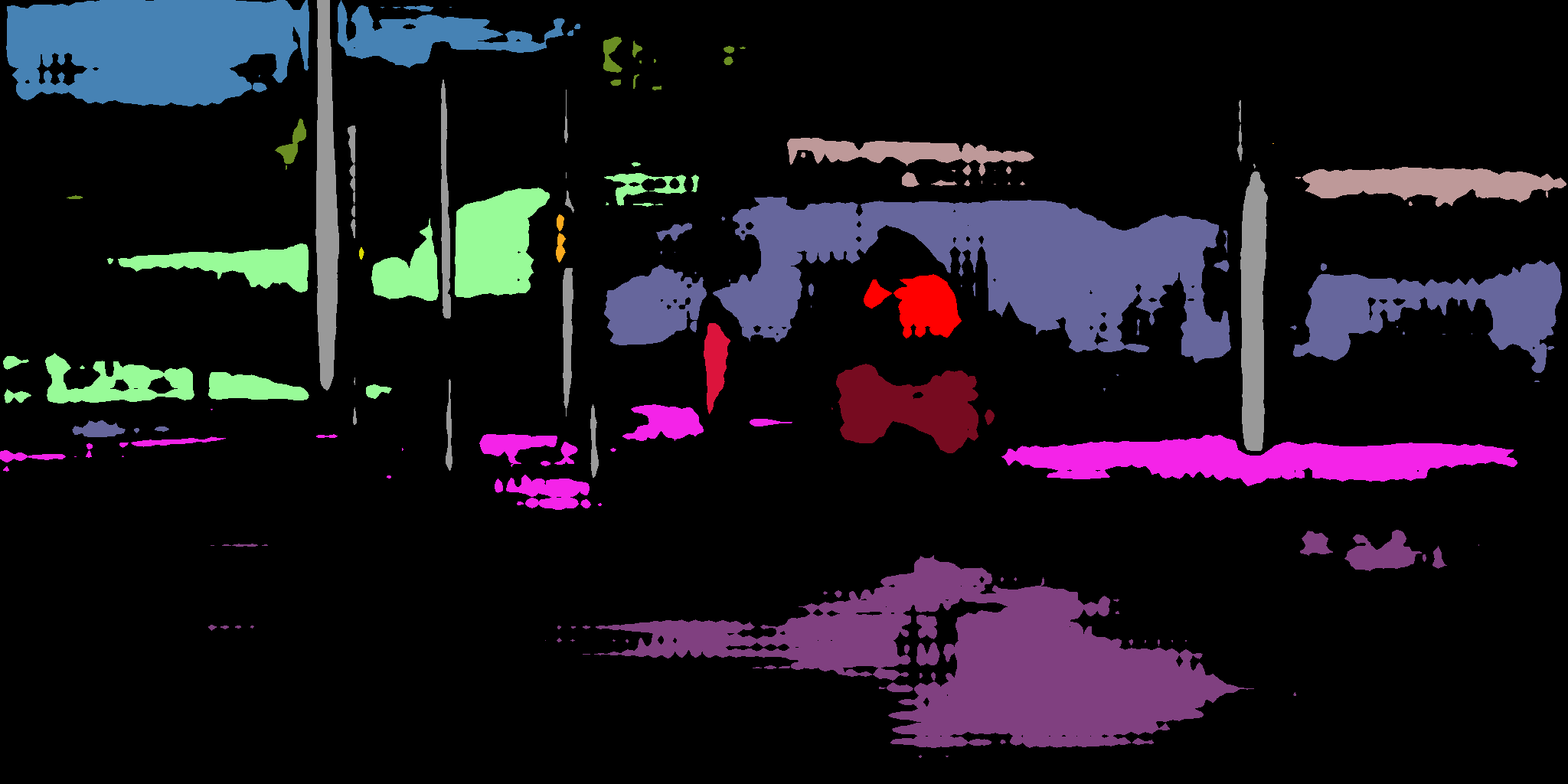}\\
	\vspace{-2mm}
	\caption{Pseudo-labels in GTA5 $\rightarrow$ Cityscapes. Rows correspond to sample images in Cityscapes. From left to right, columns correspond to original images, ground truth, and pseudo-labels of CBST, MRL2, MRENT, MRKLD, LRENT.}
	\label{fig:plgta2city}
\end{figure*}

\begin{figure}[t]
	\centering
	\includegraphics[width=0.45\linewidth]{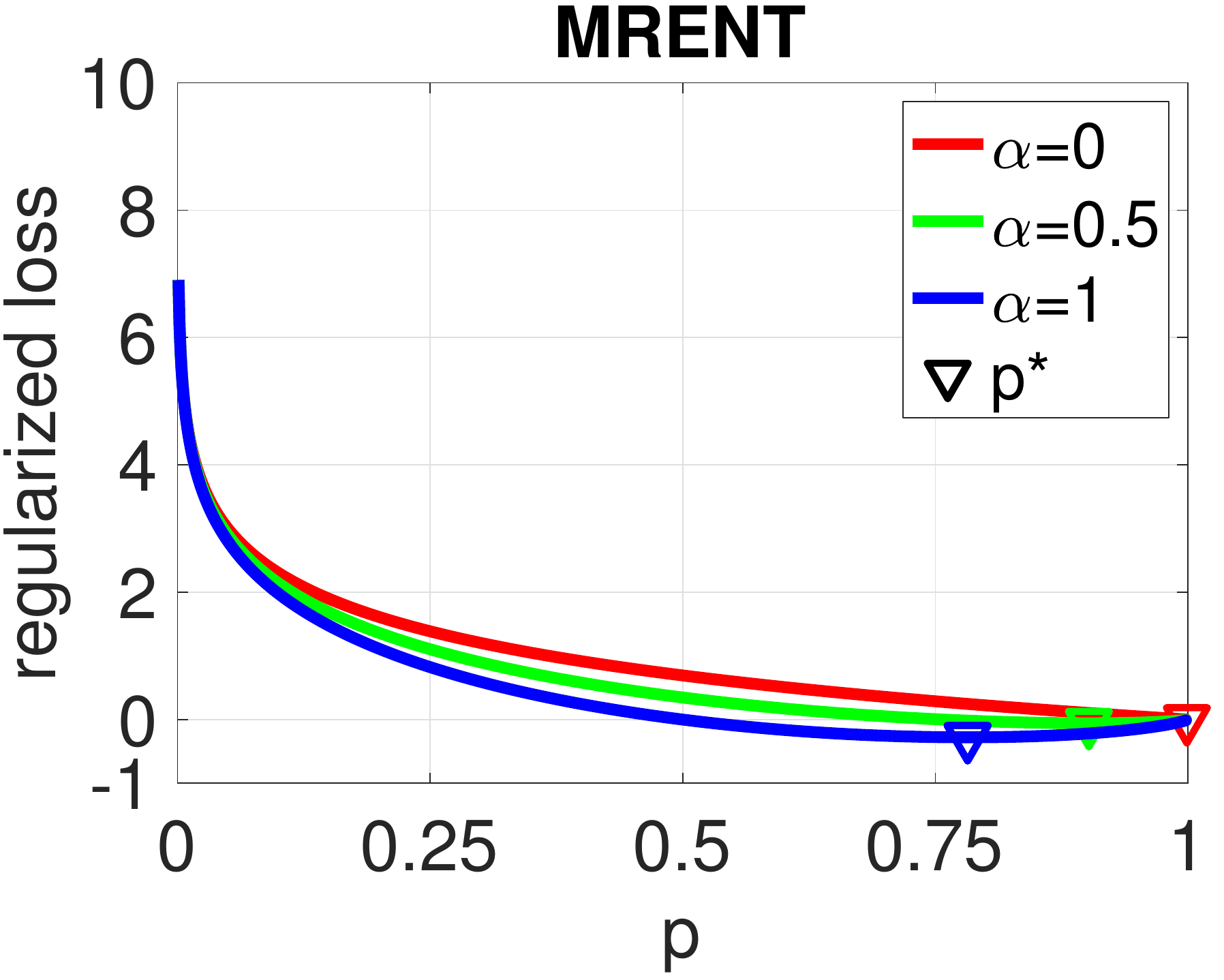}
	\includegraphics[width=0.45\linewidth]{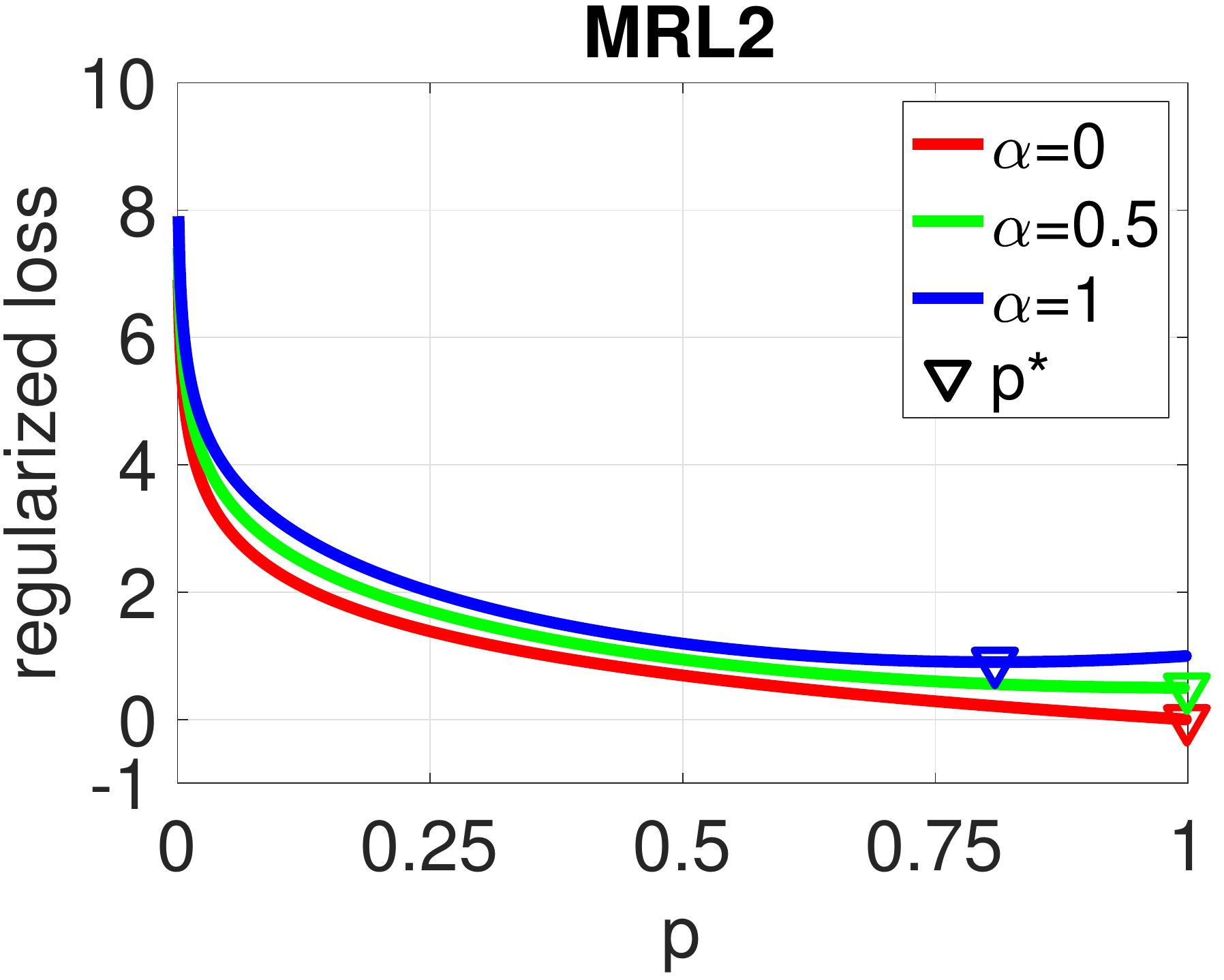}\\
	\includegraphics[width=0.45\linewidth]{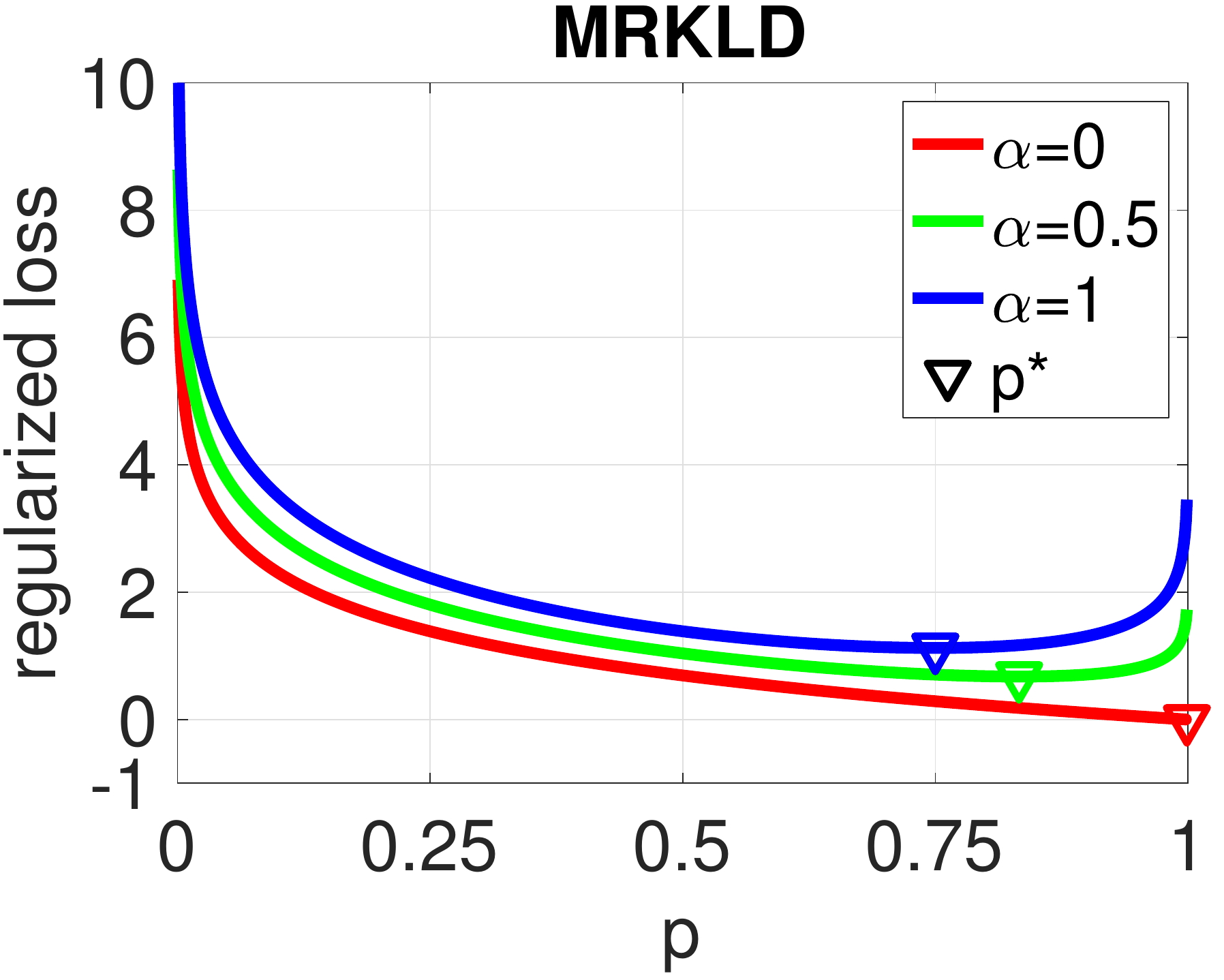}
	\includegraphics[width=0.45\linewidth]{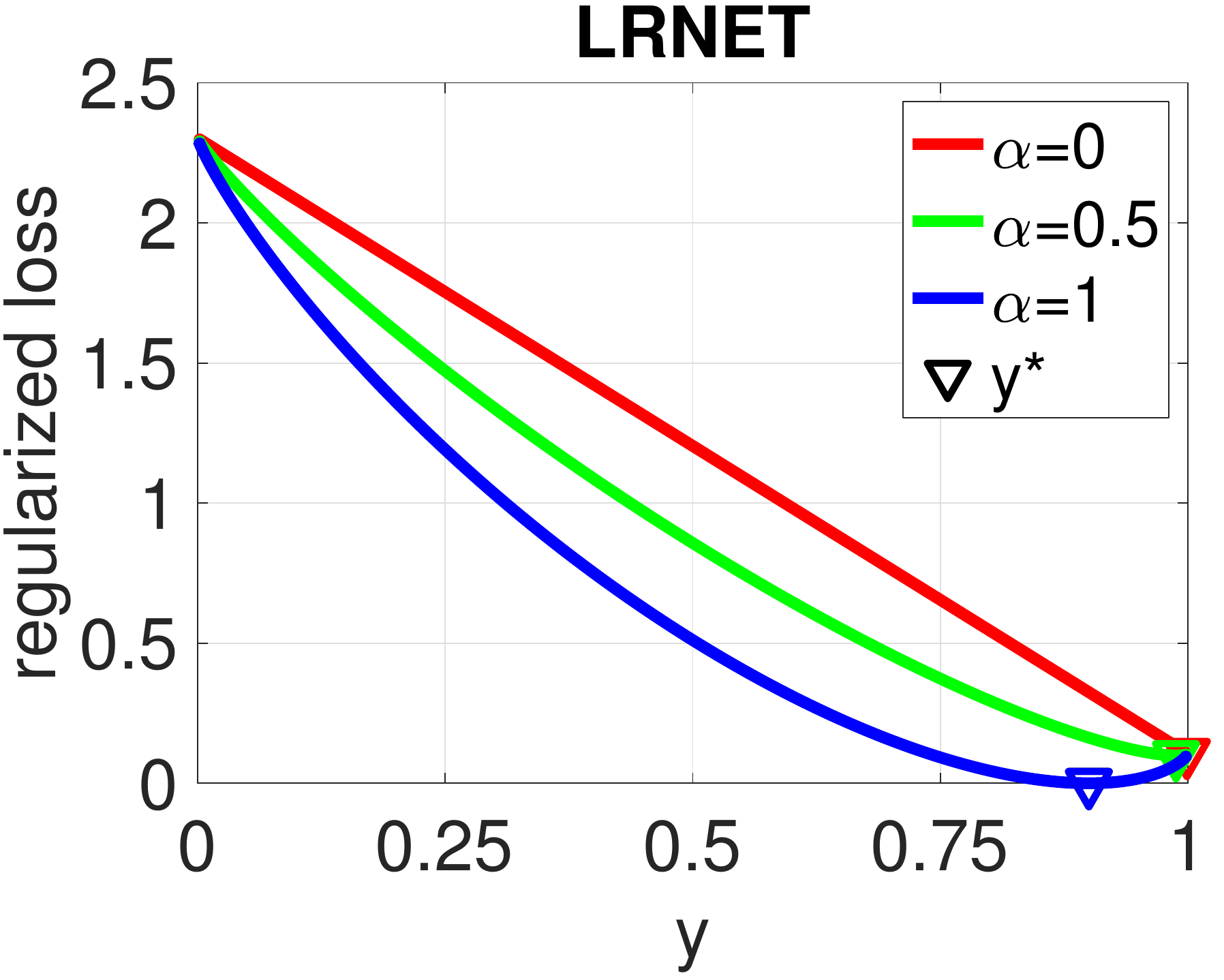}
	\vspace{-3mm}
	\caption{Loss curves regularized by different regularizers.}
	\label{fig:loss_curves}
	\vspace{-1mm}
\end{figure}

\begin{figure}[t]
	\centering
	\includegraphics[width=0.45\linewidth]{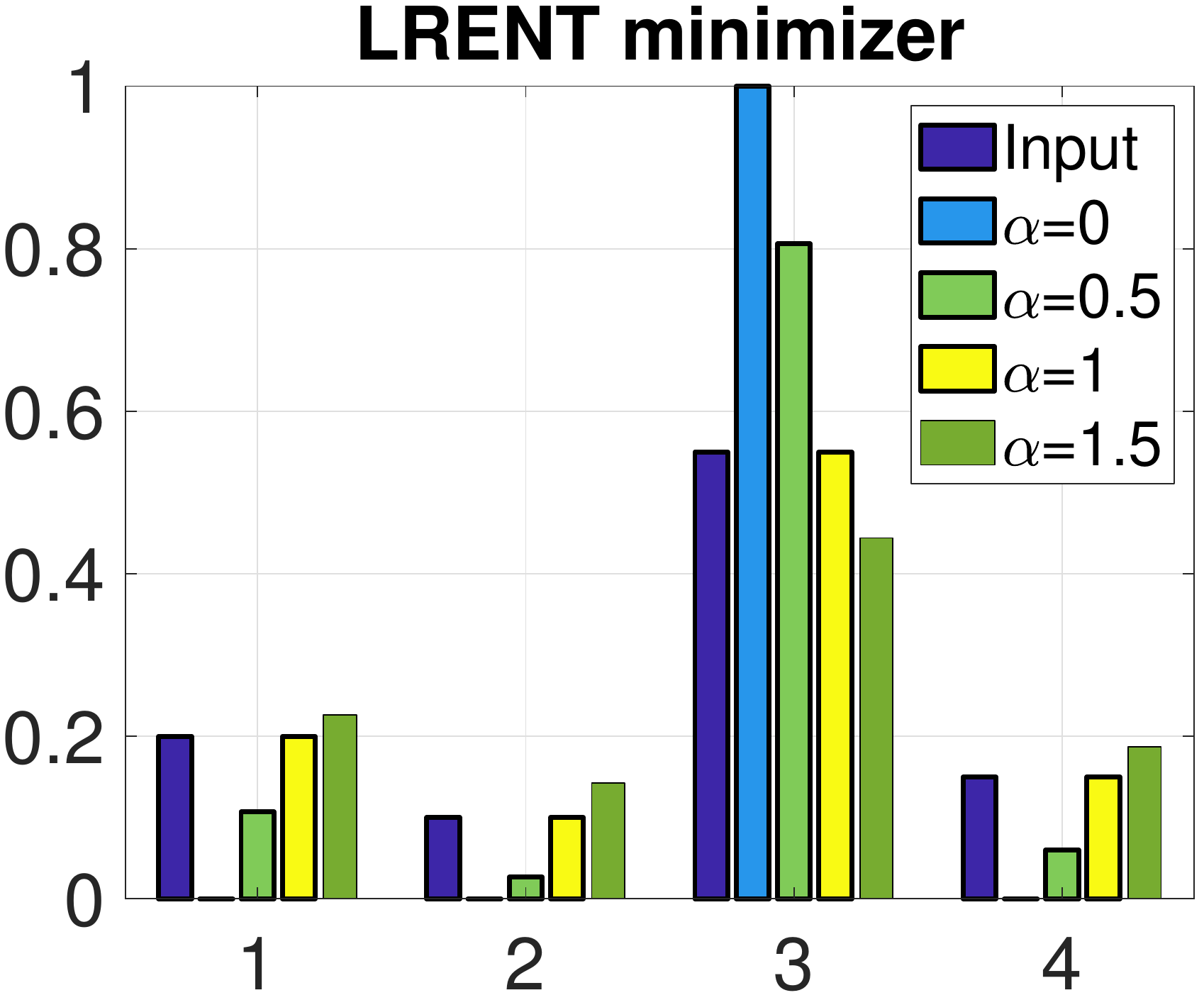}
	\includegraphics[width=0.45\linewidth]{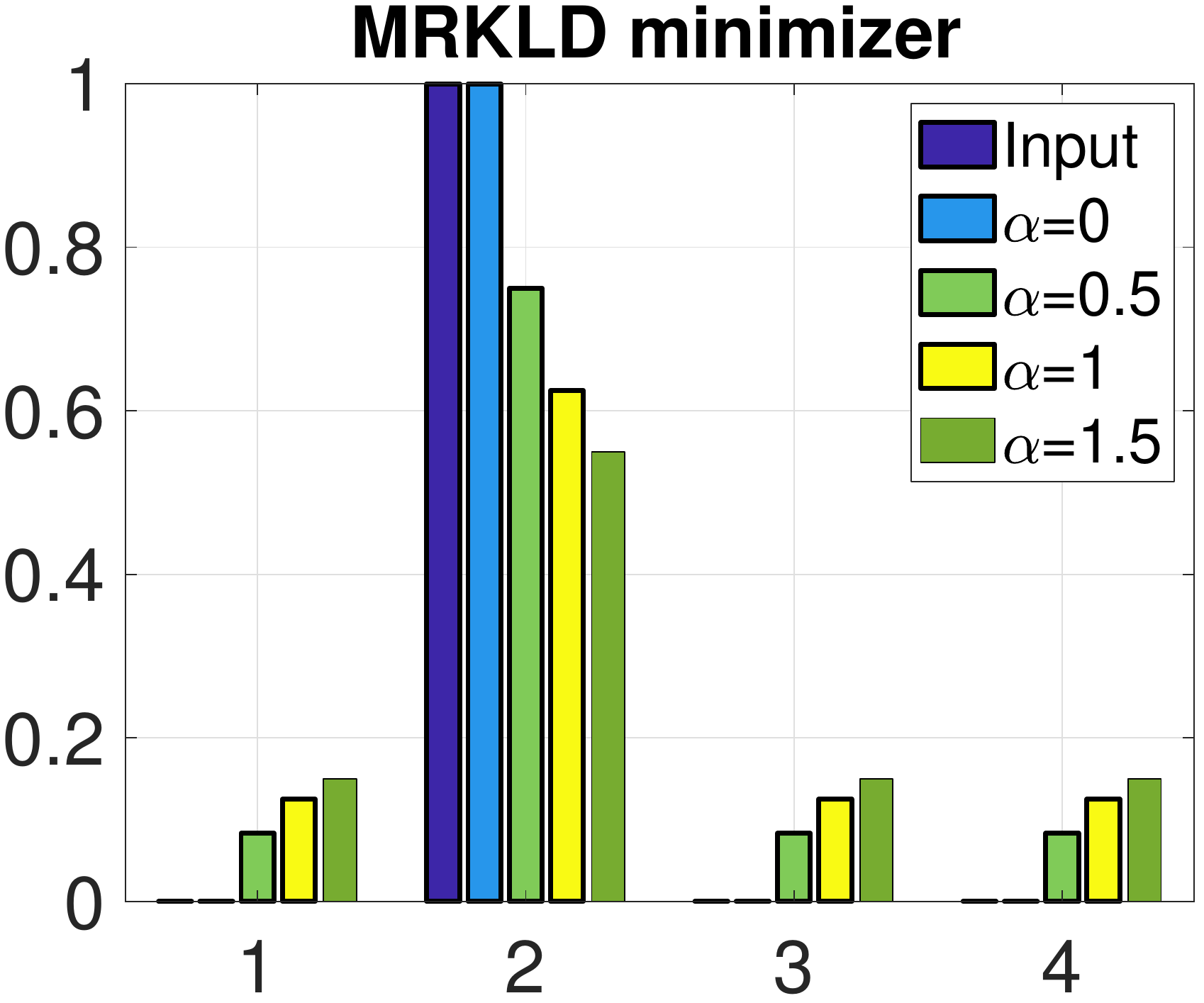}
	\vspace{-2mm}
	\caption{Minimizers of LRENT and MRKLD.}
	\label{fig:softmax_temp}
	\vspace{-3mm}
\end{figure}

\begin{table*}[!t]
	\centering
	\resizebox{0.75\textwidth}{!}{
	\begin{tabular}{c|ccccc|ccccc}
		\hline
		\multicolumn{11}{c}{W $\rightarrow$ A (Office-31)}\\
		\hline
 		& \multicolumn{5}{c|}{\textbf{MRL2}} & \multicolumn{5}{c}{\textbf{MRENT}}\\
		$p_0/\Delta p$ & 20/5 & 15/5 & 25/5 & 20/2.5 & 20/7.5 & 20/5 & 15/5 & 25/5 & 20/2.5 & 20/7.5\\
		Accuracy &  72.1$\pm$0.2 & 71.3$\pm$0.2 & 71.4$\pm$1.0 & 71.6$\pm$0.4 & 71.3$\pm$0.5 &  71.0$\pm$0.4 & 71.0$\pm$0.6 & 70.8$\pm$0.5 & 71.0$\pm$0.6 & 71.0$\pm$0.7\\
		\hline
		& \multicolumn{5}{c|}{\textbf{MRKLD}} &  \multicolumn{5}{c}{\textbf{LRENT}}\\
		$p_0/\Delta p$ & 20/5 & 15/5 & 25/5 & 20/2.5 & 20/7.5 & 20/5 & 15/5 & 25/5 & 20/2.5 & 20/7.5 \\
		Accuracy & 70.9$\pm$0.4 & 70.8$\pm$0.4 & 70.7$\pm$0.2 & 70.9$\pm$0.5 & 71.0$\pm$0.8 & 71.0$\pm$0.3 & 71.0$\pm$0.8 & 71.2$\pm$0.6 & 71.1$\pm$0.5 & 71.0$\pm$0.4 \\ \hline
	\end{tabular}
	}
	\vspace{-2mm}
	\caption{Sensitivity analysis of portion $p_0$ and portion step $\Delta p$.}
	\label{table:p}
	\vspace{-1mm}
\end{table*}

\begin{table*}[!t]
	\centering
	\resizebox{0.9\textwidth}{!}{
	\begin{tabular}{c|ccc|ccc|ccc|ccc}
		\hline
		\multicolumn{13}{c}{W $\rightarrow$ A (Office-31)}\\
		\hline
		& \multicolumn{3}{c|}{\textbf{MRL2}} & \multicolumn{3}{c|}{\textbf{MRENT}} & \multicolumn{3}{c|}{\textbf{MRKLD}} & \multicolumn{3}{c}{\textbf{LRENT}}\\
		$\alpha$ & 0.01 & 0.025 & 0.05 & 0.075 & 0.1 & 0.125 & 0.075 & 0.1 & 0.125 & 0.1 & 0.25 & 0.5\\
		Accuracy & 71.5$\pm$0.8 & 72.1$\pm$0.2 & 71.7$\pm$1.1 & 71.0$\pm$0.8 & 71.0$\pm$0.4 & 70.9$\pm$1.0 & 70.9$\pm$0.6 & 70.9$\pm$0.4 & 70.6$\pm$0.7  & 71.2$\pm$1.2 & 71.0$\pm$0.3 & 70.8$\pm$0.6\\
		\hline
	\end{tabular}
	}
	\vspace{-2mm}
	\caption{Sensitivity analysis of regularizer weight $\alpha$.}
	\label{table:alpha}
	\vspace{-1mm}
\end{table*}

\begin{table*}[!t]
	\centering
	\resizebox{\linewidth}{!}{
	\begin{tabular}{c|c|ccccccccccccccccccc|c}
		\hline
		& & Road & SW & Build & Wall & Fence & Pole & TL & TS & Veg. & Terrain & Sky & PR & Rider & Car & Truck & Bus & Train & Motor & Bike & mean\\
		\hline
		\multirow{3}{*}{CBST} & $C_{TP}$ ($\%$) & 96.2 & 86.0 & 94.6 & 83.8 & 84.9 & 84.5 & 80.4 & 78.0 & 93.9 & 87.9 & 94.5 & 90.4 & 81.4 & 95.4 & 88.4 & 85.9 & 59.5 & 78.5 & 80.6 & 85.5\\
		& $C_{FP}$ ($\%$) & 72.2 & 74.1 & 69.8 & 71.7 & 76.7 & 73.7 & 72.9 & 76.5 & 71.9 & 71.2 & 68.5 & 67.2 & 69.1 & 66.1 & 76.9 & 65.5 & 76.7 & 67.2 & 73.0 & 71.6\\
		& ${C_{TP}}/{C_{FP}}$ & 1.33 & 1.16 & 1.36 & 1.17 & 1.11 &  1.15 &  1.10 & 1.02 & 1.31 & 1.23 & 1.38 & 1.35 & 1.18 & 1.44 & 1.15 & 1.31 & 0.78 & 1.17 & 1.10 &  1.19\\
		\hline
		\multirow{3}{*}{MRKLD} & $C_{TP}$ ($\%$) & 94.7 & 82.8 & 92.4 & 81.7 & 77.8 & 84.4 & 77.0 & 76.4 & 93.4 & 86.5 & 94.4 & 88.8 & 79.7 & 93.9 & 87.0 & 84.9 & 71.9 & 77.6 & 79.2 & 84.5\\
		& $C_{FP}$ ($\%$) & 67.7 & 70.3 & 65.4 & 68.5 & 69.2 & 66.7 & 69.4 & 71.3 & 66.7 & 68.8 & 66.7 & 60.0 & 65.5 & 63.0 & 74.6 & 63.6 & 70.2 & 59.3 & 53.2 & 66.3\\
		& ${C_{TP}}/{C_{FP}}$ & 1.40 & 1.18 & 1.41 & 1.19 & 1.12 & 1.27 & 1.11 & 1.07 & 1.40 & 1.26 & 1.42 & 1.48 & 1.22 & 1.49 & 1.17 & 1.34 & 1.02 & 1.31 & 1.49 & 1.27\\
		\hline
		\multirow{3}{*}{LRENT} & $C_{TP}$ ($\%$) & 95.9 & 84.4 & 94.0 & 80.7 & 75.3 & 84.8 & 77.8 & 78.3 & 93.9 & 86.3 & 94.5 & 89.2 & 79.3 & 95.3 & 89.3 & 80.5 & 76.4 & 86.4 & 78.8  & 85.3\\
		& $C_{FP}$ ($\%$) & 69.5 & 72.1 & 68.0 & 67.8 & 71.3 & 69.7 & 71.5 & 75.4 & 69.5 & 69.9 & 70.1 & 64.1 & 67.6 & 67.3 & 77.7 & 70.3 & 63.4 & 58.6 & 55.2 & 68.4\\
		& ${C_{TP}}/{C_{FP}}$ & 1.38 & 1.17 & 1.38 & 1.19 & 1.06 & 1.22 & 1.09 & 1.04 & 1.35 & 1.23 & 1.35 & 1.39 & 1.17 & 1.42 & 1.15 & 1.15 & 1.2 & 1.47 & 1.43 & 1.25\\
		\hline
	\end{tabular}
	}
	\vspace{-2mm}
	\caption{Comparison of $C_{TP}$, $C_{FP}$ and ${C_{TP}}/{C_{FP}}$ on GTA5 $\rightarrow$ Cityscapes.}\label{ConfComp}
	\vspace{-1mm}
\end{table*}

\subsection{MR versus LR}\label{mr-lr}
We analyze MR/LR intuitively and theoretically to give suggestions for practical choice of confidence regularizers.\\
\noindent\textbf{Complexity analysis:} All model regularizers only introduce negligible extra costs for the gradient computation. Label regularizers, however, requires the storage of dataset-level soft pseudo-labels. This does not present an issue in image classification but may introduce extra I/O costs in segmentation, where labels are often too large to be stored in memory and need to be written to disk.\\
\noindent\textbf{Loss curves:} To further illustrate the different properties of regularizers, we visualize how they influence the original loss surfaces by reducing the problem into binary classification with a single sample. We assume a cross-entropy loss $-y\log p-(1-y)\log(1-p)$ plus an MR/LR weighted by $\alpha$. For MRs, we assume $y=1$ and illustrate the regularized loss curves versus $p$ in Fig. \ref{fig:loss_curves}. For all MRs, $p^{*}$ becomes smoother when $\alpha$ increases. We notice that MRKLD serves as a better barrier to prevent sharp outputs than other MRs by having steeper gradient near $p=1$. This accords with our observation that MRKLD overall works the best.
For LRENT, we assume $p=0.9$ and illustrate the regularized loss curves versus $y$ at different $\alpha$ in Fig. \ref{fig:loss_curves}. Again, $y^{*}$ becomes smoother when $\alpha$ increases.\\
\noindent\textbf{Class ranking:} Based on the closed-form solution of LR in Table \ref{table:regs}, we can prove that LR preserves the confidence ranking order between classes. On the other hand, given one-hot labels, MRs tend to discard such order information by giving equal confidences to negative classes. Taking MRKLD as example: using Lagrangian multiplier, we can prove the closed-form global minimizer for regularized cross-entropy loss as $p^{*(k)} = (y^{(k)} + {\alpha \over K}) / (1+\alpha)$, where $k=1,...,K$ is class index. With $\textbf{y}$ being one-hot, the global minimizer is uniformly smoothed on negative classes. Similar property can be also proved for MRENT/MRL2.

We illustrate two examples of LRENT and MRKLD in Fig.~\ref{fig:softmax_temp}, where we assume $\mathbf{p}=[0.2,0.1,0.55,0.15]$ for LRENT and $y^{(2)}=1$ for MRKLD. One can see, LRENT sharpens the input $\mathbf{p}$ when $\alpha\in[0, 1]$ (one-hot when $\alpha=0$), while smooths $\mathbf{p}$ when $\alpha>1$. In all cases, the inter-class confidence orders are always preserved, while the same property does not hold for MRKLD.\\
\noindent\textbf{MR+LR:} The combination of MR and LR can take advantages of both regularizers and achieve better performance compared to single regularizer, demonstrated in VisDA17 and Office-31. However, it will also introduce extra cost to validate both hyperparameters for MR and LR.\\
\noindent\textbf{Practical suggestions:} Overall, we recommend CRST-MRKLD most based on the above analysis and its better performance. Moreover, combining MR and LR may also benefit self-training at the cost of slight extra tuning.

\section{Conclusions}\label{sec:conclusion}
In this paper, we introduce a confidence regularized self-training framework formulated as regularized self-training loss minimization. Model regularization and label regularization are considered with a family of proposed confidence regularizers. We investigate theoretical properties of CRST, including its probabilistic explanation and connection to softmax with temperature. Comprehensive experiments demonstrate the effectiveness of CRST with state-of-the-art performance. We also systematically discuss the pros and cons of the proposed regularizers and made practical suggestions. We believe this work can inspire future research on novel designs of regularizations as desired inductive biases to benefit many UDA/SSL problems.

{\small
\bibliographystyle{ieee_fullname}
\bibliography{ref}
}

\clearpage

\section*{Appendix}
In this appendix, we present the additional details and results that are not covered by the main paper.

\setcounter{section}{0}
\renewcommand{\thesection}{\Alph{section}}

\section{Derivation of soft pseudo-label in LRENT}\label{sec:derivlrent}
For entropy label regularizer, the soft pseudo-label learning problem is defined as follows.
\begin{align}\label{label-ent-opt}
\underset{\hat{\mathbf{y}}_t}{\mathop{\min }}&{\sum\limits_{k=1}^{K}{-\hat{y}_{t}^{(k)}}\log \frac{p(k|{\mathbf{x}_{t}};\mathbf{w})}{\lambda_k}+\alpha \sum\limits_{k=1}^{K}{\hat{y}_t^{(k)}}{\log( \hat{y}_t^{(k)}})}\notag\\
\text{s.t.}~&~\hat{\mathbf{y}}_t\in\Delta^{(K-1)}
\end{align}
where the solution is given as below.
\begin{displaymath}
\hat{y}_{t}^{(i)\dag}=\frac{{({\frac{p(i|{\mathbf{x}_{t}})}{\lambda_k}})^{\frac{1}{\alpha }}}}{\sum\limits_{k=1}^{K}{{({\frac{p(k|{\mathbf{x}_{t}})}{\lambda_k}})^{\frac{1}{\alpha }}}}}
\end{displaymath}

It is easy to see that the optimization in (\ref{label-ent-opt}) is a convex problem. Therefore, the global optimum can be found with a Lagrangian multiplier~\cite{boyd2004convex} defined as follows:
\begin{align}
L\left( \hat{y},\beta  \right) & =\sum\limits_{k=1}^{K}{-\hat{y}_{t}^{(k)}}\log \frac{p(k|{\mathbf{x}_{t}};\mathbf{w})}{\lambda_k}\notag\\
&+\alpha \sum\limits_{k=1}^{K}{\hat{y}_{t}^{(k)}(\log (\hat{y}_{t}^{(k)})-1)}+\beta (\sum\limits_{k=1}^{K}{\hat{y}_{t}^{(k)}}-1)\notag
\end{align}
Setting the corresponding gradients equals to $0$ gives the global optimum ($k=1,...,K$).
\begin{displaymath}
\begin{split}
&~~~~~\left\{\begin{matrix}
\frac{\partial L}{\partial \hat{y}_{t}^{(i)\dag}}=-\log \frac{p(k|{\mathbf{x}_{t}};\mathbf{w})}{\lambda_i}+\alpha \log \hat{y}_{t}^{(i)\dag}+\beta =0;\text{ }  \\
\sum\limits_{k=1}^{K}{\hat{y}_{t}^{(k)\dag}}=1  \\
\end{matrix} \right. \\ 
& \Leftrightarrow \left\{ \begin{matrix}
\hat{y}_{t}^{(i)\dag}=\exp (\frac{-\beta }{\alpha })({\frac{p(i|{\mathbf{x}_{t}};\mathbf{w})}{\lambda_i})^{\frac{1}{\alpha }}};\text{ }  \\
\sum\limits_{i=1}^{K}{\hat{y}_{t}^{(i)\dag}}=1  \\
\end{matrix} \right. \\ 
& \Leftrightarrow \left\{ \begin{matrix}
\hat{y}_{t}^{(i)\dag}=\exp (\frac{-\beta }{\alpha })({\frac{p(i|{\mathbf{x}_{t}};\mathbf{w})}{\lambda_i})^{\frac{1}{\alpha }}};\text{ }  \\
\sum\limits_{i=1}^{K}{\exp (\frac{-\beta }{\alpha })({\frac{p(k|{\mathbf{x}_{t}};\mathbf{w})}{\lambda_k})^{\frac{1}{\alpha }}}}=1  \\
\end{matrix} \right. \\ 
& \Leftrightarrow \left\{ \begin{matrix}
\hat{y}_{t}^{(i)\dag}=\exp (\frac{-\beta }{\alpha })({\frac{p(i|{\mathbf{x}_{t}};\mathbf{w})}{\lambda_i})^{\frac{1}{\alpha }}};\text{ }  \\
\exp (\frac{-\beta }{\alpha })=\frac{1}{\sum\limits_{k=1}^{K}{({\frac{p(k|{\mathbf{x}_{t}};\mathbf{w})}{\lambda_k})^{\frac{1}{\alpha }}}}}  \\
\end{matrix} \right. \\ 
& \Leftrightarrow \left\{ \begin{matrix}
\hat{y}_{t}^{(i)\dag}=\frac{(\frac{p(i|{{x}_{t}};\mathbf{w})}{\lambda_i})^{{\frac{1}{\alpha }}}}{\sum\limits_{k=1}^{K}(\frac{p(k|{{x}_{t}};\mathbf{w})}{\lambda_k})^{\frac{1}{\alpha}}};\text{ }   \\
\beta =\alpha \log \sum\limits_{k=1}^{K}{(\frac{p{(k|{\mathbf{x}_{t}};\mathbf{w})}}{\lambda_k})^{\frac{1}{\alpha}}}  \\
\end{matrix} \right.
\end{split}
\end{displaymath}

\section{Theoretical properties of CRSTs}
\subsection{Proof of Proposition \ref{prop:cem}}\label{sec:proofprop1}
Classification maximum likelihood (CML) was initially proposed to model clustering tasks, and can be optimized via classification expectation maximization (CEM). Compared with traditional expectation maximization (EM) that has an ``expectation'' (E) step and a ``maximization'' (M) step, CEM has an additional ``classification'' (C) step (between E and M steps) that assigns a sample to the cluster with maximal posterior probability. In~\cite{amini2002semi}, CML is generalized to discriminant semi-supervised learning with both labeled and unlabeled data defined as follows:
\begin{displaymath}
	\label{cml}
	{\log{\mathcal{L}}_{C}} = {\log\tilde{\mathcal{L}}_{C}}  +\sum\limits_{{{i}}\in {{S,T}}}\log p(\mathbf{x}_i)
\end{displaymath}
where:
\begin{displaymath}
	\label{ecml}
	\begin{split}
		{\log\tilde{\mathcal{L}}_{C}} = &\sum\limits_{{{s}}\in {{S}}}{\sum\limits_{k=1}^{K}{y_{s}^{(k)}}\log p(k|{\mathbf{x}_{s}};\mathbf{w})} +\sum\limits_{{{t}}\in {{T}}}{\sum\limits_{k=1}^{K}{\hat{y}_{t}^{(k)}}\log p(k|{\mathbf{x}_{t}};\mathbf{w})}
	\end{split}
\end{displaymath}
Note that ${{{\hat{y}}}_{t}}\in\{0,1\}^K, \forall t$. $p(k|\mathbf{x}_t;\mathbf{w})$ is the posterior probability modeled by classifiers such as logistic classifier and neural network and $\mathbf{w}$ is the learnable weight. \cite{amini2002semi} uses a discriminant classifier which makes no assumptions about the data distribution $p(\mathbf{x}_t)$. Thus maximizing (\ref{cml}) is equal to maximizing (\ref{ecml}). Below we draw the connection of the CRST self-training algorithm to CEM. We first show that CRST can be rewritten as the following regularized classification maximum likelihood model:
\begin{displaymath}\label{rcml}
\begin{split}
	\underset{\mathbf{w},\hat{\mathbf{Y}}_T}{\mathop{\max }}\,	
	&\sum\limits_{s\in S}\sum\limits_{k=1}^K y_s^{(k)}\log p(k|{\mathbf{x}_{s}};\mathbf{w})
	+\sum\limits_{t\in T}\sum\limits_{k=1}^{K}\hat{y}_{t}^{(k)}\log p(k|{\mathbf{x}_{t}};\mathbf{w})\\
	& -\sum\limits_{t\in T}\Big[\sum\limits_{k=1}^{K}\hat{y}_{t}^{(k)}\log\lambda_k+	
	\alpha r_c(\mathbf{w},\hat{\mathbf{y}}_t)\Big]\\
	& ={\log\mathcal{\tilde{L}}_{C}}+\mathcal{R}_C\\
	\text{s.t.}~&~\mathbf{\hat{y}}_t\in~\Delta^{(K-1)}\cup\{\mathbf{0}\},~\forall t
	\end{split}
\end{displaymath}
where the above problem contains an additional regularizer term ($\mathcal{R}_C$) compared with CML, defined as:
\begin{displaymath}\label{rcml_reg}
\mathcal{R}_C = -\sum\limits_{t\in T}\Big[\sum\limits_{k=1}^{K}\hat{y}_{t}^{(k)}\log\lambda_k+	
\alpha r_c(\mathbf{w},\hat{\mathbf{y}}_t)\Big]
\end{displaymath}
In addition, the corresponding alternative self-training optimization can be written as the following CEM process:

\noindent\textbf{E-Step:} Given the model weight $\mathbf{w}$, estimate the posterior probability $p(\mathbf{x}_t;\mathbf{w}),\forall t$.

\noindent\textbf{C-Step:} Fix $\mathbf{w}$ and solve the following problem for $\hat{\mathbf{Y}}_T$:
\begin{displaymath}\label{rcml_a}
	\begin{split}
	\underset{{{{\hat{\mathbf{Y}}}}_{T}}}{\mathop{\max }}\,
	&\sum\limits_{t\in T}\Big[\sum\limits_{k=1}^{K}{\hat{y}_{t}^{(k)}}\log p(k|{\mathbf{x}_{t}};\mathbf{w})
	-\alpha r_c(\mathbf{w},\hat{\mathbf{y}}_t)\Big]\\
	\text{s.t.}~&~\hat{\mathbf{y}}_t\in~\Delta^{(K-1)}\cup\{\mathbf{0}\},~\forall t
	\end{split}
\end{displaymath}

\noindent\textbf{M-Step:} Fix $\hat{\mathbf{Y}}_T$ and use gradient ascent to solve the following problem for $\mathbf{w}$.
\begin{displaymath}\label{rcml_b}
	\begin{split}
	\underset{{{{\mathbf{w}}}}}{\mathop{\max }}\,
	&\sum\limits_{{{s}}\in {{S}}}{\sum\limits_{k=1}^{K}{y_{s}^{(k)}}\log p(k|{\mathbf{x}_{s}};\mathbf{w})}\\
	&+\sum\limits_{{{t}}\in {{T}}}\Big[{\sum\limits_{k=1}^{K}{\hat{y}_{t}^{(k)}}\log p(k|{\mathbf{x}_{t}};\mathbf{w})}
	-\alpha r_c(\mathbf{w},\hat{\mathbf{y}}_t)\Big]
	\end{split}
\end{displaymath}
We have thus shown that the CRST self-training algorithm is an instance of CEM.

\subsection{Proof of Proposition \ref{prop:convergence}}\label{sec:proofprop2}
As a brief recap, the general form of CRST in (\ref{crst}) can be optimized via the following two steps:

\noindent \textbf{a) Pseudo-label learning} \label{a)} Fix $\mathbf{w}$ and solve:
\begin{equation}
\begin{split}
\label{crst_a}
&\underset{\hat{\mathbf{Y}}_T}{\mathop{\min }}~{\sum\limits_{t\in T}\sum\limits_{k=1}^{K}-{\hat{y}_{t}^{(k)}}\log \frac{p(k|{\mathbf{{x}}_{t}};\mathbf{w})}{\lambda_k}+\alpha r_c(\mathbf{w},\hat{\mathbf{y}}_t)}\\
&~\text{s.t.}~~\hat{\mathbf{y}}_t\in\Delta^{(K-1)}\cup \{\mathbf{0}\},\forall t
\end{split}
\end{equation}
which leads to the following solver for each $\hat{\mathbf{y}}_t$:
\begin{equation}\label{crst_solver}
\hat{\mathbf{y}}_{t}^*=\left\{
\begin{aligned}
\hat{\mathbf{y}}_{t}^\dag, &~~\text{if}~~\mathcal{C}(\hat{\mathbf{y}}_t^\dag)<\mathcal{C}(\mathbf{0})\\
\mathbf{0}~~, &~~\text{otherwise}
\end{aligned}
\right.
\end{equation}
where $\mathbf{y}_{t}^\dag$ is the minimizer of (\ref{crst_a}) with the feasible set being $\Delta^{K-1}$ only, and $\mathcal{C}(\hat{\mathbf{y}}_t)$ is defined as:
\begin{displaymath}\label{plcro_0}
\mathcal{C}(\hat{\mathbf{y}}_t)= -\hat{y}_{t}^{(k)} \sum\limits_{k=1}^{K}\log \frac{p(k|{\mathbf{x}_{t}};\mathbf{w})}{\lambda_k} + \alpha r_c(\mathbf{w},\hat{\mathbf{y}}_t)
\end{displaymath}

\noindent\textbf{b) Network retraining} \label{b)} ~ Fix $\hat{\mathbf{Y}}_T$ and solve the following optimization by gradient descent:
\begin{equation}\label{crst_b}
\begin{split}
\underset{\mathbf{w}}{\mathop{\min }}& -\sum\limits_{{{s}}\in {{S}}}{\sum\limits_{k=1}^{K}{y_{s}^{(k)}}\log (p(k|{\mathbf{x}_{s}};\mathbf{w}))}\\
& -\sum\limits_{{{t}}\in {{T}}}[{\sum\limits_{k=1}^{K}{\hat{y}_{t}^{(k)}}\log (p(k|{\mathbf{x}_{t}};\mathbf{w}))-\alpha r_c(\mathbf{w},\hat{\mathbf{y}}_t)]}
\end{split}
\end{equation}

We assume $\alpha \ge 0$, and $r_c(\mathbf{w},\hat{\mathbf{y}}_t)$ is convex w.r.t. $\mathbf{w}$ and $\hat{\mathbf{y}}_t$ given the listed regularizers in Table \ref{table:regs}. Note that the definition and optimization of continuous CBST is simply a special case of CRST with $\alpha = 0$. Therefore, the convergence of CRST also indicates the convergence of CBST. With the above preliminaries, we have:

\noindent\textbf{Step a) is non-increasing:} (\ref{crst_solver}) is obtained by decomposing (\ref{crst_a}) into two subproblems with feasible sets being $\Delta^{K-1}$ and $\textbf{0}$, respectively. The former is a convex problems which gives a globally optimal solution, while (\ref{crst_solver}) is the result of comparing this solution against $\textbf{0}$ by taking the one with a smaller cost. As a result, (\ref{crst_solver}) is also a global minimizer and (\ref{crst_a}) is guaranteed to be non-increasing.

\noindent\textbf{Step b) is non-increasing:} One may use gradient descent to minimize the loss in (\ref{crst_b}). With a proper learning rate, the loss is guaranteed to decrease monotonically. In practice, network re-training is often done with mini-batch gradient descent instead of gradient descent. This may not strictly guarantee the monotonic decrease of the loss, but will almost certainly converge to a lower one.

One can prove that the self-training loss in (\ref{crst}) is lower bounded. Therefore, the optimization of (\ref{crst}) by alternatively taking step \textbf{a)} and \textbf{b)} is convergent.

\subsection{Proof of Proposition \ref{prop:prop4}}\label{sec:proofprop4}
As mentioned in~\cite{szegedy2016rethinking}, uniformly smoothed pseudo-label $\hat{\mathbf{y}}_t$ with $\epsilon=(K\alpha-\alpha)/(K+K\alpha)$ is
\begin{equation}\label{smooth_solver}
	\tilde{y}_{t}^{(k)}=\left\{
	\begin{aligned}
	1-\frac{K\alpha-\alpha}{K+K\alpha}, &~\text{if}~k=\argmax_{k}\{ {\hat{\mathbf{y}}_t}\}\\
	\frac{\alpha}{K+K\alpha}, &~\mathrm{otherwise}
	\end{aligned}
	\right.
\end{equation}
And the self-training with uniformaly smoothed pseudo-labels is defined as follows.
\begin{equation}\label{label_smooth}
	\begin{split}
	& \underset{\mathbf{w}}{\mathop{\min }}-\frac{1}{1+\alpha}\sum\limits_{{{s}}\in {{S}}}{\sum\limits_{k=1}^{K}{y_{s}^{(k)}}\log p(k|{\mathbf{x}_{s}};\mathbf{w})}\\
	&~~~~~~ -\sum\limits_{{{t}}\in {{T}}}[{\sum\limits_{k=1}^{K}{\tilde{y}_{t}^{(k)}}\log p(k|{\mathbf{x}_{t}};\mathbf{w})]}
	\end{split}
\end{equation}
where $\hat{\mathbf{y}}_t$ follows (\ref{smooth_solver}). 
	
In KLD model regularized self-training, the model retraining needs to optimize the following problem:
\begin{align}\label{kld_opt}
	& \underset{\mathbf{w}}{\mathop{\min }}-\sum\limits_{{{s}}\in {{S}}}{\sum\limits_{k=1}^{K}{y_{s}^{(k)}}\log p(k|{\mathbf{x}_{s}};\mathbf{w})}\\
	&~~~~~~~ -\sum\limits_{{{t}}\in {{T}}}[{\sum\limits_{k=1}^{K}{\hat{y}_{t}^{(k)}}\log p(k|{\mathbf{x}_{t}};\mathbf{w})+\frac{\alpha}{K}\log p(k|\mathbf{x}_t;\mathbf{w})]}\notag
\end{align}
where $\hat{\mathbf{y}}_t, \forall t$ are the fixed pseudo-labels and $\alpha$ is the regularizer weight. Here, we show the equivalence of the above two problems with the following proof:
\begin{displaymath}
	\begin{split}
	&~~~~~~ \underset{\mathbf{w}}{\mathop{\min }}-\sum\limits_{{{s}}\in {{S}}}{\sum\limits_{k=1}^{K}{y_{s}^{(k)}}\log p(k|{\mathbf{x}_{s}};\mathbf{w})}\\
	&~~~~~ -\sum\limits_{{{t}}\in {{T}}}\bigg[~{\sum\limits_{k=1}^{K}{\hat{y}_{t}^{(k)}}\log p(k|{\mathbf{x}_{t}};\mathbf{w}) +\frac{\alpha}{K}\log p(k|\mathbf{x}_t;\mathbf{w})\bigg]}\\
	& \Leftrightarrow  \underset{\mathbf{w}}{\mathop{\min }}-\sum\limits_{{{s}}\in {{S}}}{\sum\limits_{k=1}^{K}{y_{s}^{(k)}}\log p(k|{\mathbf{x}_{s}};\mathbf{w})}\\
	&~~~~~~~~~~~~~ -\sum\limits_{{{t}}\in {{T}}}\bigg[~{\sum\limits_{k=1}^{K}{(\hat{y}_{t}^{(k)}+\frac{\alpha}{K})}\log p(k|{\mathbf{x}_{t}};\mathbf{w})\bigg]}\\
	& \Leftrightarrow  \underset{\mathbf{w}}{\mathop{\min }}\,
	-\frac{1}{1+\alpha}\sum\limits_{{{s}}\in {{S}}}{\sum\limits_{k=1}^{K}{y_{s}^{(k)}}\log (p(k|{\mathbf{x}_{s}};\mathbf{w}))}\\
	&~~~~~~~~~~~~~ -\sum\limits_{{{t}}\in {{T}}}\bigg[~{\sum\limits_{k=1}^{K}{\frac{(K\hat{y}_{t}^{(k)}+\alpha)}{K+K\alpha}}\log p(k|{\mathbf{x}_{t}};\mathbf{w})\bigg]}
	\end{split}
\end{displaymath}
Replacing $\hat{\mathbf{y}}_t$ with a one-hot completes the proof.

\subsection{Proof of Proposition \ref{prop:prop5}}\label{sec:proofprop5}
In MRENT, the model retraining needs to optimize the following problem:
\begin{equation}\label{mrent}
	\begin{split}
	& \underset{\mathbf{w}}{\mathop{\min }}-\sum\limits_{{{s}}\in {{S}}}{\sum\limits_{k=1}^{K}{y_{s}^{(k)}}\log p(k|{\mathbf{x}_{s}};\mathbf{w})}\\
	&~~~~~~~ -\sum\limits_{{{t}}\in {{T}}}\bigg[~\sum\limits_{k=1}^{K}{\hat{y}_{t}^{(k)}}\log p(k|{\mathbf{x}_{t}};\mathbf{w})\\
	&~~~~~~~ -p(k|\mathbf{x}_t;\mathbf{w})\log p(k|\mathbf{x}_t;\mathbf{w})\bigg]
	\end{split}
\end{equation}
	
\noindent We will show the above problem is equivalent to the model retraining in the reverse KLD model regularized self-training, which is defined as follows.
\begin{align}\label{mrkld_rev}
	&\underset{\mathbf{w}}{\mathop{\min}}-\sum\limits_{{{s}}\in {{S}}}{\sum\limits_{k=1}^{K}{y_{s}^{(k)}}\log p(k|{\mathbf{x}_{s}};\mathbf{w})}\\
	&~~~~~~\, -\sum\limits_{{{t}}\in {{T}}}\bigg[\sum\limits_{k=1}^{K}{\hat{y}_{t}^{(k)}}\log p(k|{\mathbf{x}_{t}};\mathbf{w})+D_{KL}(p(\mathbf{x}_t)||\mathbf{u})\bigg]\notag
\end{align}
To prove the above equivalence, we have the following. 
\begin{equation}\label{kld_rev}
	\begin{split}
	& D_{KL}(p(\mathbf{x}_t)||\textbf{u})=-\sum\limits_{k=1}^{K}p(k|{\mathbf{x}_{t}})\log\frac{1/K}{p(k|{\mathbf{x}_{t}})}\\
	&~~~~~~~~~~~~~~~~~~~~~~~~~ =\log K+\sum\limits_{k=1}^{K}p(k|{\mathbf{x}_{t}})\log{p(k|{\mathbf{x}_{t}})}
	\end{split}
\end{equation}
In (\ref{kld_rev}), $K$ is a constant. Thus one can prove that the minimization in (\ref{mrent}) is equivalent to the minimization in (\ref{mrkld_rev}).

\section{Additional details on experiments}\label{sec:addexp}
\subsection{Accuracy curves}
\begin{figure}[!b]
	\centering
	\includegraphics[width=0.85\linewidth]{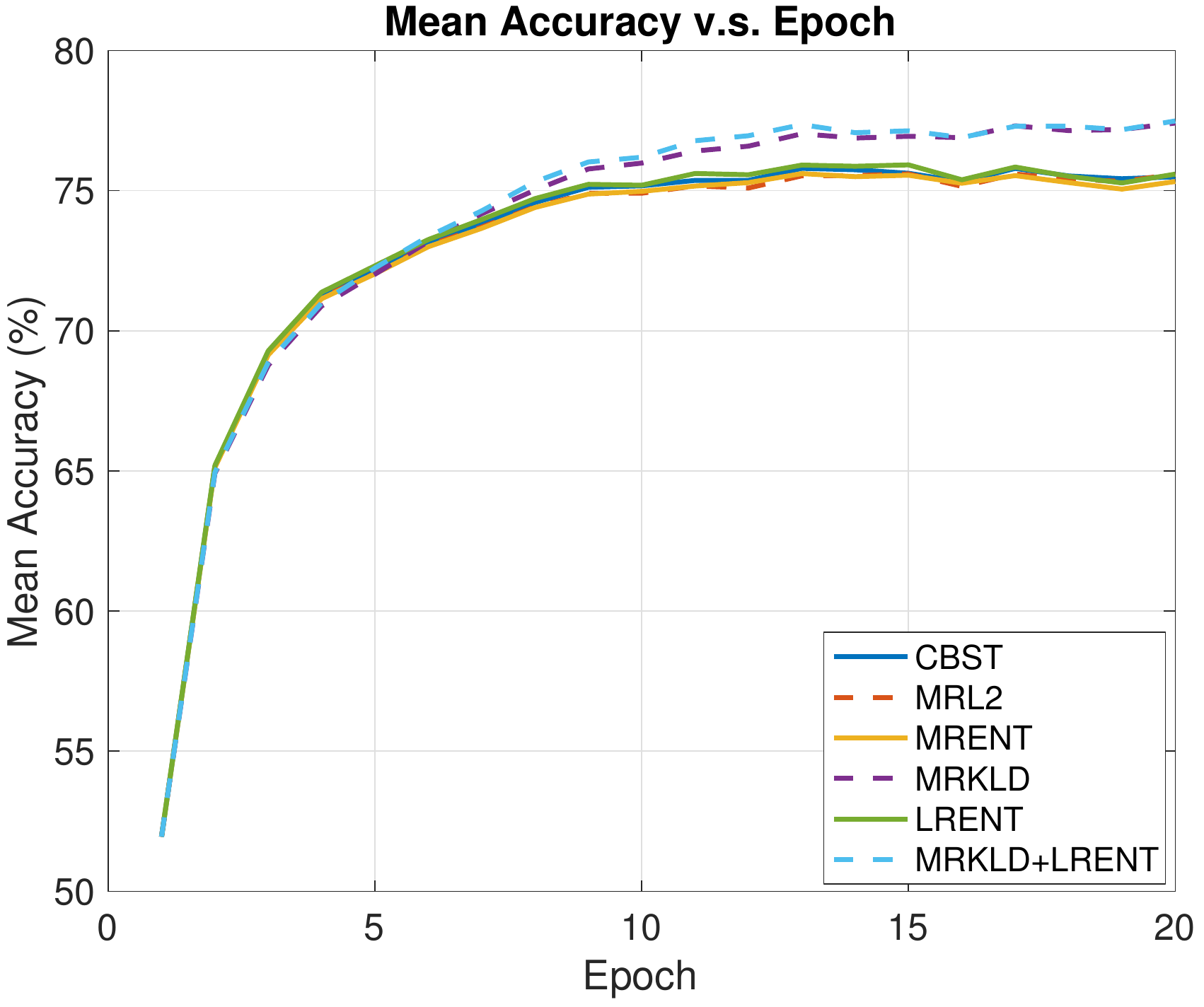}
	\caption{Mean accuracy versus number of epochs.}
	\label{learning_curves}
\end{figure}

To show the learning behaviors on VisDA17, we plot the curves of mean accuracy (averaged over 5 runs) versus epochs for CBST and CRSTs in Fig. \ref{learning_curves}. One can see, the proposed self-training methods are generally stable with only slight fluctuations after $10$ epochs. Among all comparing methods, MRKLD+LRENT gives the best performance and shows consistent improvement over the CBST baseline.

\begin{figure*}[!t]
	\centering
	\resizebox{0.94\textwidth}{!}{
		\begin{tabular}{@{}cccccccccccc@{}}
			\cellcolor{visda_color_1}{~~aero~~} &
			\cellcolor{visda_color_2}~~bike~~&
			\cellcolor{visda_color_3}{~~bus~~} &
			\cellcolor{visda_color_4}{~~car~~} &
			\cellcolor{visda_color_5}~~horse~~ &
			\cellcolor{visda_color_6}~~knife~~ &
			\cellcolor{visda_color_7}\textcolor{white}{~~motor~~} &
			\cellcolor{visda_color_8}\textcolor{white}{~~person~~} &
			\cellcolor{visda_color_9}\textcolor{white}{~~plant~~} & 
			\cellcolor{visda_color_0}\textcolor{white}{~~board~~}
			\cellcolor{visda_color_10}~~train~~ &
			\cellcolor{visda_color_11}~~truck~~ &
		\end{tabular}
	}
	\vspace{1mm}
	\includegraphics[width=0.26\textwidth]{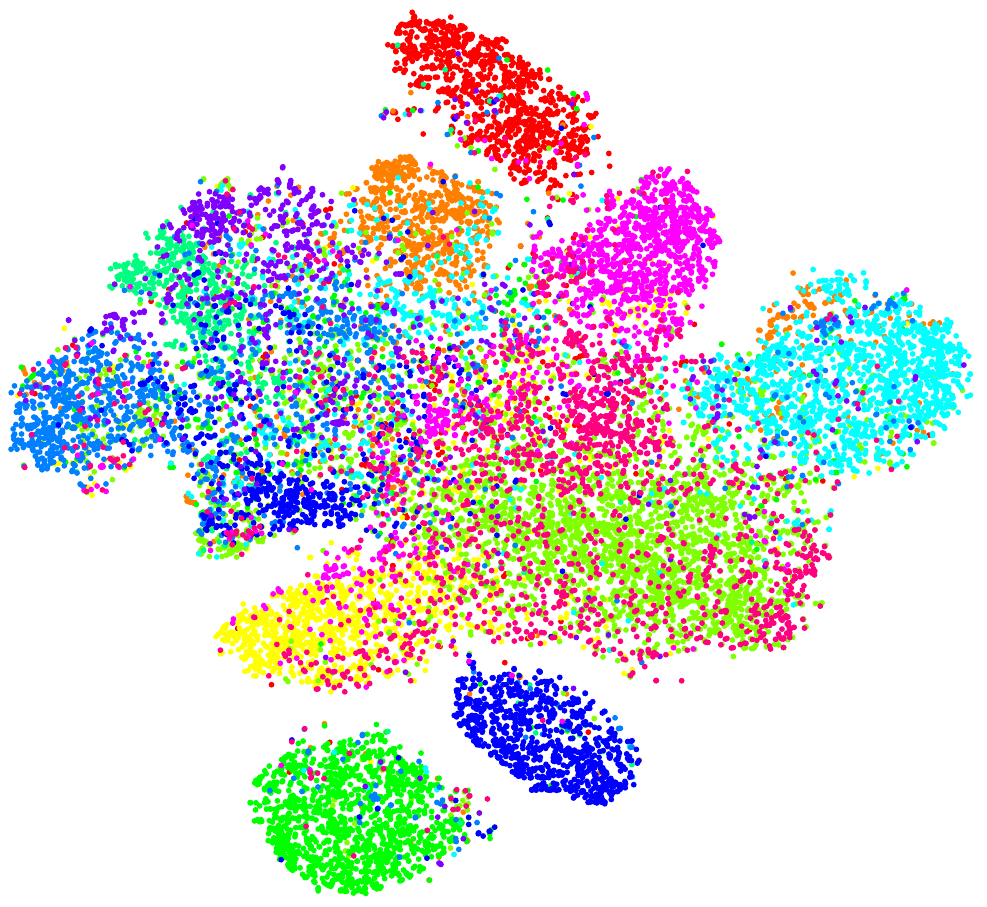}~~~~~~
	\includegraphics[width=0.26\textwidth]{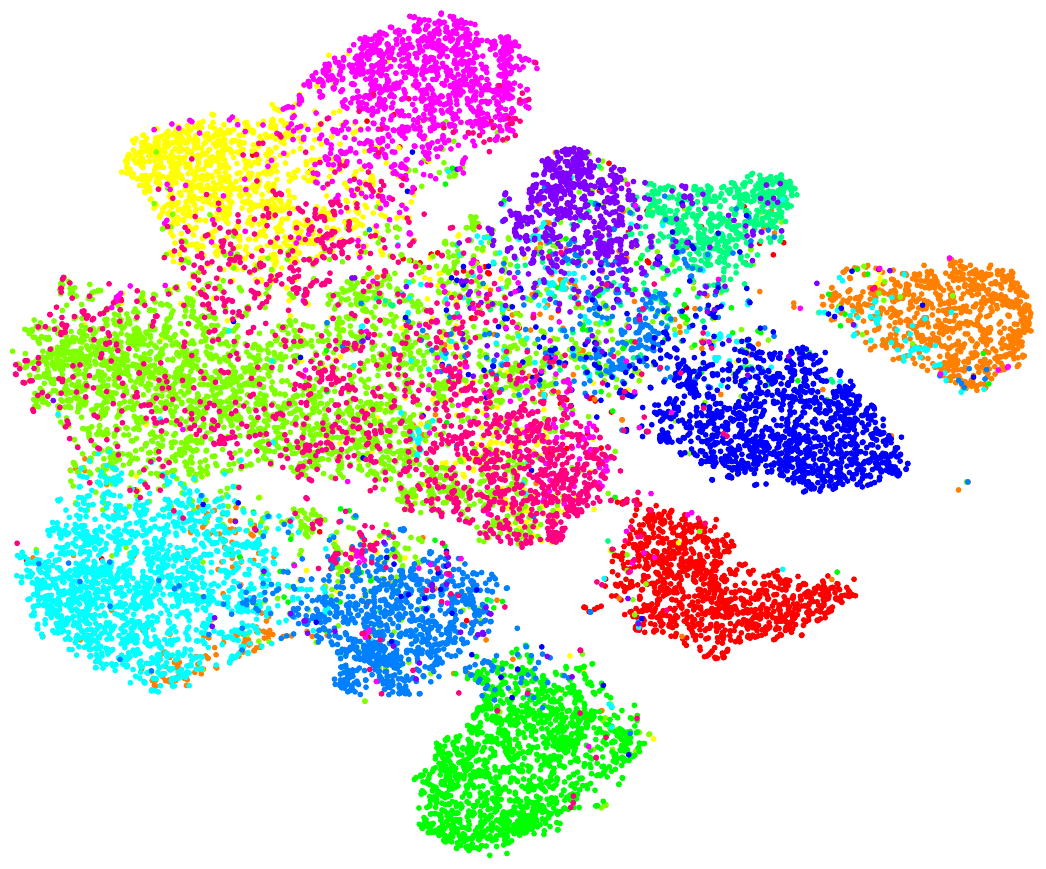}~~~~~~
	\includegraphics[width=0.26\textwidth]{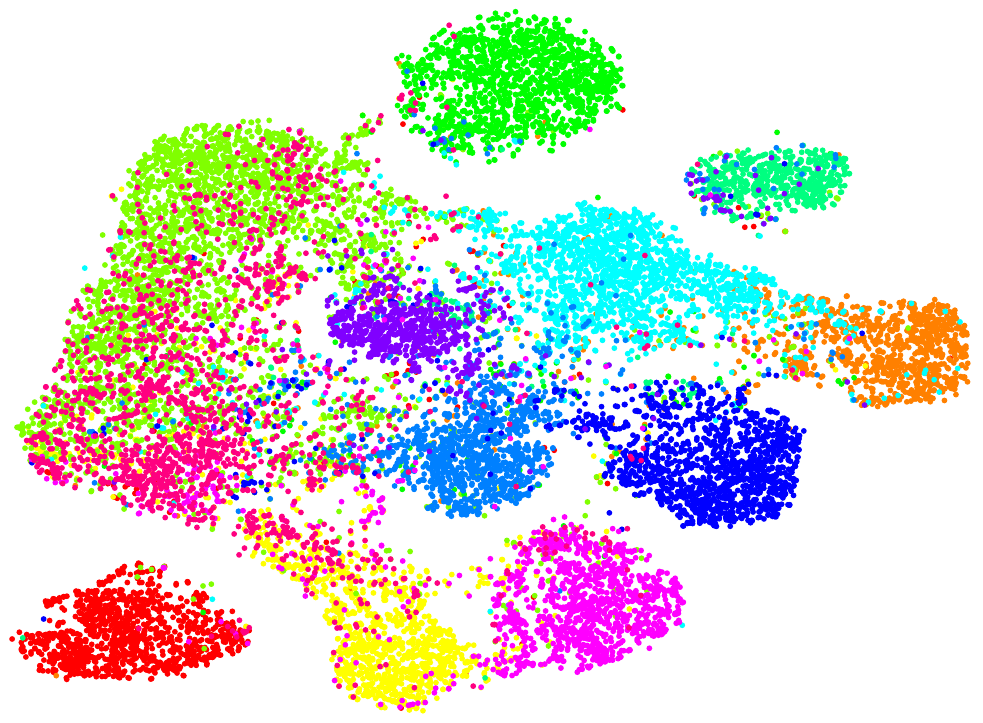}
	\caption{Feature visualization for target domain of VisDA17. From left to right: Source model, CBST, MRKLD+LRENT.}\label{fig:feat_vis}
\end{figure*}

\begin{figure*}[!t]
	\centering
	\includegraphics[height=0.3\linewidth]{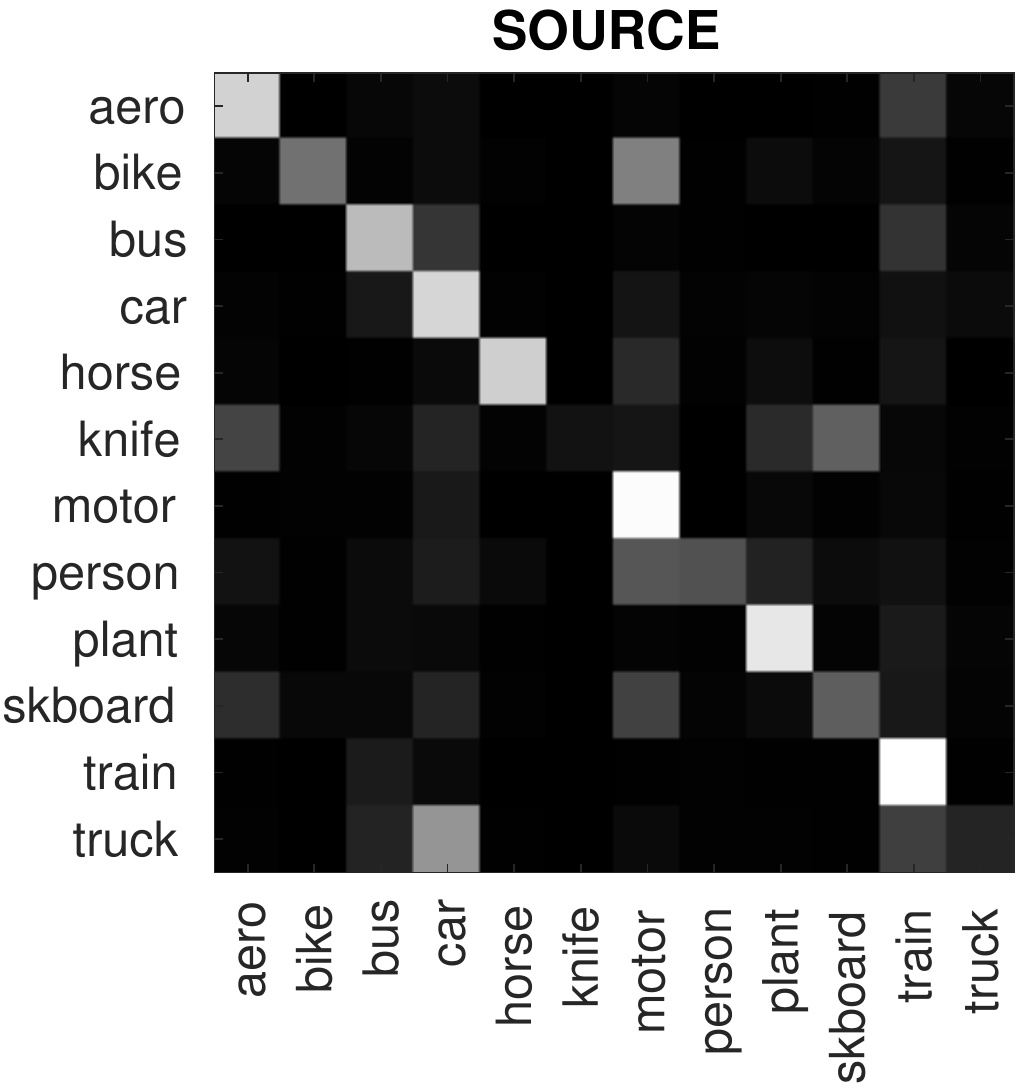}~~
	\includegraphics[height=0.3\linewidth]{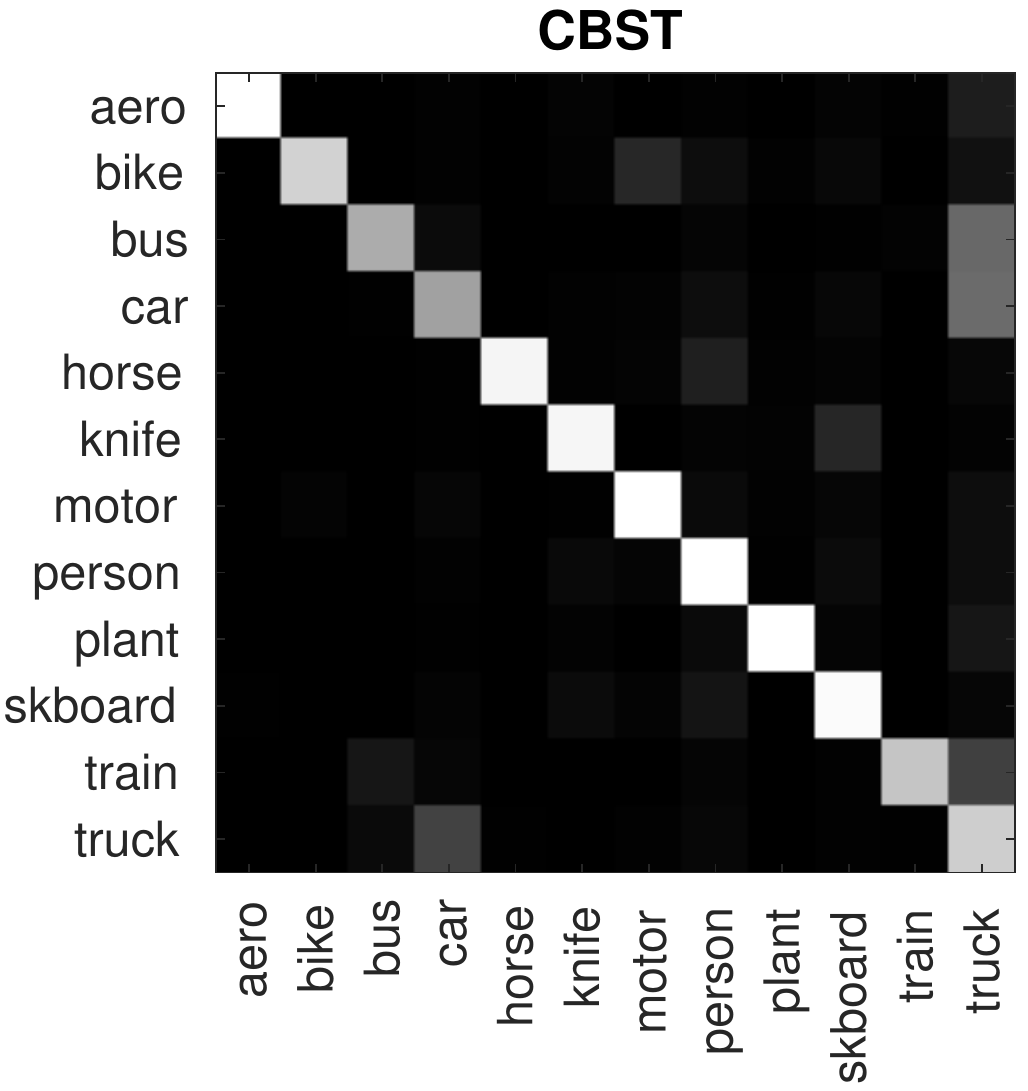}~~
	\includegraphics[height=0.3\linewidth]{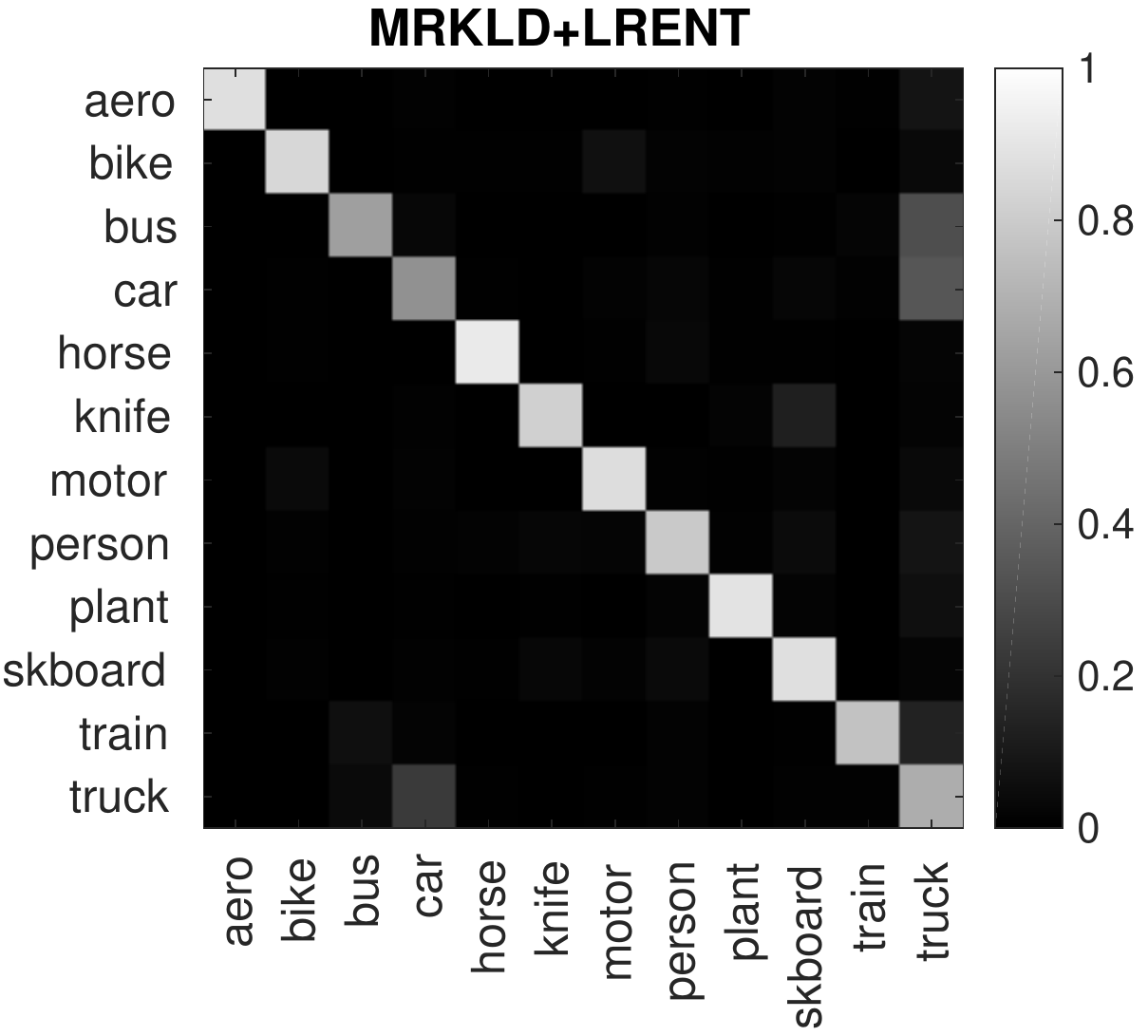}
	\caption{Confusion matrices with normalization for CBST and CRSTs.}
	\label{confus_mat}
\end{figure*}

\begin{figure*}[!t]
	\centering
	\includegraphics[width=0.19\linewidth]{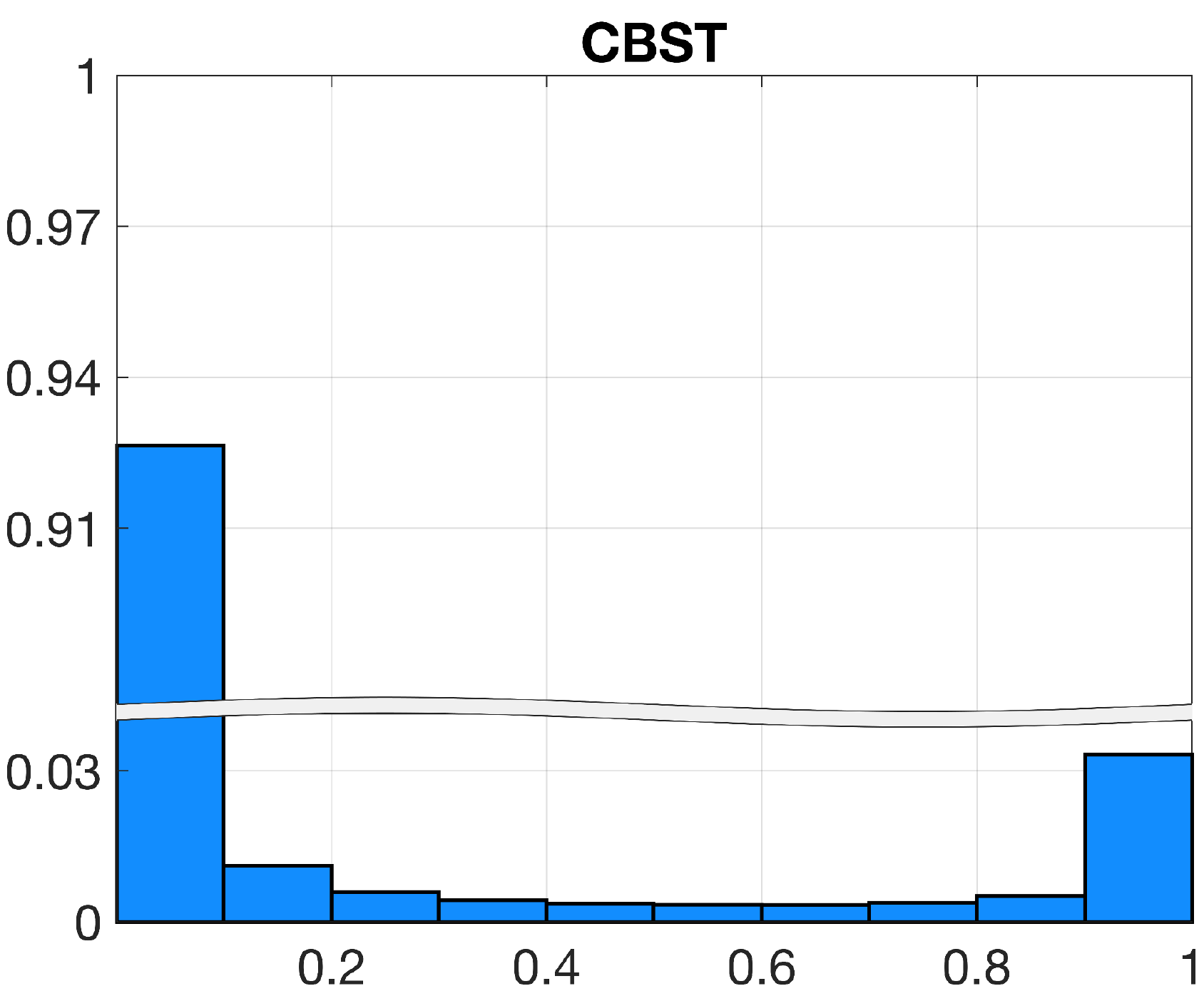}
	\includegraphics[width=0.19\linewidth]{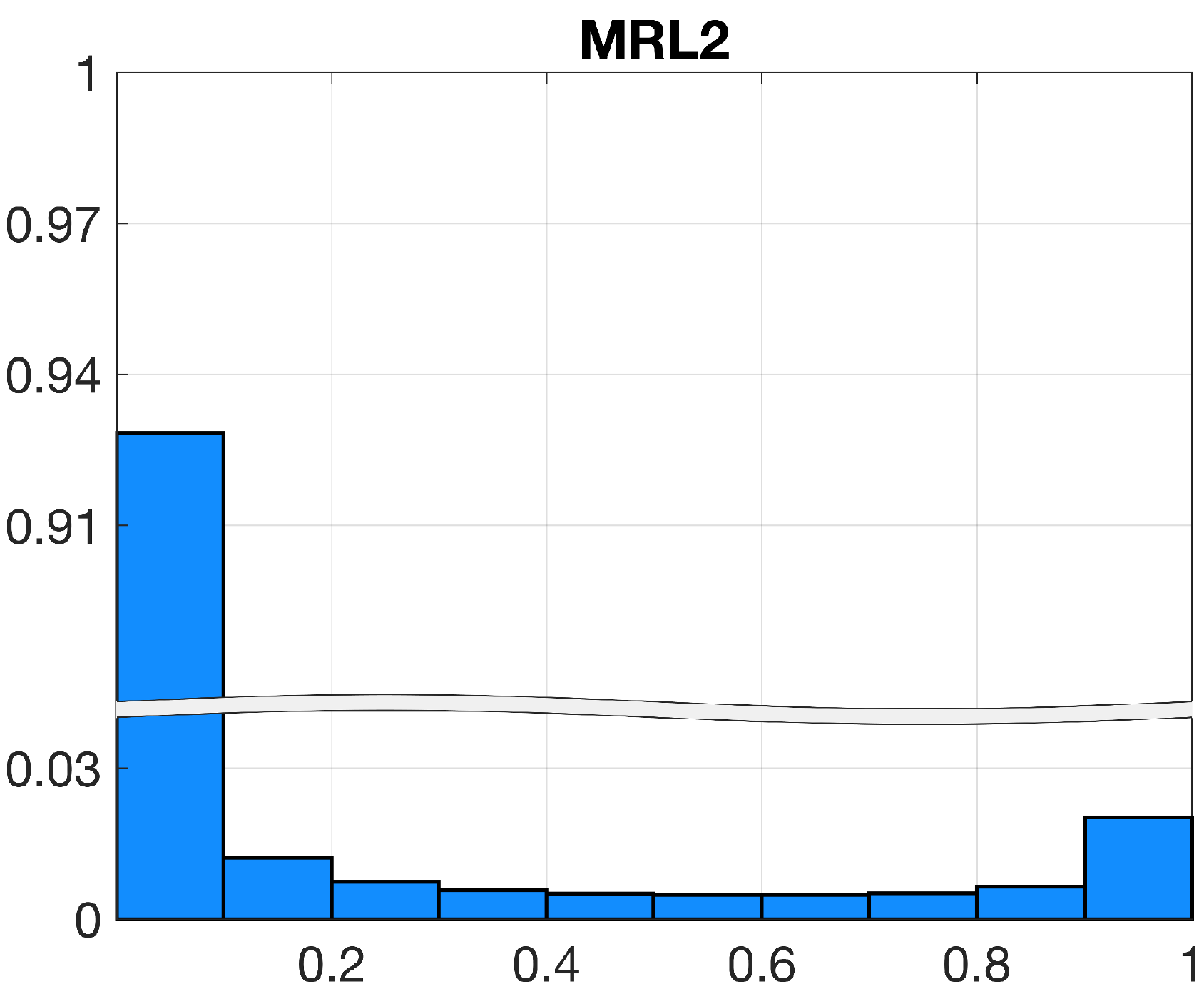}
	\includegraphics[width=0.19\linewidth]{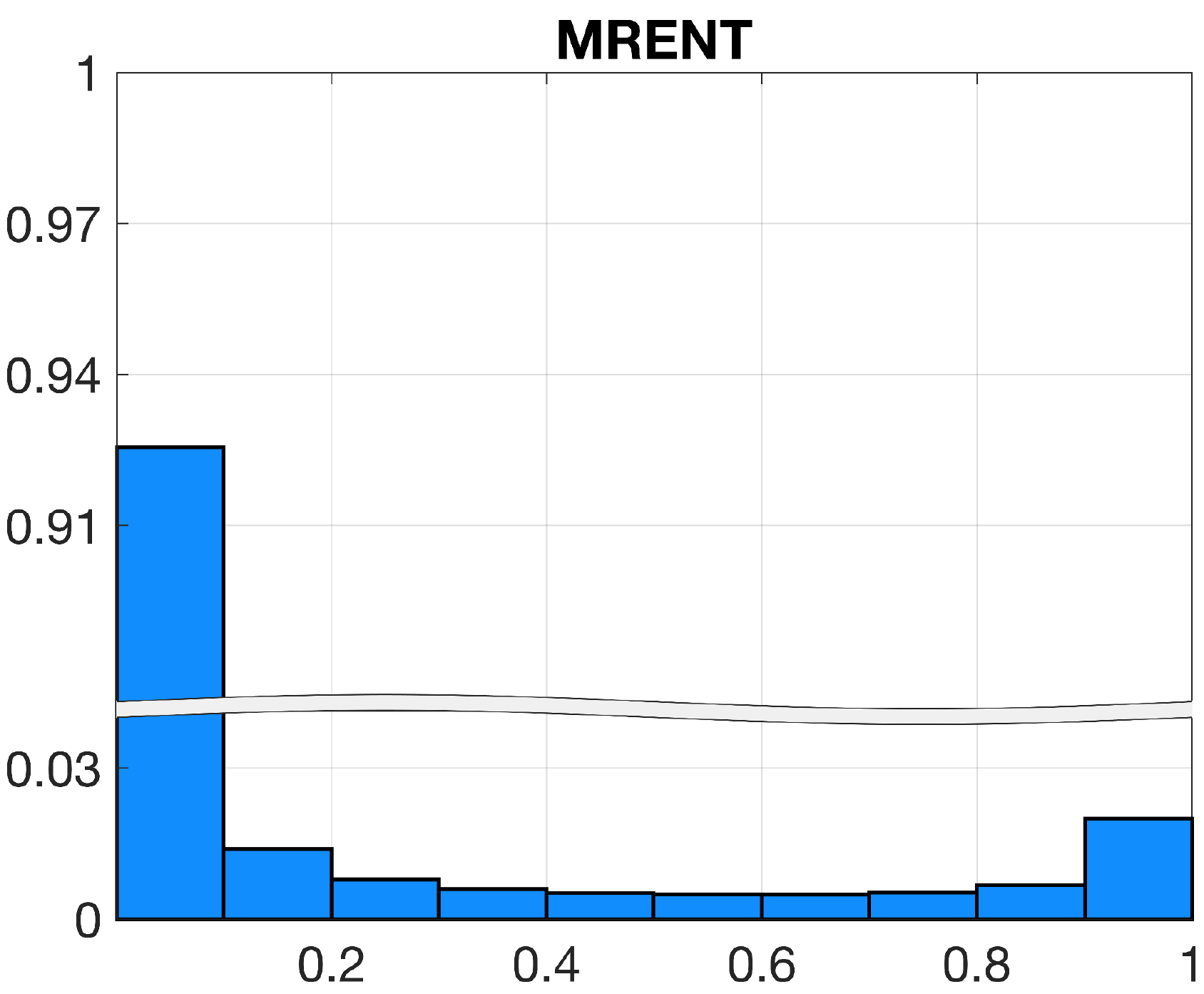}
	\includegraphics[width=0.19\linewidth]{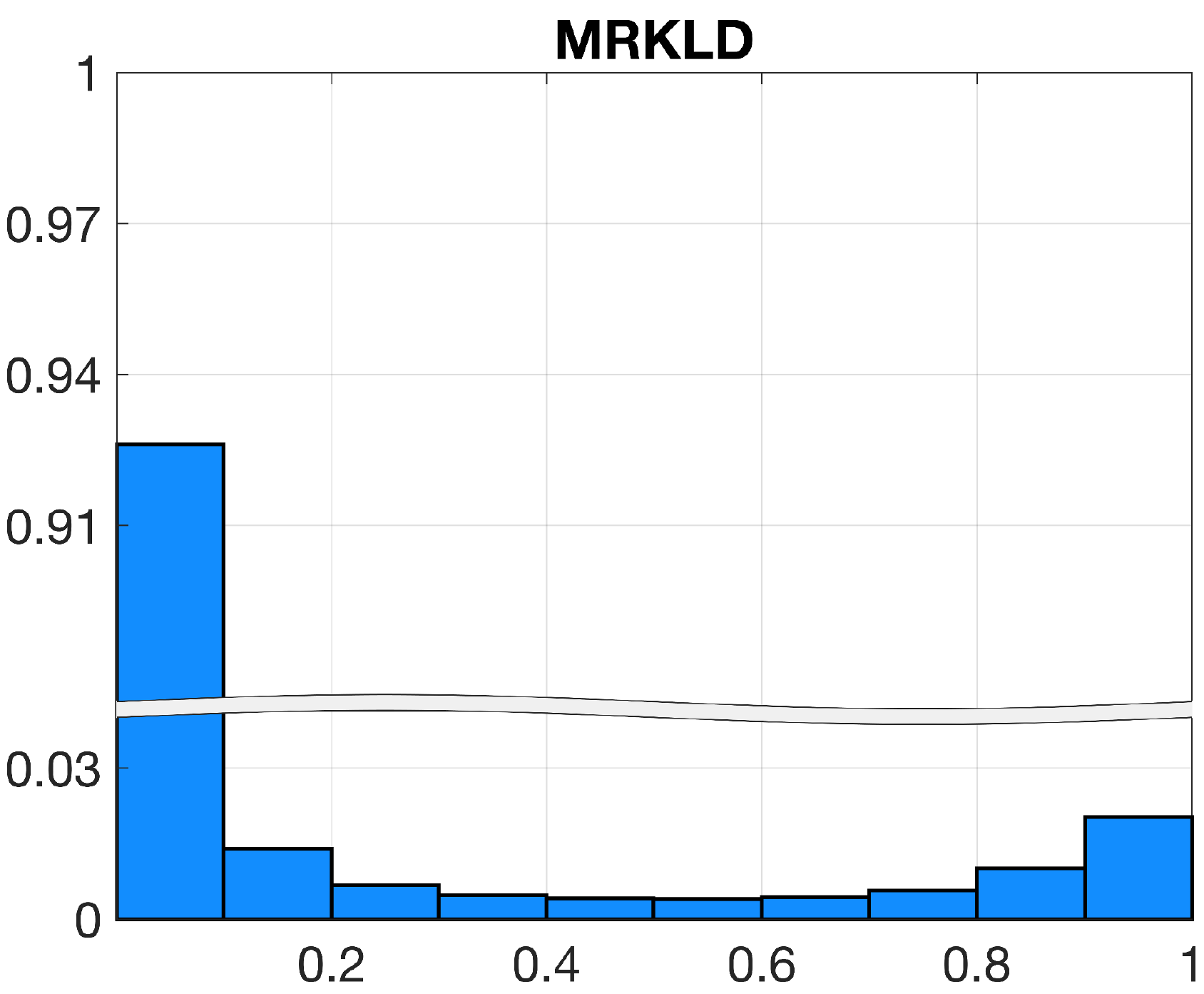}
	\includegraphics[width=0.19\linewidth]{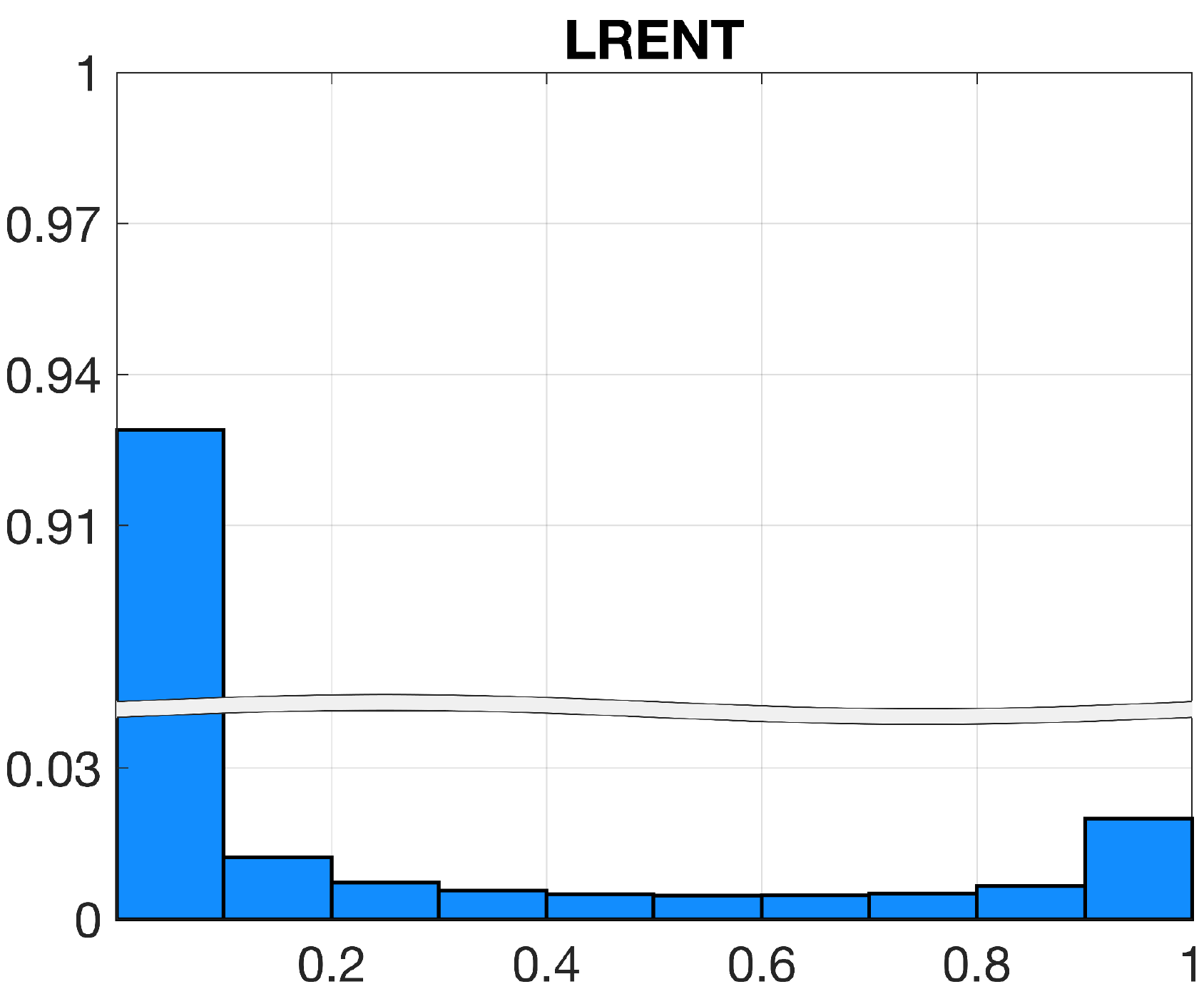}
	\caption{Histograms of softmax probability entries in target domain of GTA5 $\rightarrow$ Cityscapes.}
	\label{fig:softmax_dis}
\end{figure*}

\subsection{Feature visualization}
We also visualize the feature embeddings of the source model, CBST and MRKLD+LRENT features on VisDA17, and show them in Fig. \ref{fig:feat_vis}. Both CBST and MRKLD+LRENT obtain improved class-wise feature alignment than the source model. MRKLD+LRENT shows slightly more accurate feature alignment due to the improved performance from confidence regularization.

\subsection{Confusion matrix}
In Fig. \ref{confus_mat}, we illustrate the normalized confusion matrices of the source model, CBST and MRKLD+LRENT on VisDA17. One can see, both CBST and MRKLD+LRENT show more diagonalized confusion matrices than source model, and MRKLD+LRENT shows less mistakes. Specifically, the confusions between pairwise different classes such as ``person vs. horse'' and ``motor vs. bike'' have be reduced by confidence regularization.

\subsection{Distributions of softmax probability entries}
Following the analysis approach in~\cite{pereyra2017regularizing}, we present the distributions of predicted softmax probability entries in the target domain for different models. Specifically, we consider the ResNet-38 backbone on GTA5 $\rightarrow$ Cityscapes, with the distributions shown in Fig. \ref{fig:softmax_dis}. One could see that confidence regularization promote softer distributions by significantly reducing the proportion of highly confident entries.

\subsection{Segmentation visualization}
For qualitative evaluation, we visualize the segmentation predictions obtained by different models in Fig. \ref{fig:gta2city}. Specifically, predictions are made on sampled Cityscapes validation images by GTA5 $\rightarrow$ Cityscapes models. In Fig. \ref{fig:plgta2city}, we also visualize the pseudo-labels on sampled Cityscapes training images at the beginning second self-training round.

\begin{figure*}[!t]
	\centering
	\resizebox{0.98\textwidth}{!}{
		\begin{tabular}{@{}cccccccccc@{}}
			\cellcolor{city_color_1}\textcolor{white}{~~road~~} &
			\cellcolor{city_color_2}~~sidewalk~~&
			\cellcolor{city_color_3}\textcolor{white}{~~building~~} &
			\cellcolor{city_color_4}\textcolor{white}{~~wall~~} &
			\cellcolor{city_color_5}~~fence~~ &
			\cellcolor{city_color_6}~~pole~~ &
			\cellcolor{city_color_7}~~traffic lgt~~ &
			\cellcolor{city_color_8}~~traffic sgn~~ &
			\cellcolor{city_color_9}~~vegetation~~ & 
			\cellcolor{city_color_0}\textcolor{white}{~~ignored~~}\\
			\cellcolor{city_color_10}~~terrain~~ &
			\cellcolor{city_color_11}~~sky~~ &
			\cellcolor{city_color_12}\textcolor{white}{~~person~~} &
			\cellcolor{city_color_13}\textcolor{white}{~~rider~~} &
			\cellcolor{city_color_14}\textcolor{white}{~~car~~} &
			\cellcolor{city_color_15}\textcolor{white}{~~truck~~} &
			\cellcolor{city_color_16}\textcolor{white}{~~bus~~} &
			\cellcolor{city_color_17}\textcolor{white}{~~train~~} &
			\cellcolor{city_color_18}\textcolor{white}{~~motorcycle~~} &
			\cellcolor{city_color_19}\textcolor{white}{~~bike~~}
		\end{tabular}
	}
	
	\vspace{1mm}
	\includegraphics[width=0.24\textwidth]{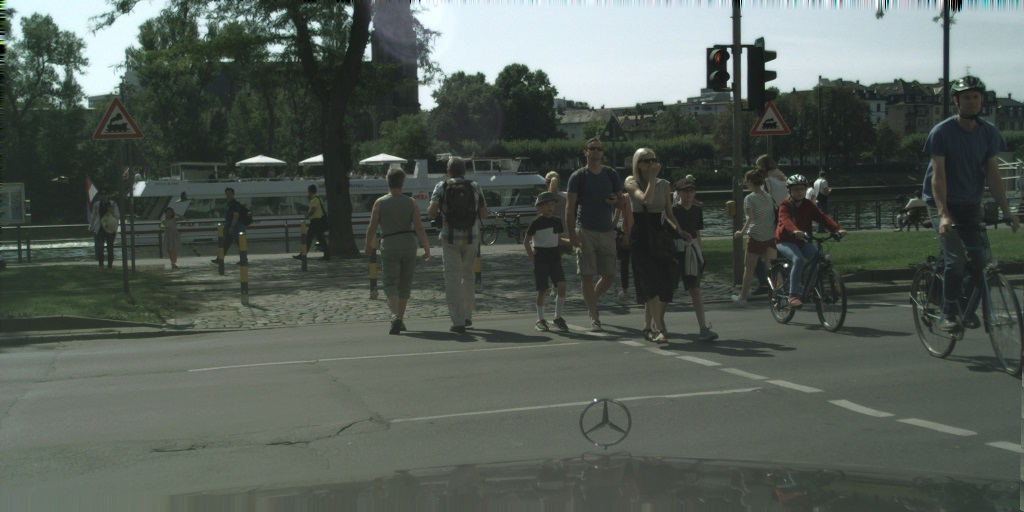}
	\includegraphics[width=0.24\textwidth]{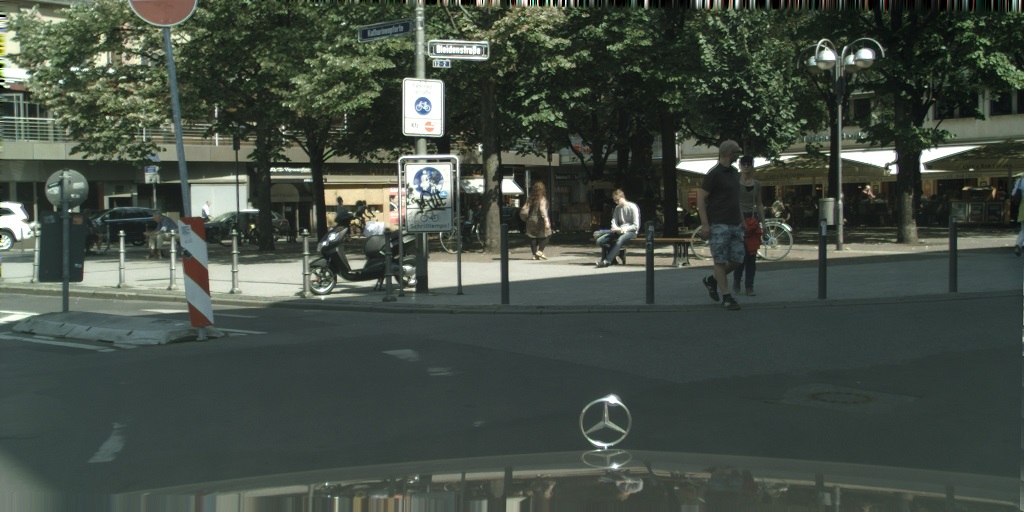}
	\includegraphics[width=0.24\textwidth]{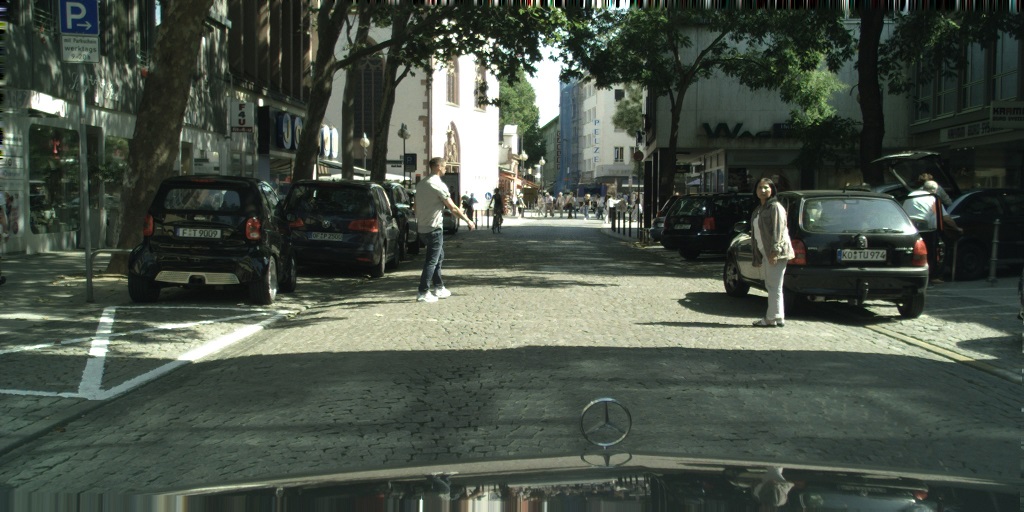}
	\includegraphics[width=0.24\textwidth]{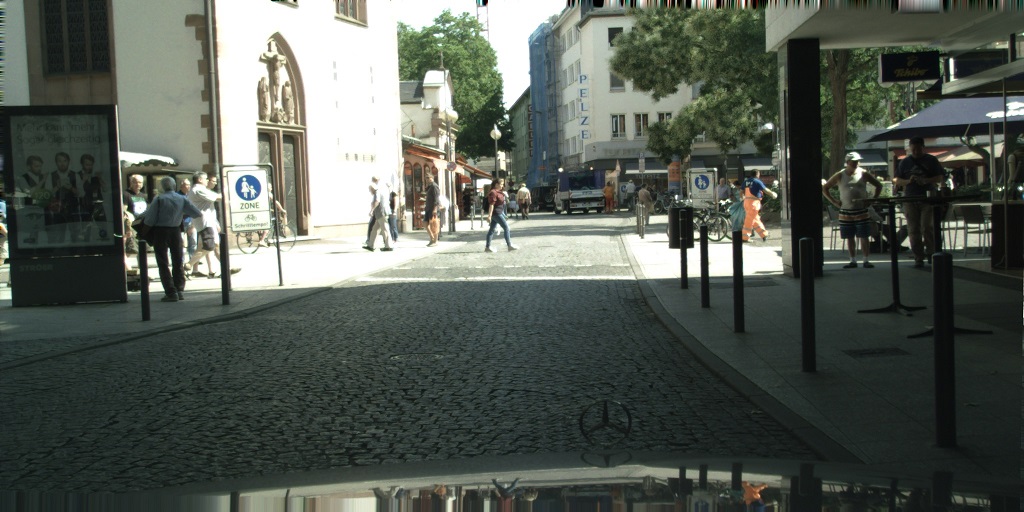}
	\quad\\\vspace{0.5mm}
	\includegraphics[width=0.24\textwidth]{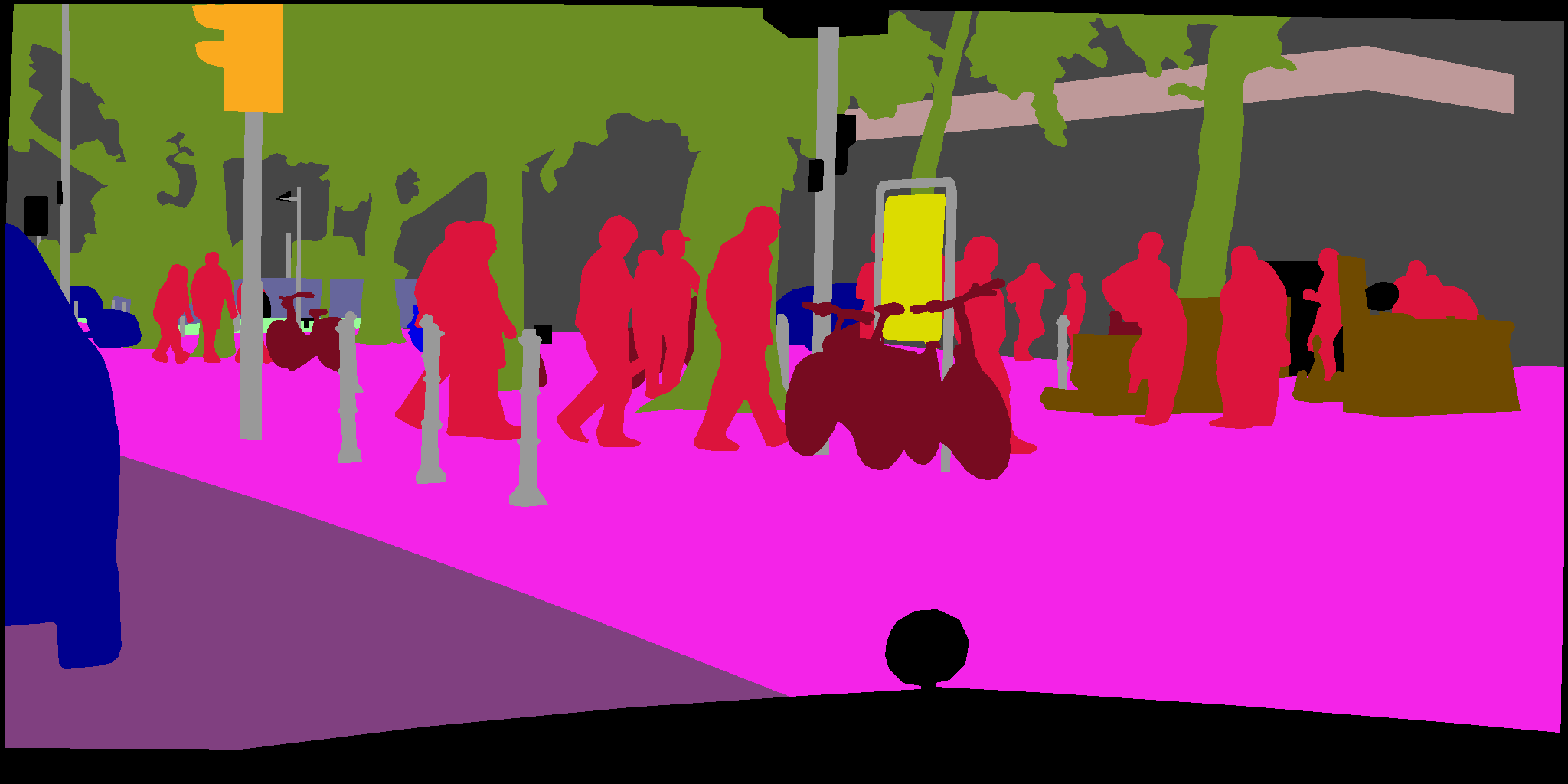}
	\includegraphics[width=0.24\textwidth]{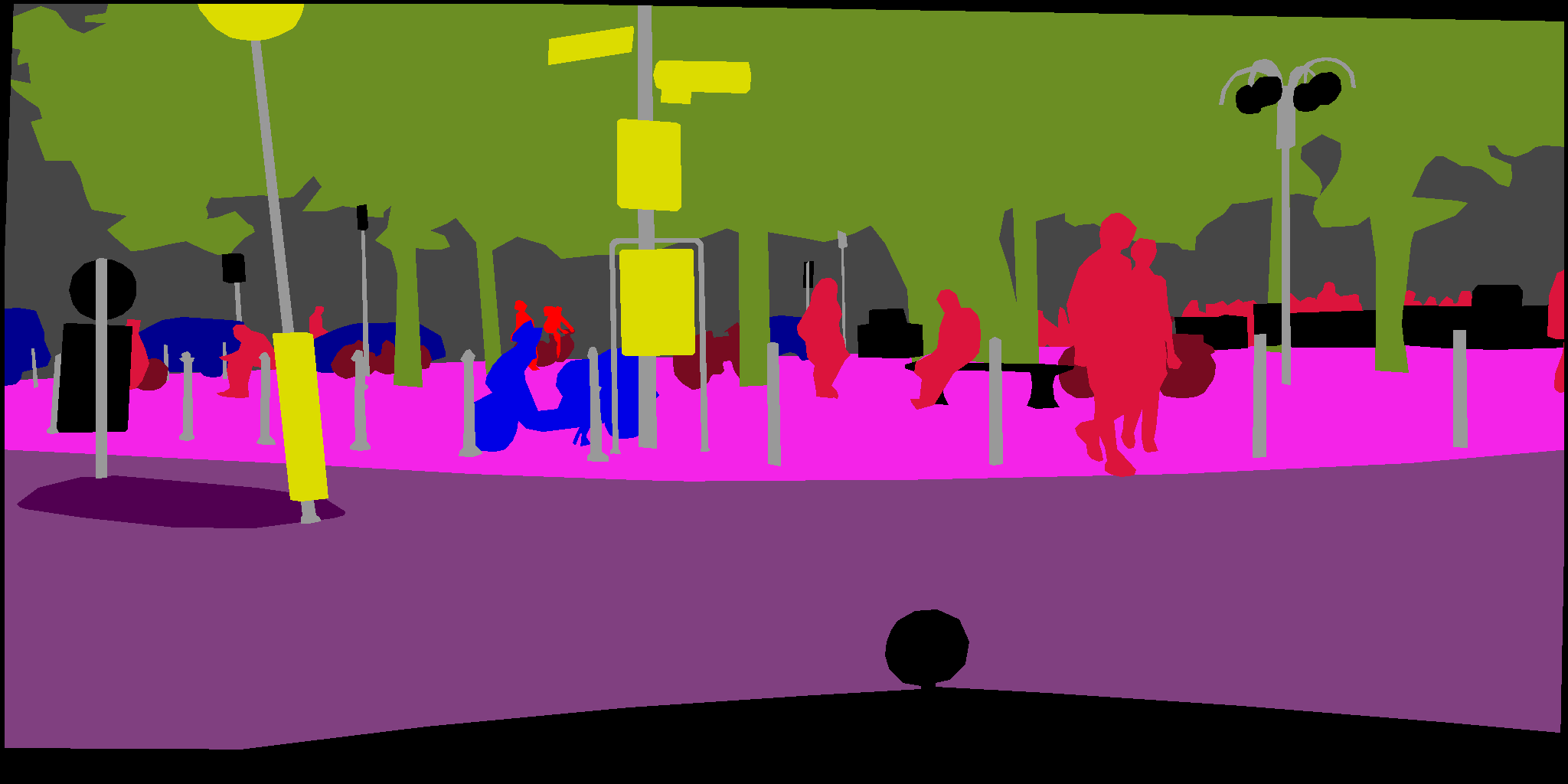}
	\includegraphics[width=0.24\textwidth]{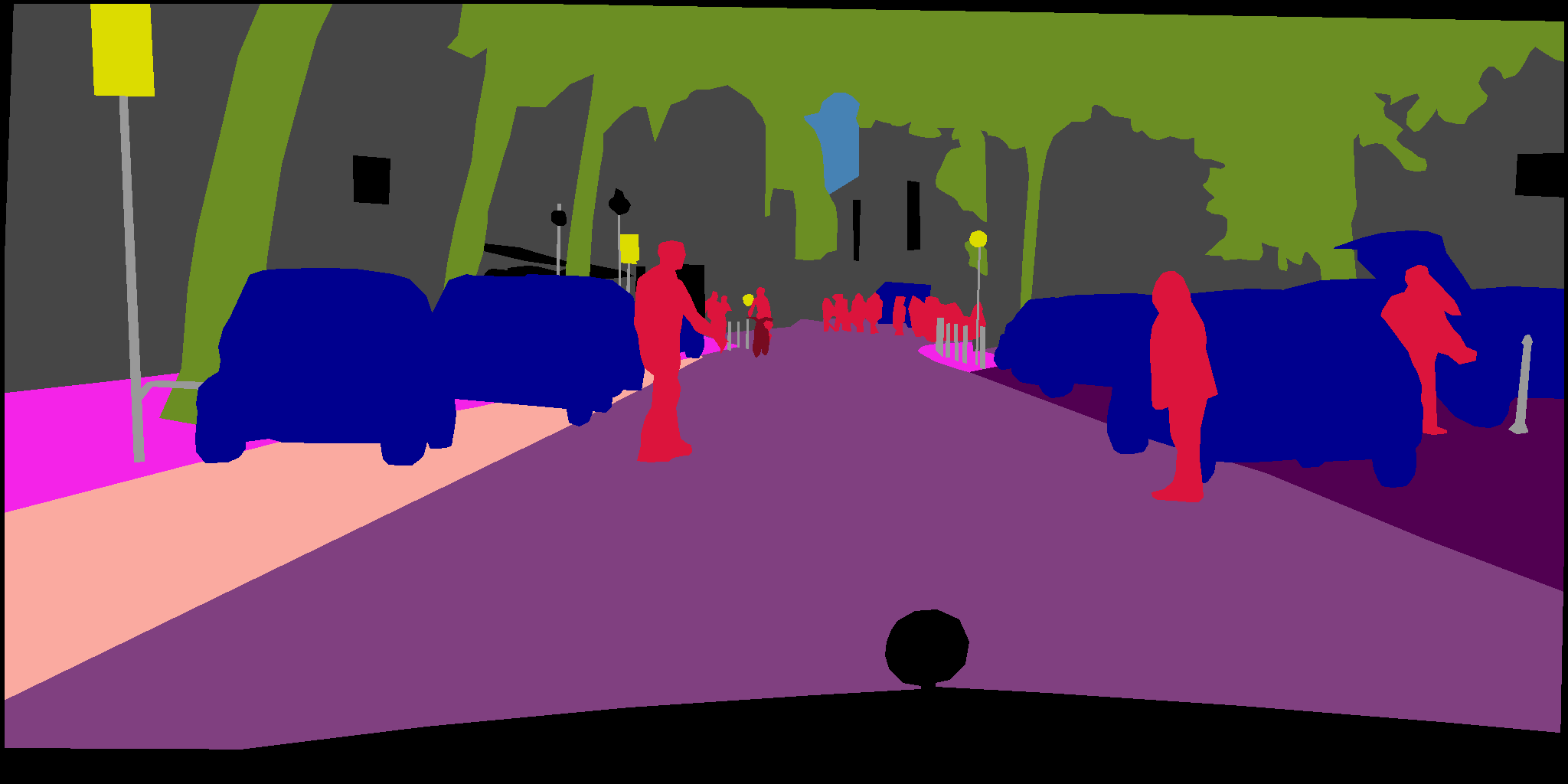}
	\includegraphics[width=0.24\textwidth]{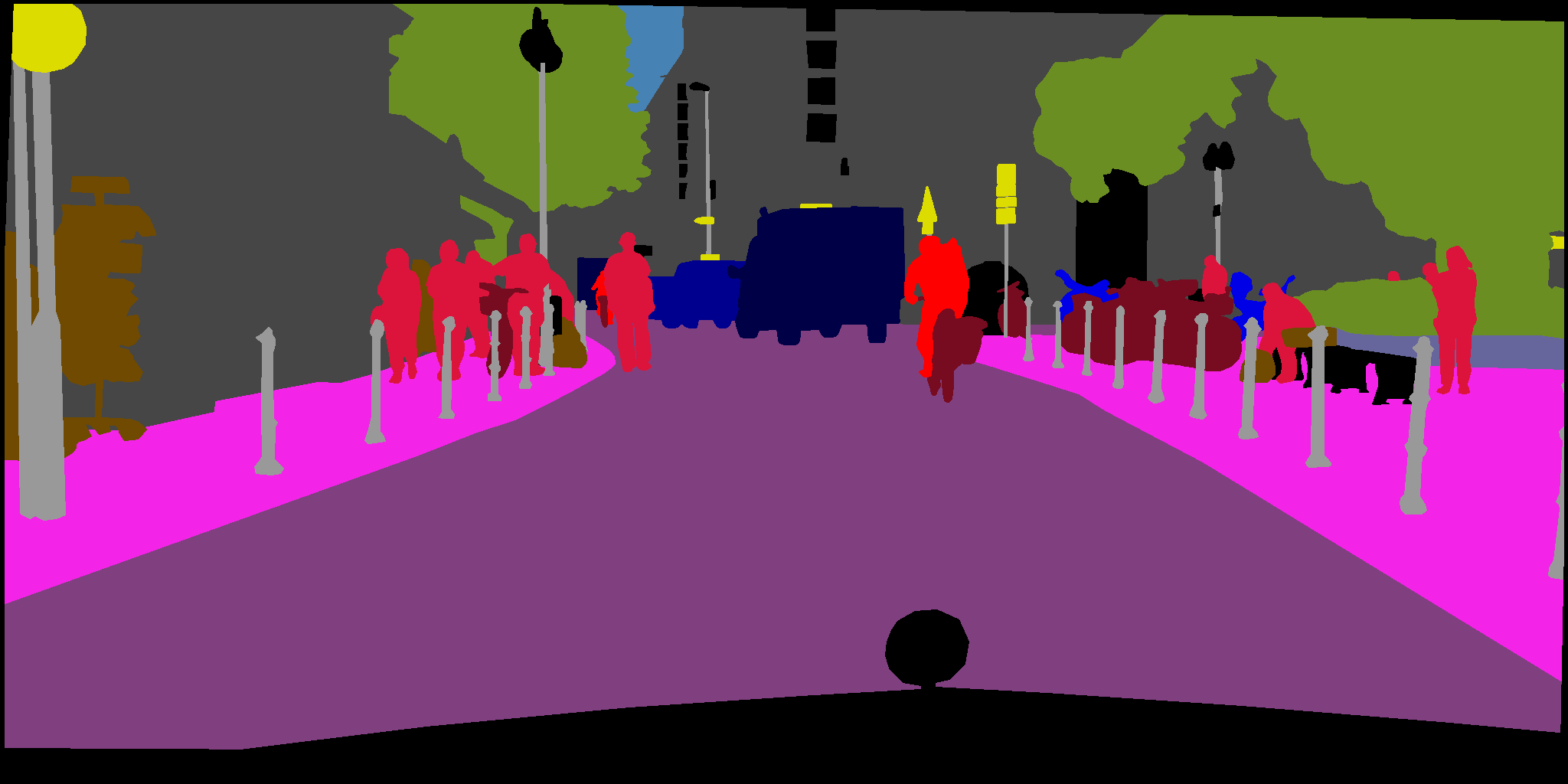}
	\quad\\\vspace{0.5mm}
	\includegraphics[width=0.24\textwidth]{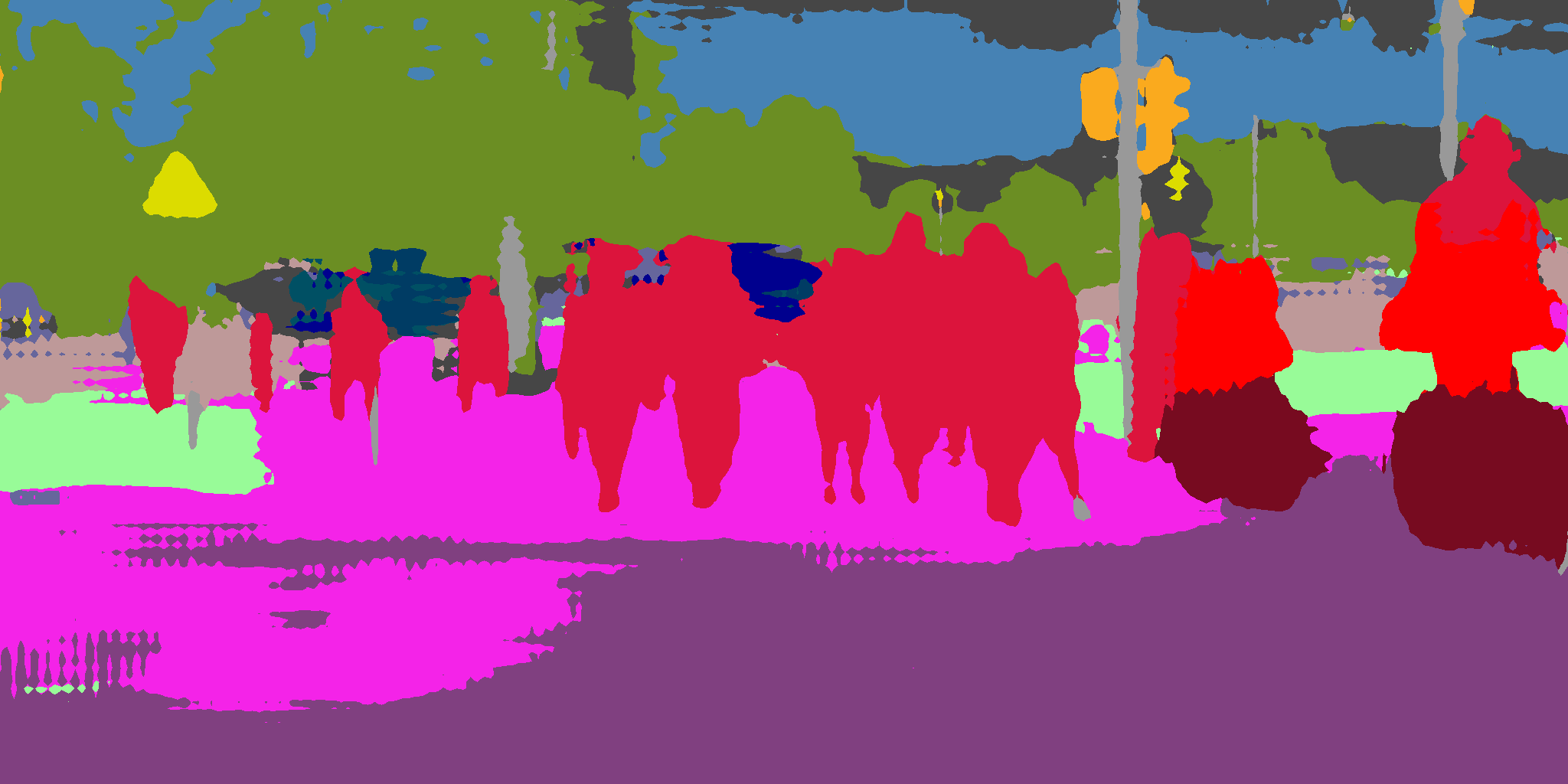}
	\includegraphics[width=0.24\textwidth]{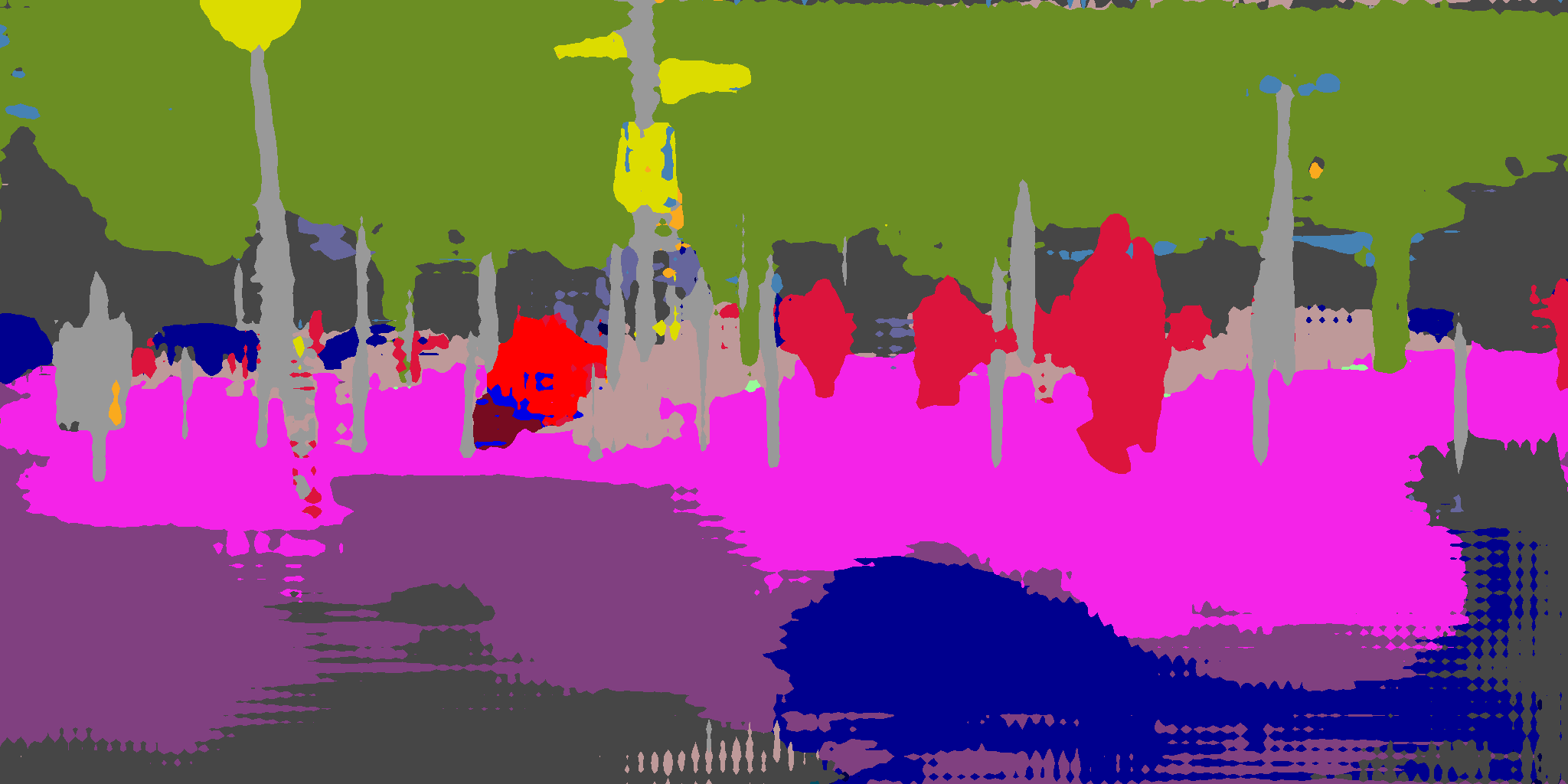}
	\includegraphics[width=0.24\textwidth]{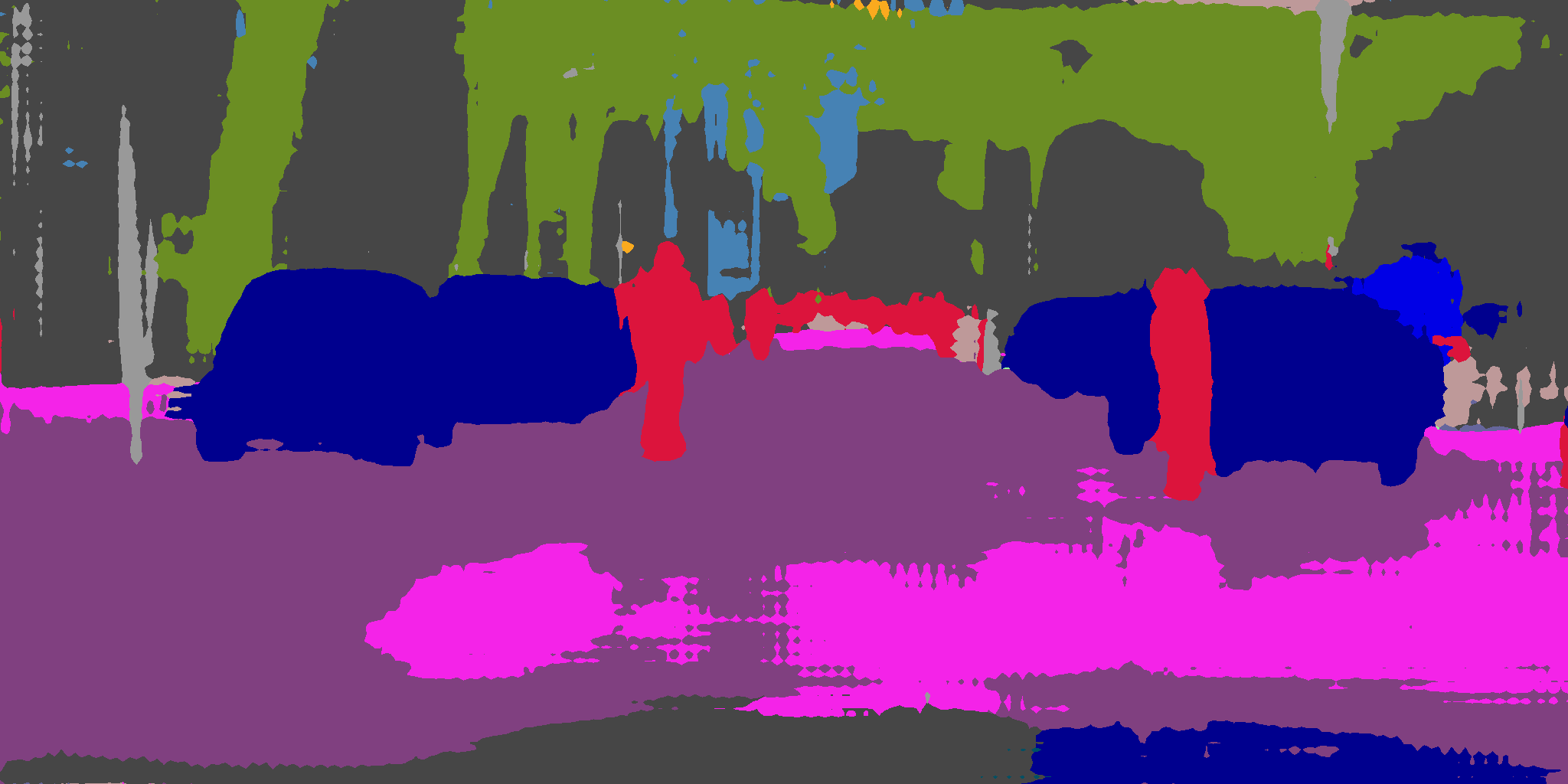}
	\includegraphics[width=0.24\textwidth]{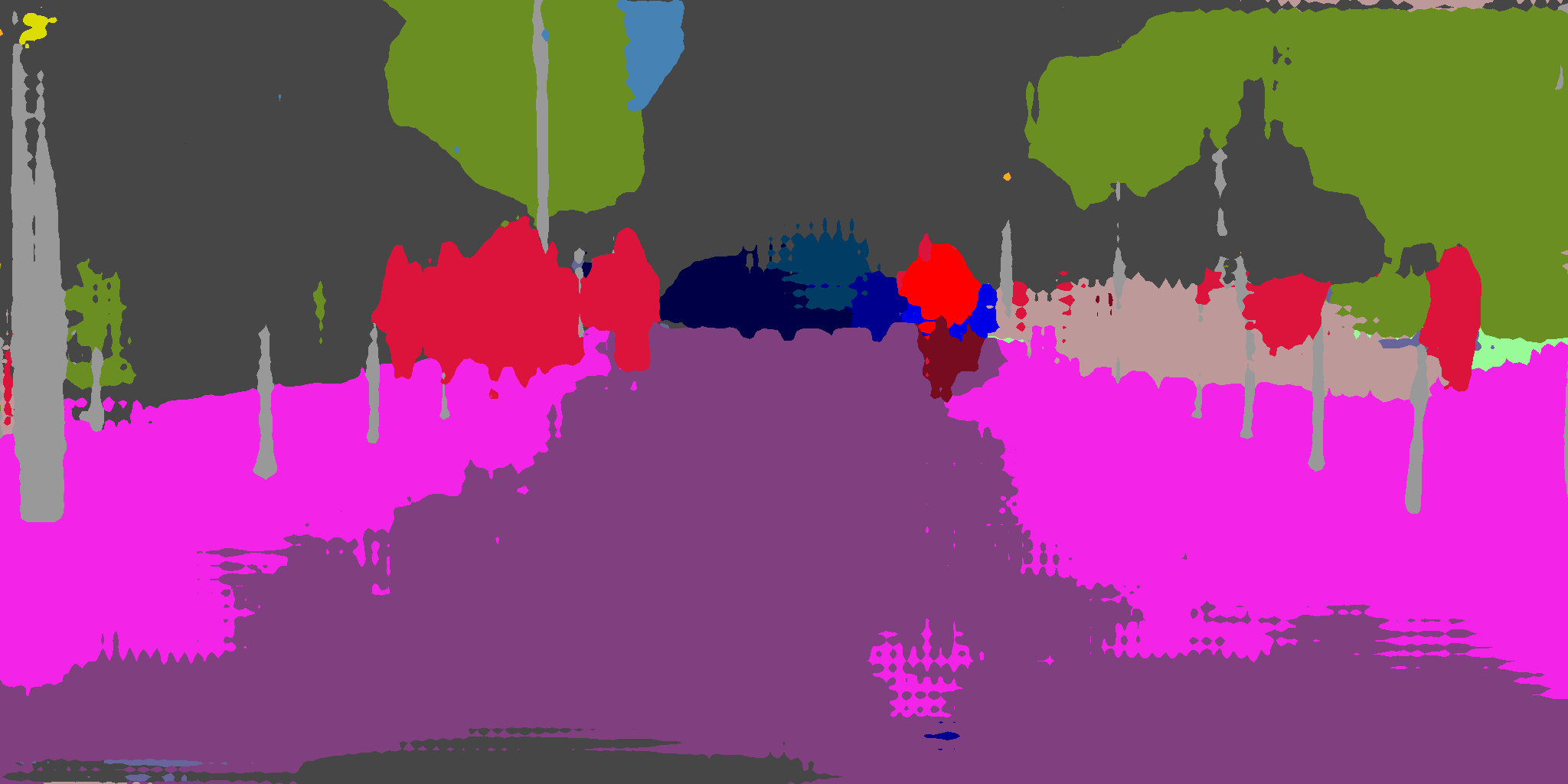}
	\quad\\\vspace{0.5mm}
	\includegraphics[width=0.24\textwidth]{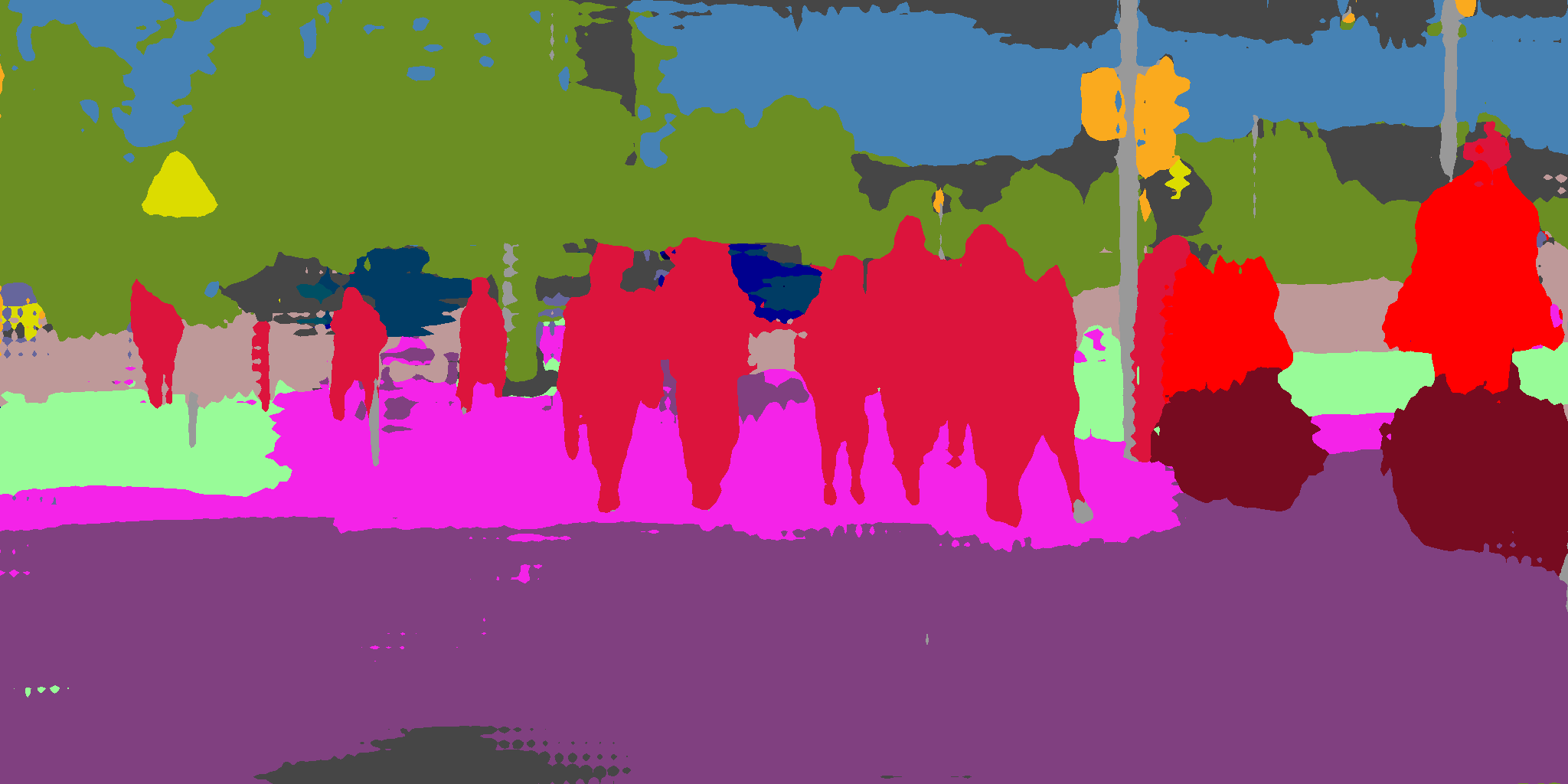}
	\includegraphics[width=0.24\textwidth]{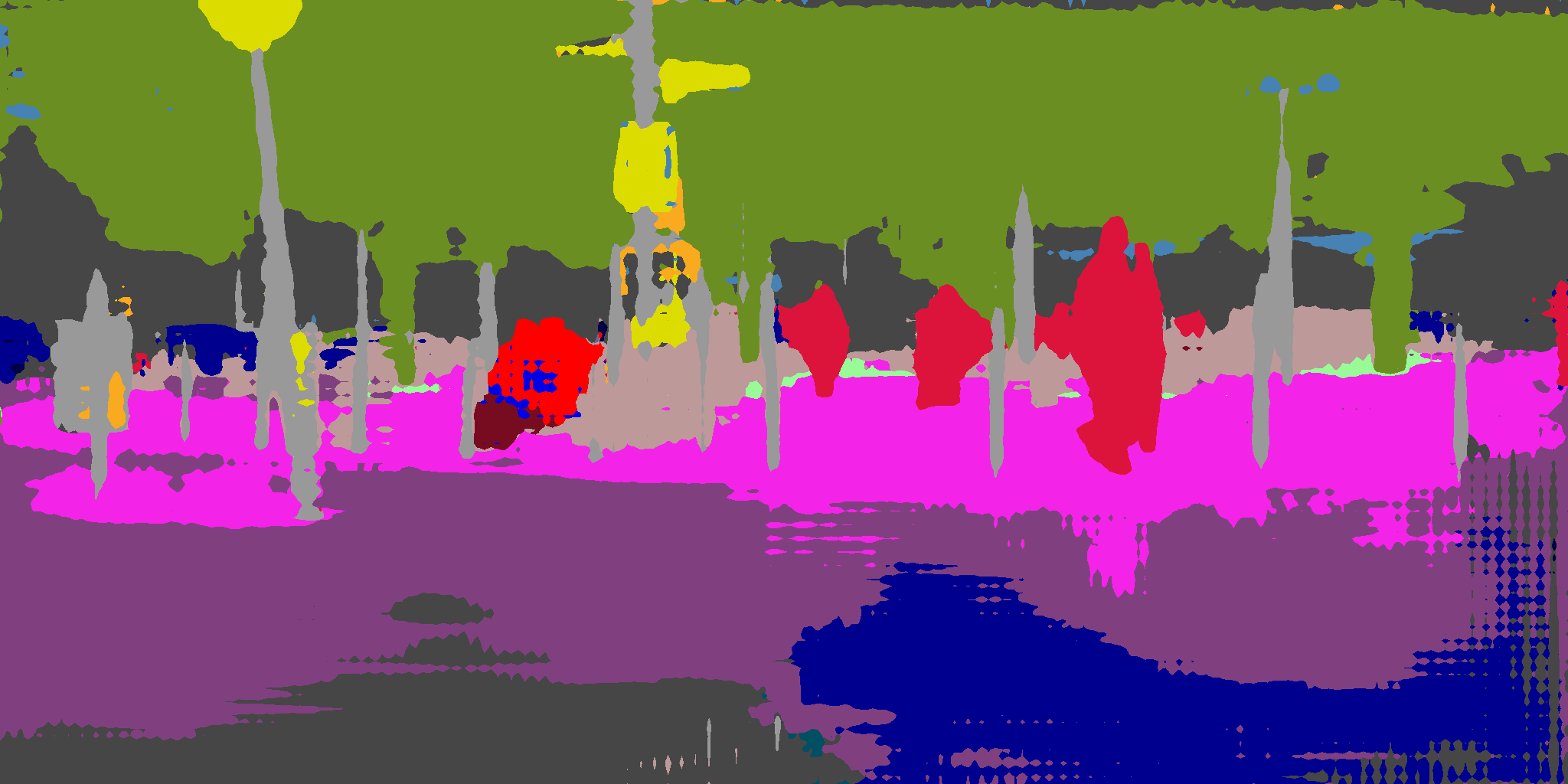}
	\includegraphics[width=0.24\textwidth]{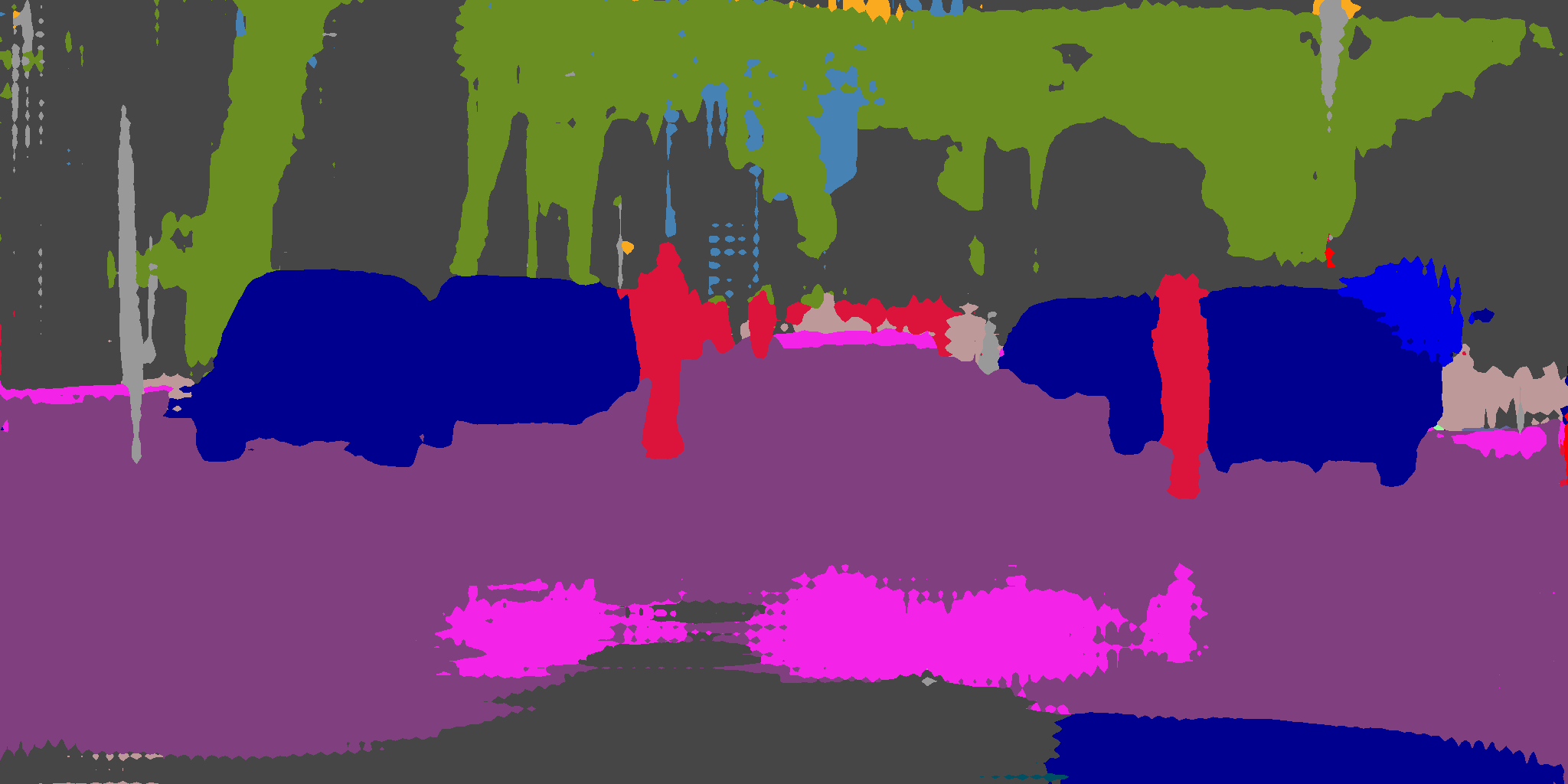}
	\includegraphics[width=0.24\textwidth]{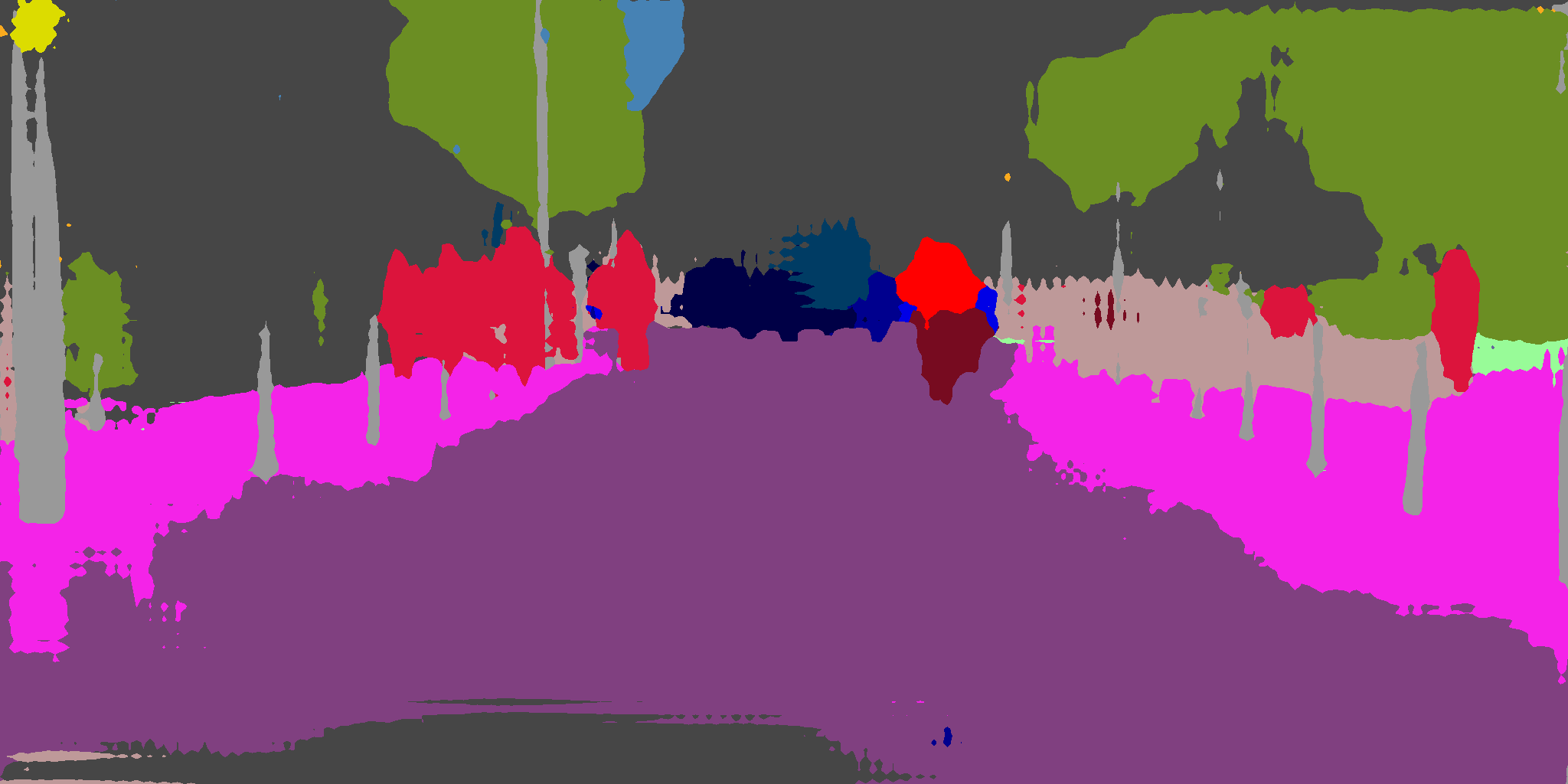}
	\quad\\\vspace{0.5mm}
	\includegraphics[width=0.24\textwidth]{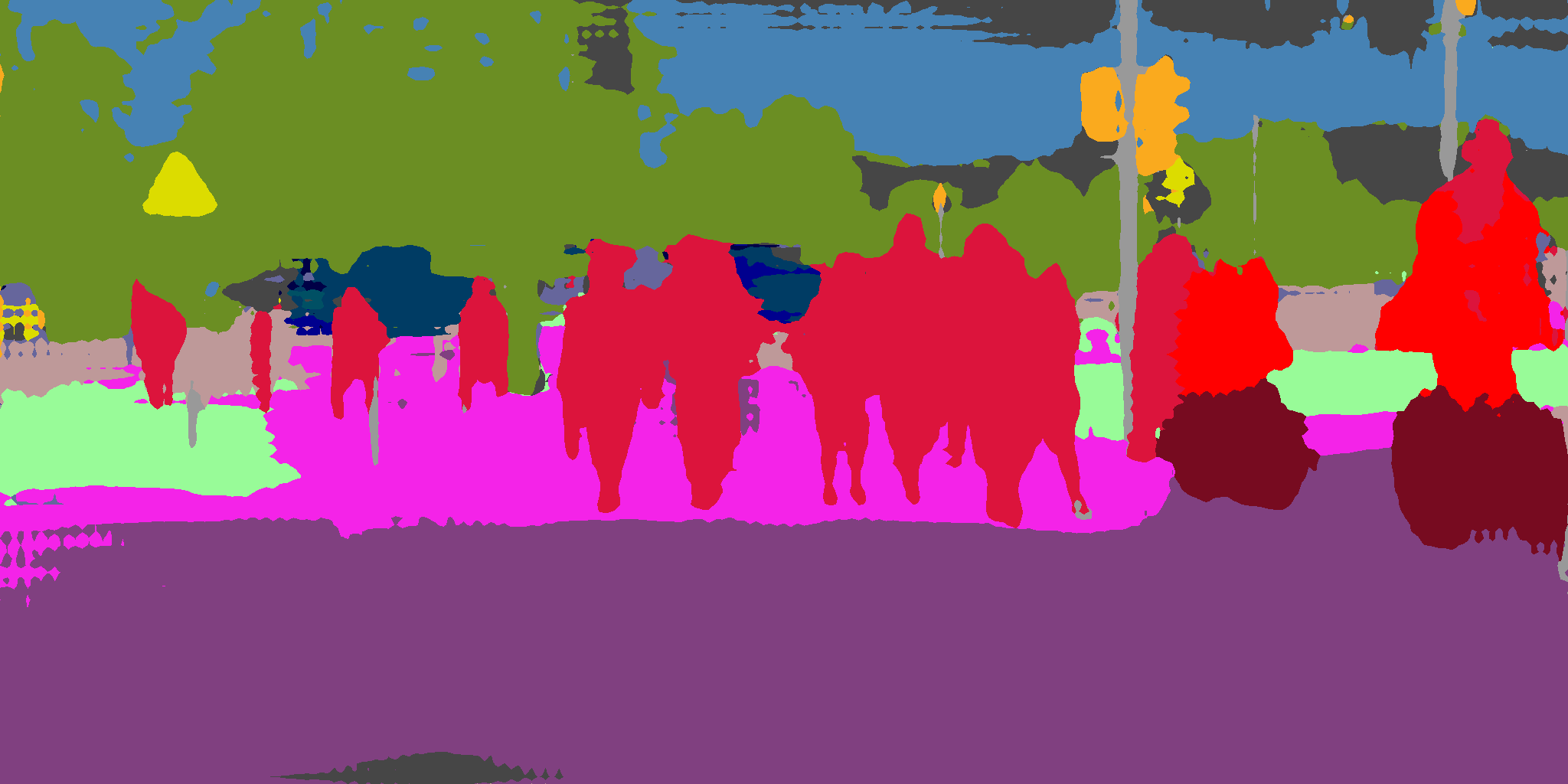}
	\includegraphics[width=0.24\textwidth]{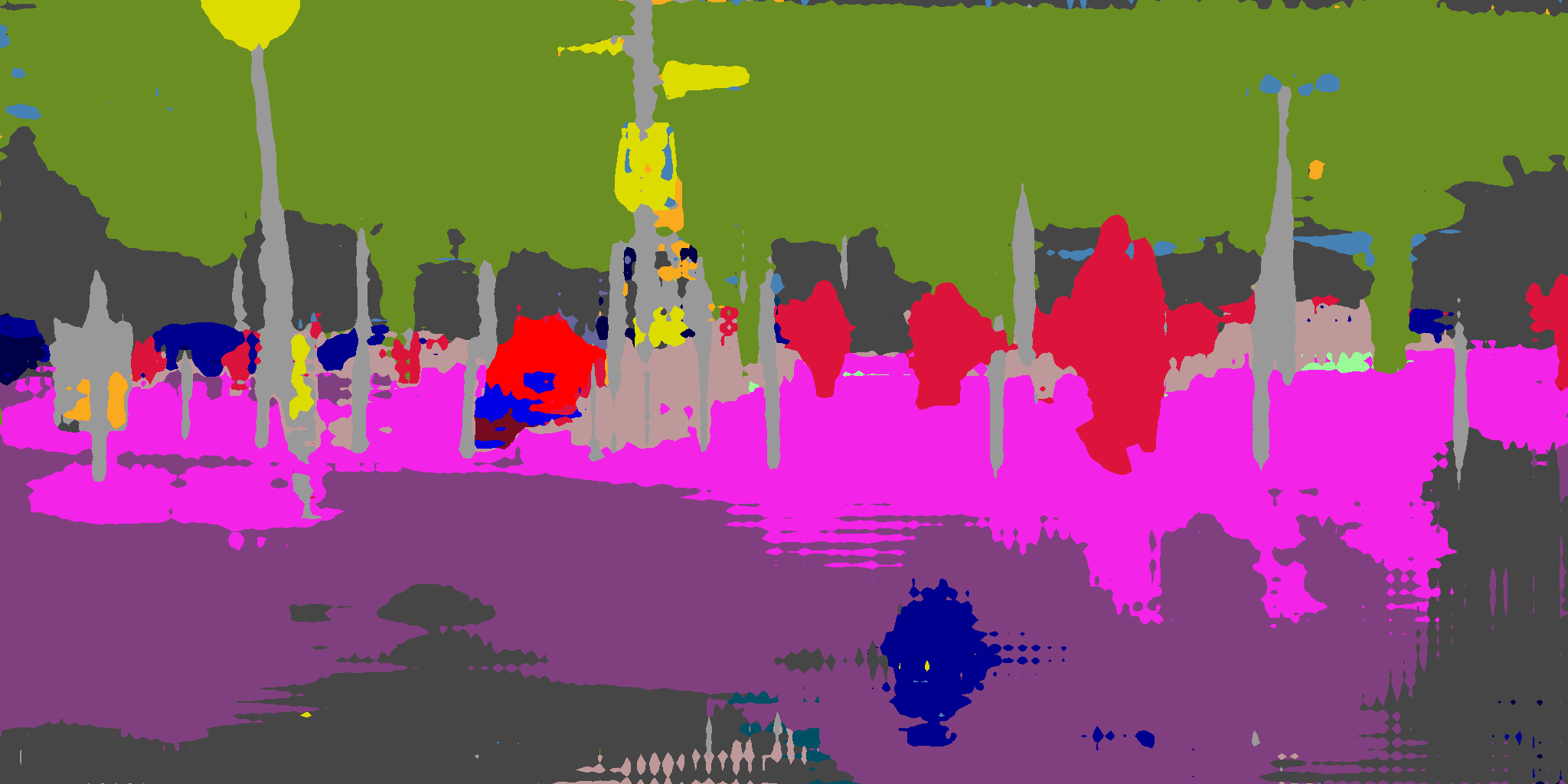}
	\includegraphics[width=0.24\textwidth]{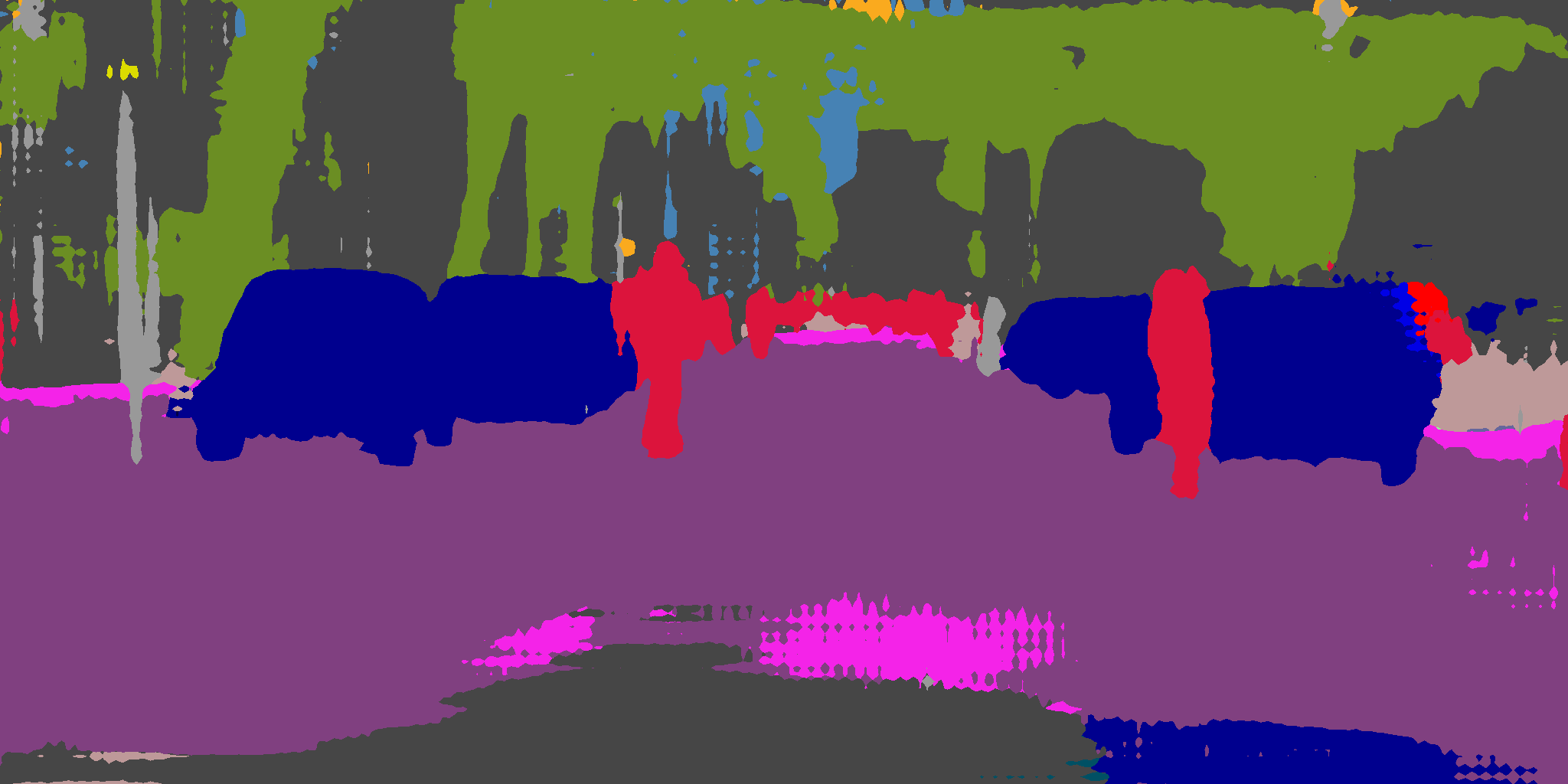}
	\includegraphics[width=0.24\textwidth]{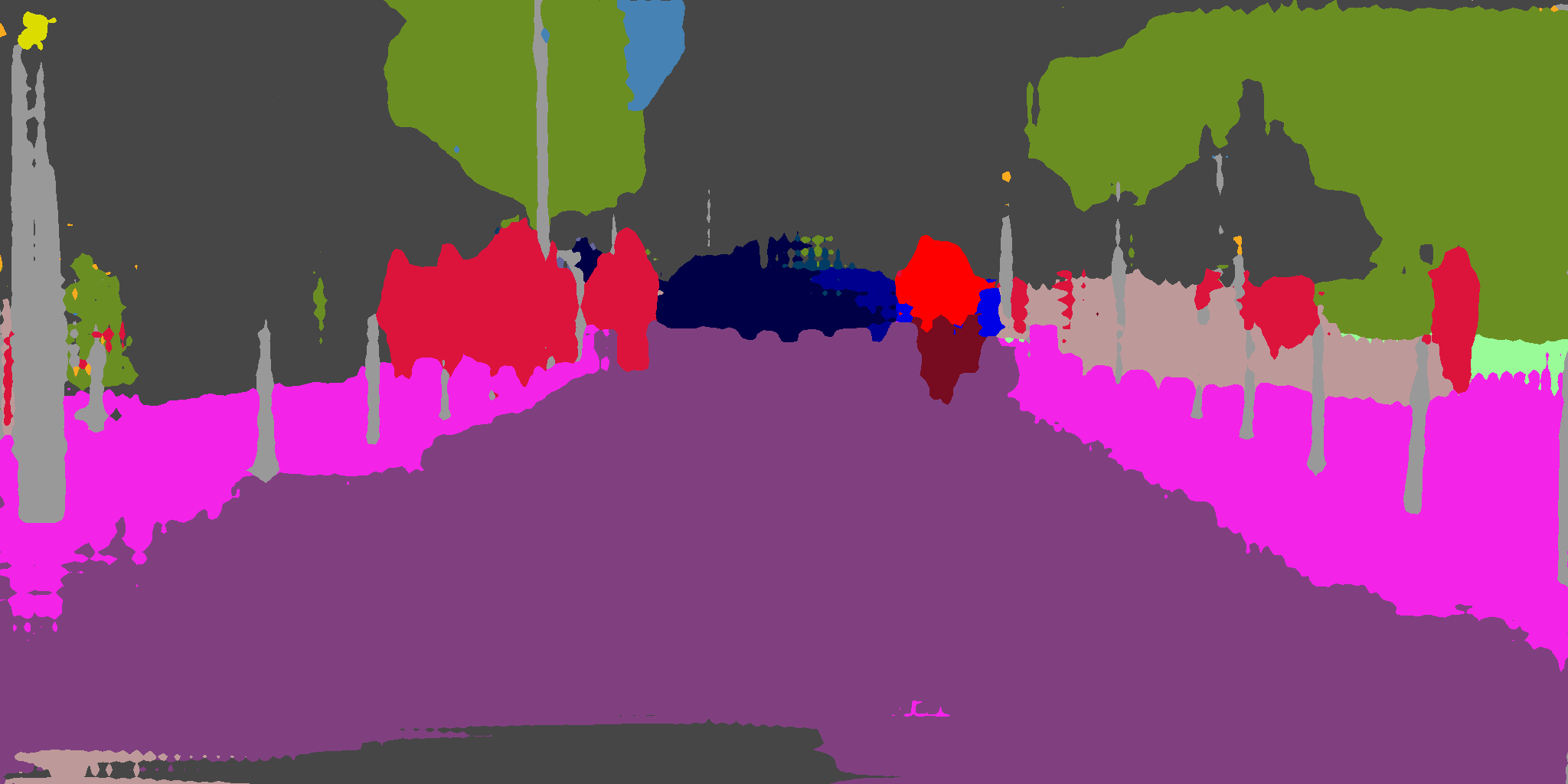}
	\quad\\\vspace{0.5mm}
	\includegraphics[width=0.24\textwidth]{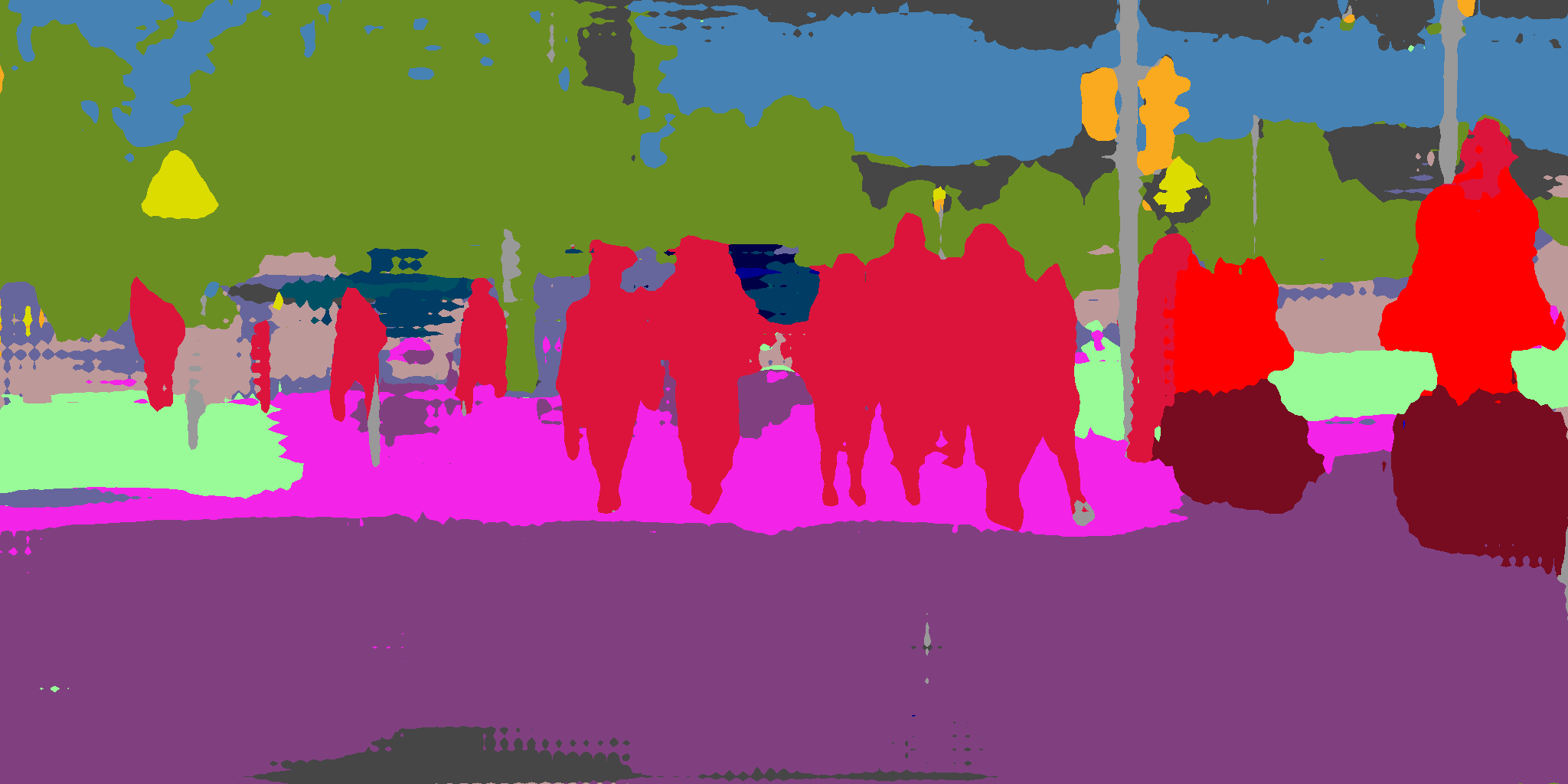}
	\includegraphics[width=0.24\textwidth]{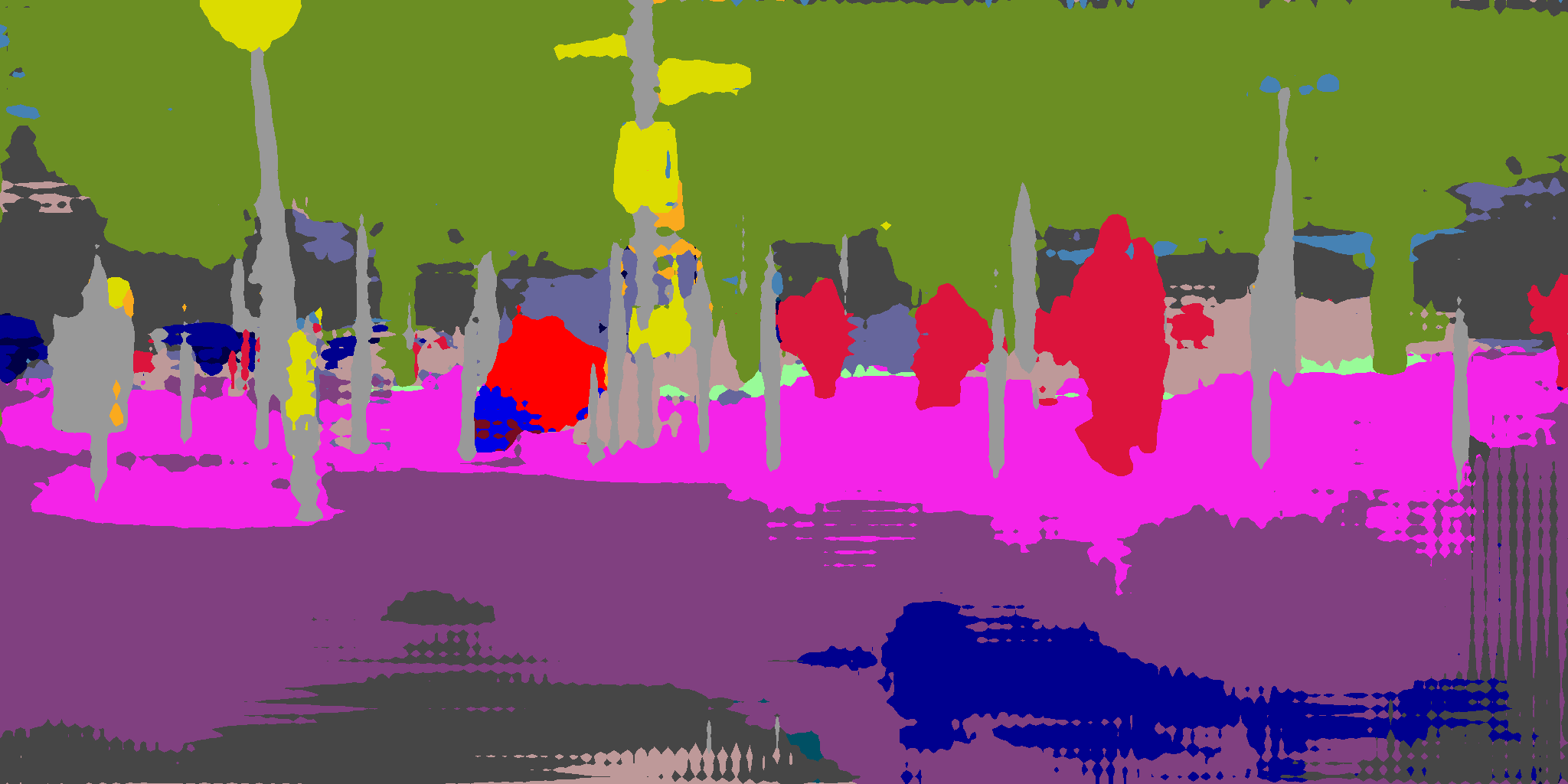}
	\includegraphics[width=0.24\textwidth]{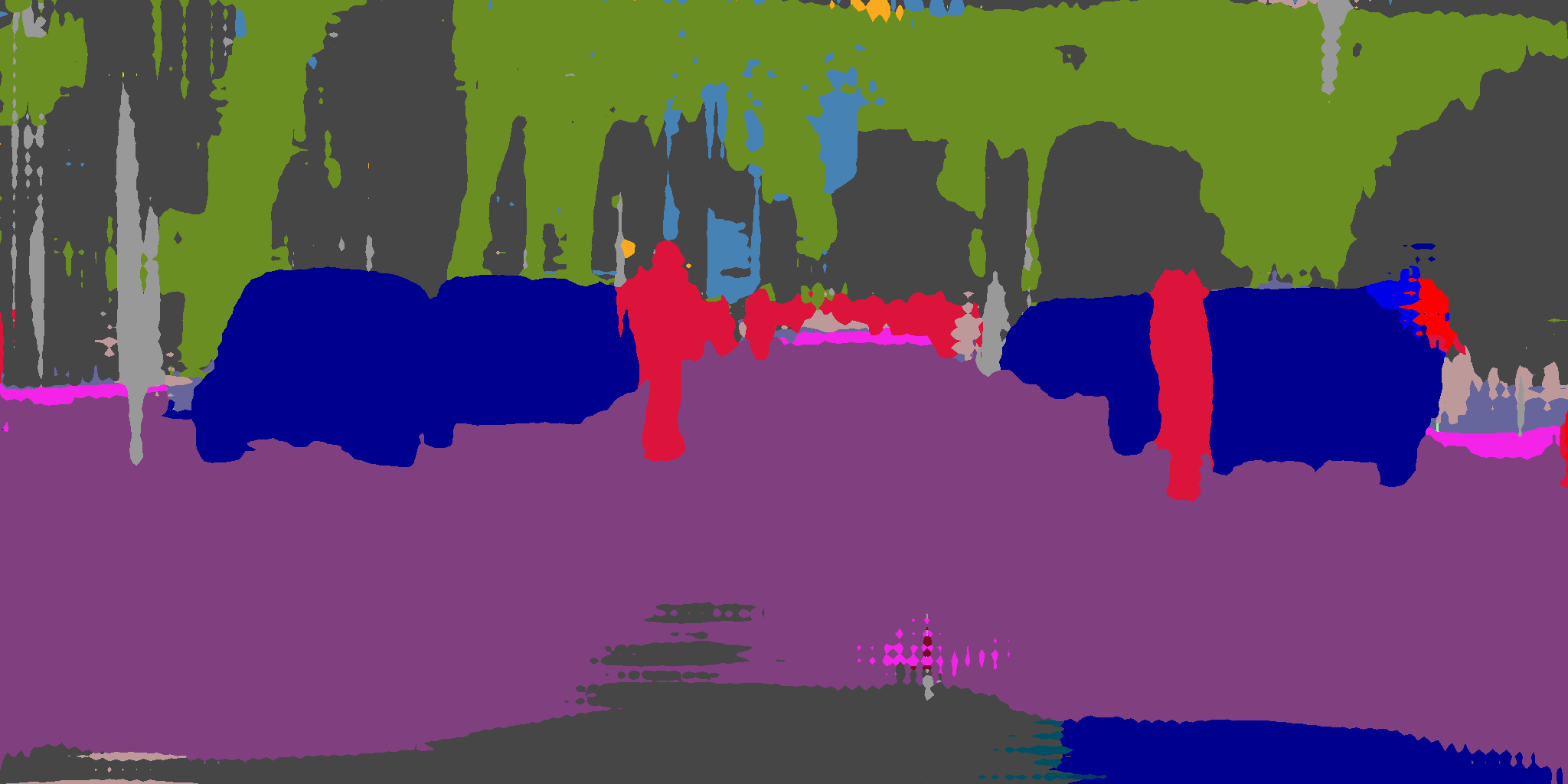}
	\includegraphics[width=0.24\textwidth]{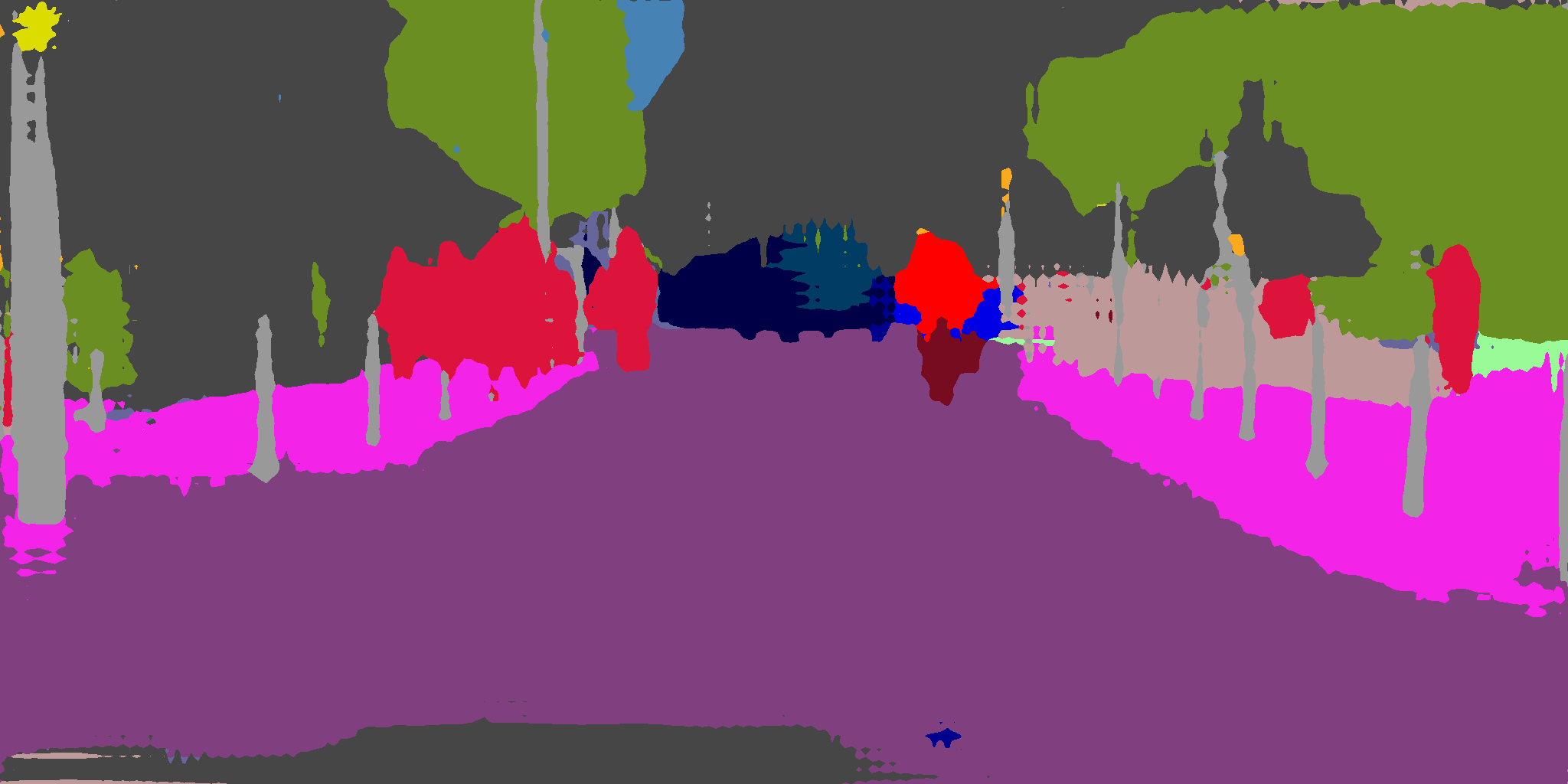}
	\quad\\\vspace{0.5mm}
	\includegraphics[width=0.24\textwidth]{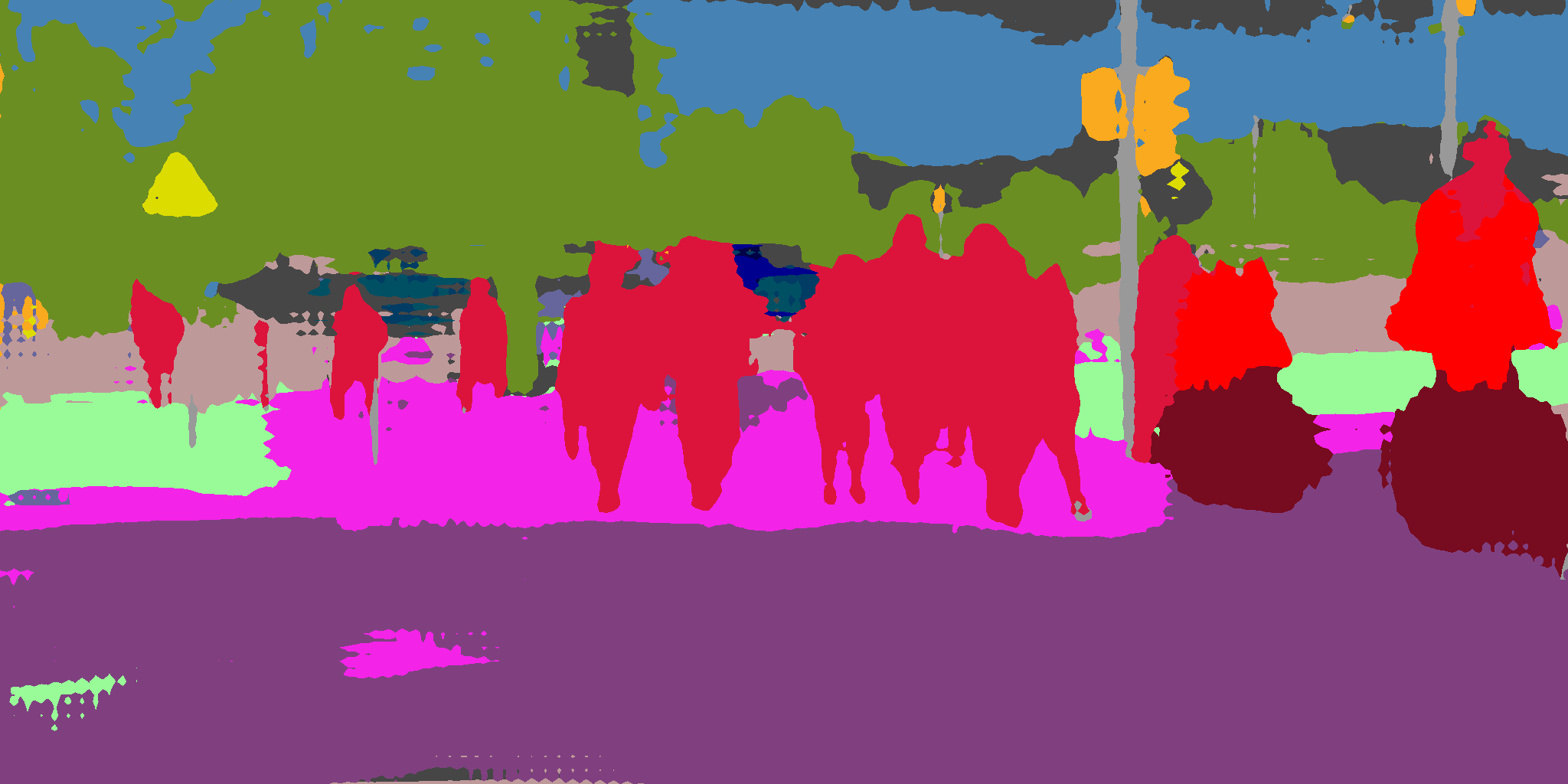}
	\includegraphics[width=0.24\textwidth]{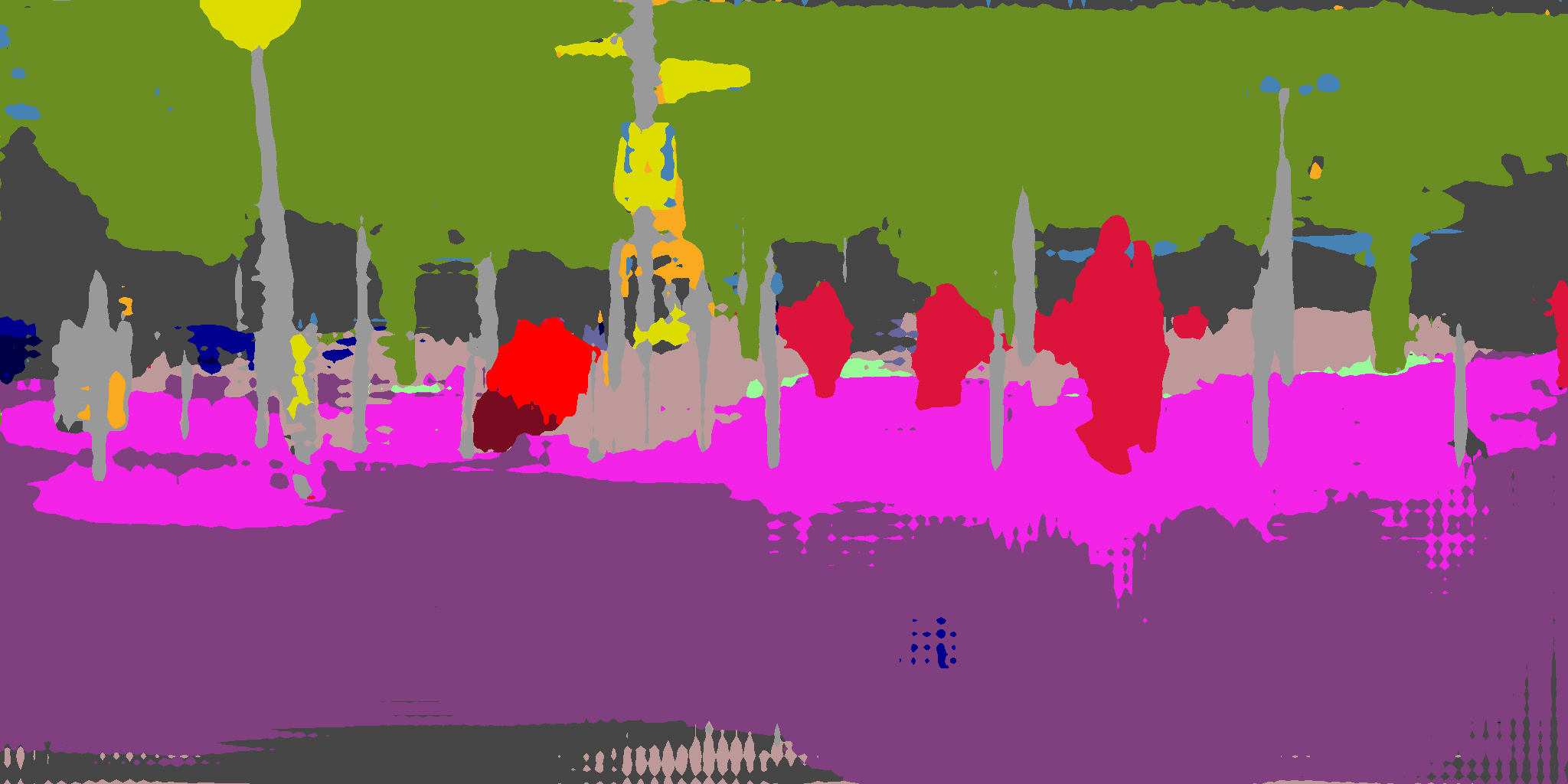}
	\includegraphics[width=0.24\textwidth]{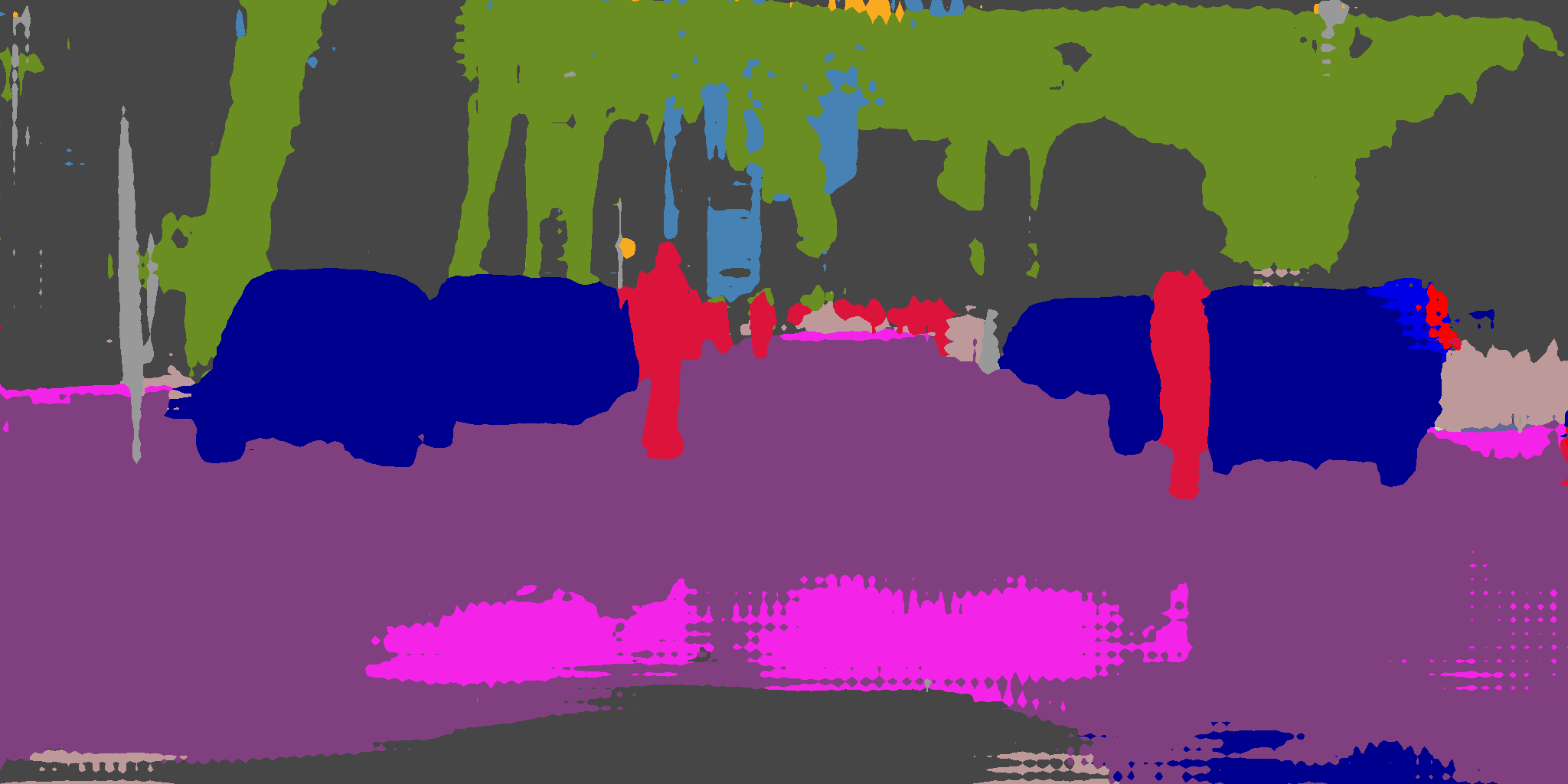}
	\includegraphics[width=0.24\textwidth]{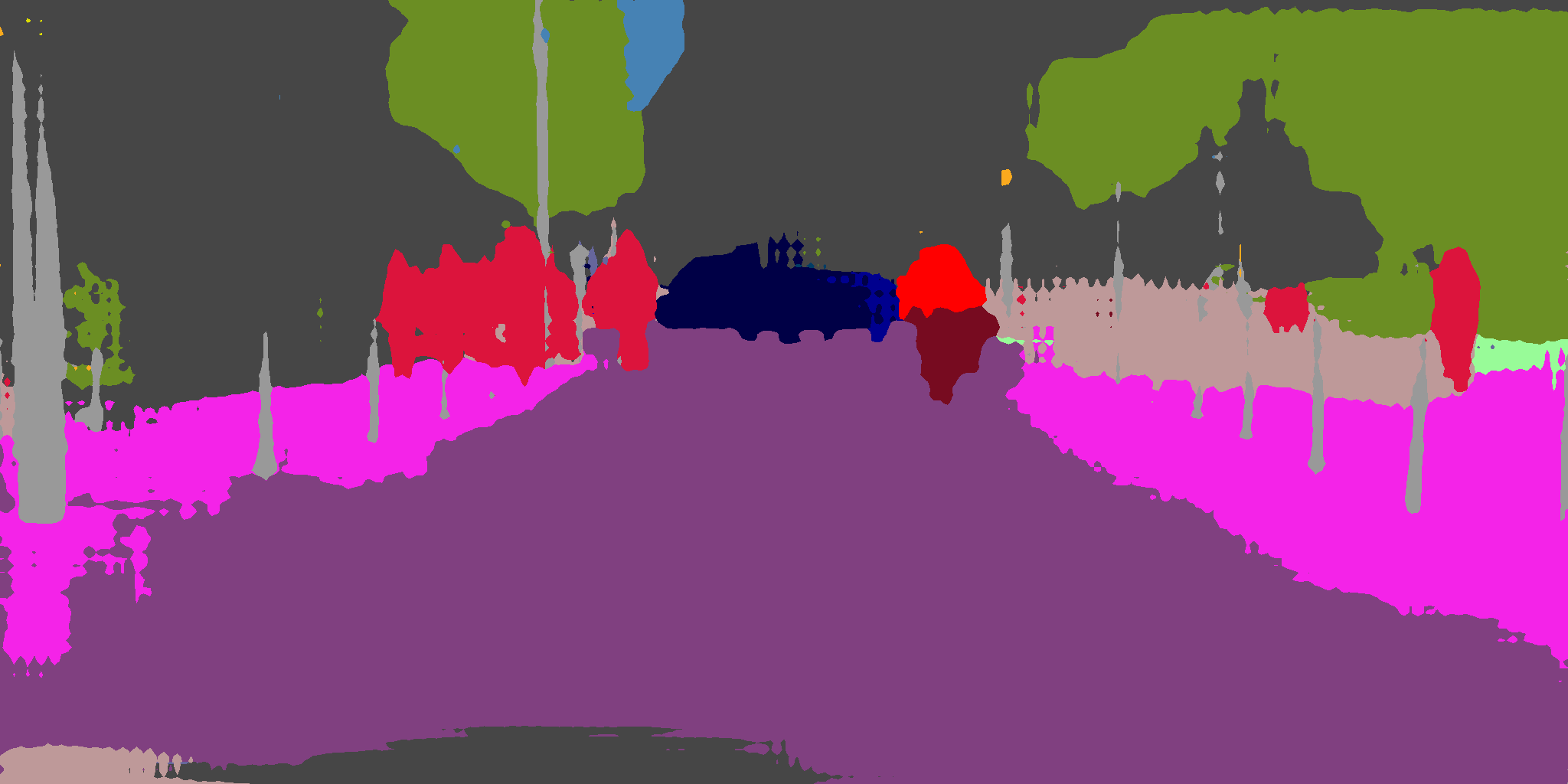}
	\caption{Adaptation results on GTA5 $\rightarrow$ Cityscapes. Rows correspond to sample images in Cityscapes. From top to bottom, rows correspond to original images, ground truth, and predication results of CBST, MRL2, MRENT, MRKLD, LRENT.}
	\label{fig:gta2city}
\end{figure*}

\begin{figure*}[!t]
	\centering
	\resizebox{0.98\textwidth}{!}{
		\begin{tabular}{@{}cccccccccc@{}}
			\cellcolor{city_color_1}\textcolor{white}{~~road~~} &
			\cellcolor{city_color_2}~~sidewalk~~&
			\cellcolor{city_color_3}\textcolor{white}{~~building~~} &
			\cellcolor{city_color_4}\textcolor{white}{~~wall~~} &
			\cellcolor{city_color_5}~~fence~~ &
			\cellcolor{city_color_6}~~pole~~ &
			\cellcolor{city_color_7}~~traffic lgt~~ &
			\cellcolor{city_color_8}~~traffic sgn~~ &
			\cellcolor{city_color_9}~~vegetation~~ & 
			\cellcolor{city_color_0}\textcolor{white}{~~ignored~~}\\
			\cellcolor{city_color_10}~~terrain~~ &
			\cellcolor{city_color_11}~~sky~~ &
			\cellcolor{city_color_12}\textcolor{white}{~~person~~} &
			\cellcolor{city_color_13}\textcolor{white}{~~rider~~} &
			\cellcolor{city_color_14}\textcolor{white}{~~car~~} &
			\cellcolor{city_color_15}\textcolor{white}{~~truck~~} &
			\cellcolor{city_color_16}\textcolor{white}{~~bus~~} &
			\cellcolor{city_color_17}\textcolor{white}{~~train~~} &
			\cellcolor{city_color_18}\textcolor{white}{~~motorcycle~~} &
			\cellcolor{city_color_19}\textcolor{white}{~~bike~~}
		\end{tabular}
	}
	
	\vspace{1mm}
	\includegraphics[width=0.24\textwidth]{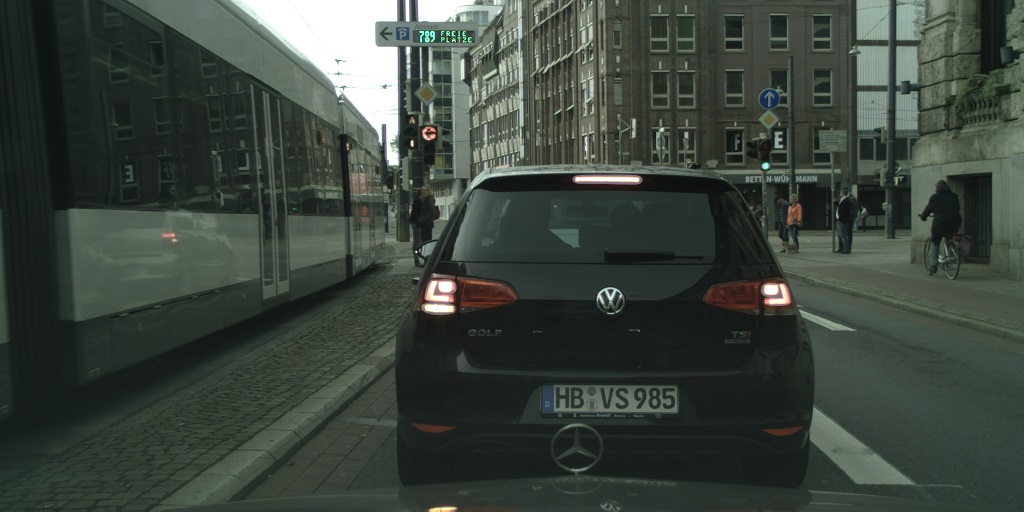}
	\includegraphics[width=0.24\textwidth]{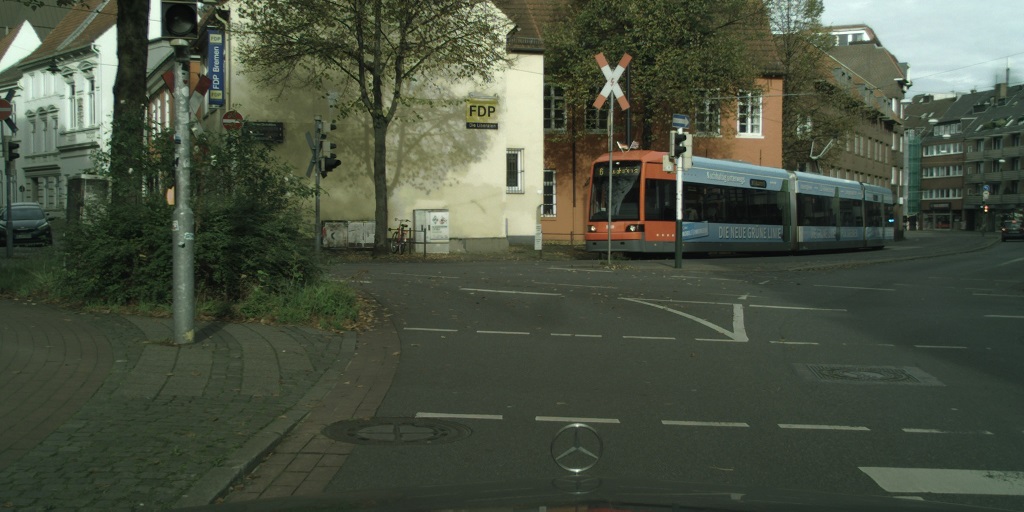}
	\includegraphics[width=0.24\textwidth]{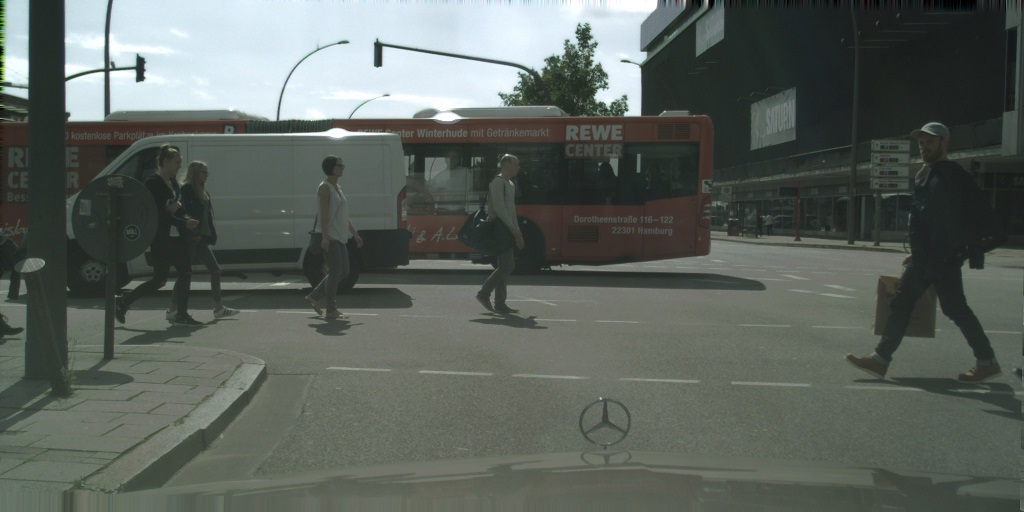}
	\includegraphics[width=0.24\textwidth]{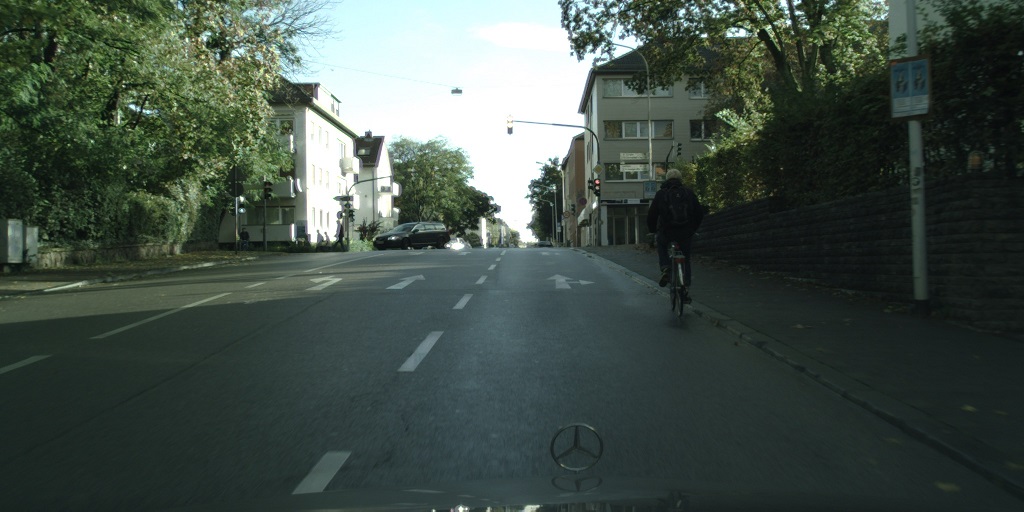}
	\quad\\\vspace{0.5mm}
	\includegraphics[width=0.24\textwidth]{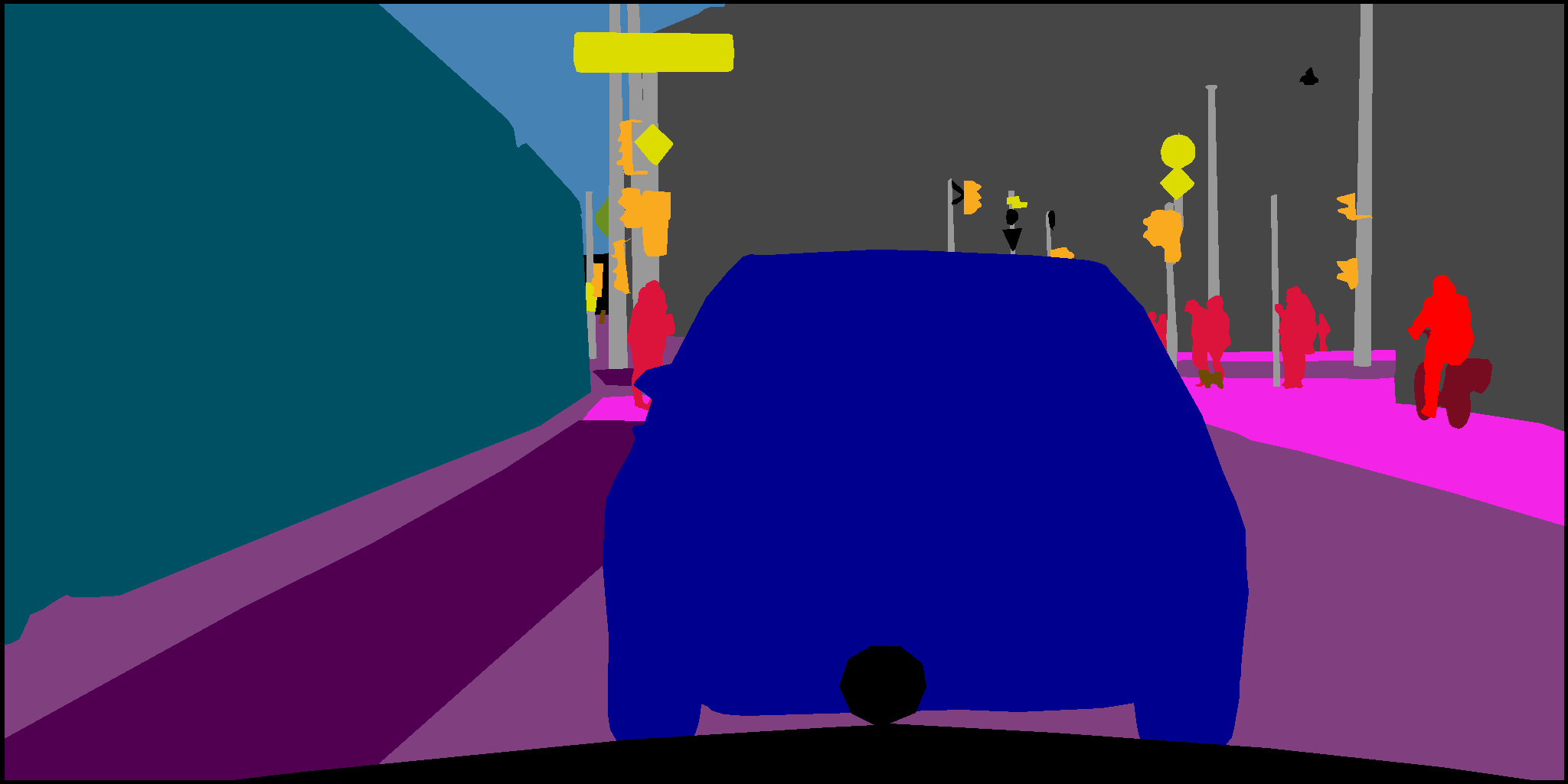}
	\includegraphics[width=0.24\textwidth]{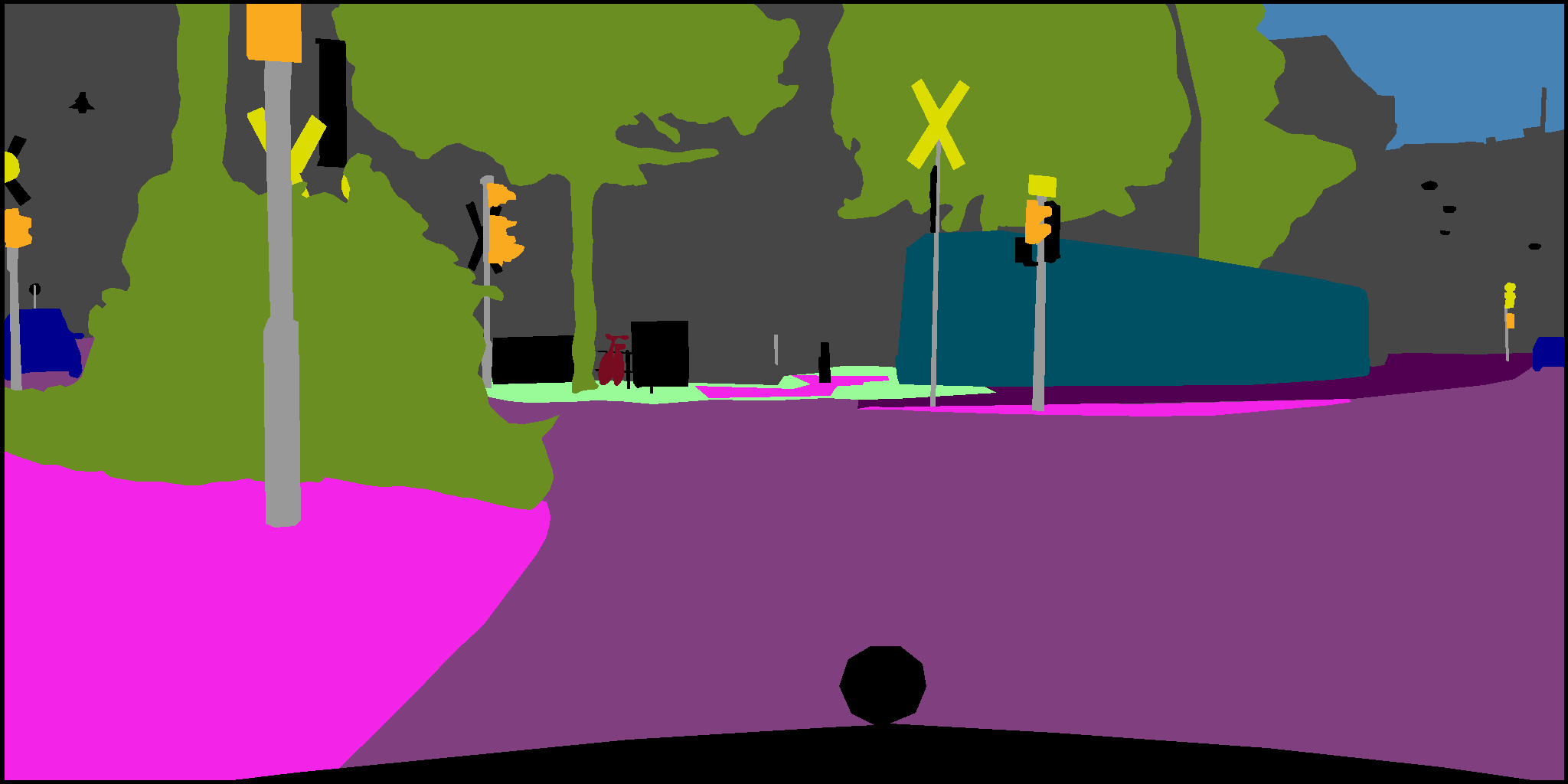}
	\includegraphics[width=0.24\textwidth]{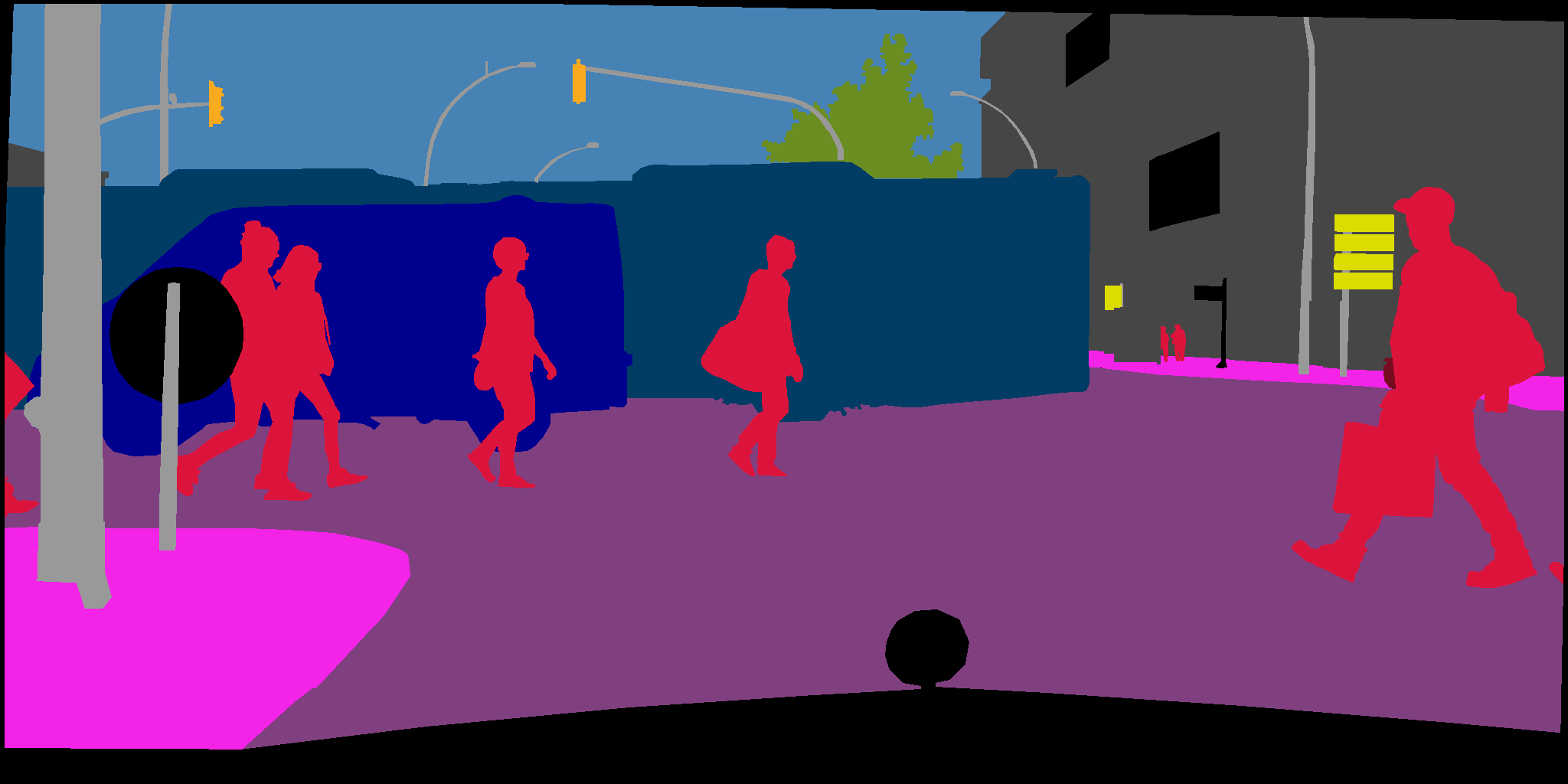}
	\includegraphics[width=0.24\textwidth]{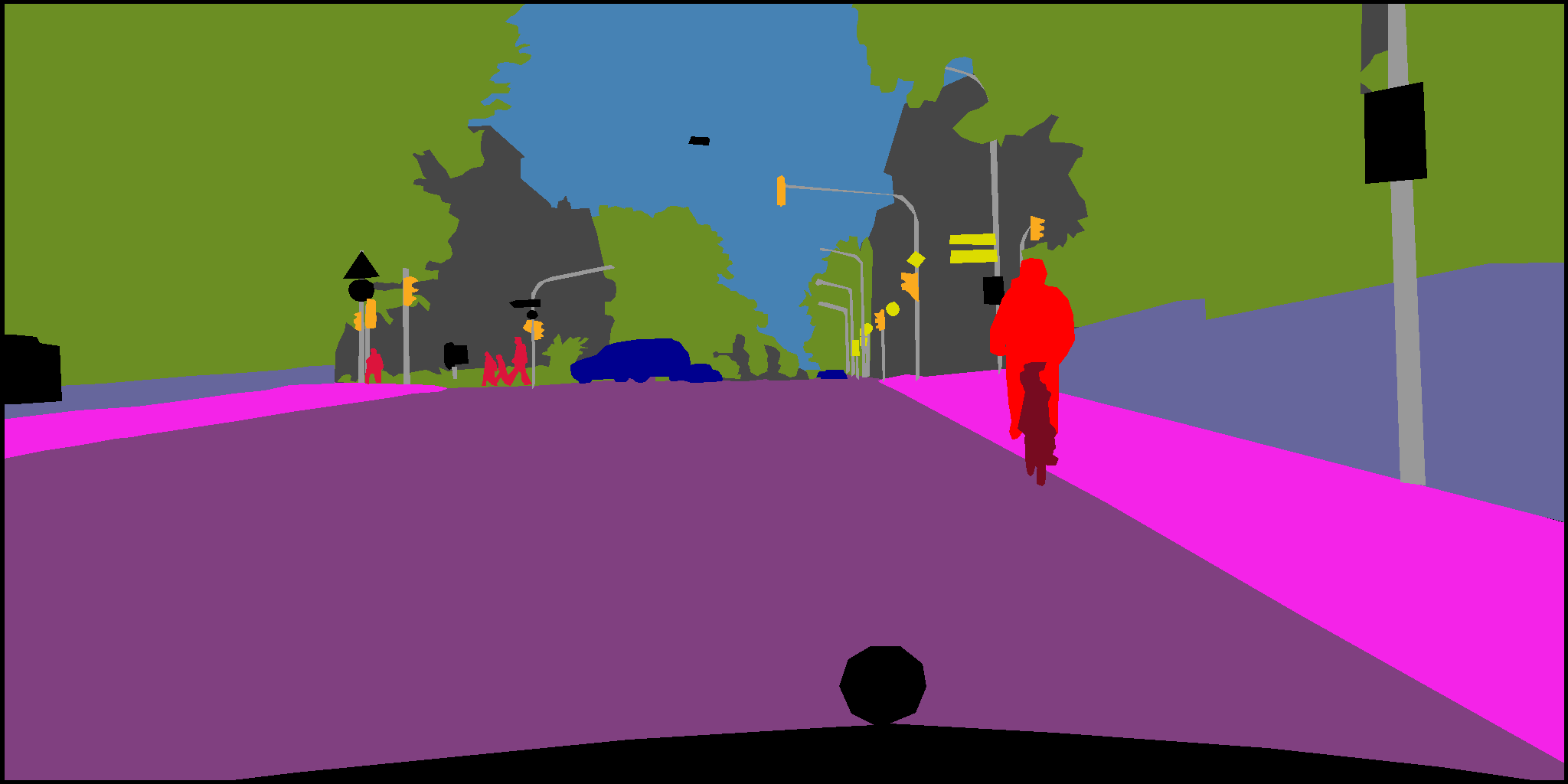}
	\quad\\\vspace{0.5mm}
	\includegraphics[width=0.24\textwidth]{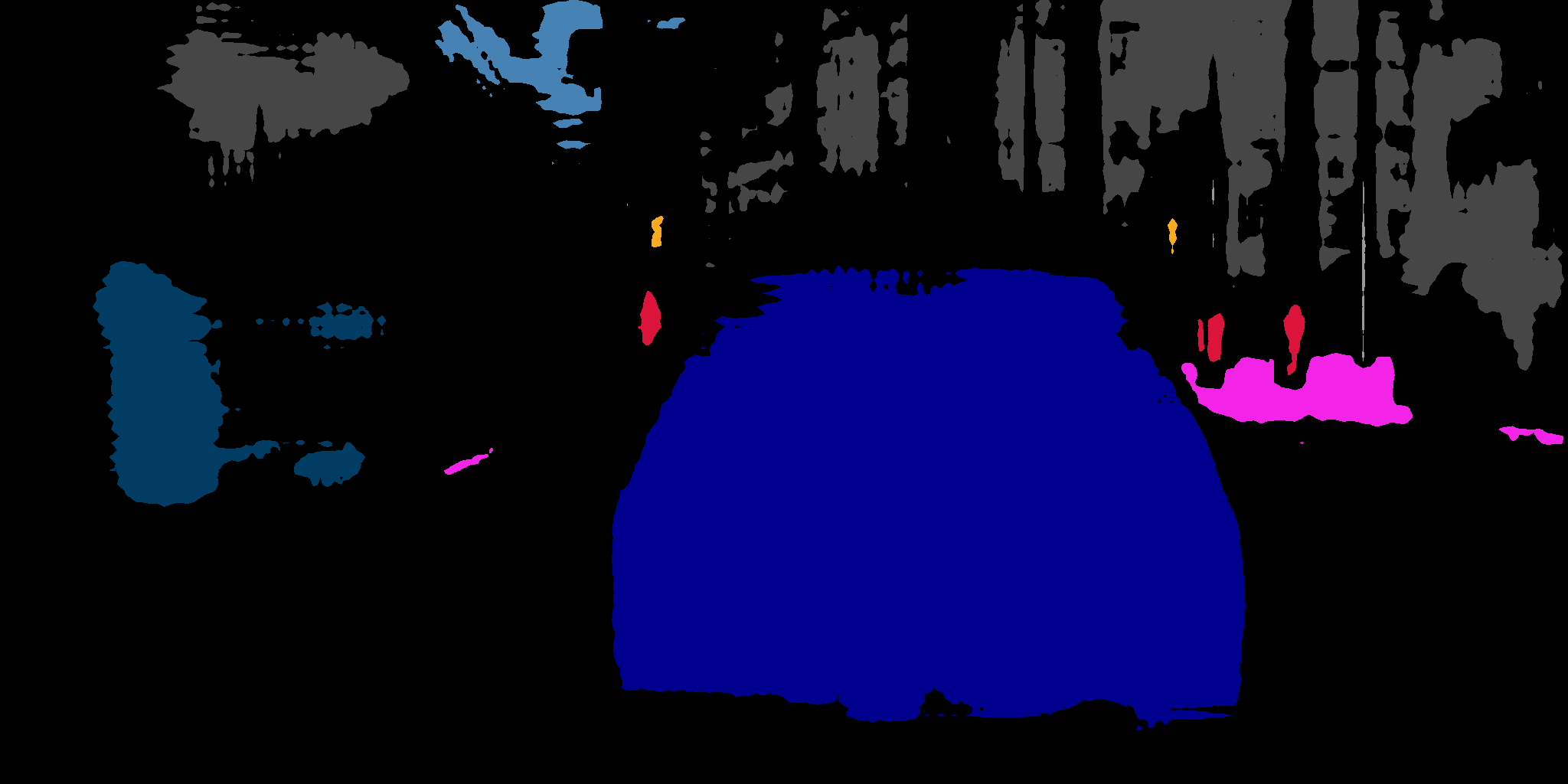}
	\includegraphics[width=0.24\textwidth]{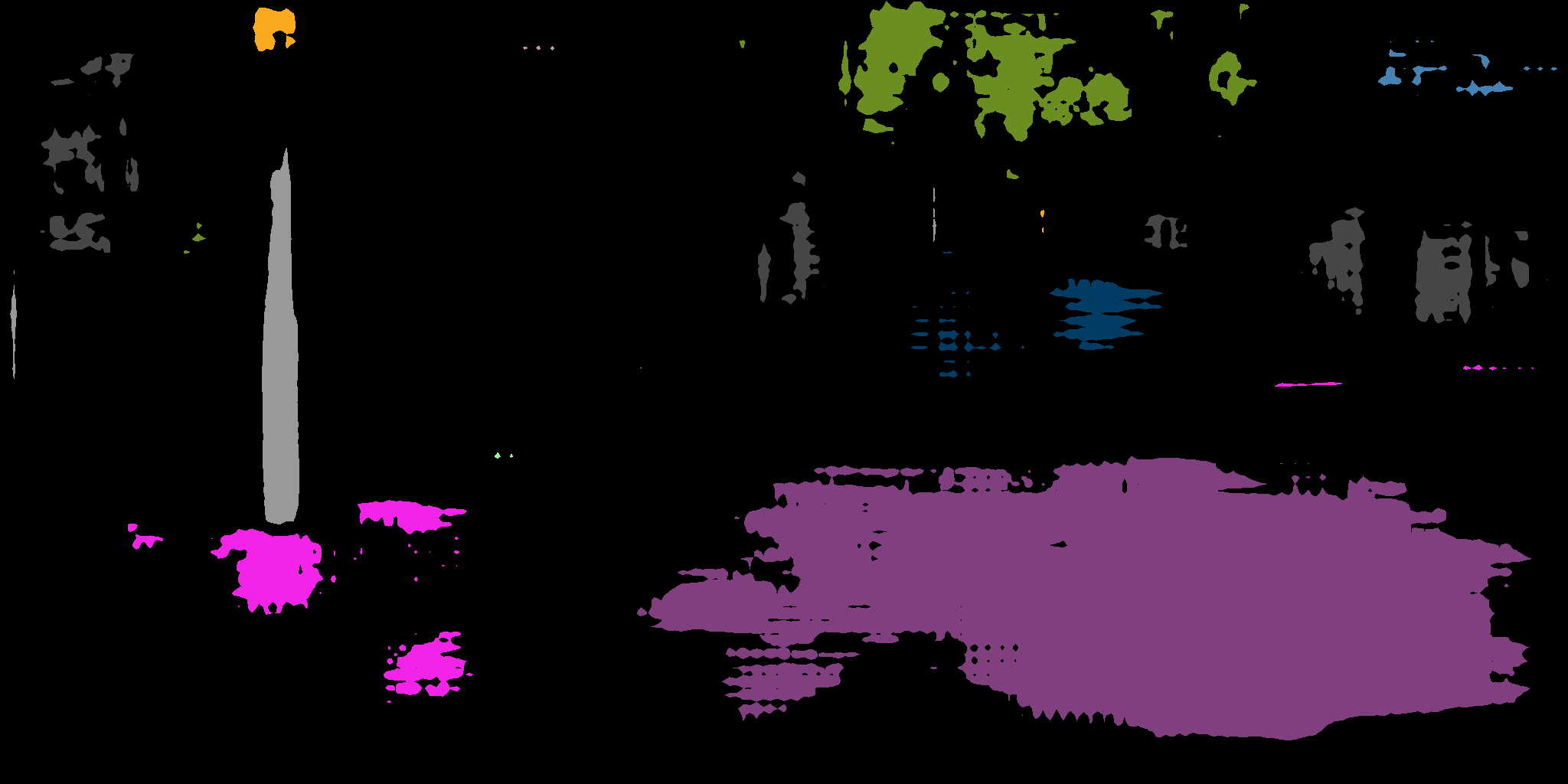}
	\includegraphics[width=0.24\textwidth]{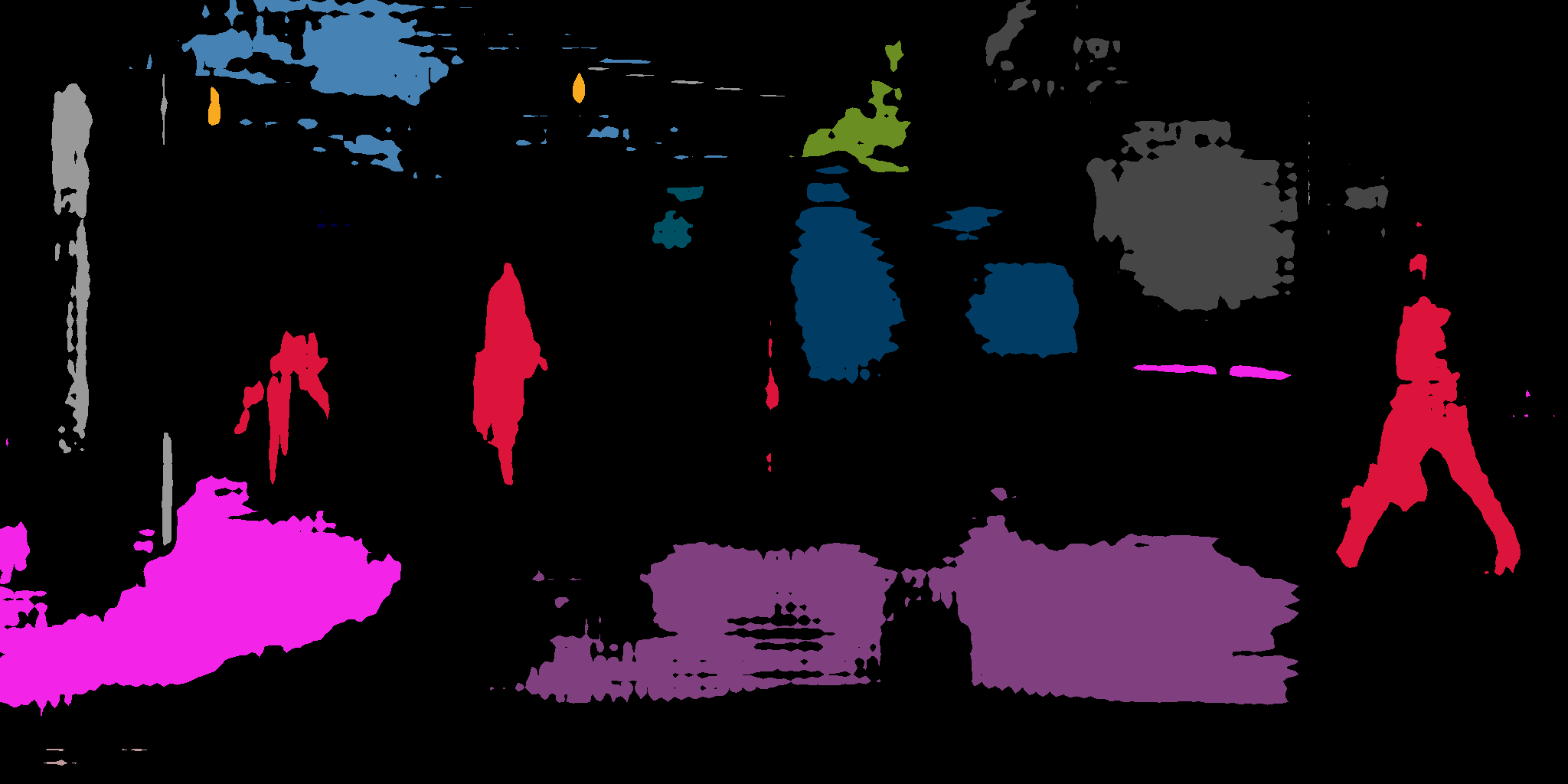}
	\includegraphics[width=0.24\textwidth]{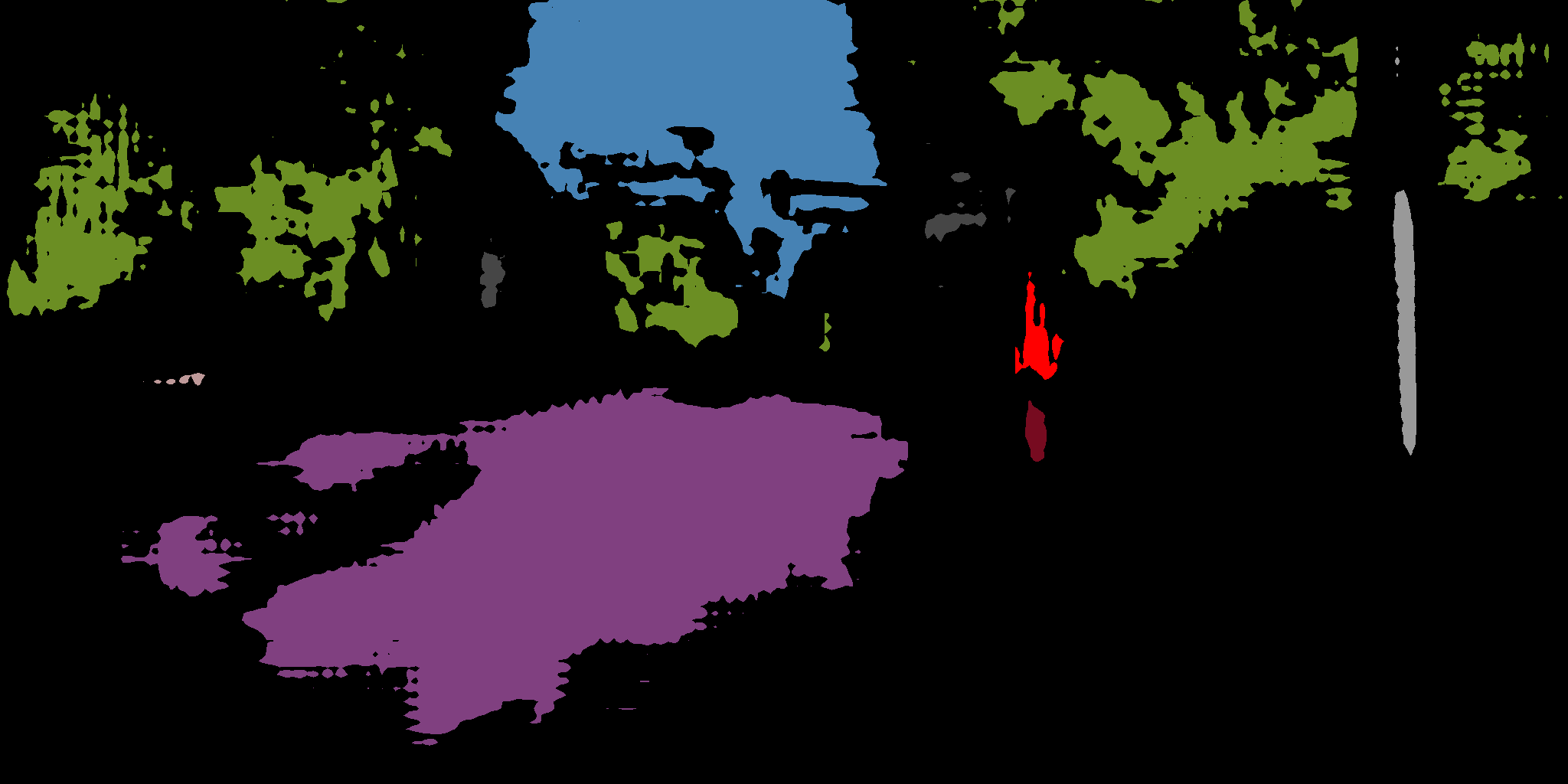}
	\quad\\\vspace{0.5mm}
	\includegraphics[width=0.24\textwidth]{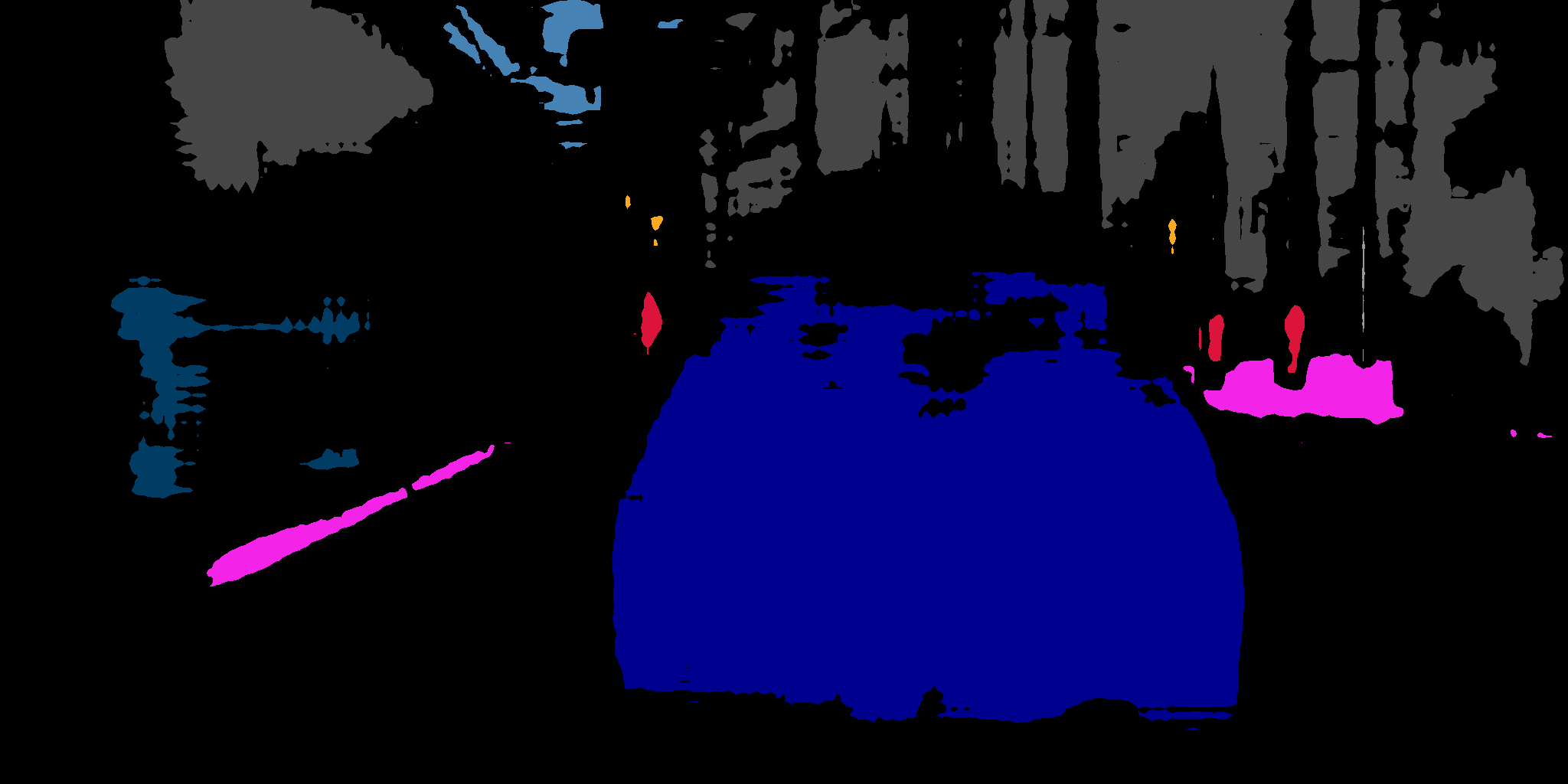}
	\includegraphics[width=0.24\textwidth]{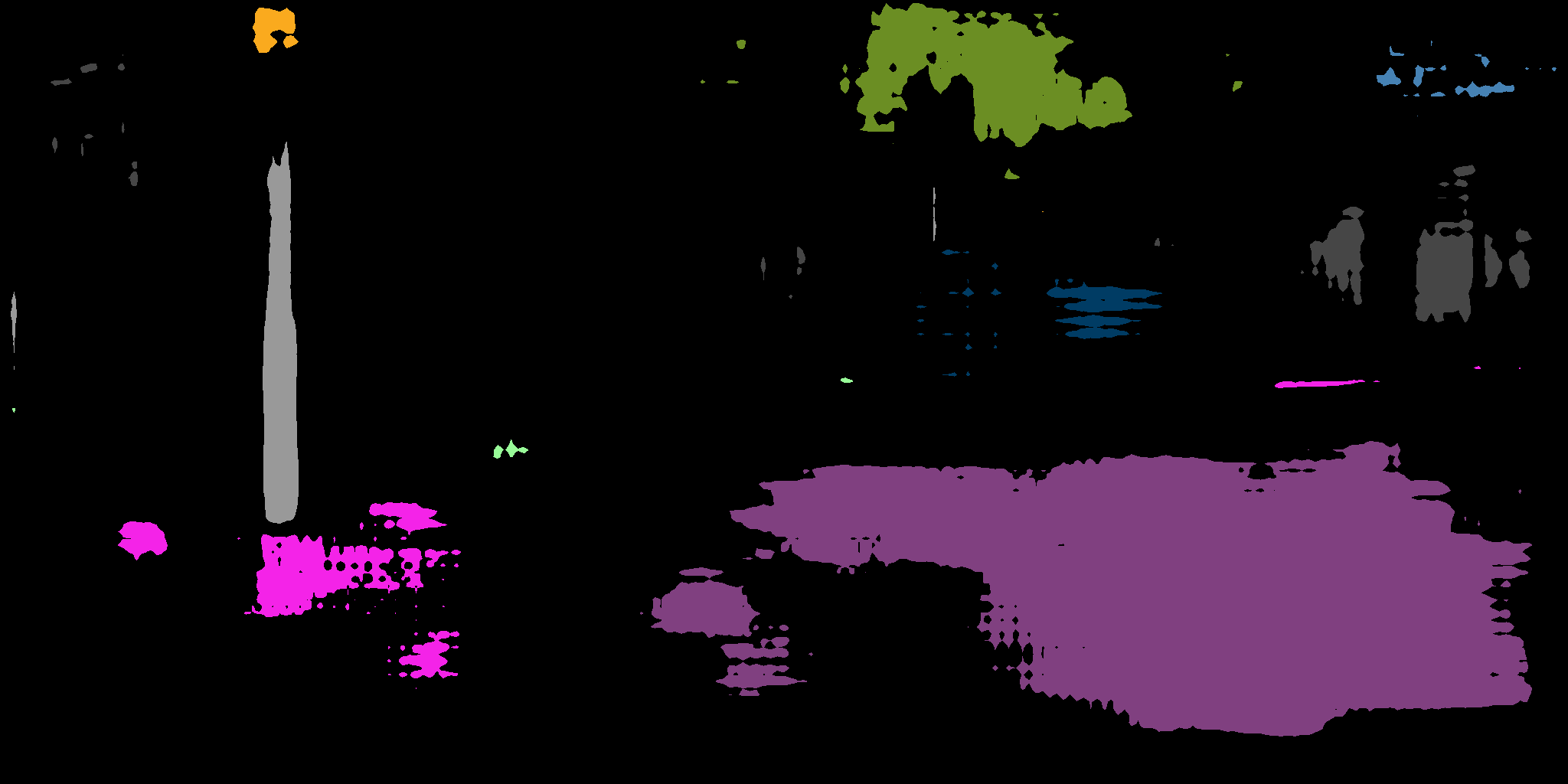}
	\includegraphics[width=0.24\textwidth]{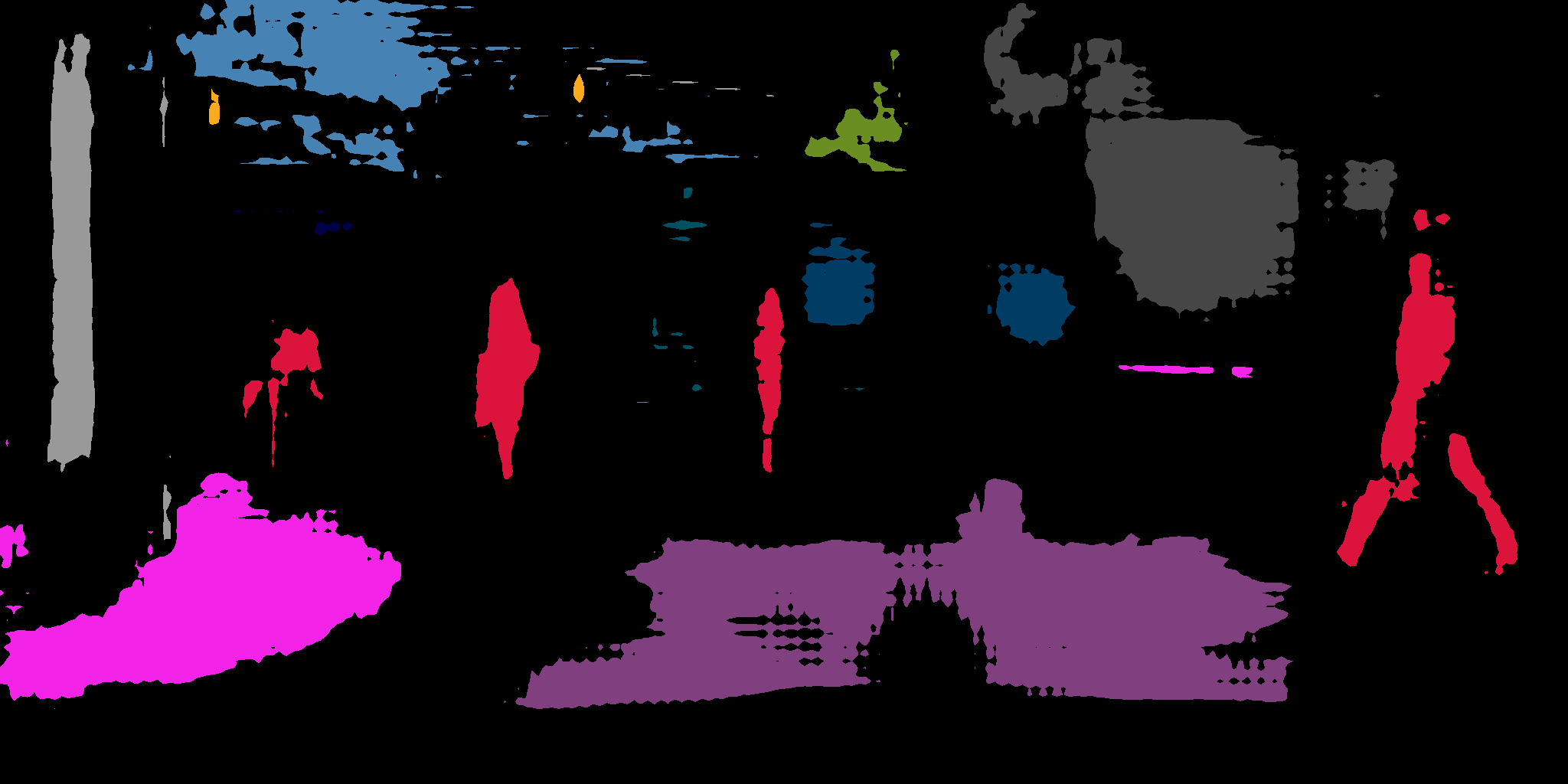}
	\includegraphics[width=0.24\textwidth]{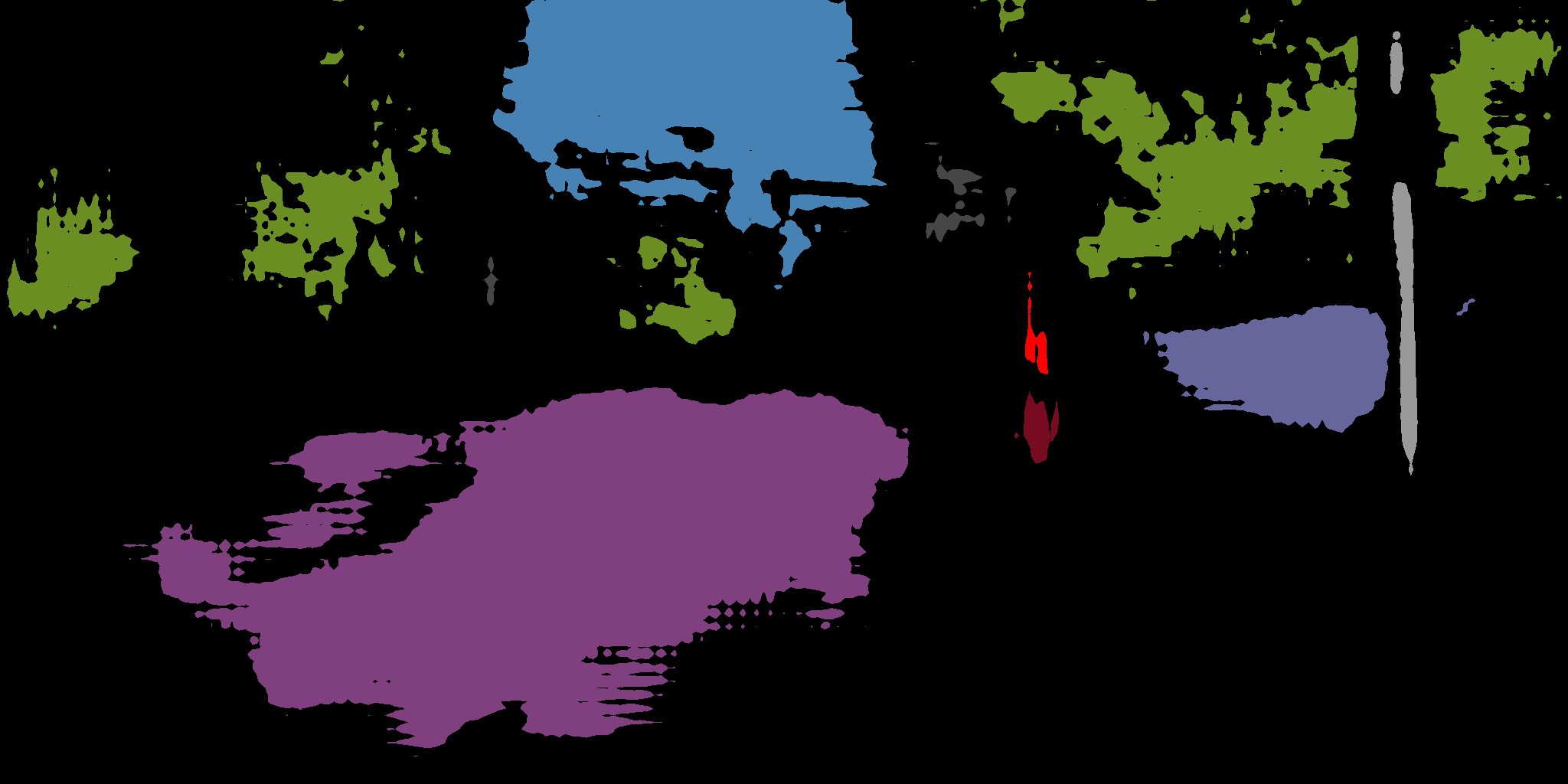}
	\quad\\\vspace{0.5mm}
	\includegraphics[width=0.24\textwidth]{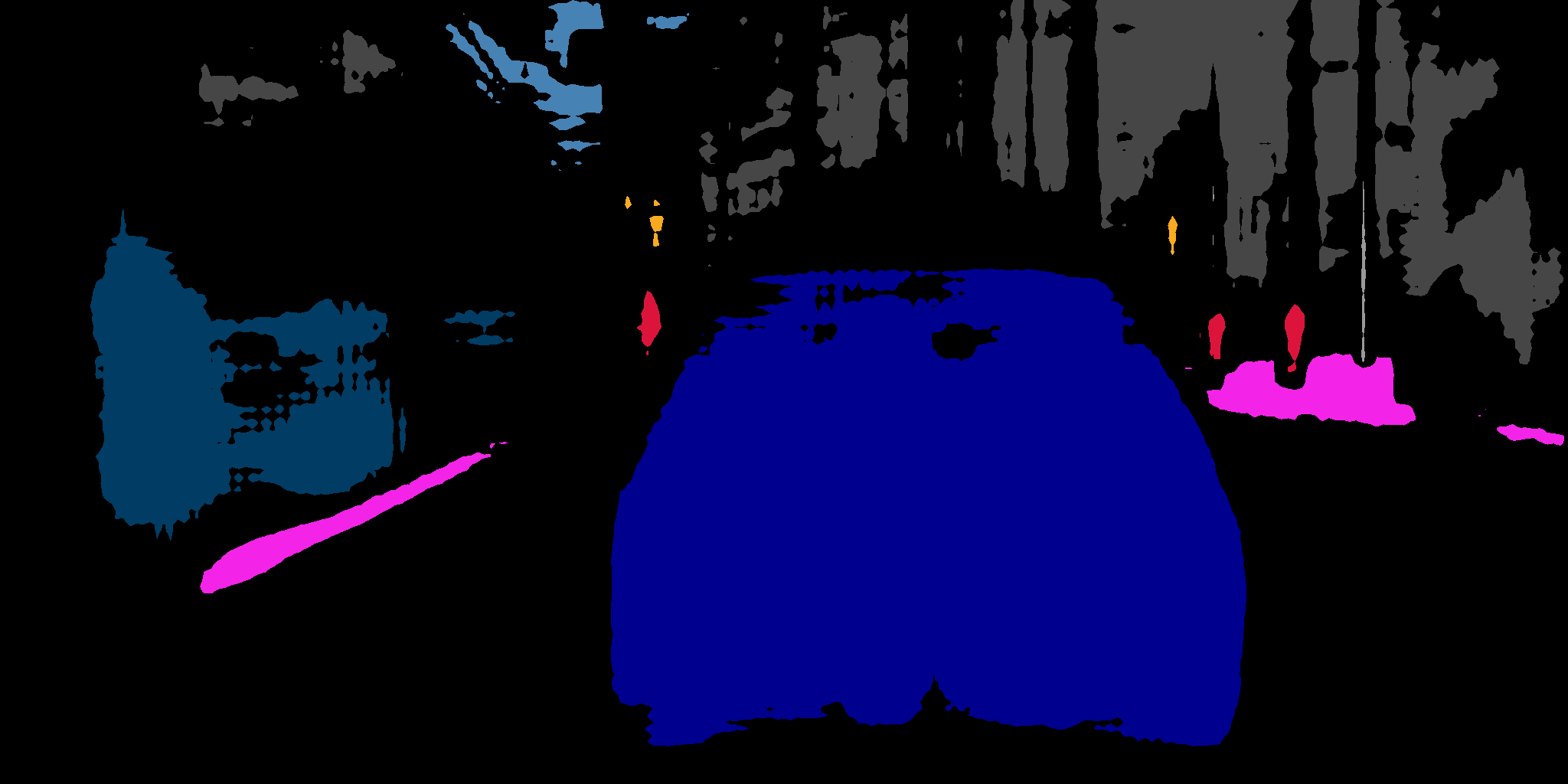}
	\includegraphics[width=0.24\textwidth]{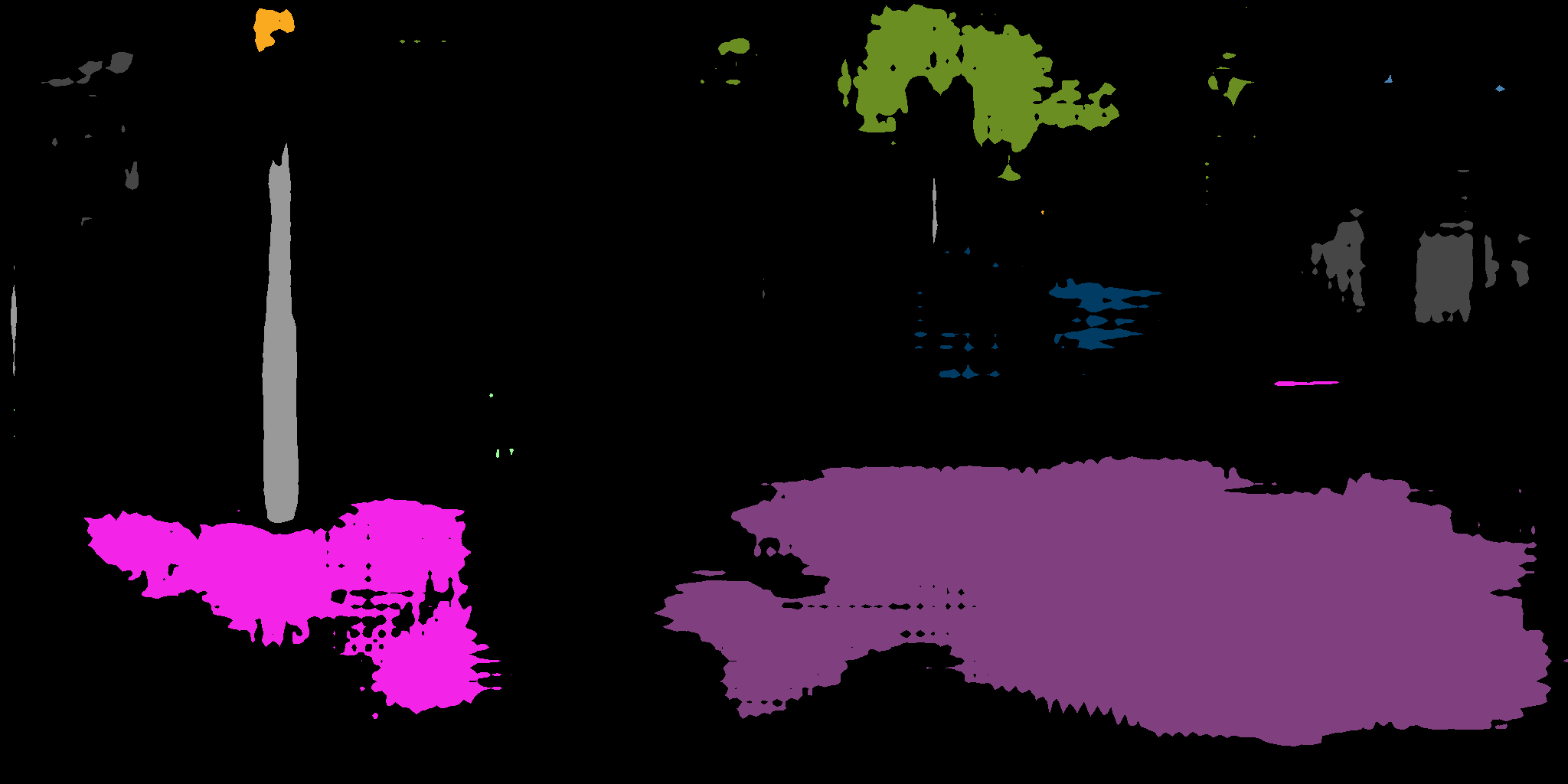}
	\includegraphics[width=0.24\textwidth]{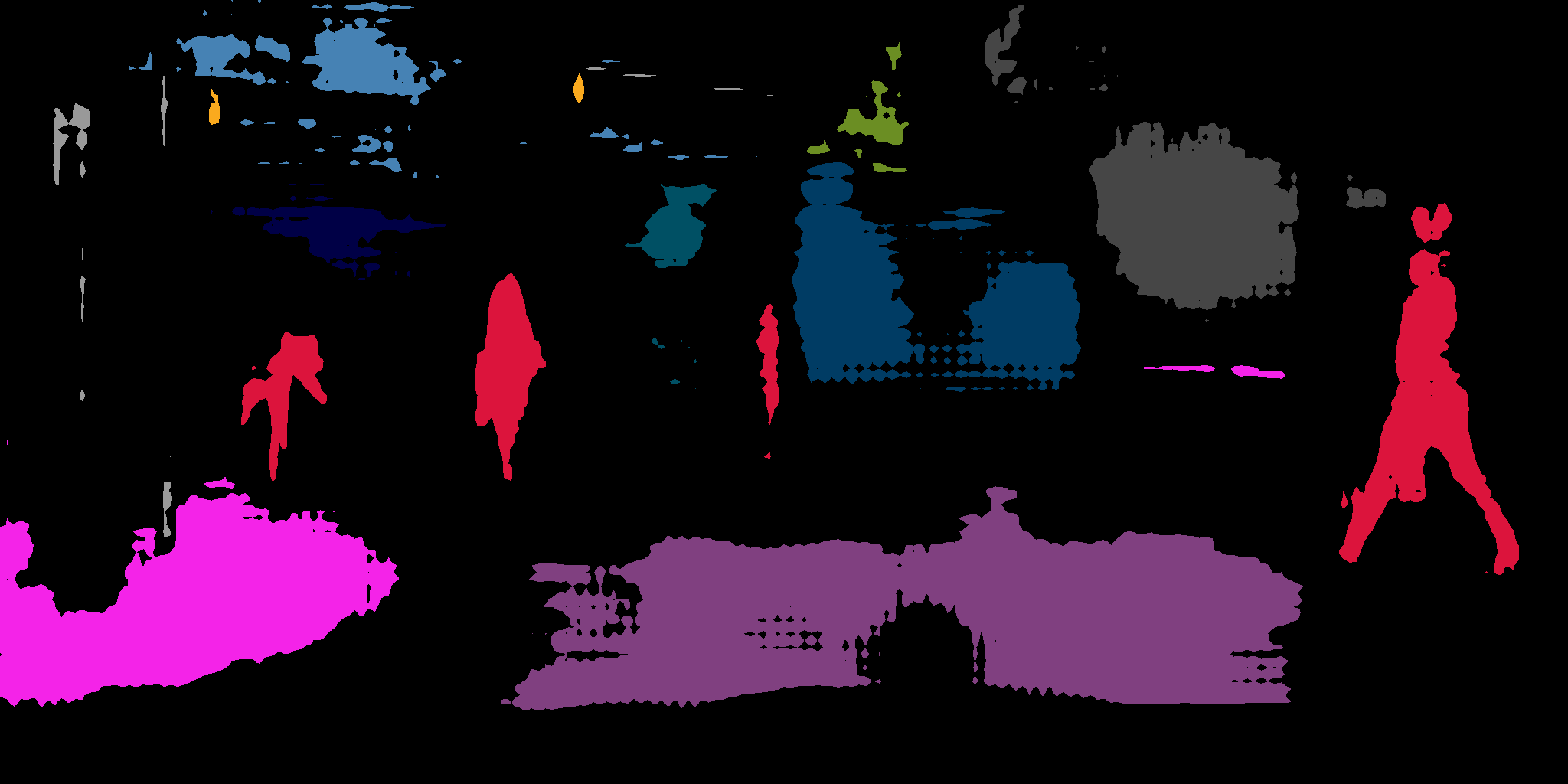}
	\includegraphics[width=0.24\textwidth]{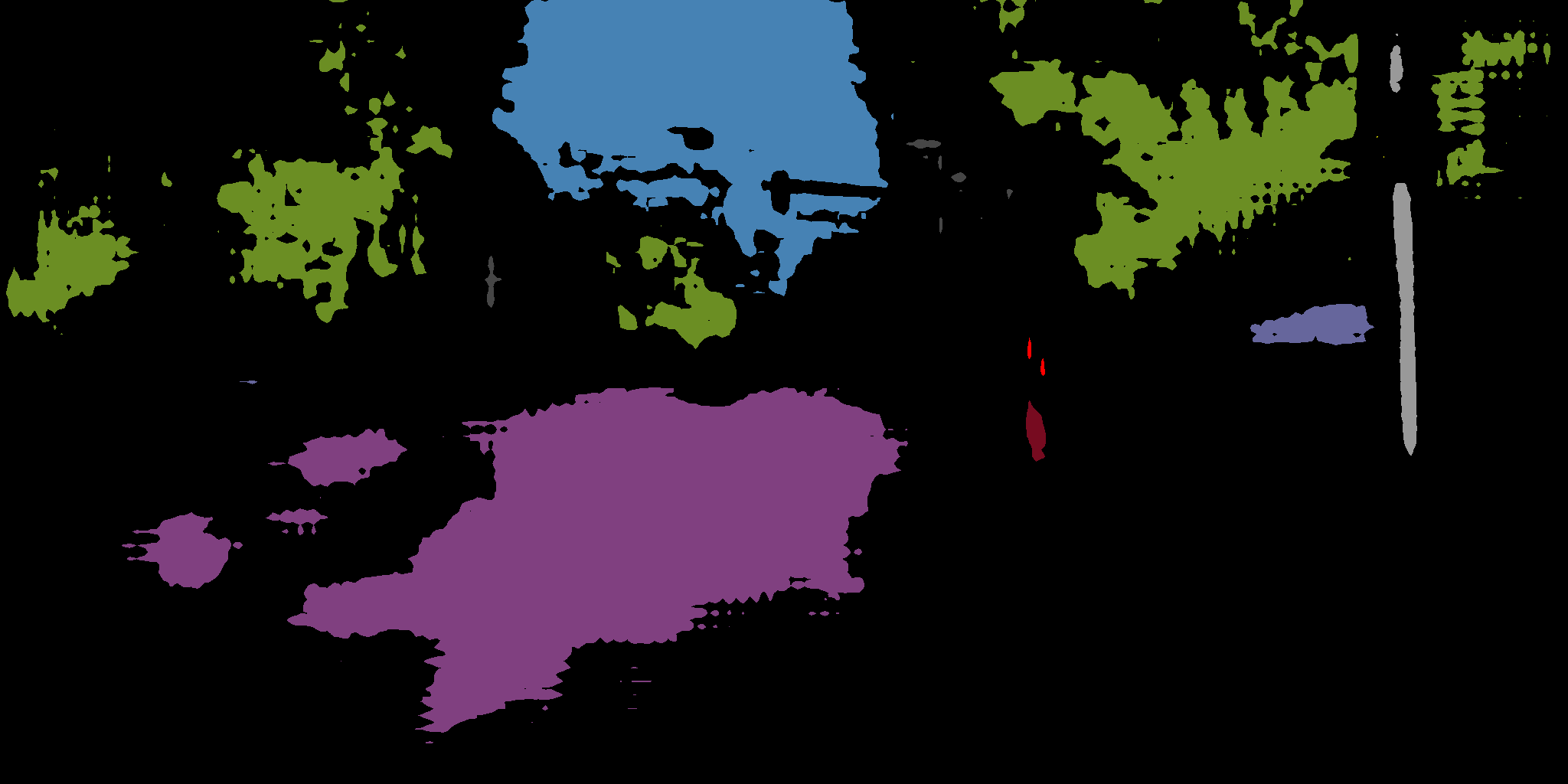}
	\quad\\\vspace{0.5mm}
	\includegraphics[width=0.24\textwidth]{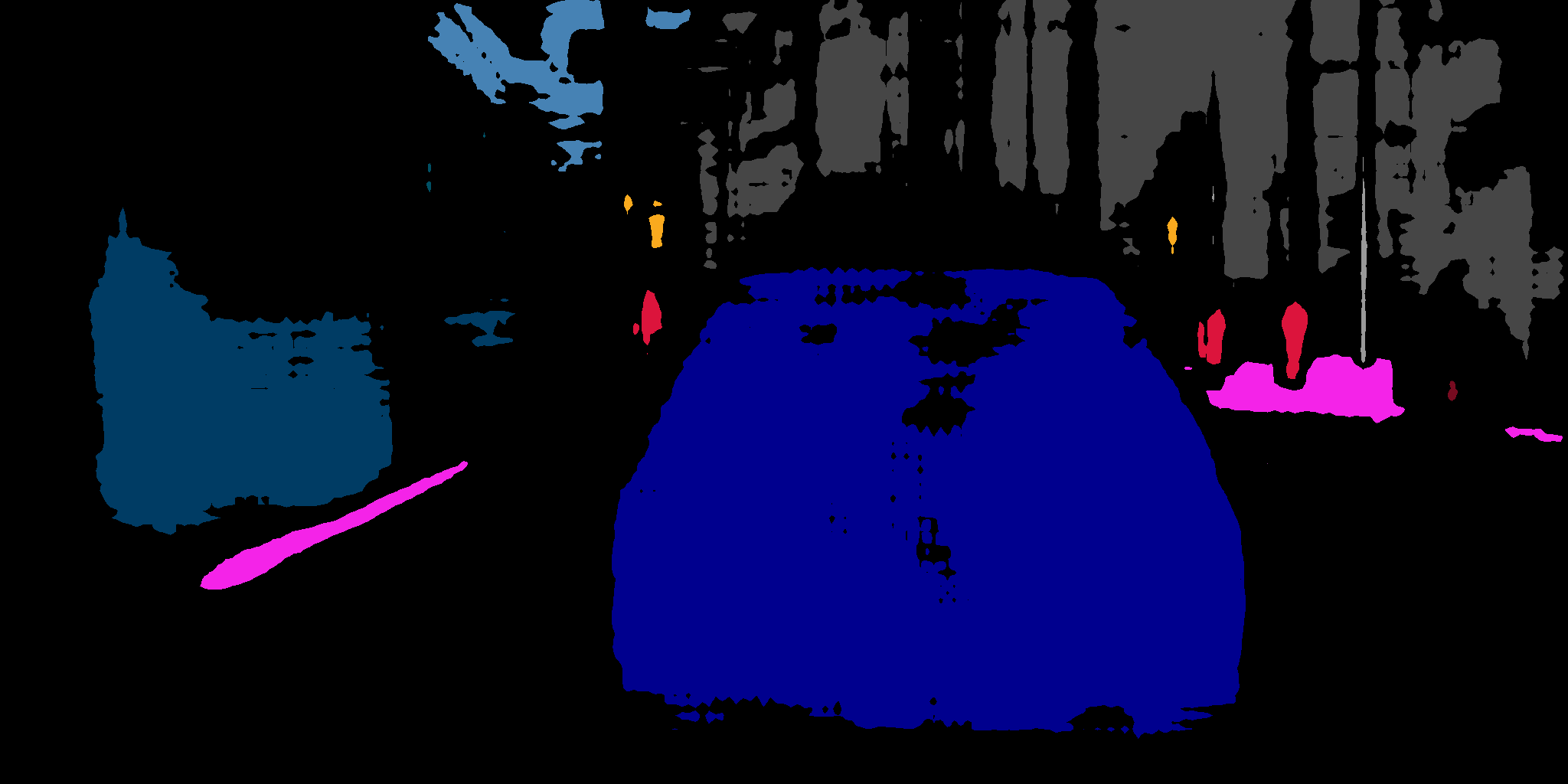}
	\includegraphics[width=0.24\textwidth]{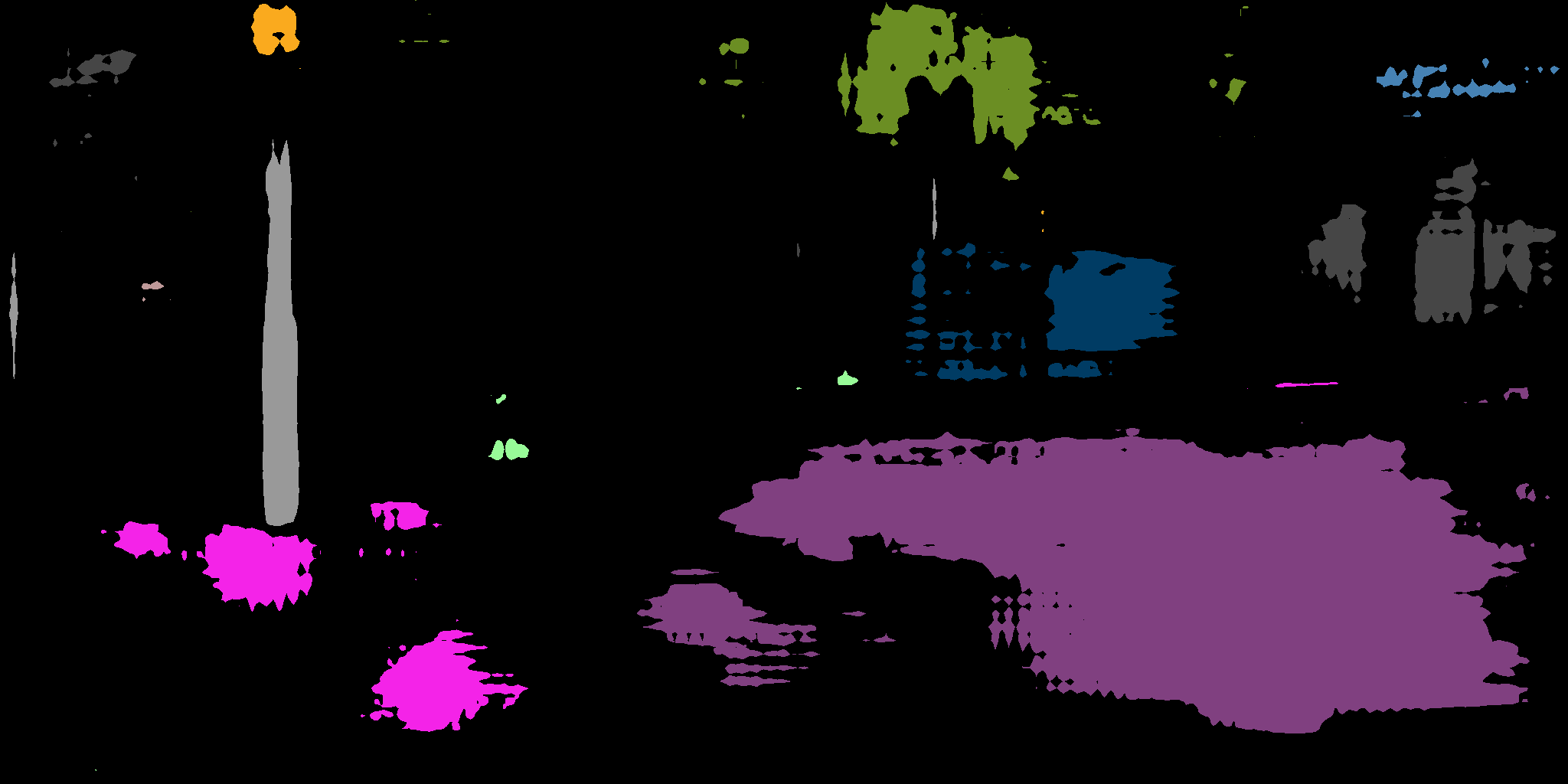}
	\includegraphics[width=0.24\textwidth]{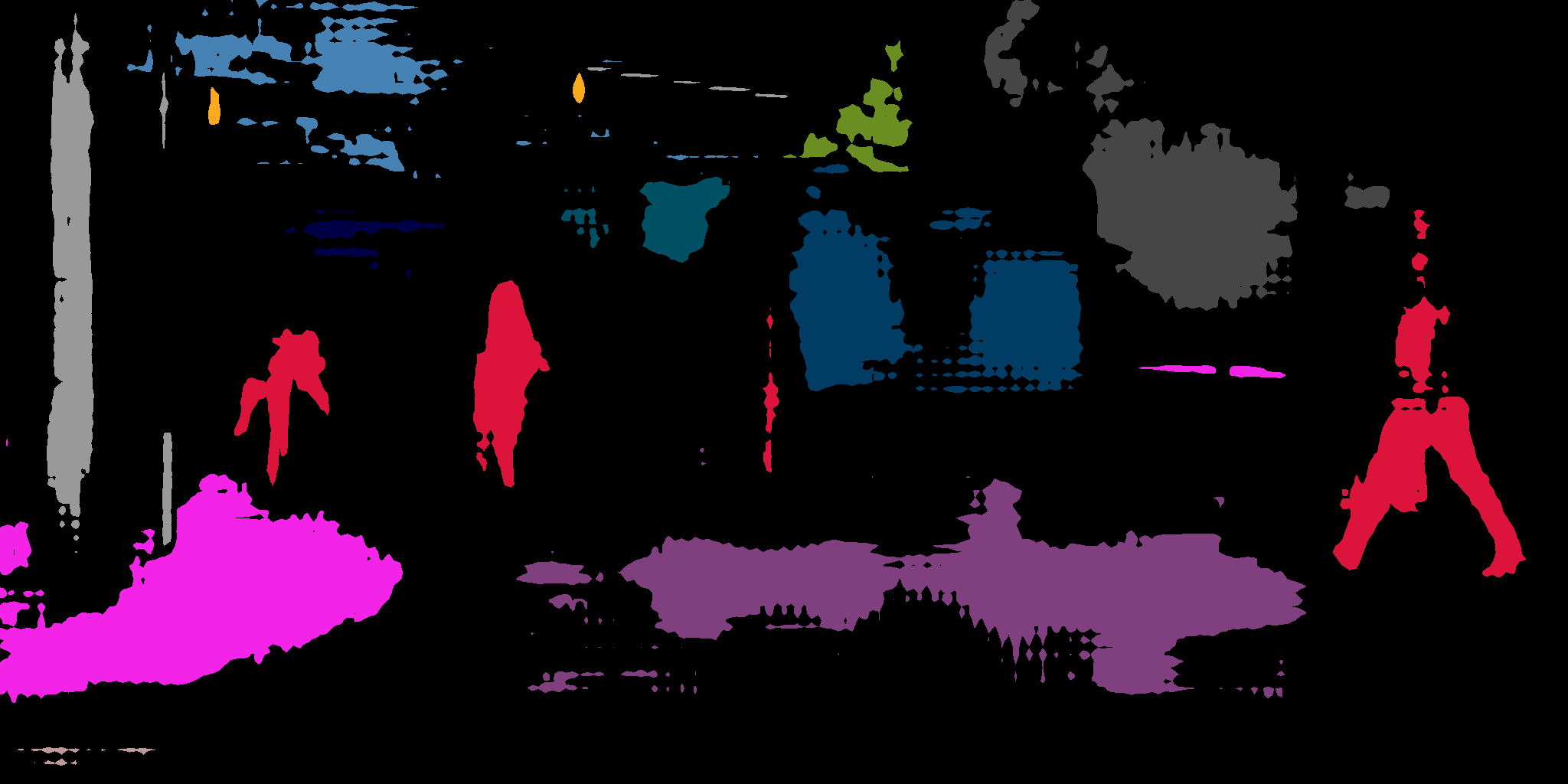}
	\includegraphics[width=0.24\textwidth]{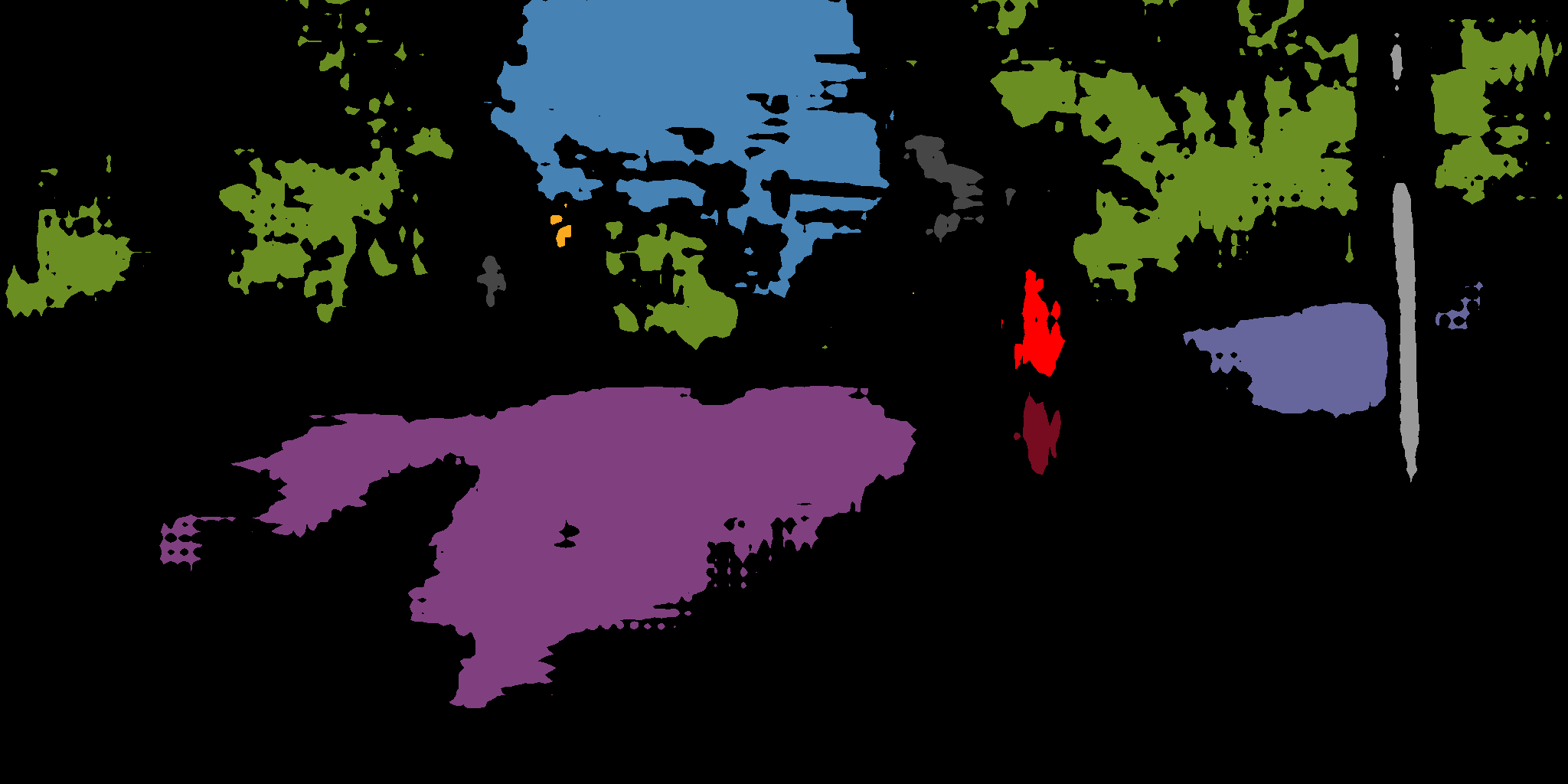}
	\quad\\\vspace{0.5mm}
	\includegraphics[width=0.24\textwidth]{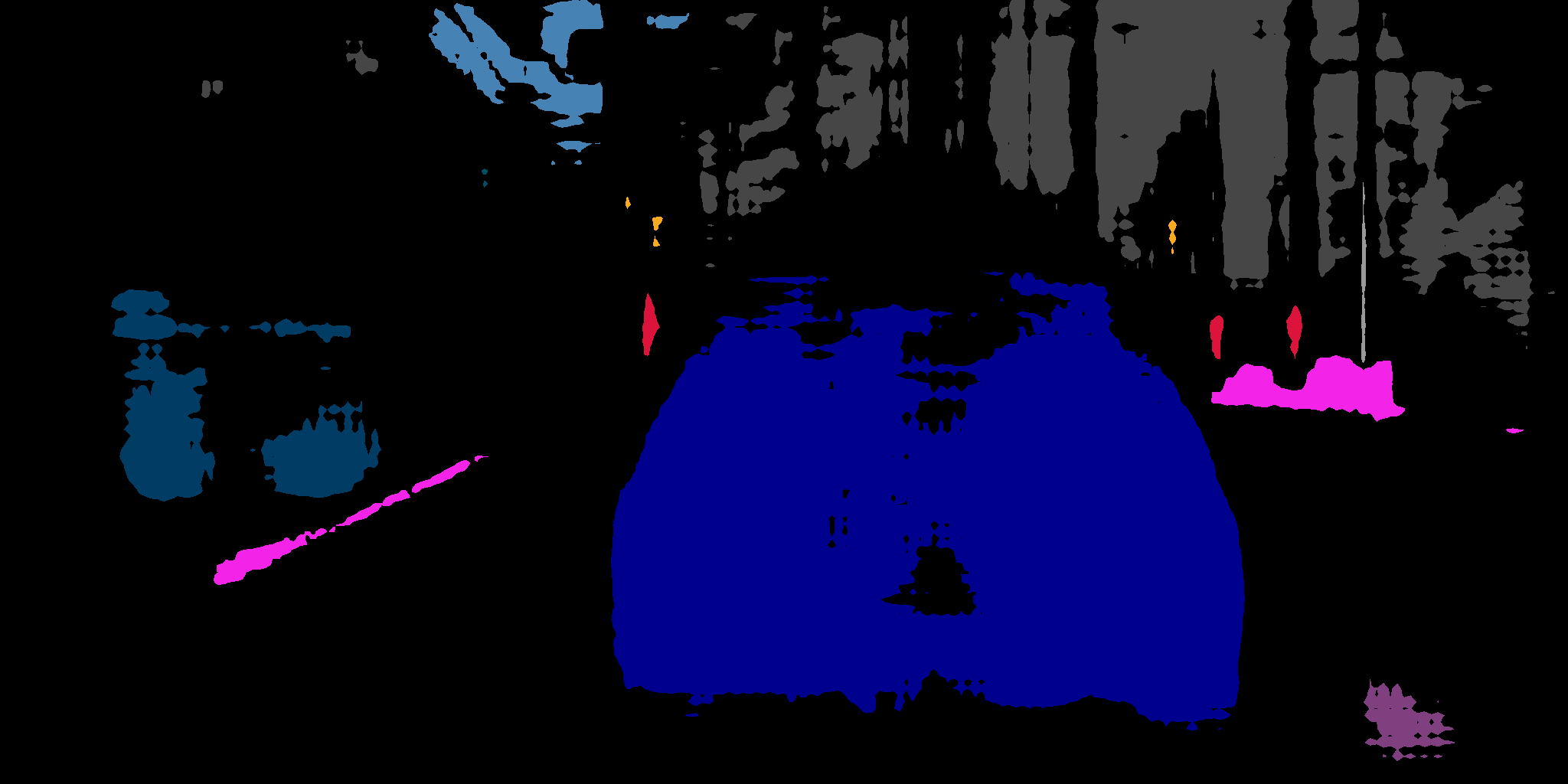}
	\includegraphics[width=0.24\textwidth]{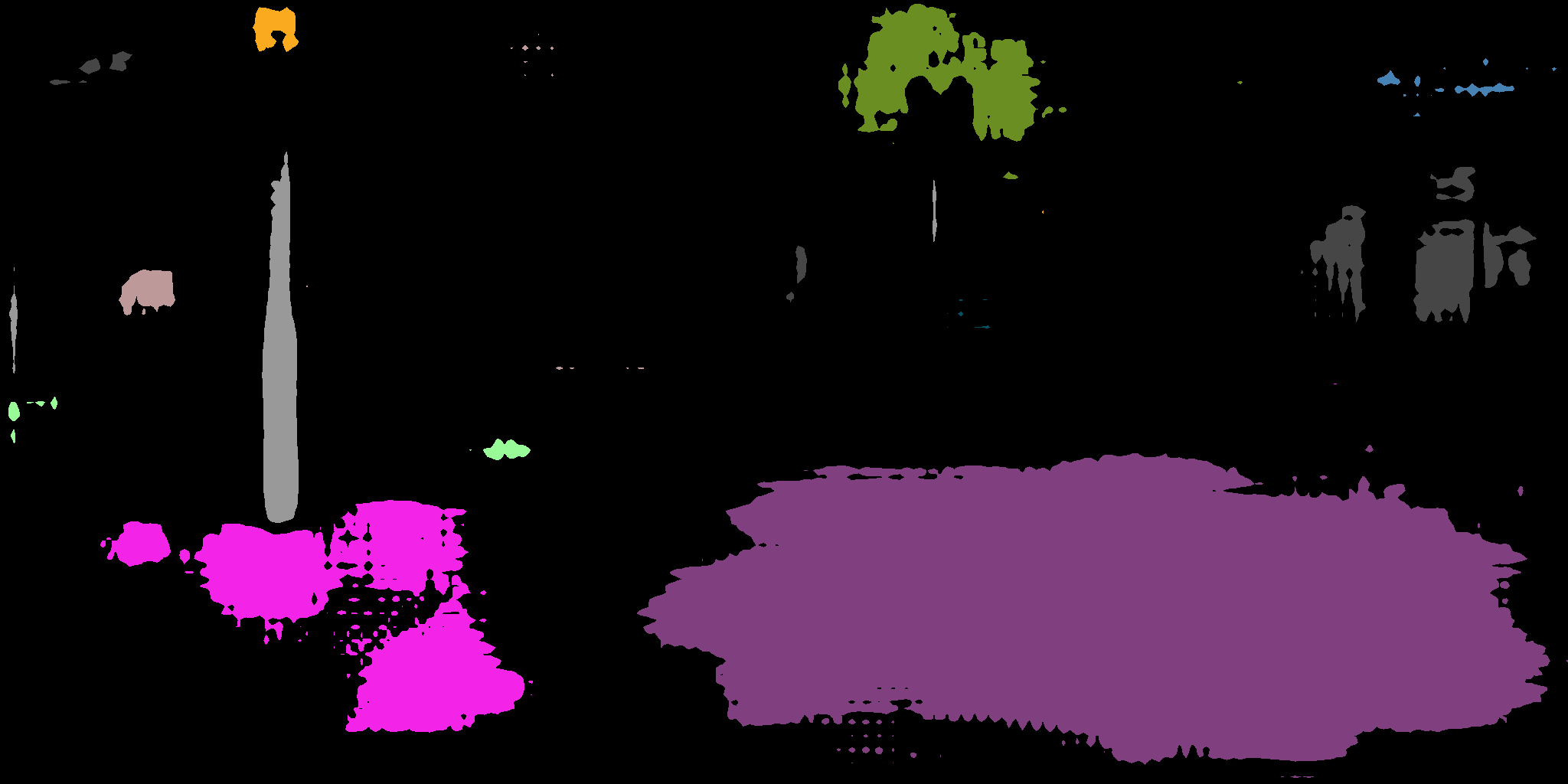}
	\includegraphics[width=0.24\textwidth]{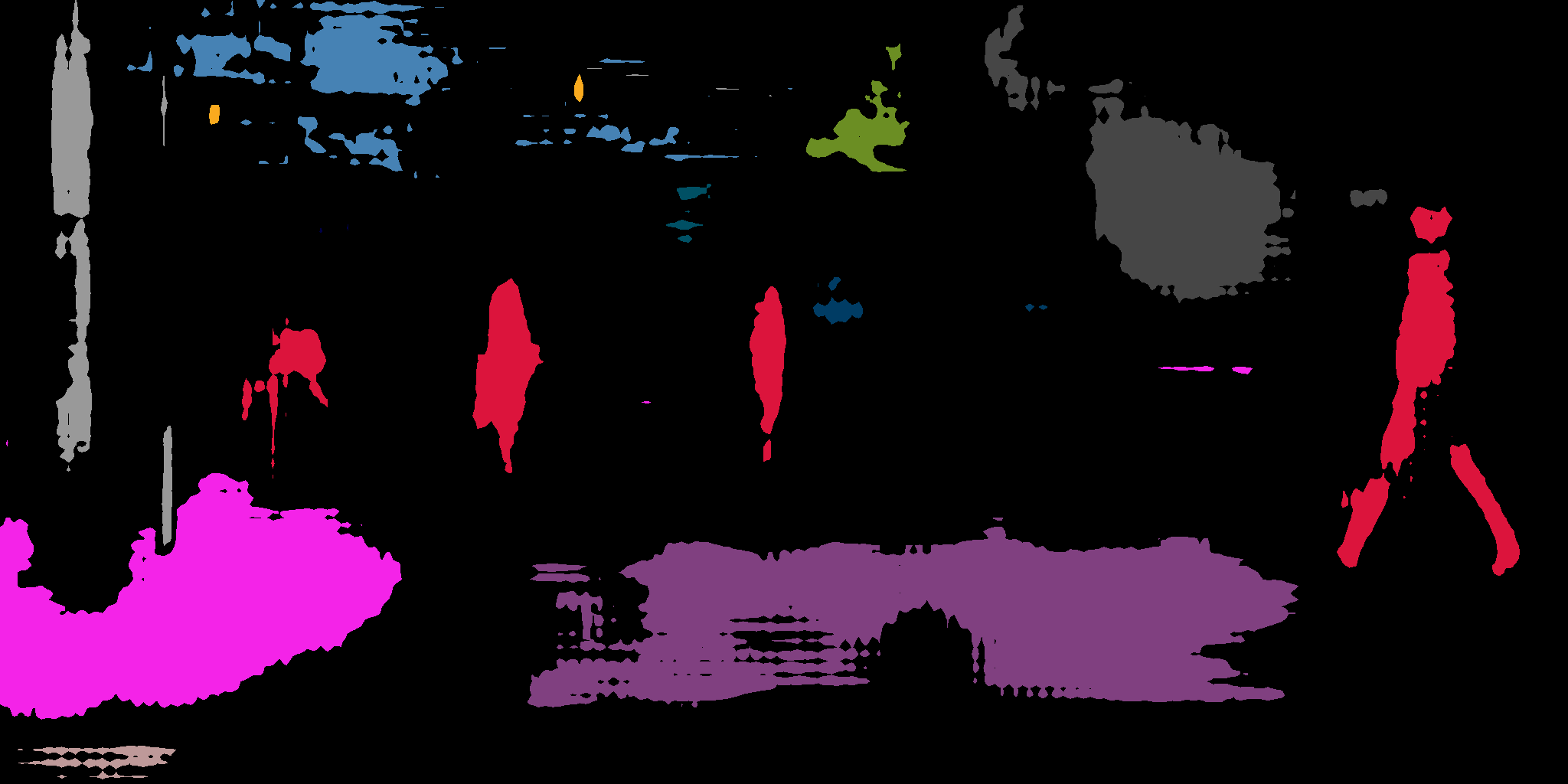}
	\includegraphics[width=0.24\textwidth]{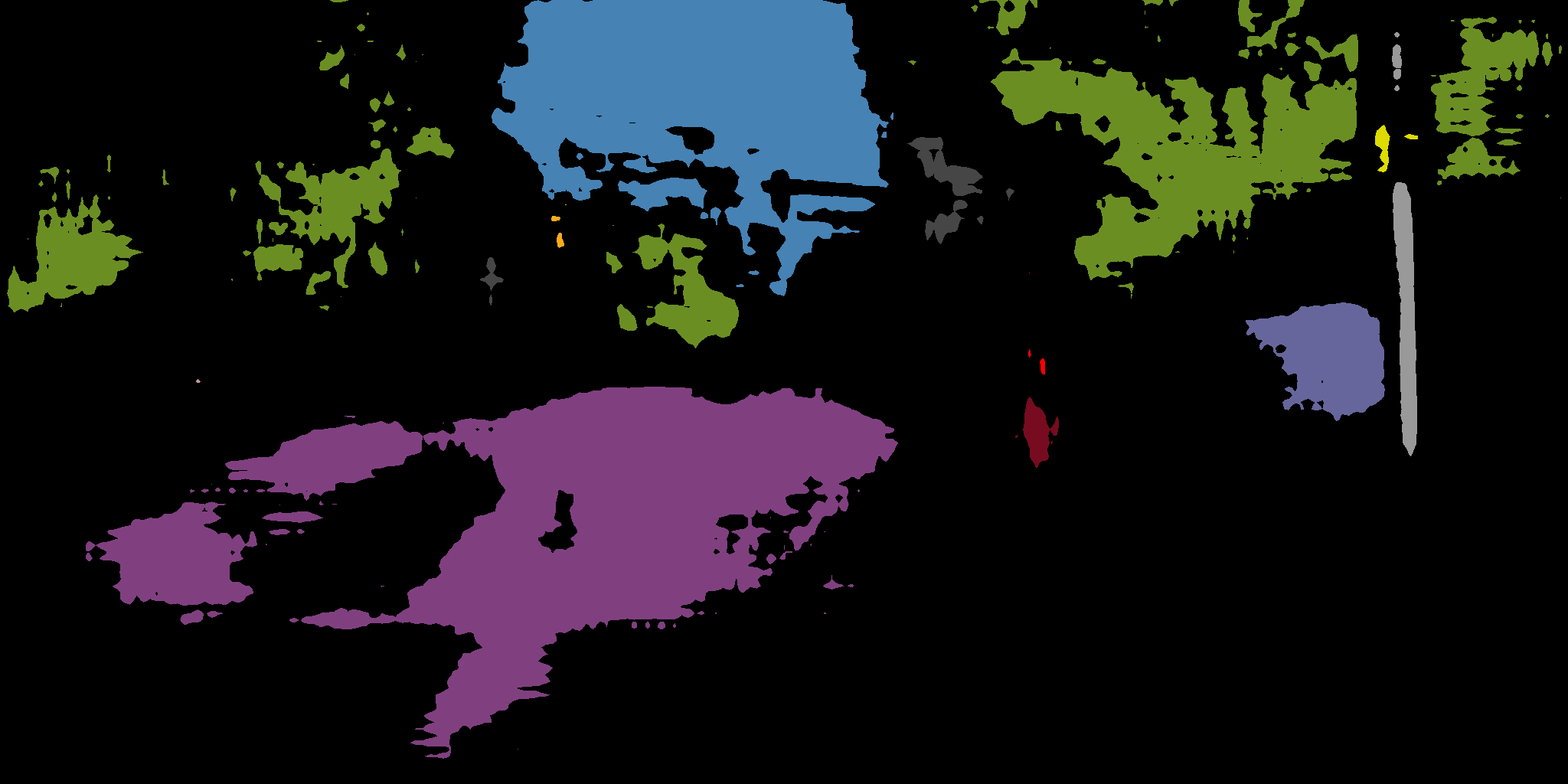}
	\caption{Adaptation results on GTA5 $\rightarrow$ Cityscapes. Rows correspond to sample images in Cityscapes. From top to bottom, rows correspond to original images, ground truth, and pseudo-label maps of CBST, MRL2, MRENT, MRKLD, LRENT.}
	\label{fig:plgta2city}
\end{figure*}

\end{document}